\documentclass[a4paper,fleqn]{cas-dc}

\usepackage[numbers]{natbib}
\usepackage{soul}
\usepackage{algorithm}
\usepackage{algpseudocode}
\usepackage{graphicx}
\usepackage{subcaption}
\usepackage[utf8]{inputenc}
\usepackage{textgreek}

\def\tsc#1{\csdef{#1}{\textsc{\lowercase{#1}}\xspace}}
\tsc{WGM}
\tsc{QE}
\tsc{EP}
\tsc{PMS}
\tsc{BEC}
\tsc{DE}

\begin{document}
\let\WriteBookmarks\relax
\def\floatpagepagefraction{1}
\def\textpagefraction{.001}

\shorttitle{Adaptive Surrogate-Based Strategy for Accelerating Convergence Speed in MOEAs}


\title [mode = title]{Adaptive Surrogate-Based Strategy for Accelerating Convergence Speed when Solving Expensive Unconstrained Multi-Objective Optimisation Problems}



\author{Tiwonge Msulira Banda}[type=editor,
                        auid=000,bioid=1,
                        orcid=0000-0002-2344-0397]
\cormark[1] 

\ead{t.banda@rgu.ac.uk} 



\affiliation{organization={School of Computing, Engineering and Technology, Robert Gordon University},
    city={Aberdeen},
    postcode={AB10 7QB}, 
    country={United Kingdom}}

\author{Alexandru-Ciprian Z\u{a}voianu}[type=editor,
                        auid=001,bioid=2,
                        orcid=0000-0003-1003-7504]
\ead{c.zavoianu@rgu.ac.uk} 
\cortext[cor1]{Corresponding author}




\begin{abstract}
\noindent Multi-Objective Evolutionary Algorithms (MOEAs) have proven effective at solving Multi-Objective Optimisation Problems (MOOPs). However, their performance can be significantly hindered when applied to computationally intensive industrial problems. To address this limitation, we propose an adaptive surrogate modelling approach designed to accelerate the early-stage convergence speed of state-of-the-art MOEAs. This is important because it ensures that a solver can identify optimal or near-optimal solutions with relatively few fitness function evaluations, thereby saving both time and computational resources. Our method employs a two-loop architecture. The outer loop runs a (baseline) host MOEA which carries out true fitness evaluations. The inner loop contains an Adaptive Accelerator that leverages data-driven machine learning (ML) surrogate models to approximate fitness functions. Integrated with NSGA-II and MOEA/D, our approach was tested on 31 widely known benchmark problems and a real-world North Sea fish abundance modelling case study. The results demonstrate that by incorporating Gaussian Process Regression, one-dimensional Convolutional Neural Networks, and Random Forest Regression, our proposed approach significantly accelerates the convergence speed of MOEAs in the early phases of optimisation.


\end{abstract}




\maketitle{}

\section{Introduction}
\label{sec:Introduction}
Multi-Objective Optimisation Problems (MOOPs) are a class of optimisation problems that involve multiple and often conflicting objectives. These are commonly encountered in industrial applications such as manufacturing and product design \cite{Deb2011b}, logistics \cite{Zhu2015, Bevilacqua2012}, etc. MOOPs rarely have a single solution (i.e., that is optimal across all the objectives simultaneously), as a gain in one objective results in a decline on another. Therefore, the general goal in solving MOOPs is to identify a  set of Pareto-optimal solutions (PS) that describe the best trade-offs between the considered objectives. This endeavour is complex as the PS can have an arbitrarily large size, especially when the number of objectives increases. For many problems, the PS is unknown and so a solver aims to find high-quality Pareto non-dominated solutions (PNs) that provide an approximation of the PS. We are particularly interested in problems that have 2 to 3 objectives, while those with 4 or more objectives are called many-objective optimisation problems \cite{Purshouse2007, Ishibuchi2008, Chand2015}. 

Over the decades, population-based nature-inspired Multi-Objective Evolutionary Algorithms (MOEAs) have proved effective at solving MOOPs because they are able to identify PNs in a single optimisation run for decision makers to review and choose from \cite{Coello2007}. Since the mid 70s, when John Holland popularised the Genetic Algorithm in his book \textit{Adaptation in Natural and Artificial Systems} \cite{Holland1975}, later updated in 1992 \cite{Holland1992}, effective and powerful MOEAs have been developed, improved upon and used in various research and industrial applications \cite{Deb2011b, Zhu2015, Bevilacqua2012, Bramerdorfer2018, Abdulrahman2023}. State-of-the-art MOEAs use three distinct approaches:
\begin{itemize}
\item MOEAs that use Pareto-dominance to guide the search process, e.g. the Strength Pareto Evolutionary Algorithm 2 (SPEA2) \cite{Zitzler1999}; the Nondominated Sorting Genetic Algorithm 2 (NSGA-II) \cite{Deb2002a}; and the Generalised Differential Evolution 3 solver (GDE3) \cite{Kukkonen2005}. 

\item Indicator-based MOEAs, e.g. Indicator-Based Evolutionary Algorithm (IBEA) \cite{ Zitzler2004}; the S-Metric Selection Evolutionary Multiobjective Optimisation Algorithm (SMS-EMOA) \cite{Beume2007}; and Hypervolume-based Estimation of Distribution Algorithm (HypE) \cite{ Bader2011}. 

\item Reference-point based decomposition MOEAs, e.g. Multi-Objective Evolutionary Algorithm based on Decomposition (MOEA/D) \cite{Zhang2007}; and NSGA-III \cite{Deb2014}. 
\end{itemize}

The effectiveness of MOEAs however comes into question when they are used to solve computationally intensive MOOPs (CI-MOOPs). The criticism arises from the fact that MOEAs typically need to evaluate (tens of) thousands of candidate solutions for them to find optimal or near-optimal solutions. While this is reasonable for problems whose fitness function is less time consuming to evaluate, it is not feasible for CI-MOOPs. Fitness functions for the majority of industrial optimisation problems take significant amounts of time to evaluate. Others require specialised software with limited licenses. For example, the optimisation of the design of an electric motor involves optimising the geometry of the motor, selection of materials, and thermal aspects, among others. Evaluation of each candidate design solution involves finite element analysis or computational fluid dynamics, which take a lot of time. For such a problem, one may only be able to give the MOEA a computational budget of a few hundred or thousand evaluations and hope that the algorithm would find good solutions \cite{Bramerdorfer2017a, Huber2022}.

To address the challenges above, the use of surrogate models has been proposed \cite{ SantanaQuintero2010}. Surrogate models (or meta models) are incorporated into a MOEA to speed up the optimisation run, either by replacing the expensive fitness function with an estimate \cite{Hardy1971, Myers2016, Andres2012, Ratle2001} or pre-selecting viable solutions \cite{Loshchilov2010a, Pan2019, Li2022, Yuan2022, Banda2024}. In the former, the goal is to reduce the number of expensive fitness function calls, whereas, in the latter, the aim is to improve efficiency by only expending the expensive fitness function call on viable solutions. While dozens, perhaps hundreds of surrogate models have been proposed and keep being published, there are still challenges associated with them, which present opportunities for further research. For example, there are questions such as: a) How can training data be efficiently created on-the-fly for the surrogate model? b) When is the right time to introduce and exit a surrogate model during an evolutionary run? c) What are efficient ways of integrating the surrogate into a MOEA? d) What data-driven model is efficient for a surrogate?

In this paper, we propose a strategy for constructing on-the-fly surrogate models to accelerate the convergence speed of existing MOEAs. Our approach addresses key challenges in surrogate-assisted optimisation above in the following unique ways.
\begin{itemize}
    \item To efficiently generate training data, we use solutions from the immediately preceding generation, eliminating the need for extensive archives or complex data filtering mechanisms (e.g., clustering). This reduces model training time and simplifies integration. In this way, no additional data-driven models or filtering procedures are required for selecting solutions for model training.

    \item The surrogate is introduced early in the run (i.e. after 2 generations) and remains active only while it continues to contribute to the discovery of promising solutions. A built-in adaptive exit mechanism automatically disables the surrogate once it stops offering improvement, ensuring that the surrogate does not unnecessarily delay the optimisation.

    \item Integration with a MOEA is done by attaching the surrogate as a modular accelerator, which estimates objective values for solutions and uses a simple random selection strategy for passing surrogate solutions to the MOEA for re-evaluation with the true fitness function. This minimises disruption and avoids the overhead of complex surrogate management. 

    \item The approach is flexible in terms of model choice; any effective data-driven regression model can be used to approximate the objective function.
\end{itemize}

While similar methods have been proposed in recent years \cite{Chugh2018, Yagoubi2023, Zhao2022, Zavoianu2022}, our approach is unique in its simplicity, plug-and-play modularity, and the minimal overhead it imposes on the base MOEA. The rest of the paper is structured as follows: In section \ref{sec:RelatedWork}, we go through some published literature that is related to our work. In section \ref{sec:ProposedApproach}, we describe our proposed approach. Section \ref{sec:ExperimentalDesign} contains the description of the experimental design. In section \ref{sec:Results}, we present our results and discuss them. Lastly, in section \ref{sec:Conclusion}, we conclude and highlight areas for further research.

\section{Related Work}
\label{sec:RelatedWork}

An unconstrained MOOP with $n$ real-valued variables and $m$ objectives can be defined as:
\begin{equation}\label{eq:equation1}
{Minimise\ F(x)=(f_1(x), … f_m(x)), \quad m \leq 3}
\end{equation} 
where $x\in \mathbb{R}^n$ represents a candidate solution and is subject to $x_l\leq x\leq x_u$, the lower and upper boundaries, and $f_1$ to $f_m$ represent individual objectives (usually not more than 3) that must be optimised simultaneously. In this paper, we assume that the objectives $f_1(x)$ to $f_m(x)$ are expensive to evaluate.

Numerous studies have explored the use of surrogate models to accelerate optimisation in Multi-Objective Evolutionary Algorithms (MOEAs). One of the earliest and most influential works is Jones et al. (1998) \cite{Jones1998}, who introduced Kriging to optimisation. Kriging, which is closely related to Gaussian Process Regression (GPR), is a method for approximating unknown functions by predicting values at unobserved points based on historical data. The Algorithm was called Efficient Global Optimisation (EGO). The method uses a weighted average of data points, with weights determined by the spatial correlation between them, captured by a covariance function. In optimisation, Kriging serves as a surrogate model for CI-MOOPs, providing not only predictions but also uncertainty estimates. This uncertainty helps balance exploration and exploitation, enabling more efficient optimisation by reducing the number of expensive fitness function evaluations. 

Originally, Kriging was used on single-objective optimisation problems, but later its use was extended to MOOPs. In this case, a surrogate model is constructed for each objective, for example in Pareto Efficient Global Optimization (ParEGO) \cite{Knowles2006}. Other than Kriging, various machine learning algorithms have been used as surrogates over the years, including support vector machines \cite{Andres2012}, radial basis functions \cite{Hardy1971}, polynomial regression \cite{Myers2016} and artificial neural networks (ANNs) \cite{Haykin1999} . These methods are effective, but the surrogates risk taking too long to train models as the number of objectives increases.

Graphical Processing Units (GPUs) have made it increasingly feasible to construct surrogate models based on deep neural networks (DNNs). Although DNNs can achieve higher accuracy than traditional machine learning models, their adoption as surrogate models has been limited due to their computational complexity and the longer training times required. With the advent of powerful GPUs, however, this burden has been significantly reduced, leading to the emergence of deep learning–based surrogate models. Examples include \cite{Wang2025}, where a multi-output deep neural network is used to approximate objective functions and constraints, and \cite{Guo2022}, which employs a dropout neural network.

Most surrogate models are attached to an existing MOEA to assist it. Usually, state-of-the-art MOEAs are used.  This approach serves well for comparison as it is easy to compare the surrogate-assisted MOEA with the unassisted baseline MOEA. Then there is the question of when it is best to introduce the surrogate. One approach is one where the MOEA alternates with the surrogate. After initialisation, the MOEA runs for several generations with the true fitness function evaluations before handing over to the surrogate when enough training data has been collected. Then the surrogate kicks in, running for several generations, producing a final set of solutions which are re-evaluated by the true fitness function. This was used in \cite{Zavoianu2013}, where ANN-based surrogate was attached to NSGA-II in what the authors called HybridOpt. In their experimentation, NSGA-II ran for 25 generations, and thereafter, the ANN-based surrogate was initialised, taking over from the MOEA, producing a final set of Pareto non-dominated solutions that was re-evaluated by the true fitness function. A similar approach was used by Nain and Deb in \cite{Nain2003} where an ANN-based surrogate was paired with a Genetic Algorithm (GA). The key difference with the HybridOpt is that in Nain and Deb, the two alternate multiple times. Upon initialisation, the GA runs with true fitness functions multiple times, then the surrogates are built and run for several times, handing back to the GA to run for several generations. Each time the surrogate kicks in, it is trained on updated data. The final population is evaluated with the true fitness function. Although effective, this alternating framework is not ideal when the goal is achieving early convergence as the MOEA needs to run for several generations before the surrogate can influence it. 

Another approach for attaching surrogates to MOEAs uses a two-loop architecture, where the MOEA runs in the main or outer loop and the surrogate runs in the secondary or inner loop. There are several papers using this approach with important differences in the way training data is constructed and how the resulting output from the surrogate is treated. In Yagoubi and Baderina \cite{Yagoubi2023}, support vector regression (SVR) based surrogate is attached to NSGA-II. After the initial population is created by NSGA-II in the main loop, the solutions are evaluated with the true function and passed on to the inner loop for training of surrogate models. The inner loop runs for several generations, using the surrogate models, after which a k-means clustering algorithm is used to cluster solutions, then non-dominated ranking is applied in each cluster to determine a winner from each cluster to be passed on to the main loop for re-evaluation with the true fitness function. Another algorithm, Two-stage infill Strategy and surrogate-ensemble assisted Expensive Many-objective evolutionary Optimization algorithm (TSEMO) \cite{Zhao2022} works in a similar fashion. While the algorithms outperformed the standard NSGA-II, the introduction of the clustering algorithm and another instance of non-dominated ranking increases the complexity and runtime of the solver.

Kriging Assisted Reference Vector Evolutionary Algorithm (K-RVEA) \cite{Chugh2018}  also incorporates Gaussian Processes Regression (GPR) as a surrogate to assist the Reference Vector guided Evolutionary Algorithm (RVEA) \cite{Cheng2016} in a two-loop architecture. The main loop uses RVEA and evaluates solutions using the true fitness function. The inner loop uses GPR to estimate the fitness functions. The algorithm uses uncertainty information from the Gaussian Processes to determine which surrogate solutions should be passed on to the main loop for re-evaluation with the true fitness function. A key distinction is in the way K-RVEA creates training data. The algorithm stores training data in archive with a pre-determined size. Individual solutions are added to the archive from recently evaluated solutions minus duplicates. If the number is large, reference point information and a clustering approach are used to select viable solutions. Again, the filtering of surrogate solutions is complex.

A slightly different, but important surrogate-assisted MOEA that uses the two-loop architecture is the Classification-based Surrogate-Assisted Evolutionary Algorithm (CSEA) \cite{Pan2019}. We label CSEA as different because the surrogate model in the solver does not approximate the objective functions directly; instead, it acts as a classifier that classifies candidate based on their estimated viability. The algorithm constructs training data by selecting a subset of reference solutions that have been evaluated using the true objective functions. The reference solutions are then labelled to form a classification boundary separating promising and unpromising regions of the search space. In its original formulation, CSEA employed a Feedforward Neural Network (FFNN) \cite{Svozil1997} as the surrogate classifier. The inner loop uses the classifier to guide the evolutionary search by rapidly screening candidate solutions, while the outer loop periodically updates the classifier using newly acquired true evaluations. Only the most promising candidate solutions, as predicted by the classifier, are selected for evaluation with the true/expensive fitness functions.

While a majority of surrogate-assisted MOEAs overwhelmingly use machine learning algorithms as surrogate models, a recent publication proposed a new way of lightweight surrogate models that are based on interpolation functions that do not require any training \cite{Zavoianu2022}. The approach aims to accelerate the convergence speed of MOEAs by promoting the early creation of high-quality candidate solutions through pre-emptive evaluation (PE) and speculative exploration (SE) -- which we describe below. Both strategies rely on the fitness approximation capabilities of lightweight interpolation models based on Shepard’s inverse distance weighting function \cite{Shepard1968}. By leveraging this function, the interpolation-based strategies reduce the number of fitness evaluations needed to generate high-quality Pareto front approximations during the MOEA process.

\begin{itemize} 
\item \textbf{Pre-emptive Evaluation (PE):} This approach has a filter that is used to pre-emptively evaluate a solution once it is created using Shepard’s inverse distance weighting function instead of the true fitness function. An individual that passes a certain threshold is selected, whereas one that doesn’t is discarded. The algorithm includes a mechanism for accepting solutions if the number of failed consecutive attempts to generate a high-quality offspring exceeds a certain threshold. This way, the algorithm forces the creation of high-quality solutions, which are passed on for evaluation with the true fitness function. 

\item \textbf{Speculative Exploration (SE):} This approach follows a two-loop architecture as described before. A key feature is that a surrogate multi-objective interpolated continuous optimisation problem (MO-ICOP) that mirrors the definition of the original problem to be solved is constructed in the inner loop and evaluated using Shepard’s inverse distance weighting function instead of the true fitness function. After a predefined number of generations, the inner loop outputs surrogate solutions for evaluation with the true fitness function. 
\end{itemize}

In \cite{Zavoianu2022}, the two approaches were incorporated into NSGA-II and the Differential Evolution-based, Coevolutionary Multi-objective Optimization (DECMO++) \cite{Zavoianu2018}. An instance of NSGA-II enhanced with both strategies (NSGA-II PE+SE) significantly sped up the convergence speed of NSGA-II, whilst only using the PE strategy induced a smaller performance boost. 

In this paper, we adopt the two-loop architecture and,  building upon  recent developments in \cite{Banda2024}, introduce a novel general-purpose method for on-the-fly surrogate-based MOEA convergence acceleration featuring adaptive surrogate activation and streamlined training data generation and surrogate result integration. As such, we only use solutions from the previous generation to build the surrogate models and, instead of employing clustering algorithms, we randomly select surrogate solutions to pass to the main loop for re-evaluation using the true fitness function. Surrogates are activated as early as possible, but their usage interval is governed by an adaptive performance indicator. By limiting total surrogate usage as well as their training data to the previous generation evaluated with the true fitness function, our approach aims to strike a balance between more intensive ML models that require several generations to collect sufficient training data and/or rely on complex filtering algorithms on one hand, and the recent interpolation-based (light) surrogate models that do not require training but might be more prone to underfitting on the other hand. To illustrate the pros and cons of the proposed approach, we integrated the proposed adaptive on-the-fly surrogate accelerator strategy with NSGA-II and MOEA/D, which are dominance-based and reference-point-based algorithms, and carried out both comprehensive benchmark testing and a case study analysis.

\section{Proposed Approach}
\label{sec:ProposedApproach}

\begin{figure}
    \centering
    \includegraphics[width=0.49\textwidth]{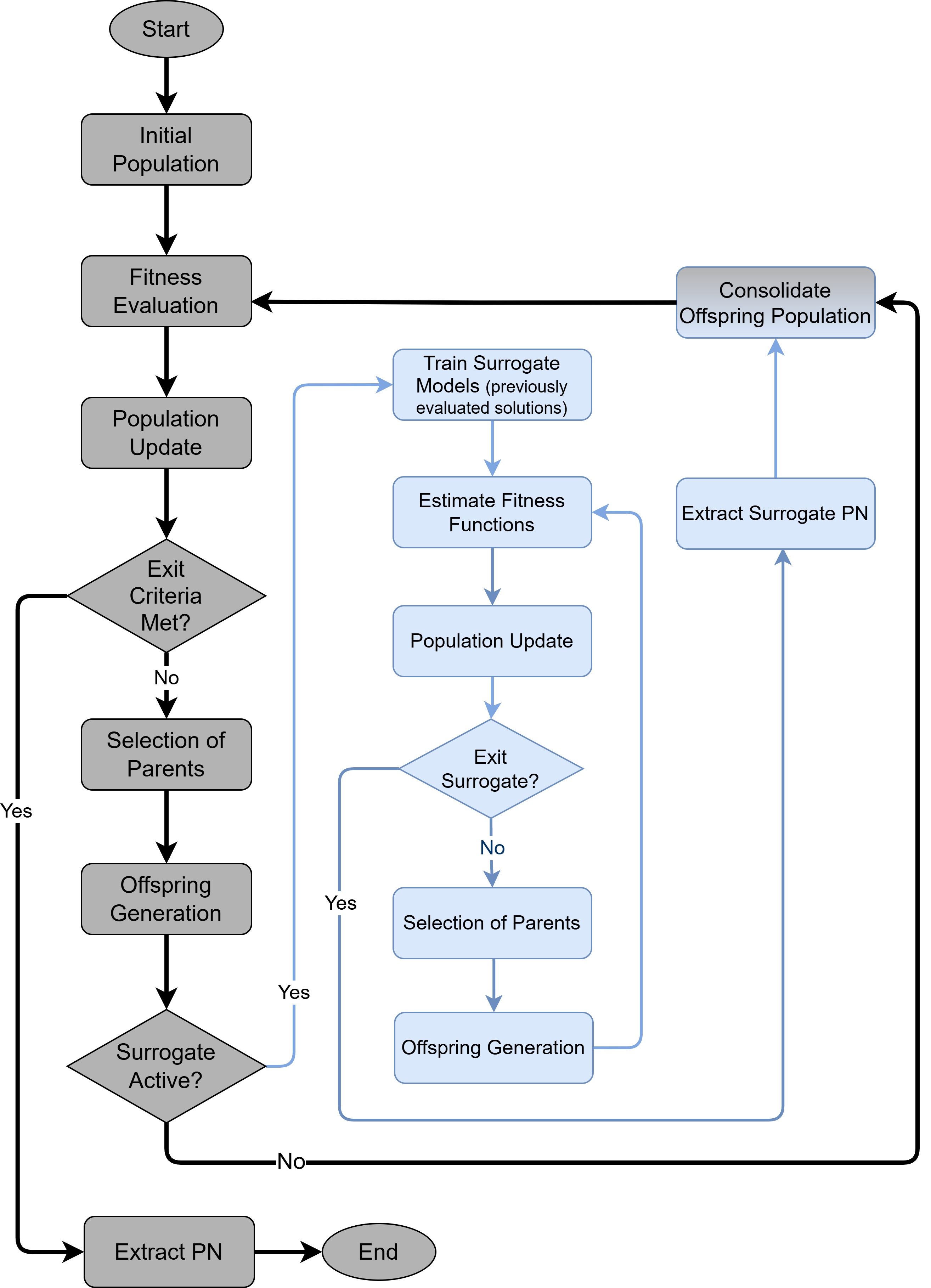}
    \caption{Overview of the Adaptive Accelerated MOEA showing the two loops. The standard (host) MOEA is shown in black whereas the surrogate-based accelerator is shown in blue.}
    \label{fig:RegSurrogate}
\end{figure}

In this section, we describe our proposed framework for accelerating the convergence speed of MOEAs when solving computationally intensive, unconstrained MOOPs. As mentioned in section \ref{sec:RelatedWork}, our approach adopts a two-loop architecture. A standard / host MOEA, in our case, NSGA-II or MOEA/D (shown in grey in Figure \ref{fig:RegSurrogate}) runs in the outer loop and performs true (but expensive) fitness function evaluations. The Adaptive Accelerator, which runs in the inner loop (highlighted in light blue in Figure \ref{fig:RegSurrogate} and detailed in lines \hyperref[alg:acceleratedMOEA]{8-17} of Algorithm \ref{alg:acceleratedMOEA}), is responsible for generating and evaluating candidate solutions using a surrogate model. To avoid ambiguity, we refer to solutions generated by the standard / host MOEA as real solutions, and those generated by the Adaptive Accelerator as surrogate solutions. An MOEA equipped with the Adaptive Accelerator will be referred to as an Accelerated MOEA or a surrogate variant (of the standard MOEA). In terms of integration, the Adaptive Accelerator is a customised extension of its host MOEA and is positioned between the Offspring Generation and Fitness Evaluation phases of the host MOEA.

\begin{algorithm*}
\caption{Accelerated Multi-Objective Evolutionary Algorithm (Accelerated MOEA)}
\label{alg:acceleratedMOEA}
\begin{algorithmic}[1]

\Statex \textbf{Input:} Population size $N$, number of main generations $G$
\Statex \textbf{Output:} Non-dominated solution set

\State $mainPopulation_0 \gets \text{CreateInitialPopulation}(N)$
\State $mainPopulation_0 \gets \text{FitnessEvaluation}(mainPopulation_0)$
\State $mainGen \gets 1$
\State $G_s \gets \lfloor G / 2 \rfloor$ 

\While{$mainGen < G$}
    \State $mainOffspring \gets \text{SelectParents}(mainPopulation_{mainGen - 1})$ 
    \State $mainOffspring \gets \text{OffspringGeneration}(mainOffspring)$ 
    
    \State $surrogateActive \gets \text{EvaluateSurrogateStatus()}$
    \If{$surrogateActive$}
        \State $surrogateModel \gets \text{TrainSurrogates}(mainPopulation_{mainGen - 1})$
        \State $surrogatePopulation_0 \gets \text{EstimateFitness}(surrogateModel, mainOffspring)$
        \State $surrGen \gets 1$
        
        \While{$surrGen < G_s$}
            \State $surrogateParents \gets \text{SelectParents}(surrogatePopulation_{surrGen - 1})$
            \State $surrogateOffspring \gets \text{OffspringGeneration}(surrogateParents)$
            \State $surrogateOffspring \gets \text{EstimateFitness}(surrogateModel, surrogateOffspring)$
            \State $surrogatePopulation_{surrGen} \gets \text{PopulationUpdate}(surrogatePopulation_{surrGen - 1}, surrogateOffspring)$
            \State $surrGen \gets surrGen + 1$
        \EndWhile

        \State $mainOffspring \gets \text{ConsolidateOffspring}(mainOffspring, surrogatePopulation_{surrGen - 1})$
    \EndIf

    \State $mainOffspring \gets \text{FitnessEvaluation}(mainOffspring)$ 
    \State $mainPopulation_{mainGen} \gets \text{PopulationUpdate}(mainPopulation_{mainGen - 1}, mainOffspring)$
    \State $mainGen \gets mainGen + 1$
\EndWhile

\State \Return $\text{nonDominatedSolutions}(mainPopulation_{mainGen - 1})$

\end{algorithmic}
\end{algorithm*}

The Accelerated MOEA operates as follows: When initialised, the standard / host MOEA runs normally for the first two generations (\hyperref[alg:acceleratedMOEA]{
lines 1-7 and 23-25}). This initial phase allows the evolutionary process to start shaping the Pareto front, as the initial generation (generation 0) consists of randomly generated solutions which are typically too scattered in the objective space to be effectively used by the subsequent acceleration mechanism. Solutions from the next generation onwards (generation 1+) are created by the solver’s genetic operators and it is at this point that the Pareto Front begins to evolve. We hypothesise that for surrogates that aim at fast initial convergence, it is beneficial to introduce them into the host MOEA as early as possible to get maximum benefit and consider generation=2 is an appropriate surrogate initialisation threshold. The training data creating mechanism we propose is very straightforward. When the Adaptive Accelerator is activated (line 8), the recently evaluated population (\textit{mainPopulation}) and the new offspring population  (\textit{mainOffspring}), which is unevaluated, are sent to the Adaptive Accelerator. \emph{\textbf{Only the recently evaluated population is used to train surrogate models}} (\hyperref[alg:acceleratedMOEA]{line 10}) for each objective, whereas the unevaluated offspring population is loaded as the initial population for the Adaptive Accelerator and evaluated using the trained surrogate models (\hyperref[alg:acceleratedMOEA]{line 11}). The Accelerator then proceeds with its own evolutionary process for several generations using the trained surrogate models for objective function evaluation (\hyperref[alg:acceleratedMOEA]{lines 12-19}). At the end, the Accelerator outputs a set of surrogate non-dominated solutions (Surrogate PN or \textit{surrogatePopulation}), some of which are fed back into the host MOEA in the main loop for re-evaluation using the actual fitness function. Extensive experiments we report on in section \ref{sec:Results} demonstrate that this approach allows the Adaptive Accelerator to significantly speed up the convergence of its host MOEA at a reduced computational cost.

\subsection{Data collection and pre-processing}
\label{sec:DataCollection}
As described earlier, the outer loop passes on the most recently evaluated population of solutions to the Adaptive Accelerator component. The solutions are processed accordingly to create training data, assigning the solution variables as input features and solution objective function values as outputs or targets. The size of the training set is always equal to the population size, which helps reduce the amount of time required to train the surrogate models, even when employing more complex ML architectures and best parameter grid searches. The training data is then normalised using minimax normalisation, which involves rescaling data range between 0 and 1, by subtracting the minimum value and dividing by the range of the feature. 

\subsection{Model training and fitness function estimation}
\label{sec:ModelTrainig}
The success of any surrogate model lies in its ability to accurately estimate the objective function values of new solutions. Due to the fact that the objectives of a given problem can be quite heterogenous and the surrogate models need to be created on-the-fly, the Accelerator creates an individual surrogate model for each objective. Technically, any regression model can be integrated and used for this purpose, and we experimented with Random Forest Regression (RFR), Gaussian Processes Regression (GPR), and 1-Dimension Convolution Neural Network (CNN). We chose to experiment with these as surrogate models due to their established use in surrogate-assisted optimisation. The three, as with many ML models are non-parametric, i.e., they do not assume a specific data distribution. This is important for our modelling strategy because we cant not make any assumptions about the distribution of the training data, especially in the early phases of the optimisation. Below, we provide some details about the three chosen models:
\begin{itemize}
\item \textbf{Random Forest Regression (RFR):} This is a robust non-parametric supervised machine learning algorithm used for classification and regression problems. Originally proposed by Breiman in 2001 \cite{Breiman2001}, the algorithm is based on the principle of ensemble learning through bagging (Bootstrap Aggregating). The algorithm builds numerous independent Decision Trees during training, each trained on a random subset of data and features. For classification tasks, the final output is determined by majority voting among the trees, while for regression, it is the average of all tree predictions. An individual Decision Trees works by creating hierarchical tree-like structures that partition the data based on the values of features, leading to a prediction about a target variable. The fundamental idea behind a Decision Tree is to create a set of rules that can be used to classify or predict the value of a target variable based on the features of the data. These rules are learned by recursively partitioning the data into smaller and smaller subsets based on the most significant features \cite{Probst2019}. RFR is robust to noise and capable of capturing complex, non-linear interactions.
\item \textbf{Gaussian Processes Regression (GPR):} Gaussian Process \cite{Rasmussen2006} is also a non-parametric, supervised machine learning algorithm for both regression and classification. It differs from most other machine learning models in that, in addition to making an estimate, it also provides a measure of uncertainty, which quantifies how confident the model is of its estimate. Conceptually, a Gaussian Process can be viewed as a distribution over possible functions, fully defined by a mean function (often assumed to be zero) and a covariance function or kernel, which measures the similarity between input points. GPR is widely used in surrogate modelling, particularly when training data is limited. It operates by considering an infinite set of potential functions that could explain the observed data and represents relationships between data points using a covariance matrix derived from the kernel. Similar points have high covariance values, while dissimilar ones have low covariance. When making predictions, the GPR uses the kernel to evaluate how closely a new point relates to the training data and produces a predictive distribution. The mean represents the expected output and the variance indicates the model’s confidence.

\item \textbf{1-Dimension Convolution Neural Network (CNN):} This is a type of deep learning model primarily designed to process sequential or time-series data. Unlike traditional fully connected CNNs, a 1D-CNN applies convolutional filters along one spatial dimension, enabling it to automatically extract local patterns and dependencies within the sequence without the need for manual feature engineering \cite{Kiranyaz2015}. For a surrogate model this is important  because the algorithm learns the relationships between the decision variables. The model consists of convolutional layers that slide learnable filters across the input sequence to produce feature maps, followed by pooling layers that down sample these maps to reduce dimensionality and highlight the most important features. These extracted features are then passed through fully connected layers for final prediction.  Due to its ability to learn spatial hierarchies of features directly from raw data, the 1D-CNN is both computationally efficient and highly effective in capturing local correlations and temporal structures, making it a powerful choice for a surrogate \cite{ Zhang2017}.

The architecture of the 1D-CNN comprised of an input layer accepting sequences of length corresponding to the number of the problem’s decision variables, followed by two 1D convolutional layers, each with 64 filters, ReLU activation, L2 regularisation, and a kernel size of 3. Each 1D convolutional layer was followed by a dropout layer with a dropout rate of 0.2. The convolutional outputs are flattened and passed through a dense layer with 32 neurons for feature abstraction, followed by a single output neuron to produce the final prediction. The small kernel size biases the network to predominantly focus on local variable interaction patterns (dependencies between adjacent decision variables), whilst the robust multi-layer architecture enables it to capture wider / more complex interaction patterns. An adaptive kernel size might improve performance across problem classes, but it will also increase the complexity and computational cost of the surrogate model, which will further increase the surrogate overheads on the base MOEA.

\end{itemize}

The first stage in model training is to find the best hyper-parameters for the model and this stage is known to have a high time complexity. Apart from dataset size, the time complexity for hyper-parameter search depends on the number of parameter combinations sampled, the number of cross-validation (CV) folds, and the time complexity of fitting the estimator itself. To reduce the overall time complexity of the algorithm, the Adaptive Accelerator uses randomised search cross-validation, which has a reduced number of parameter combinations. When the best parameters are identified, a final model is trained and is used to estimate the fitness function values of the particular objective. After constructing all required surrogates, the Adaptive Accelerator proceeds with the rest of the evolutionary steps, population update, selection of parents and reproduction of offspring until its exit criterion is met (i.e., a pre-defined number of inner loop generations has been evolved). Numerical experiments on standard PCs/CPUs show that given the limited sizes of the training sets and usage of randomised search CV, the full training of a surrogate model takes on average 205.4 seconds (3 minutes and 25 seconds). This can be undoubtedly improved on using GPU-based parallelisation for certain classes of models. As inference times are negligible, with a basic objective-wise training parallelisation, the total surrogate usage overhead is limited by the adaptive deactivation mechanism to between 17 minutes (GPR) and 36 minutes (RFR). The full breakdown of the surrogate computational burden is presented in Table \ref{tab:surrogate_overheads}.  

\begin{table}[width=.99\linewidth,cols=4,pos=h]
\caption{Details of average surrogate usage overheads (in seconds). The single inner loop sums up all inference times for one run of the inner loop whilst total usage shows the average overhead of applying the Adaptive Accelerator.}
\label{tab:surrogate_overheads}
\begin{tabular*}{\tblwidth}{@{} LLLL@{} }
\toprule
\textbf{Model} &  \textbf{Single training}& \textbf{Single inner loop} & \textbf{Total usage}\\ \midrule
RFR & 260.18 & 1.24 & 2091.47\\ 
GPR & 208.86 & 0.27 & 1018.49\\ 
1D-CNN & 147.05 & 7.94 & 1586.10\\ 
\bottomrule
\end{tabular*}
\end{table} 

\subsection{Parameters}
\label{sec:Parameters}

Having described how the algorithm works, it is time to introduce the key parameters of the proposed algorithm in more detail. While we describe them as parameters that may require tuning to improve algorithm performance, we are aware that introducing user-defined parameters increases algorithm complexity. In \cite{Deb2001} and \cite{ Zitzler2000}, it is established that minimising user-defined parameters enhances the algorithm’s robustness, and makes comparative evaluation complicated. The evolution from NSGA \cite{Srinivas1994} to NSGA-II is a good example of how removing user-defined parameters can improve usability and performance stability. Similarly, the decomposition-based \cite{Zhang2007} and reference-based \cite{Deb2014} solvers demonstrate that adaptive or structural design choices can replace the need for manually tuned parameters, aligning with the broader trend towards self-adaptive and parameterless evolutionary algorithms. 

To avoid introducing additional user-defined parameters, we adopt a "rule of 1/2" strategy that links the three parameters described below to existing parameters of the host algorithm. This approach renders the proposed surrogate modelling technique effectively adaptive and apparently parameterless, as the new parameters don't need to be manually specified.

\begin{itemize}
\item \textbf{Number of inner loop surrogate generations:} When the algorithm enters the inner loop, i.e., the Adaptive Accelerator, a question arises: \emph{how many generations should it run for?} A key aspect of answering this question lies in the fact that surrogate models are trained each time the Adaptive Accelerator is activated and used to estimate objective functions until the end of the inner loop run. The challenge is that if the surrogate runs only for a few inner generations, we limit its acceleration capability. On the other hand, if we run it for too long, the performance of the surrogate models degrade given that the search space shifts as the optimisation progresses. In machine learning terms, there is a data shift that happens when the distribution of data used for training models differs from that of the data used in deployment \cite{Rahmani2023} (i.e., very late inner loop evaluations). Based on preliminary experiments, we decided to fix the number of surrogate generations by setting it to be equal to half of the generations used by the outer loop (line 4 of Algorithm \ref{alg:acceleratedMOEA}).

\item \textbf{Surrogate integration threshold:} When the Adaptive Accelerator exits, it outputs a set of surrogate PN solutions that is equal to or less than the offspring population size of the host solver. The question is, \emph{how many of these surrogate PN solutions should be transferred to the offspring population of the main (host) loop for re-evaluation with the true fitness function?} We noted during our preliminary experiments that letting the offspring population of the host MOEA comprise of surrogate solutions only led to loss of diversity and degraded performance, but if we only took a fraction of the surrogate PNs, the MOEA's performance would improve.  To ensure that the surrogate effectively plays the role of an accelerator to the host MOEA, we introduce a threshold for the maximum number of surrogate solutions to add to the offspring population for re-evaluation. The rest of the solutions are taken from the original offspring population generated by the host MOEA before sending them to the Accelerator. This is depicted by the step  "Consolidate Offspring Population" in Figure \ref{fig:RegSurrogate} and line 20 in Algorithm \ref{alg:acceleratedMOEA}. We experimented with 25\%, 50\%, 75\% and 100\% of the offspring population size and settled on a constant threshold value of 50\%.

\item \textbf{Adaptive surrogate deactivation:} Whilst, all the experiments that we carried out indicate that for each tested problem there is a tipping point after which our Adaptive Accelerator no longer benefits its host MOEA, the exact placement (in terms of generation no.) of the tipping point is problem and even optimisation run specific. As such, we propose an adaptive strategy to answer the key emerging question of \emph{when is the best time to deactivate the surrogate?}
The idea is that whilst the surrogate is active, we monitor the host solver's archive and automatically disable the Adaptive Accelerator when its contribution declines. Specifically, when evaluating the surrogate status at host generation $t$ (line 8 of Algorithm \ref{alg:acceleratedMOEA}), we track the proportion of surrogate solutions currently in the archive that were generated at host generation $t-2$, and once this percentage falls to or below half of its historical maximum (i.e., its "half-life"), the Accelerator is deactivated. Let $\mu$ denote the historical maximum percentage of surrogate solutions from the penultimate generation and $\chi_{t}$ the current percentage of penultimate surrogate solutions in the archive. The surrogate is deactivated when:
\begin{equation}
(\chi_{t} \leq \frac{\mu}{2} \land \mu \neq 0) \lor (t=50).
\label{eq:AdaptiveExitCriterion}
\end{equation}
The reason for using the survivability of surrogate-based solutions in the host archive over two generations as a proxy for their usefulness is aligned with the very rationale for creating them: obtaining superior solutions that can leapfrog the current state of the evolutionary process and hopefully accelerate its convergence once introduced in the population pool. Once surrogate-based solutions no longer retain a minimal survivability advantage over regular offspring, their speed-up potential is deemed insufficient for accelerating the search. It is noteworthy that according to Equation \ref{eq:AdaptiveExitCriterion}, there is a forced Adaptive Accelerator exit at generation $t=50$. The hard stop threshold was decided based on two factors: (i) the design of our Adaptive Accelerator targets the very early host MOEA convergence stage which is likely to be reached by generation no. 50 \cite{Zavoianu2015} and (ii) as shown by the comprehensive benchmark tests in section \ref{sec:ExitMechanism}, on average the Adaptive Accelerator is deactivated much earlier across all MOOPs.    

\end{itemize}

\begin{algorithm*}
\caption{Adaptive Surrogate Exit Mechanism}
\label{alg:AdaptiveSurrogateExit}
\begin{algorithmic}[1]
\State \textbf{Input:} Archive $\mathcal{A}_t$ at generation $t$, historicalMaximum $\mu$
\State \textbf{Output:} Accelerator status (active/inactive)
\State
\State $s_t \gets$ count of surrogate solutions in $\mathcal{A}_t$ generated at generation $t-2$
\State $|\mathcal{A}_t| \gets$ total number of solutions in archive
\State
\State $\chi_t \gets \frac{s_t}{|\mathcal{A}_t|}$ \Comment{Current proportion of surrogate solutions from penultimate generation}
\State
\If{$p_t > \mu$}
    \State $\mu \gets \chi_t$ \Comment{Update historical maximum}
\EndIf
\State
\If{$\mu \neq 0$ \textbf{and} $\chi_t \leq \frac{\mu}{2}$}
    \State \textbf{deactivate} Accelerator \Comment{Half-life threshold reached}
\ElsIf{$t = 50$}
    \State \textbf{deactivate} Accelerator \Comment{Generation limit reached}
\Else
    \State \textbf{keep} Accelerator active
\EndIf 
\end{algorithmic}
\end{algorithm*}

It is important to note that our strategy and fixed parameter choices promote very early surrogate usage (and deactivation) as well as a fairly balanced integration/transfer of surrogate-based results. There are two main reasons for this. Firstly, surrogates that are very well correlated with the true fitness function, but not necessarily extremely accurate are likely to still advance the search early in the run (i.e., the "exploration" stage) as most MOEAs have an evolutionary  logic ultimately grounded on decisioning regarding the "order" between solution candidates (e.g., Pareto dominance / non-dominance, better / not better than previous candidate solution for a given decomposition vector). Towards the end of the runs (i.e., the "exploitation" stage), errors in surrogate accuracy are far more likely to translate into ordering errors when contrasting with alternative decisions based on the true fitness function. Secondly, by using surrogates to accelerate the search during "exploration", one runs the risk of exacerbating linkage disequilibrium \cite{altenberg1995} by "jumping over" key sections of the search space where the population of the base solver would normally “gather” building blocks that are critical for high-quality mid/end of the run solutions. To mitigate this, in our approach, 50\% of offspring are generated as per the base solver logic. 

\section{Experimental Design}
\label{sec:ExperimentalDesign}

\subsection{Incorporation into Solvers}
\label{IncorporationIntoSolvers}
We incorporated the Adaptive Accelerator into two very well-known and widely used MOEAs: NSGA-II \cite{Deb2002a} and MOEA/D \cite{Zhang2007}. The two are distinct in that the former is an elitist algorithm that uses Pareto dominance as a search strategy, whereas the later uses reference-point based decomposition. 

\begin{itemize}
\item [$\blacksquare$] \textbf{NSGA-II:} NSGA-II is the updated version of the Non-dominated Sorting Genetic Algorithm originally proposed in 1994 \cite{Srinivas1994}. The algorithm employs a non-dominated sorting procedure to categorise solutions into hierarchical non-domination fronts, ranking solutions based on Pareto dominance, with the most optimal ones placed in the leading fronts. To maintain population diversity within each front, NSGA-II utilises a crowding distance mechanism, which penalises solutions that are too closely spaced in the objective space, thus encouraging the internal storage of a well-distributed set of solutions. In light of its efficiency and robustness, NSGA-II has been widely adopted across diverse industrial applications \cite{Verma2021}. We integrated the Adaptive Accelerator into NSGA-II exactly as described in section \ref{sec:ProposedApproach}. We parameterised the host NSGA-II  based on literature recommended settings. Specifically, the population and offspring population sizes were both set to 200. We used Simulated Binary Crossover (SBX) \cite{Deb1995} with a crossover probability rate of 0.8 and a crossover distribution index of 20, and Polynomial Mutation (PM) \cite{Deb1996} with a mutation probability of $1/n$ and a mutation distribution index of 20. The same settings were set for the Adaptive Accelerator NSGA-II component in the inner loop, except that the number of evaluations was set to half of that in the outer loop. 

\item [$\blacksquare$] \textbf{MOEA/D:} MOEA/D solves MOOPs by decomposing them into a set of single-objective sub-problems, each capturing a distinct trade-off among objectives. These sub-problems are optimised simultaneously, with each one guided not only by its own search but also by information shared from neighbouring sub-problems. The neighbourhood structure is defined based on the relative distances between sub-problems in the objective space. At each generation, the algorithm maintains a population comprising the best-known solution for each sub-problem \cite{Zhang2007}. We used a variant of MOEA/D with dynamic resource allocation (DRA). We parameterised the host MOEA/D with  a population size of 300, the Differential Evolution Crossover (with a crossover rate of 0.2, and a scaling factor of 0.5), and Polynomial Mutation (with probability rate of $1/n$ and distribution index of 20) and the Tschebycheff aggregation function with the dimension set equal to the number of objectives). We also set the neighbour size to 20, the neighbourhood selection probability to 0.9, and the maximum number of replaced solutions to 2. The same settings were used for the Adaptive Accelerator MOEA/D component inside the inner loop, with the number of evaluations set to half of that in the outer loop.
\end{itemize}

Given that the two solvers work differently, i.e. NSGA-II is generational and MOEA/D is a steady-state MOEA, we modified MOEA/D slightly in the way we capture a generation. After the initial population, an iteration in the main step only contains a single solution. For us to capture a generation comparable to NSGA-II, we created an external store for these offspring solutions and count a generation when the number of individuals reached the desired population size (i.e., 300 for the results reported on in section \ref{sec:Results}).

\subsection{Benchmark Problems}
\label{sec:BenchmarkProblems}
We evaluated the efficacy of the proposed Accelerated MOEAs on a test harness consisting of 31 benchmark problems drawn from established suites as summarised in Table \ref{tab:benchmark_problems}. The test suites are: DTLZ \cite{Deb2002} (all 7 problems), KSW10 \cite{Kursawe1991} (1 problem), LZ09 \cite{Li2008} (all 9 problems), WFG \cite{Huband2006} (all 9 problems), and ZDT \cite{Zitzler2000} (5 problems -- ZDT5 excluded because it is binary). Performance of our proposed Accelerated NSGA-II and Accelerated MOEAD/DRA was mainly compared against the standard NSGA-II and MOEA/D-DRA variants, but we also contrasted our results to those of interpolation-based surrogates reported in \cite{Zavoianu2022}. The algorithms were given a fixed computational budget of 50,000 fitness function evaluations for every test problem. For each benchmark problem, 100 independent runs of each solver were performed, and the mean performance was recorded to mitigate the influence of stochastic variability. The test harness was implemented using jMetalPy version 1.7.0, a comprehensive Python framework for single- and multi-objective optimisation problems with metaheuristics \cite{BenitezHidalgo2019}.

\begin{table}[width=.99\linewidth,cols=3,pos=h]
\caption{Details of the 31 benchmark problems used for performance comparison.}
\label{tab:benchmark_problems}
\begin{tabular*}{\tblwidth}{@{} LLL@{} }
\toprule
\textbf{Problem} &  \textbf{No. of variables}& \textbf{No. of objectives}\\ \midrule
DTLZ1 & 7 & 3 \\ 
DTLZ2-6 & 12 & 3 \\ 
DTLZ7 & 22 & 3 \\ 
KSW10 & 10& 2 \\ 
LZ09\_F1-F5, F9 & 30 & 2 \\ 
LZ09\_F6 & 30 & 3 \\ 
LZ09\_F7-F8 & 10 & 2 \\ 
WFG1-9 & 6 & 2 \\ 
ZDT1, 2 & 30 & 2 \\ 
ZDT3, 4, 6 & 10& 2 \\ 
\bottomrule
\end{tabular*}
\end{table}

\subsection{Performance Indicator}
\label{sec:PerformanceIndicator}
 We used the hypervolume indicator ($H$) \cite{Zitzler1998}, a unary quality metric for evaluating Pareto fronts, as our main metric for measuring the performance of the solvers. The hypervolume $H(PF_c)$ quantifies the volume of the objective subspace dominated by a candidate Pareto front $PF_c$, relative to a specified anti-optimal reference point. A higher $H$ indicates better performance. The $H$ is commonly used in the MOEA community because it has theoretical proof of monotonic convergence \cite{Auger2009} and characterises the $PF_c$ both in terms of diversity (i.e., spread across) and convergence (i.e., distance from) the true Pareto Front of the MOOP. In itself, $H$ is not intuitive as it just comes in the form of a numerical value that depends on the choice of the anti-optimal reference point. To make it easy to interpret, we compute the relative hypervolume, defined as: 
 \begin{equation}
    {Hv}(PF_c) = 100 \cdot \frac{H(PF_c)}{H(PF_t)}, 0 \leq {Hv}(PF_c) \leq 100.  
 \end{equation}
 \noindent where $H(PF_t)$ denotes the hypervolume of the true Pareto front for each benchmark problem. We calculated $Hv(PF_t)$ for all 31 benchmark problems using their known true Pareto fronts. Using this formulation, for each problem and solver pairing, we computed the mean of the relative hypervolume achieved at the end of each generation across the 100 independent runs.

As a secondary general metric for measuring solver performance on benchmark problems, we have used the inverse generational distance $IGD(PF_c)$ \cite{coello2005}. Like the hypervolume, the $IGD$ assesses both diversity and convergence of a candidate Pareto front. A smaller $IGD$ indicates better performance and, in order to enable averaging across the benchmark, we have minmax normalised $IGD$ values between 0 and 1 for each test problem.

For a particular comparison, we have also employed the spread metric $\Delta(PF_c)$ \cite{Deb2002a}. In the case of $\Delta$, smaller values indicate that the solutions in the $PF_c$ are more equally distributed along the true Pareto front, the minimal value is problem specific and values larger than 1 are possible.  

\section{Results and Discussion}
\label{sec:Results}

\subsection{Adaptive Surrogate Deactivation}
\label{sec:ExitMechanism}

\begin{figure*}
    \centering
    \begin{subfigure}[b]{0.455\textwidth}
        \centering
        \includegraphics[width=\textwidth]{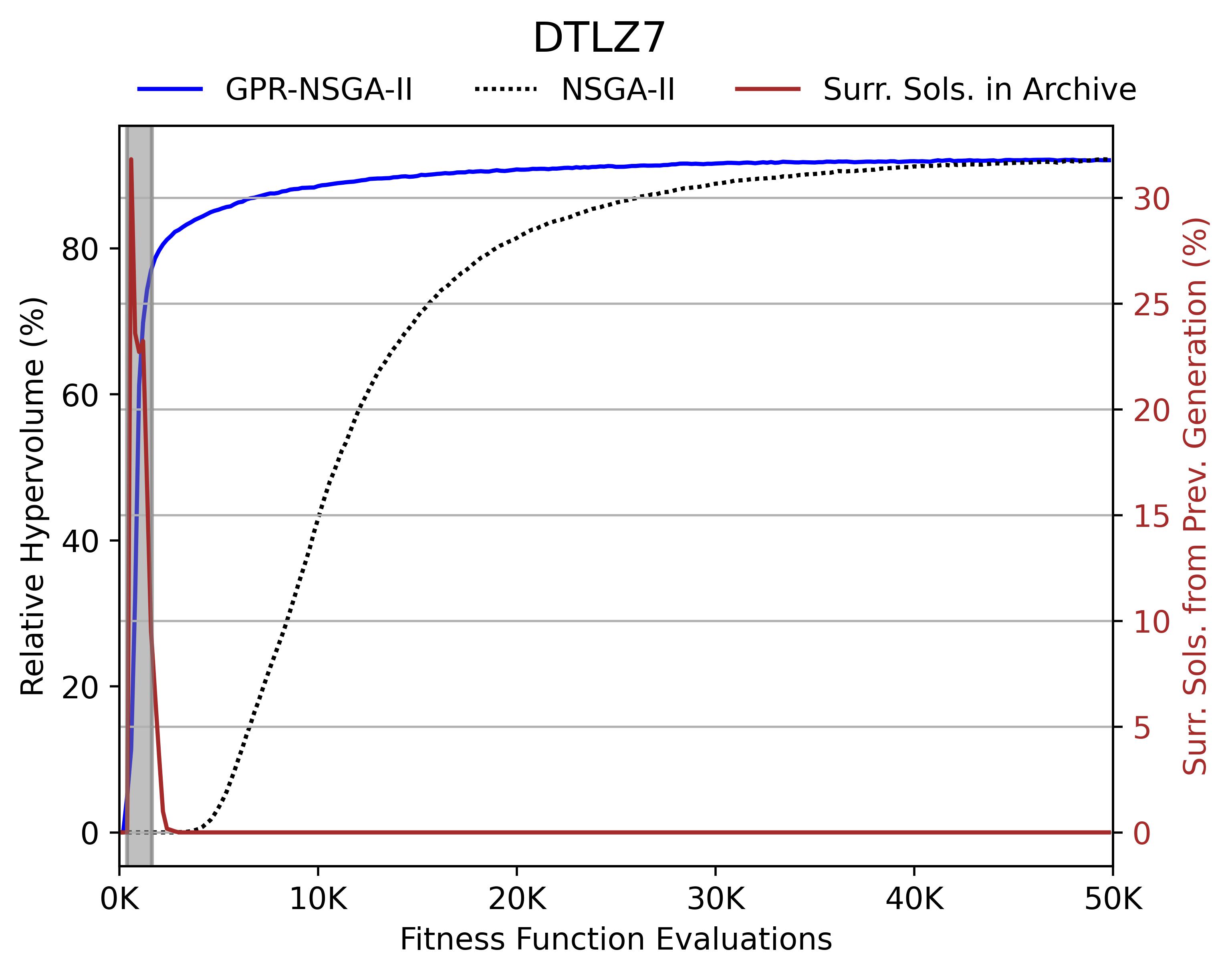}
        \caption{NSGA-II}
    \end{subfigure}
    \begin{subfigure}[b]{0.455\textwidth}
        \centering
        \includegraphics[width=\textwidth]{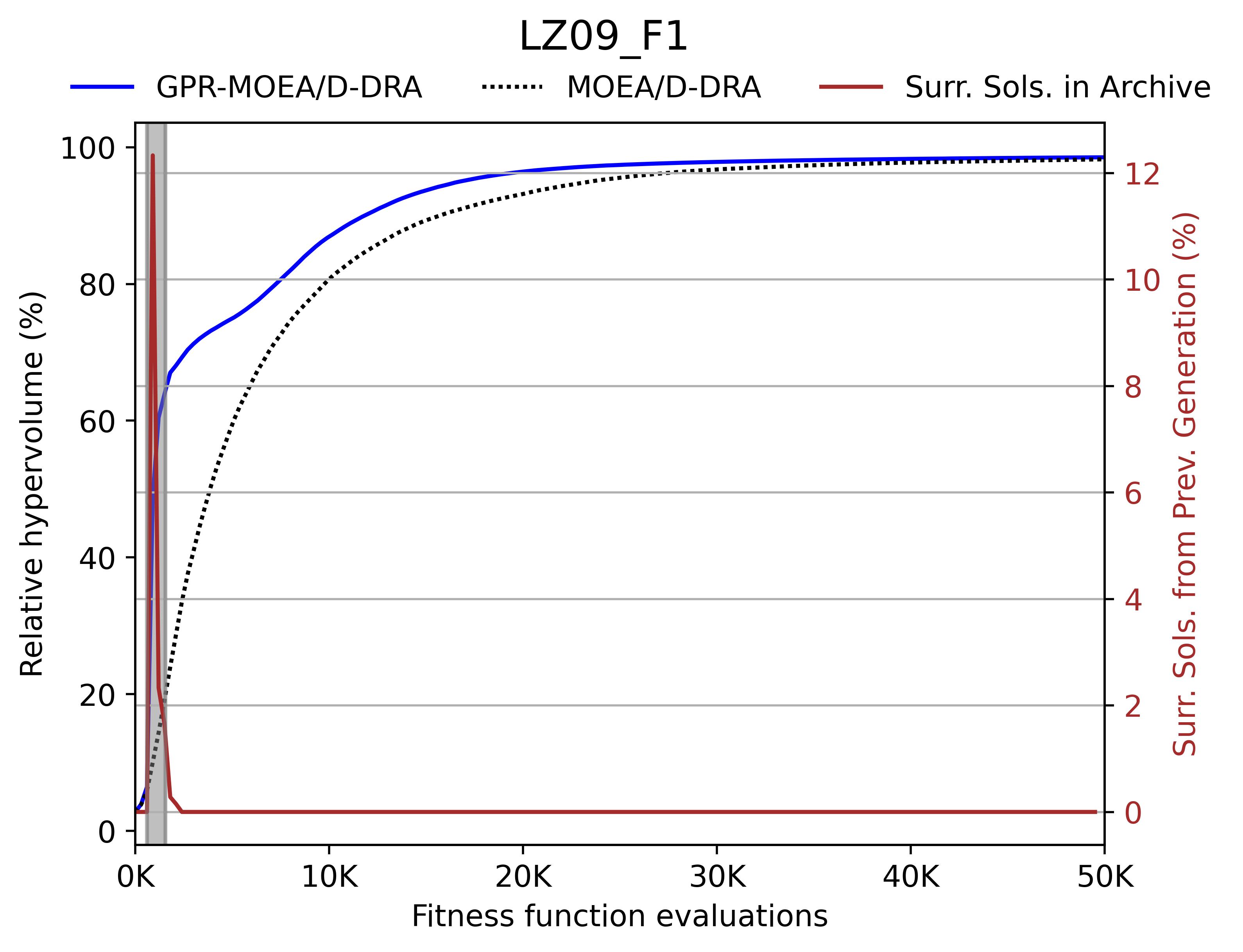}
        \caption{MOEA/D-DRA}
    \end{subfigure}
     \caption{Mean performance of GPR-based Adaptive Accelerator for NSGA-II on DTLZ7 and MOEA/D-DRA on LZ09\_F1 across the 100 independent runs. The shaded part is the mean interval the surrogate was active. The brown line indicates the mean value of $\chi_{t}$ -- the percentage of surrogate solutions from the penultimate generation that are in the archive at generation $t$ (read from the right y-axis).}
    \label{fig:surrogate_impact}
\end{figure*}

We use Figure  \ref{fig:surrogate_impact} to illustrate the impact of the proposed Adaptive Accelerator strategy for our two considered host solvers on the DTLZ7 and LZ09\_F1 problems. On average, in both cases, after activation at generation no. 2, the GPR surrogate is only active for: 6 generations in the case of NSGA-II on DTLZ7 and 3 generations in the case of MOEA/D-DRA on LZ09\_F1. Nevertheless, by the time the surrogate is deactivated based on the half-life criterion, the accelerated GPR-NSGA-II reaches a relative hypervolume of 75\%+ whilst the accelerated GPR-MOEA/D-DRA  reaches a $Hv(PF_c)$ of 60\%+. On average, their respective baselines,  (black dotted lines) are able to match this performance after 20k (for NSGA-II) and 5k (for MOEA/D-DRA) extra fitness evaluations with the initial surrogate advantage being preserved throughout the run. When the GPR surrogate is active, the maximum proportion of archive surrogate solutions from the penultimate generation (i.e., $\mu$ from Equation \ref{eq:AdaptiveExitCriterion}) reaches 32\% for NSGA-II and 12\% for MOEA/D-DRA.

In Table \ref{tab:surrogate_exit_points} we present statistics of the surrogate deactivation points (i.e., generations) for the three NSGA-II-incorporated surrogates on the 31 benchmark problems. We remind the reader that counting starts from generation 0 (as the initial population of the solver) and the surrogate-assisted module (i.e., Adaptive Accelerator) is activated at generation 2. Generally, the minimum deactivation point for the three surrogate-assisted internal solvers across all 31 benchmark problems is at generation 5 (i.e., the Adaptive Accelerator is only active for 3 generations). The mean and maximum deactivation generation vary significantly. The highest orderly maximum deactivation points were observed for RFR-NSGA-II on the LZ09 and DTLZ problems. There are also four problems where we observed a forced deactivation (at generation 50) of the CNN surrogate. Thus, out of 100 independent runs, the forced deactivation occurred in 6 runs on LZ09\_F1, 7 runs on LZ09\_F3, 10 runs on LZ09\_F4 and 5 runs on LZ09\_F5 and this also increases the relative mean deactivation generation of CNN-NSGA-II for these four benchmark MOOPs. In general, the very small likelihood of a forced deactivation and the low mean deactivation points support the effectiveness of our proposed "half-life" survivability criterion.   

\begin{table*}
\caption{Adaptive Accelerator deactivation point statistics for the surrogate-enhanced NSGA-II variants on the 31 benchmark problems.}
\label{tab:surrogate_exit_points}
\centering
\small
\begin{tabular}{l|ccc|ccc|ccc}
\toprule
\textbf{Problem} & \multicolumn{3}{c|}{\textbf{RFR-NSGA-II}} & \multicolumn{3}{c|}{\textbf{GPR-NSGA-II}} & \multicolumn{3}{c}{\textbf{CNN-NSGA-II}} \\
 & Min & Mean & Max & Min & Mean & Max & Min & Mean & Max \\
\midrule
DTLZ1 & 5 & 10.11 & 26 & 5 & 6.21 & 14 & 5 & 6.48 & 9 \\
DTLZ2 & 5 & 7.5 & 12 & 5 & 5.31 & 9 & 5 & 5.1 & 7 \\
DTLZ3 & 5 & 8.91 & 19 & 5 & 6.15 & 9 & 5 & 6.13 & 9 \\
DTLZ4 & 5 & 7.79 & 15 & 5 & 6.0 & 11 & 5 & 6.34 & 10 \\
DTLZ5 & 5 & 8.01 & 13 & 5 & 5.53 & 7 & 5 & 5.12 & 7 \\
DTLZ6 & 9 & 16.0 & 25 & 5 & 7.74 & 13 & 6 & 6.95 & 8 \\
DTLZ7 & 5 & 11.21 & 18 & 5 & 8.19 & 13 & 5 & 7.05 & 9 \\
KSW & 5 & 7.18 & 13 & 5 & 5.88 & 9 & 5 & 5.79 & 9 \\
LZ09\_F1 & 5 & 8.24 & 16 & 5 & 6.46 & 12 & 5 & 12.42 & 50 \\
LZ09\_F2 & 5 & 8.25 & 22 & 5 & 5.48 & 9 & 5 & 5.67 & 8 \\
LZ09\_F3 & 5 & 7.74 & 16 & 5 & 5.64 & 8 & 5 & 8.61 & 50 \\
LZ09\_F4 & 5 & 7.32 & 20 & 5 & 5.73 & 7 & 5 & 10.32 & 50 \\
LZ09\_F5 & 5 & 6.82 & 14 & 5 & 5.3 & 8 & 5 & 7.38 & 50 \\
LZ09\_F6 & 5 & 9.58 & 20 & 5 & 6.04 & 9 & 5 & 6.86 & 11 \\
LZ09\_F7 & 5 & 7.98 & 23 & 5 & 5.42 & 9 & 5 & 6.1 & 9 \\
LZ09\_F8 & 5 & 7.43 & 18 & 5 & 5.52 & 9 & 5 & 6.06 & 9 \\
LZ09\_F9 & 5 & 8.07 & 22 & 5 & 5.64 & 10 & 5 & 5.66 & 12 \\
WFG1 & 5 & 8.67 & 15 & 5 & 5.22 & 6 & 5 & 6.44 & 8 \\
WFG2 & 5 & 7.15 & 15 & 5 & 5.43 & 8 & 5 & 5.75 & 8 \\
WFG3 & 5 & 6.81 & 11 & 5 & 5.2 & 8 & 5 & 5.27 & 8 \\
WFG4 & 5 & 5.83 & 9 & 5 & 5.43 & 8 & 5 & 5.58 & 8 \\
WFG5 & 5 & 6.92 & 11 & 5 & 6.0 & 9 & 5 & 6.0 & 7 \\
WFG6 & 5 & 6.89 & 11 & 5 & 6.0 & 9 & 5 & 5.97 & 7 \\
WFG7 & 5 & 5.3 & 7 & 5 & 5.82 & 8 & 5 & 5.36 & 7 \\
WFG8 & 6 & 6.95 & 8 & 5 & 6.85 & 7 & 6 & 6.91 & 7 \\
WFG9 & 5 & 6.04 & 9 & 5 & 5.56 & 7 & 5 & 5.91 & 8 \\
ZDT1 & 5 & 9.37 & 21 & 5 & 5.13 & 10 & 5 & 6.04 & 8 \\
ZDT2 & 5 & 6.53 & 11 & 5 & 7.97 & 16 & 5 & 6.64 & 8 \\
ZDT3 & 5 & 9.31 & 17 & 5 & 6.26 & 10 & 5 & 5.78 & 7 \\
ZDT4 & 5 & 6.63 & 10 & 5 & 7.04 & 10 & 5 & 6.28 & 8 \\
ZDT6 & 5 & 6.53 & 19 & 5 & 7.22 & 18 & 5 & 6.43 & 11 \\
\bottomrule
\end{tabular}
\end{table*}

\begin{figure}
    \centering
    \includegraphics[width=0.455\textwidth]{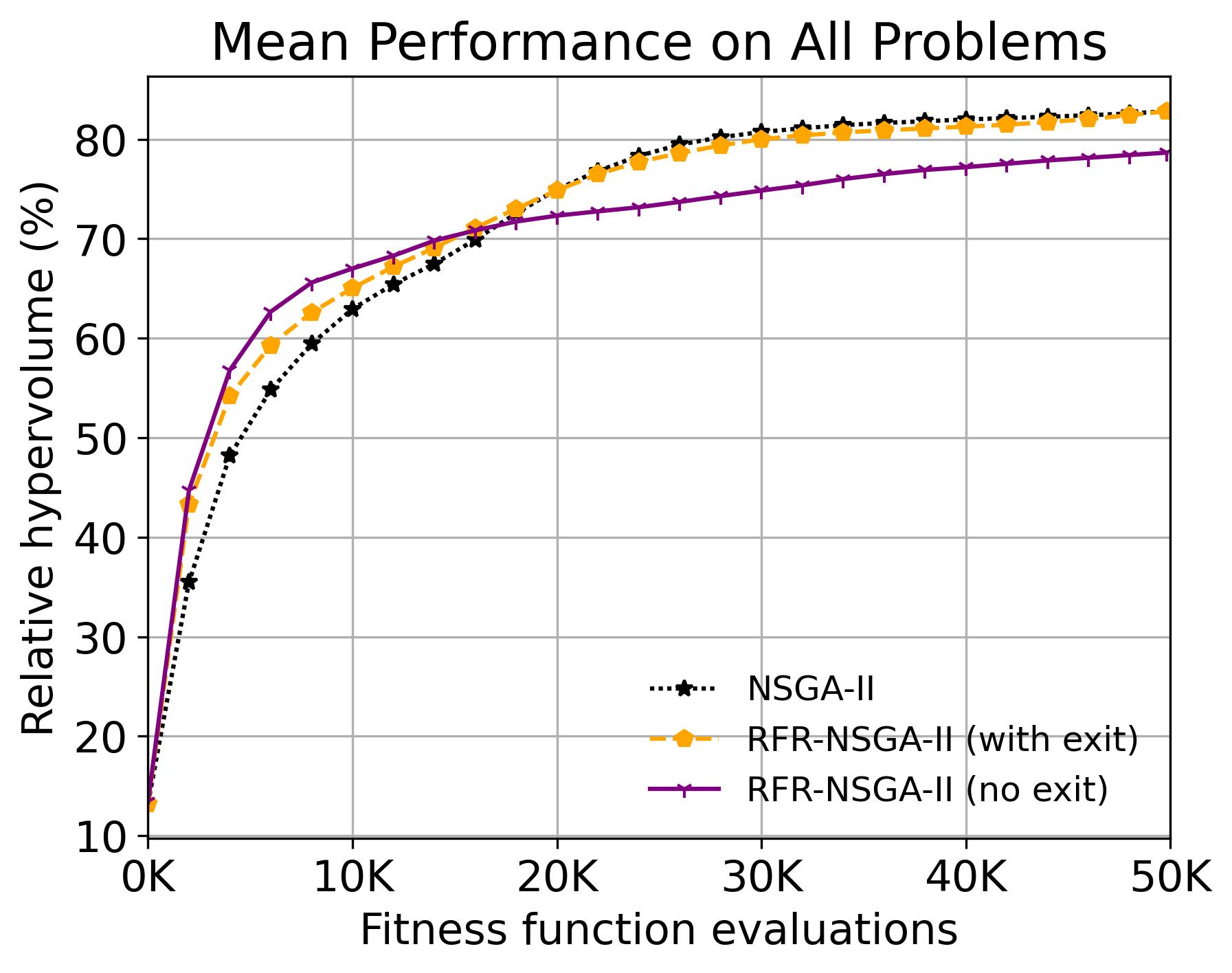}
    \caption{Comparison with full-on surrogate version (i.e., no exit) for the RFR-NSGA-II variant.}
    \label{fig:ablationResults}
\end{figure}

We present in Figure \ref{fig:ablationResults} the comparative average performance of two RFR-NSGA-II variants: a standard one with an adaptive surrogate deactivation mechanism (i.e., with exit) and one where the deactivation mechanism is turned off (i.e., no exit). Whilst always using the surrogate does provide some early convergence boost, this does come at the expense of late stage performance. Furthermore, the version without adaptive surrogate deactivation has a computational overhead that is more than 25 times higher than the adaptive variant. These ablation results empirically confirm the usefulness of our proposed adaptive surrogate deactivation strategy based on the half-life criterion.

\begin{figure*}
    \centering
    \begin{subfigure}[t]{0.455\textwidth}
        \centering
        \includegraphics[width=\textwidth]{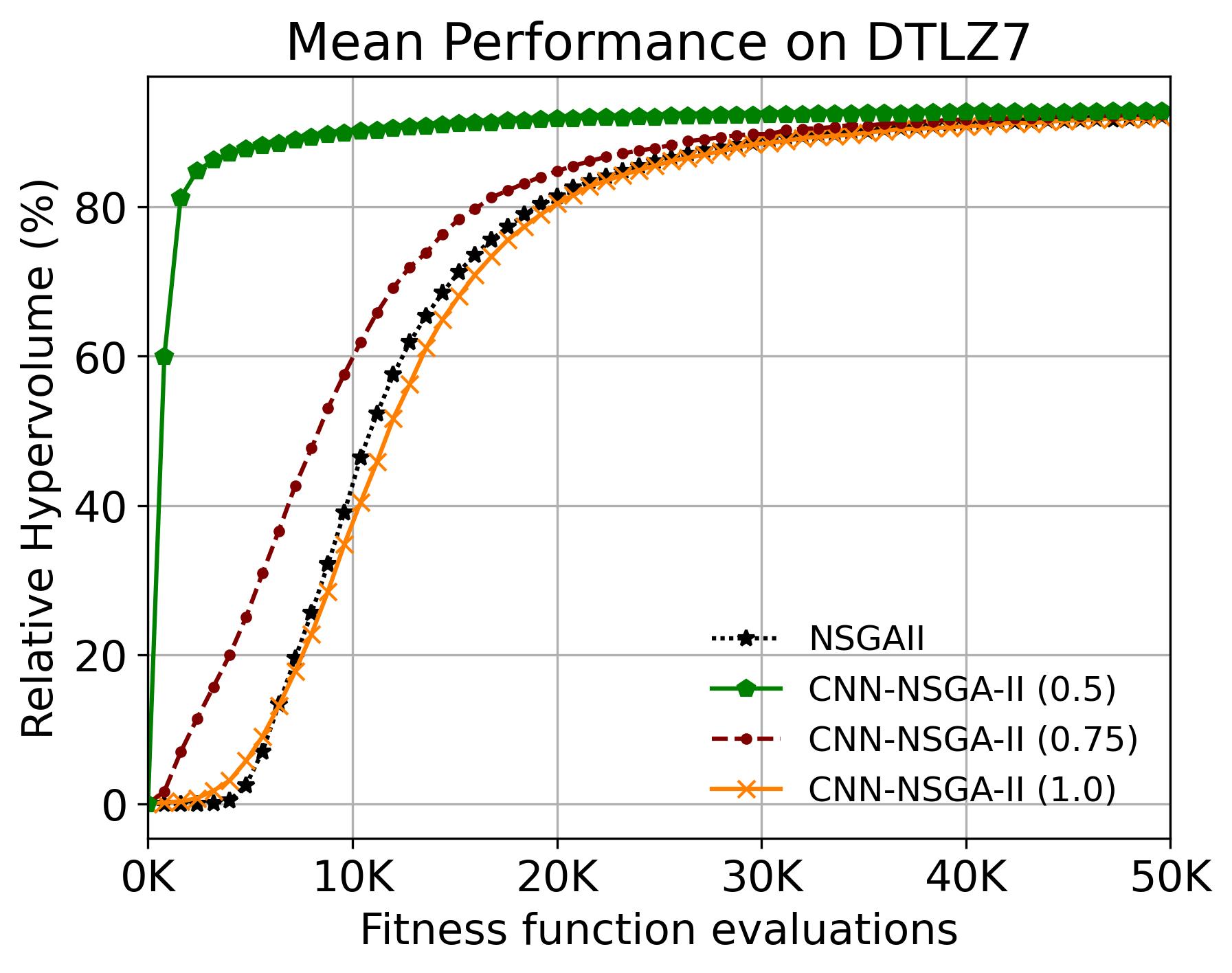}
        \caption{$Hv(PF_c)$-measured convergence}
    \end{subfigure}
    \begin{subfigure}[t]{0.455\textwidth}
        \centering
        \includegraphics[width=\textwidth]{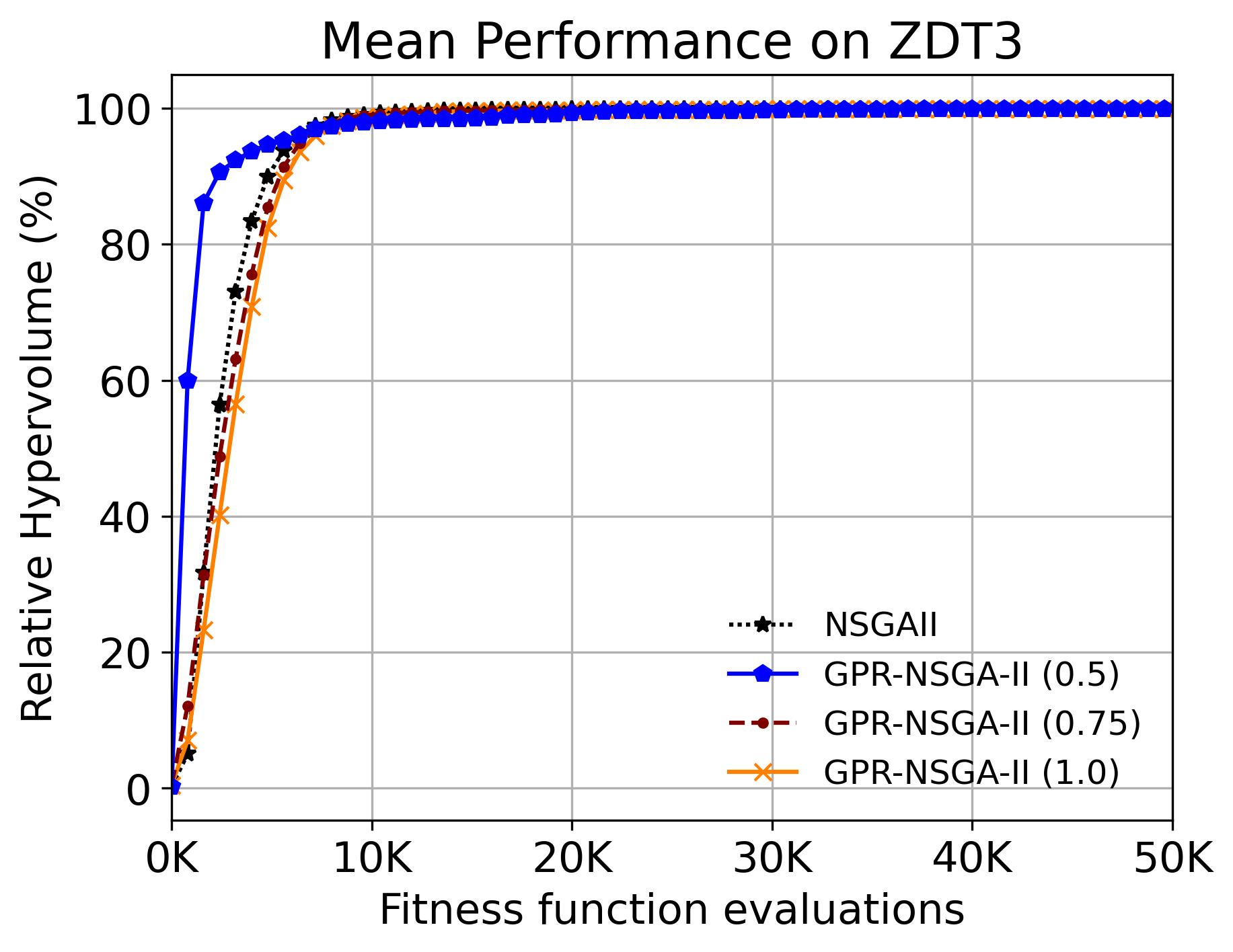}
        \caption{$Hv(PF_c)$-measured convergence}
    \end{subfigure}
    \begin{subfigure}[b]{0.455\textwidth}
        \centering
        \includegraphics[width=\textwidth]{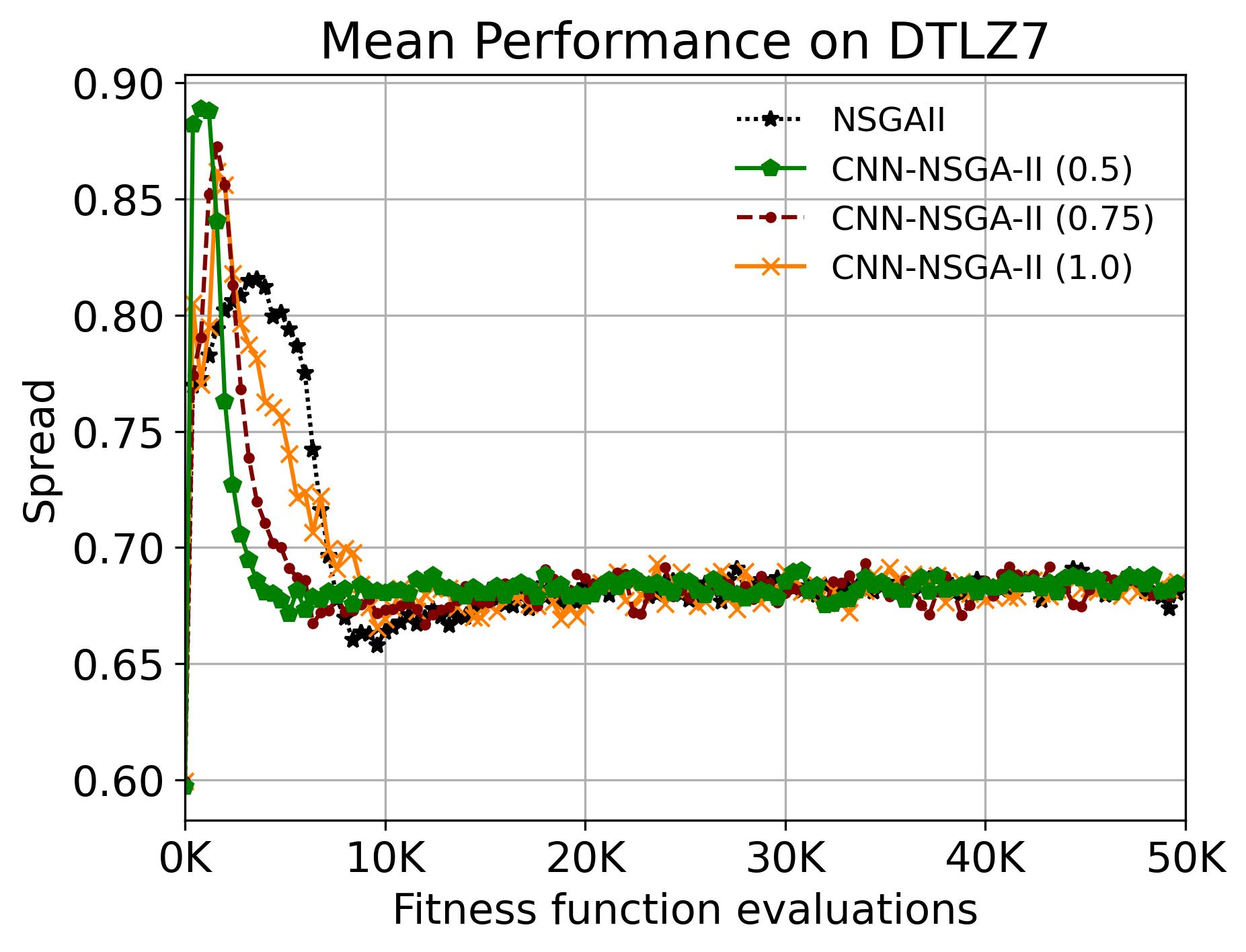}
        \caption{Associated $\Delta(PF_c)$}
    \end{subfigure}
    \begin{subfigure}[b]{0.455\textwidth}
        \centering
        \includegraphics[width=\textwidth]{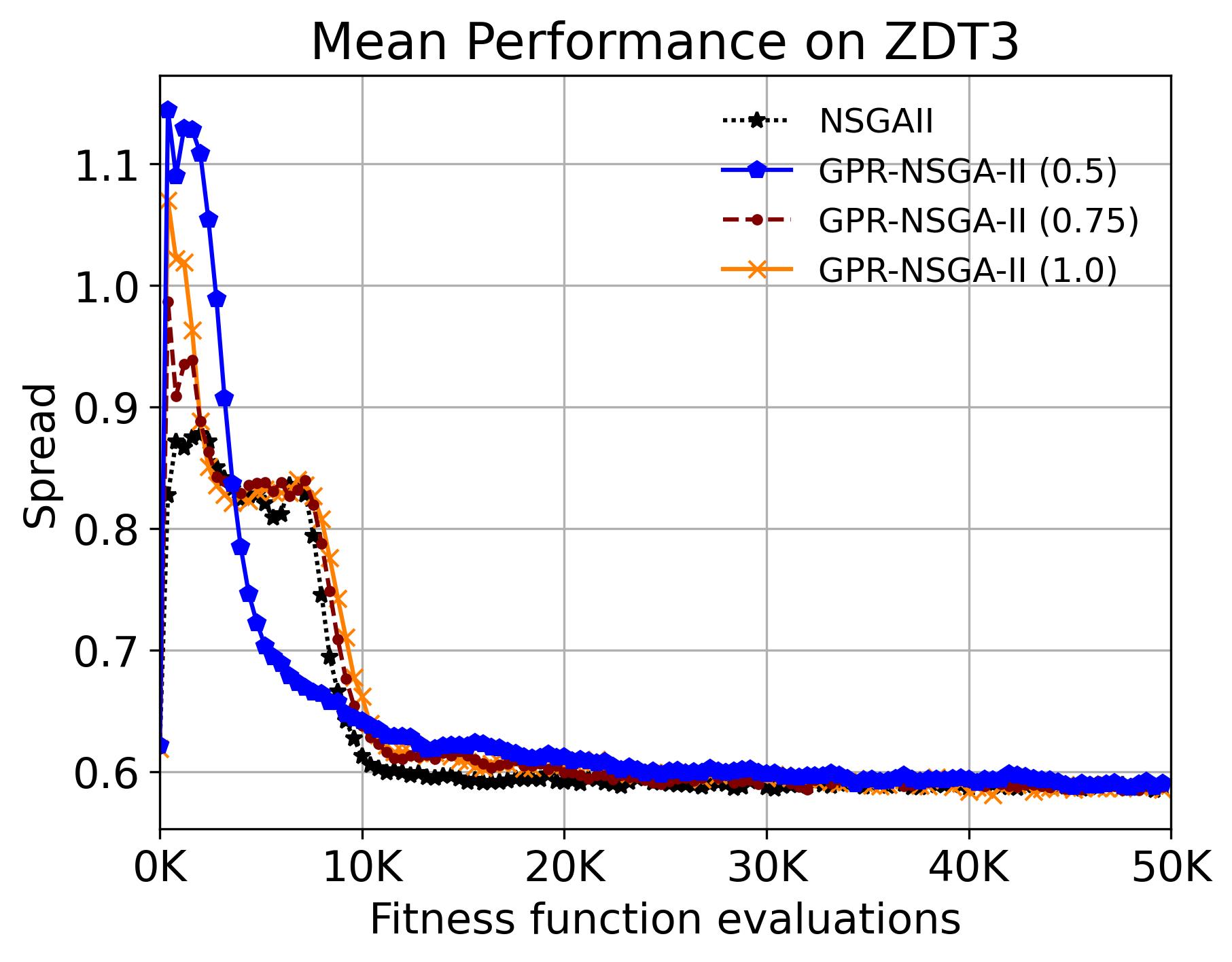}
        \caption{Associated $\Delta(PF_c)$}
    \end{subfigure}
     \caption{Mean comparative performance (convergence and spread) of CNN-NSGA-II and GPR-NSGA-II variants with surrogate integration thresholds of 50\%, 75\% and 100\%.}
    \label{fig:integration_thresholds}
\end{figure*}

In Figure \ref{fig:integration_thresholds} we present the comparative performance of different surrogate integration thresholds on DTLZ7 and ZDT3 -- two benchmark MOOPs that feature disconnected true Pareto Fronts which test the ability of solvers to maintain diversity across different optimal regions of the search space. For this test, the average $Hv(PF_c)$ convergence performance is complemented by the associated $\Delta(PF_c)$ spread plots. In the case of the baseline solver (i.e., NSGA-II), the $\Delta$ plots indicate that spread tends to increase or stabilise early on at a relatively high level ($>0.75$) and then flatten out. This is intuitive as $PF_c$ solutions tend to be far apart in objective space during the early and mid convergence stages (when progress towards the true Pareto Front can increase $PF_c$ gaps) and then coalesce during late convergence (when the solver focus shifts from discovering the true Pareto Front to better approximating it). In case of the NSGA-II surrogate variants, all surrogate integration thresholds (i.e., 50\%, 75\% and 100\%) are associated with noticeable early spikes of $\Delta(PF_c)$. This is a strong indicator that surrogate-derived solutions are able to leapfrog the search when compared with the baseline. However, the magnitudes of the $\Delta(PF_c)$ peaks is not proportional to that of the integration thresholds and only thresholds $<100\%$ are also associated with $Hv(PF_c)$-measured convergence boosts: i.e., 75\% and especially 50\% for the CNN-NSGA-II on DTLZ7 and 50\% for GPR-NSGA on ZDT3. This suggests that combining surrogate-based solutions with solutions generated by the host MOEA's native operators is crucial for ensuring diversity retention (i.e., mitigating potential linkage disequilibrium), whilst also motivating our fixed surrogate integration threshold of 50\%.     

\subsection{Mean Performance on MOOP Benchmark Set}
\label{sec:AllProblems}
\begin{figure*}
    \centering
    \begin{subfigure}[t]{0.455\textwidth}
        \centering
        \includegraphics[width=\linewidth]{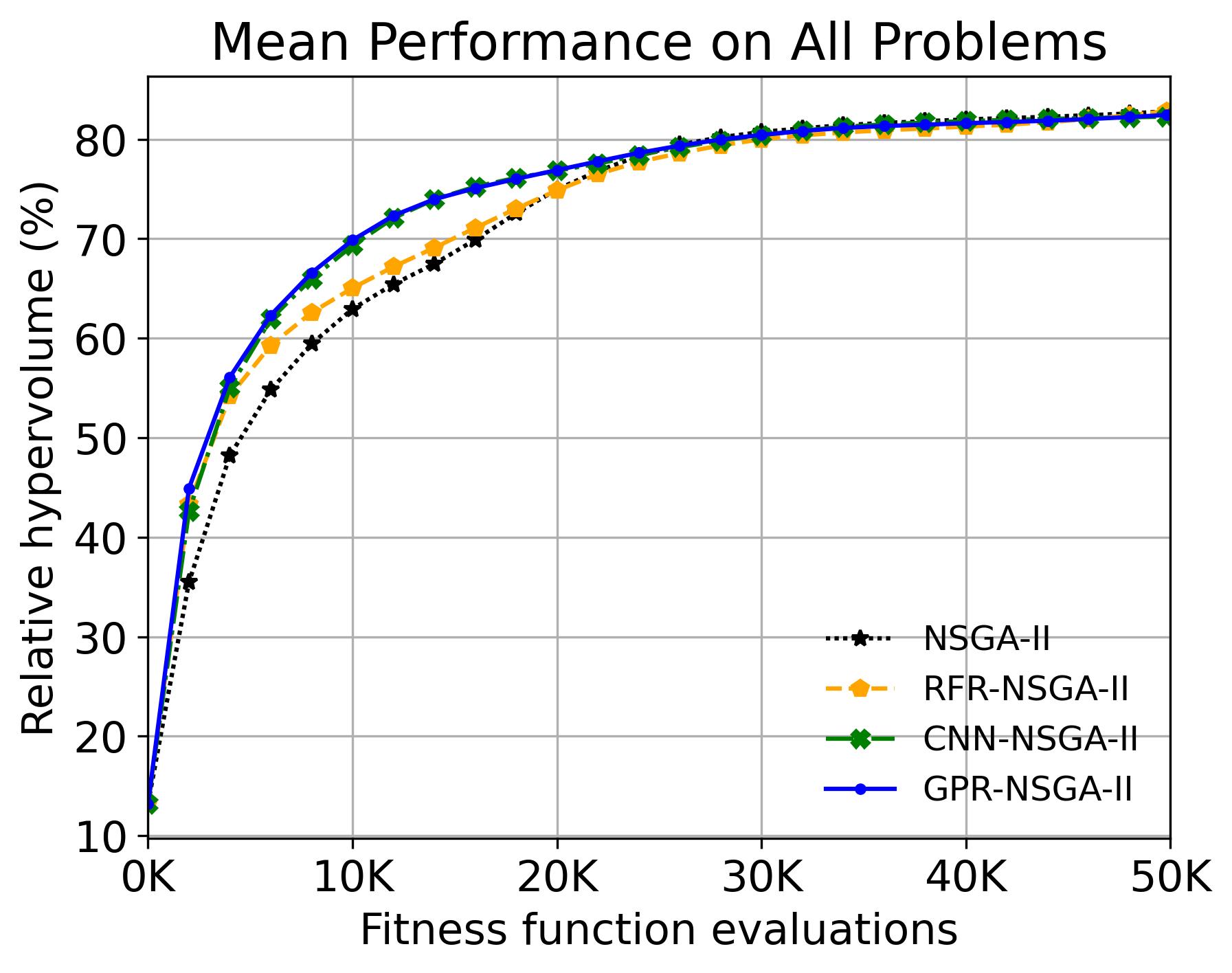}
        \caption{NSGA-II - $Hv(PF_c)$}
    \end{subfigure}
    \begin{subfigure}[t]{0.455\textwidth}
        \centering
        \includegraphics[width=\linewidth]{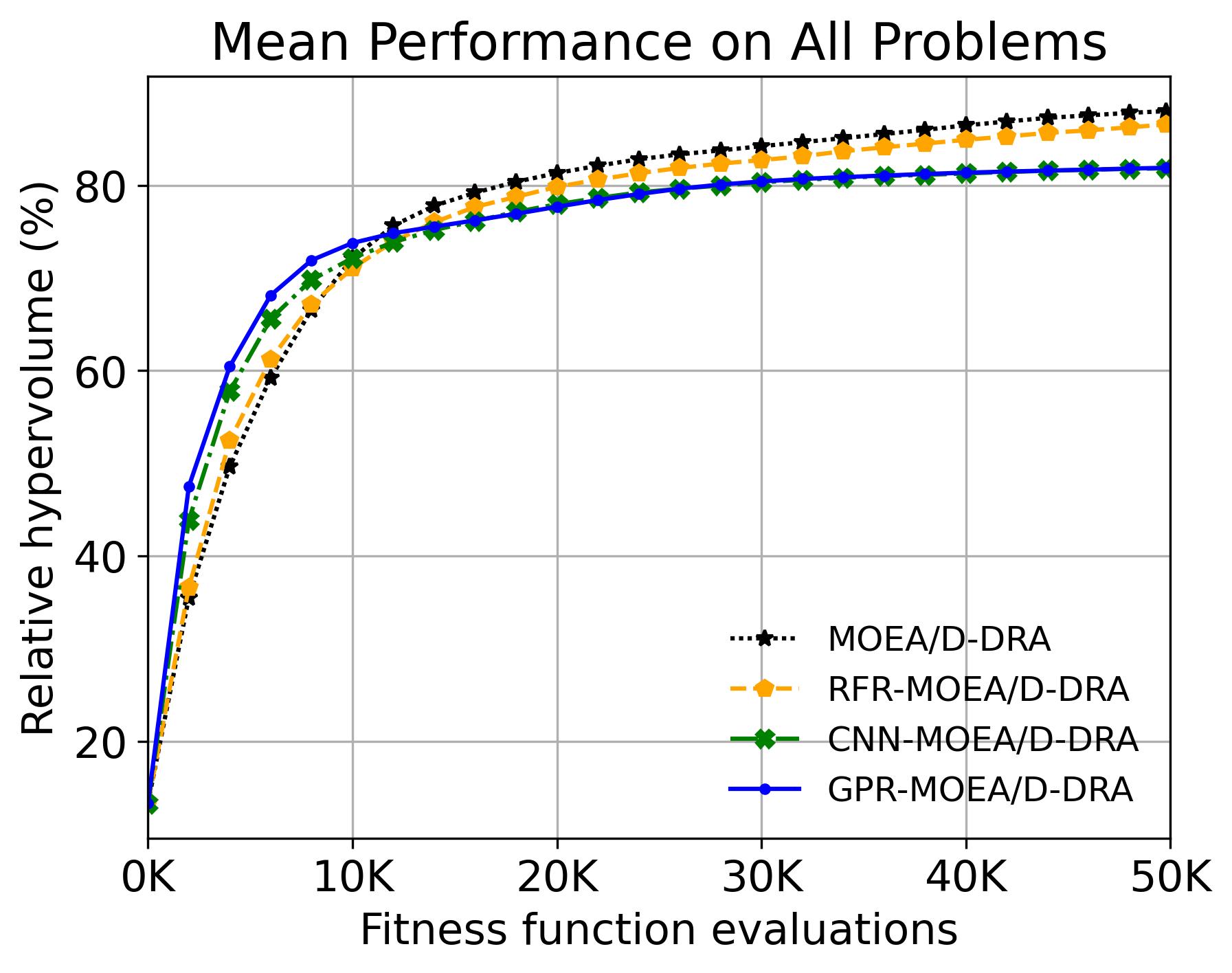}
        \caption{MOEA/D-DRA - $Hv(PF_c)$}
    \end{subfigure}
        \caption{Mean comparative performance of the surrogate-enhanced NSGA-II and MOEA/D-DRA and their standard versions on all 31 problems using the hypervolume ($Hv(PF_c)$) metric.}
        \label{fig:ALLperf}
\end{figure*}

\begin{figure*}
    \centering
    \begin{subfigure}[t]{0.95\textwidth}
        \centering
        \includegraphics[width=\linewidth]{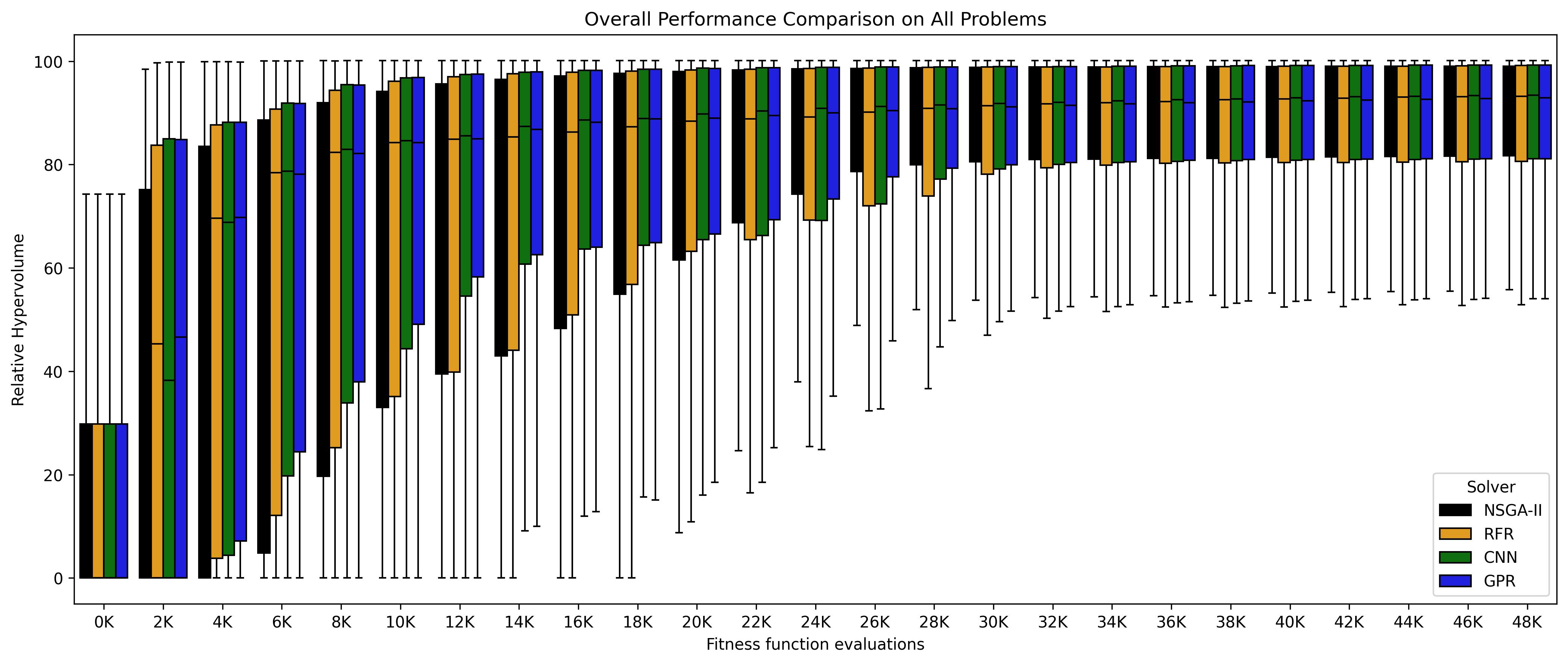}
        \caption{NSGA-II - $Hv(PF_c)$}
    \end{subfigure}
    \begin{subfigure}[t]{0.95\textwidth}
        \centering
        \includegraphics[width=\linewidth]{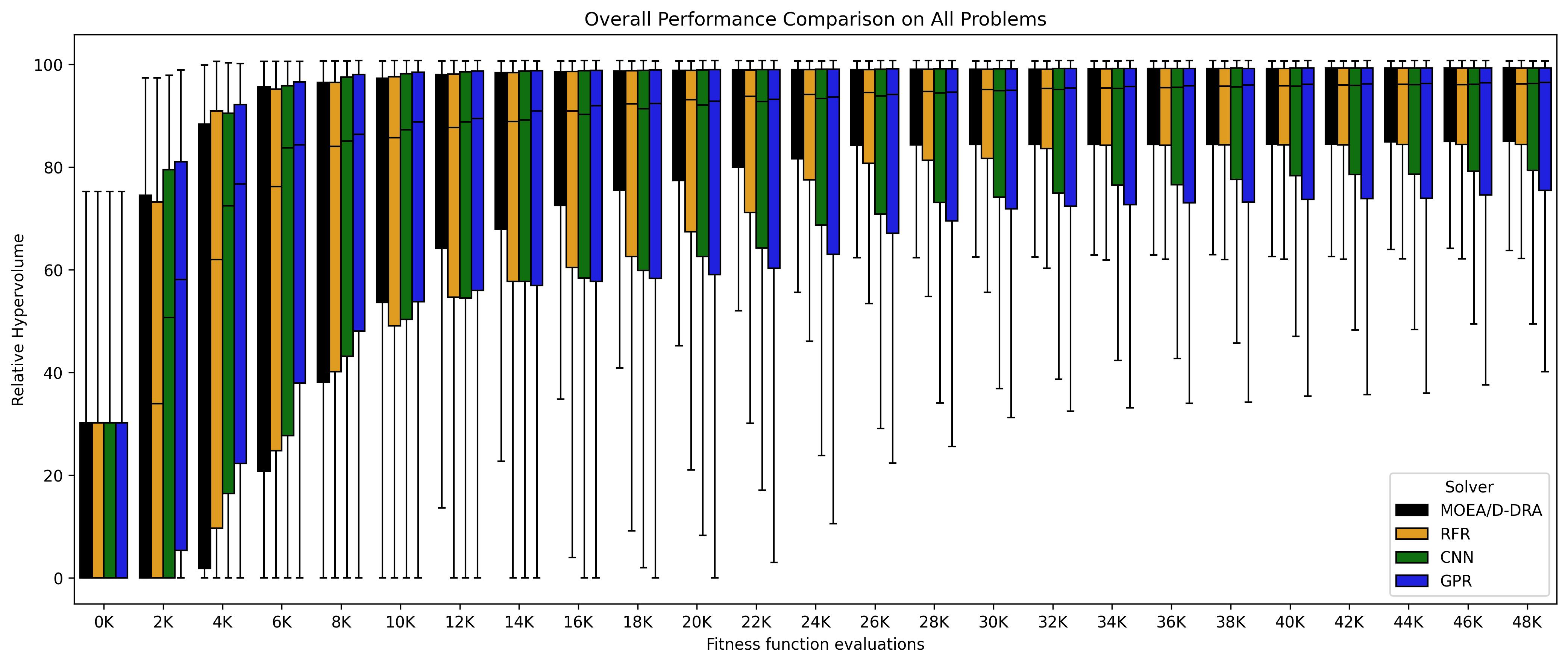}
        \caption{MOEA/D-DRA - $Hv(PF_c)$}
    \end{subfigure}
        \caption{$Hv(PF_c)$ comparative performance of the surrogate-enhanced NSGA-II and MOEA/D-DRA and their standard versions after every 2000 fitness evaluations. Each box plot is based on 3100 values (31 MOOPs times 100 runs per problem).}
        \label{fig:ALLperfBoxPlots}
\end{figure*}

We present the mean $Hv(PF_c)$ performance across all 31 benchmark problems that we have considered in Figures \ref{fig:ALLperf} and \ref{fig:ALLperfBoxPlots}. The results indicate that our Adaptive Accelerators successfully speed up the general convergence of NSGA-II, with GPR-NSGA-II and CNN-NSGA-II notably doing so, followed by RFR-NSGA-II. The same is largely true for MOEA/D-DRA early convergence as both GPR-MOEA/D-DRA and CNN-MOEA/D-DRA  outperform the baseline solver whilst the early boost provided by RFR-MOEA/D-DRA is more subdued. However, the overall MOEA/D-DRA Adaptive Accelerator performance is more nuanced as the early performance improvement of the GPR and RFR variants come at the expense of late-stage performance attainment whilst the RFR variant tracks the end-of-run baseline solver performance much better. All results and the associated code base used to produce them are available online at <web respository withheld>. 

\begin{figure*}[t]
    \centering
    \begin{subfigure}[b]{0.455\textwidth}
        \centering
        \includegraphics[width=\textwidth]{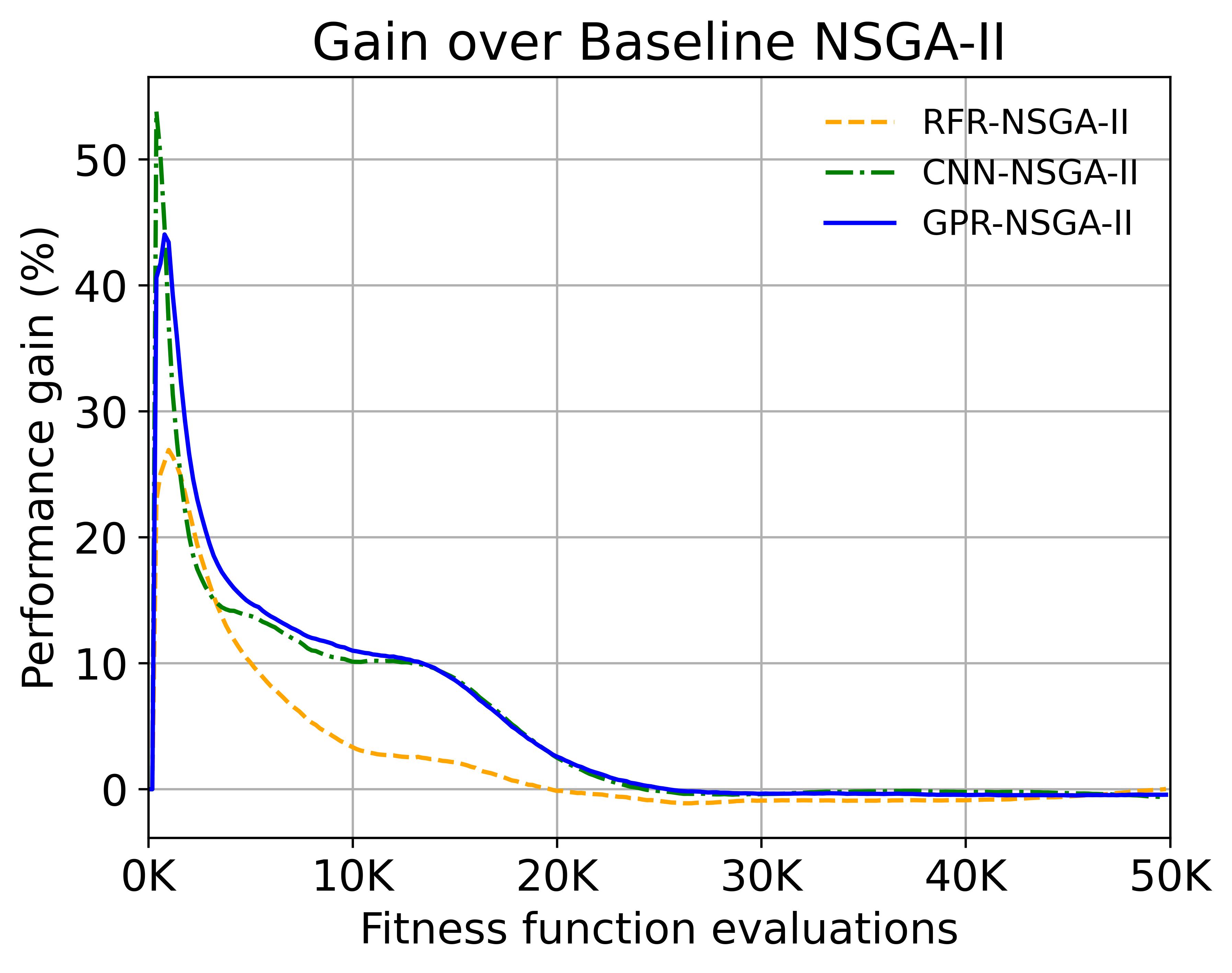}
        \caption{NSGA-II}
        \label{fig:gain_NSGAII}
    \end{subfigure}
    \begin{subfigure}[b]{0.455\textwidth}
        \centering
        \includegraphics[width=\textwidth]{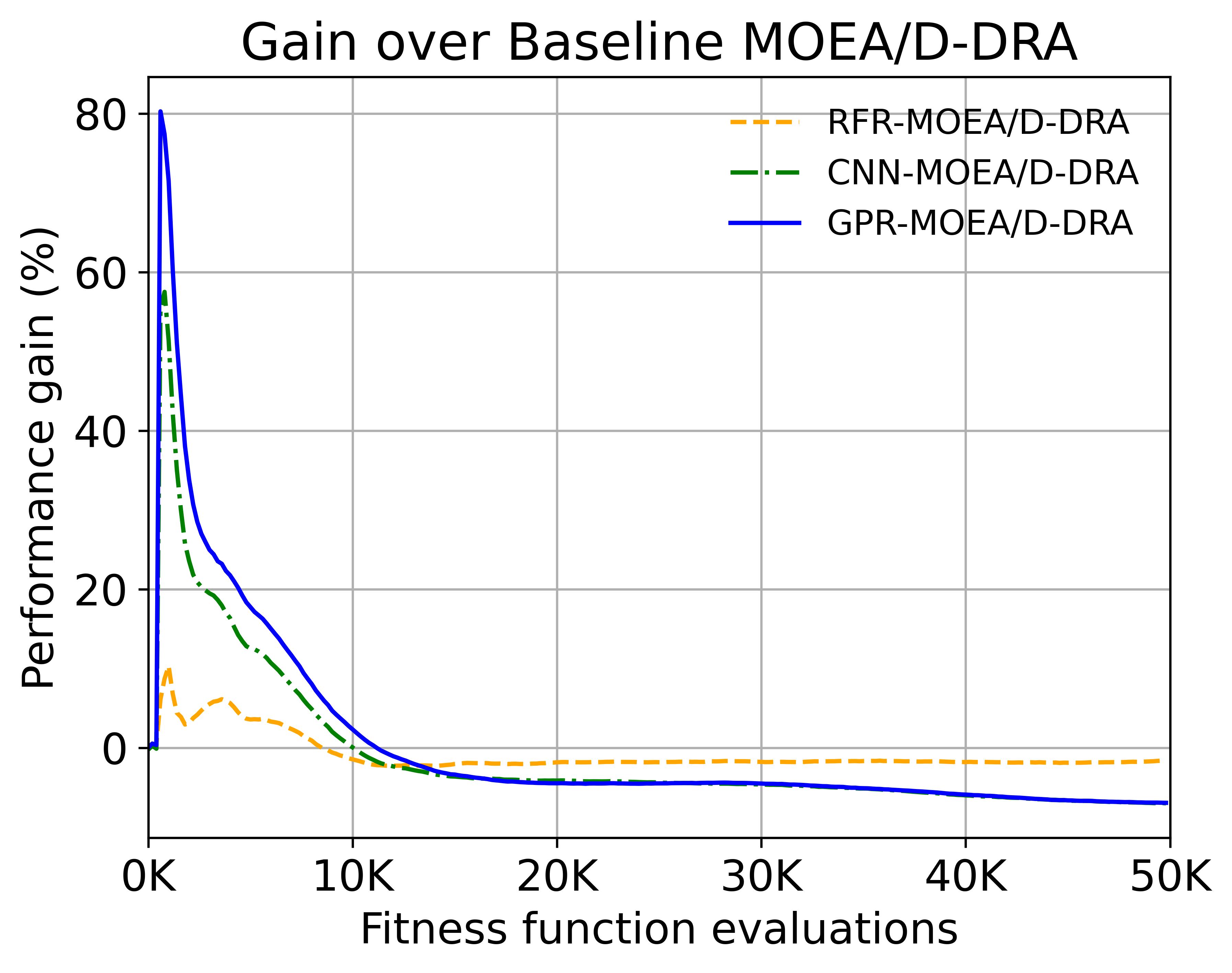}
        \caption{MOEA/D-DRA}
        \label{fig:gain_MOEAD}
    \end{subfigure}
    \caption{Performance gains of surrogate-enhanced MOEAs compared to their respective baseline solvers.}
    \label{fig:gain_surrogates}
\end{figure*}

In order to better illustrate the convergence behaviour of the Accelerated MOEAs, we computed their respective $Hv(PF_c)$ performance gain (\%) by subtracting the mean relative hypervolume of the baseline MOEA from that obtained by each accelerated variant at each generation. The results are presented in Figure \ref{fig:gain_surrogates} and provide further support that our surrogate-based strategy is successful at speeding up convergence, especially during the early generations.

For instance, in the case of NSGA-II (Figure \ref{fig:gain_NSGAII}), all three surrogate models accelerate the benchmark-wide average convergence by at least 20\% in the first 10 generations and by at least 10\% by the 20th generation. The CNN and GPR surrogates are able to induce a mean performance gain of about 10\% until the 70th generation. 

The performance gain for MOEA/D-DRA surrogates (Figure \ref{fig:gain_MOEAD}) are in line with previous observations, with GPR-MOEA/D-DRA attaining noticeable mean gains of about 80\% over its baseline very early in the runs. However, despite the initial positive impact, both the the GPR and CNN variants of MOEA/D-DRA start to underperform the baseline after the 50th generation, ending the runs at a 10\% mean deficit. The end-of-the-run performance drop of the RFR variant is less pronounced, but its early generations boost is also lackluster (max 10\%).  

\begin{figure}
    \centering
    \includegraphics[width=0.455\textwidth]{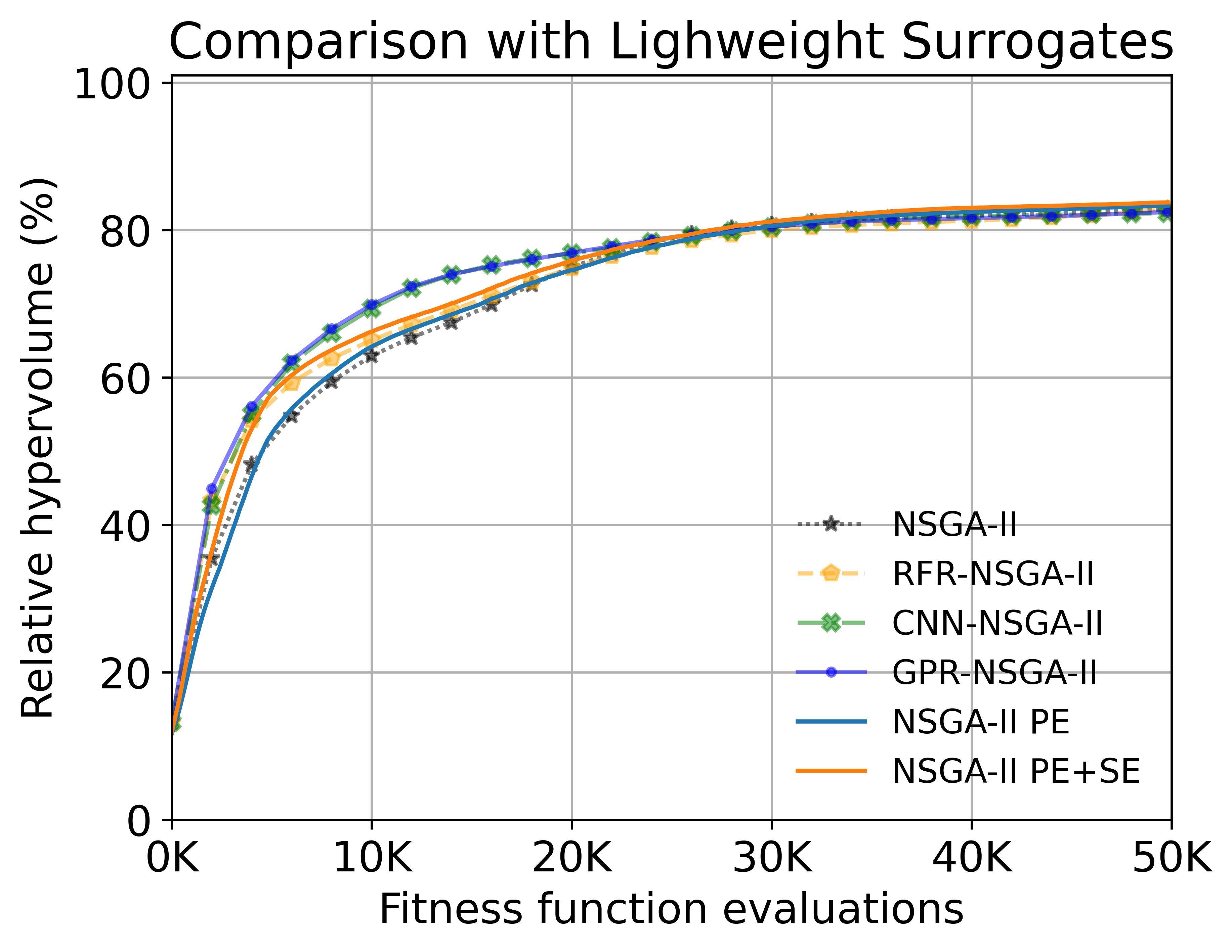}
    \caption{Comparison with lightweight surrogate models proposed in \cite{Zavoianu2022}.}
    \label{fig:compare_lightweight}
\end{figure}

In the case of the three surrogate-accelerated variants of NSGA-II, we also performed a $Hv(PF_c)$ comparison with the lightweight interpolation-based surrogate modelling strategy proposed in \cite{Zavoianu2022}. The results we obtained are shown in Figure \ref{fig:compare_lightweight} and indicate that our Adaptive Acceleration strategy using GPR and CNN surrogate models consistently achieves faster convergence than NSGA-II PE+SE -- the best performing interpolation-based approach. The RFR surrogate also initially outperforms both NSGA-II PE and NSGA-II PE+SE but is overtaken by the latter around generation 20. The mean end-of-the-run $Hv(PF_c)$ attainment is roughly similar  across all surrogate-enhanced MOEAs after equalising around 25k fitness evaluations (i.e., as the solvers enter the late convergence stage). These results indicate that, when compared with a simpler strategy, the increased complexity of our on-the-fly surrogate models (coupled with an effective adaptive deactivation strategy) is justified by their superior capability to generally accelerate NSGA-II convergence. 

\subsection{Statistical Analysis of MOOP Benchmark Set Performance}
\label{sec:StatisticalAnalysis}

We applied the Mann-Whitney U Test \cite{Mann1947} to determine if the mean relative hypervolumes attained by the surrogate variants on each of the 31 benchmark problems at every generation are statistically different from those obtained by their respective baseline solvers. The Mann-Whitney U test is a non-parametric statistical test used to compare two independent groups. Using a significance level of 0.05, we carried out a one-sided test with a null hypothesis that there is no statistical difference between the observed mean $Hv(PF_c)$ of the baseline solvers and their accelerated variants. The alternative was that the mean $Hv(PF_c)$ values achieved by the baseline solver was lower than that of the tested accelerated variant.

\begin{figure*}
    \centering
    \begin{subfigure}[b]{0.455\textwidth}
        \centering
        \includegraphics[width=\textwidth]{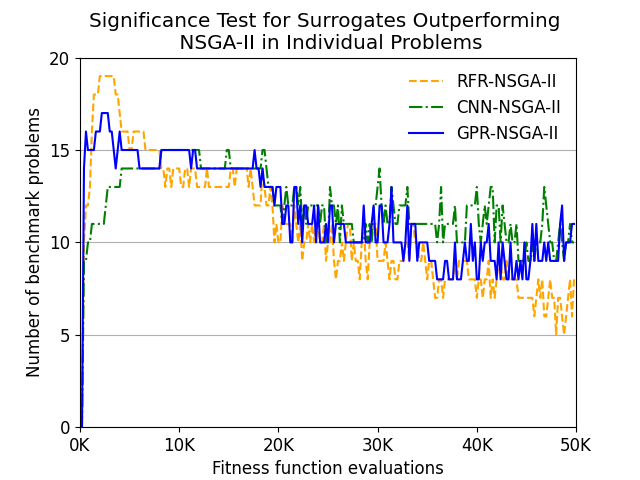}
        \caption{NSGA-II}
        \label{fig:nsgaii_sig_counter}
    \end{subfigure}
    \begin{subfigure}[b]{0.455\textwidth}
        \centering
        \includegraphics[width=\textwidth]{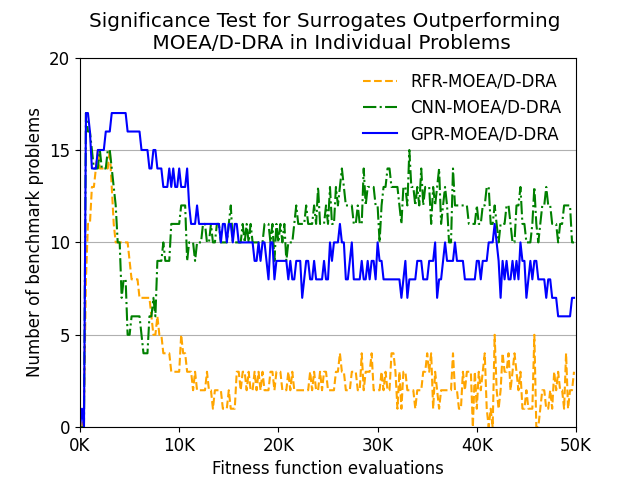}
        \caption{MOEA/D-DRA}
        \label{fig:moead_sig_counter}
    \end{subfigure}
    \caption{Comparative statistical test results showing the number of benchmark problems in which surrogate-assisted variants statistically outperformed their baseline solver at each generation.}
    \label{fig:sig_counters}
\end{figure*}

The statistical significance results we obtained are aggregated in Figure \ref{fig:nsgaii_sig_counter} for NSGA-II and Figure \ref{fig:moead_sig_counter} for MOEA/D-DRA. In these plots we show the number of benchmark problems where we rejected the null hypothesis in favour of the alternative hypothesis (i.e., that the surrogate variant outperform the baseline solver). The results follow an intuitive pattern for both the NSGA-II and the MOEA/D-DRA surrogate-accelerated variants, confirming that surrogate-driven outperformance of the baseline solvers occurs across many problems during the early stages of the optimisation. For instance, Figure \ref{fig:nsgaii_sig_counter} indicates that, in generation 11 (i.e., after 2400 fitness function evaluations), RFR-NSGA-II outperformed NSGA-II in 19 problems, CNN-NSGA-II in 17 problems and GPR-NSGA-II in 11.  
As expected, as the generations progress and the solvers reach their late convergence stage, the number of problems where the accelerated solvers outperform the baseline tends to decline from approx 1/2 of the total number of problems to approx 1/4 of the benchmark set (especially for RFR variants). In this respect, CNN variants (especially CNN-MOEA/D-DRA) demonstrate a noteworthy stability between 1/2 and 1/3 benchmark set outperformance.

\subsection{Performance on Individual Problems and Problem Suites}
\label{sec:ProblemSuites}
\begin{figure*}
    \centering
    \begin{subfigure}[t]{0.455\textwidth}
        \centering
        \includegraphics[width=\linewidth]{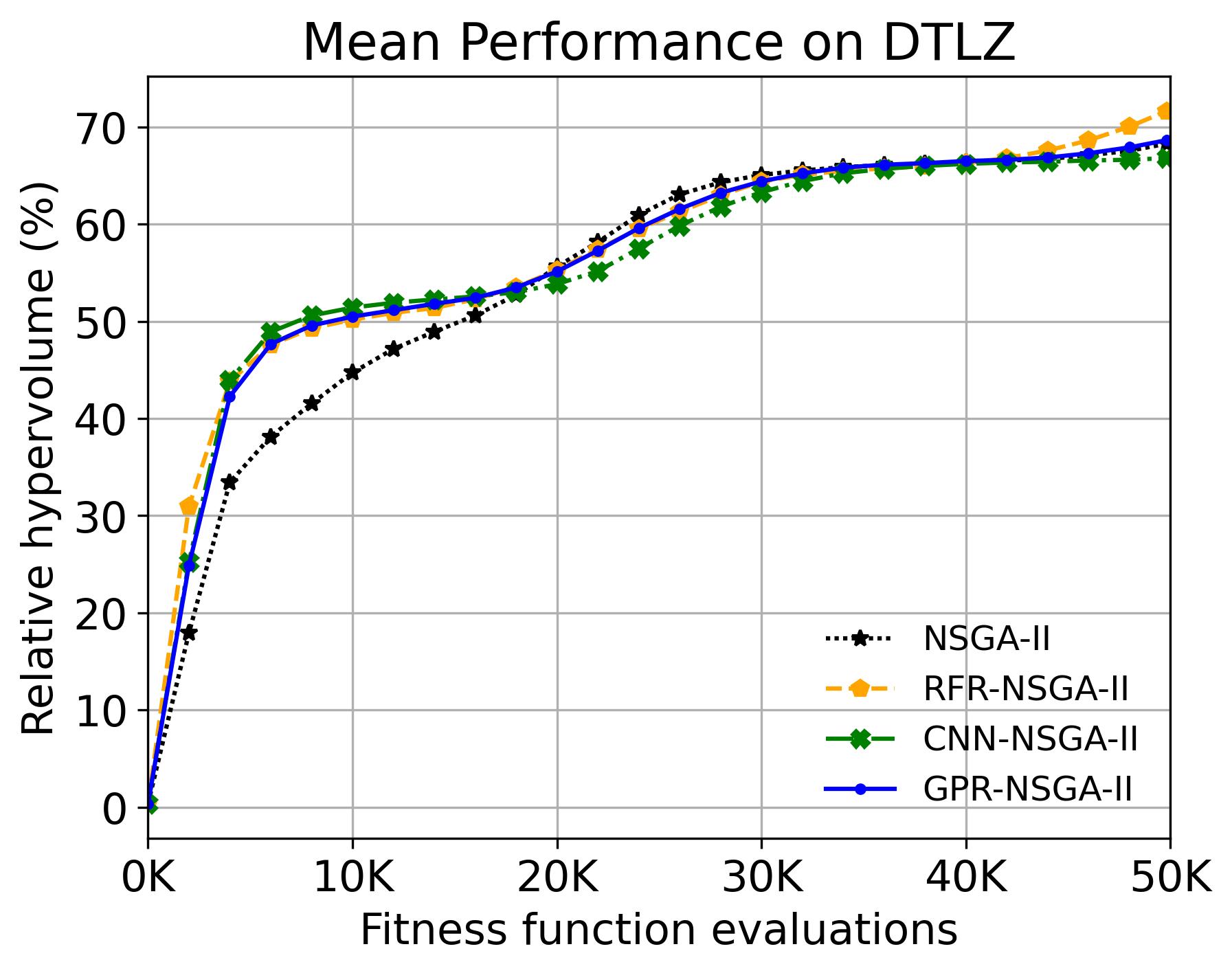}
        \caption{NSGA-II}
    \end{subfigure}
    \begin{subfigure}[t]{0.455\textwidth}
        \centering
        \includegraphics[width=\linewidth]{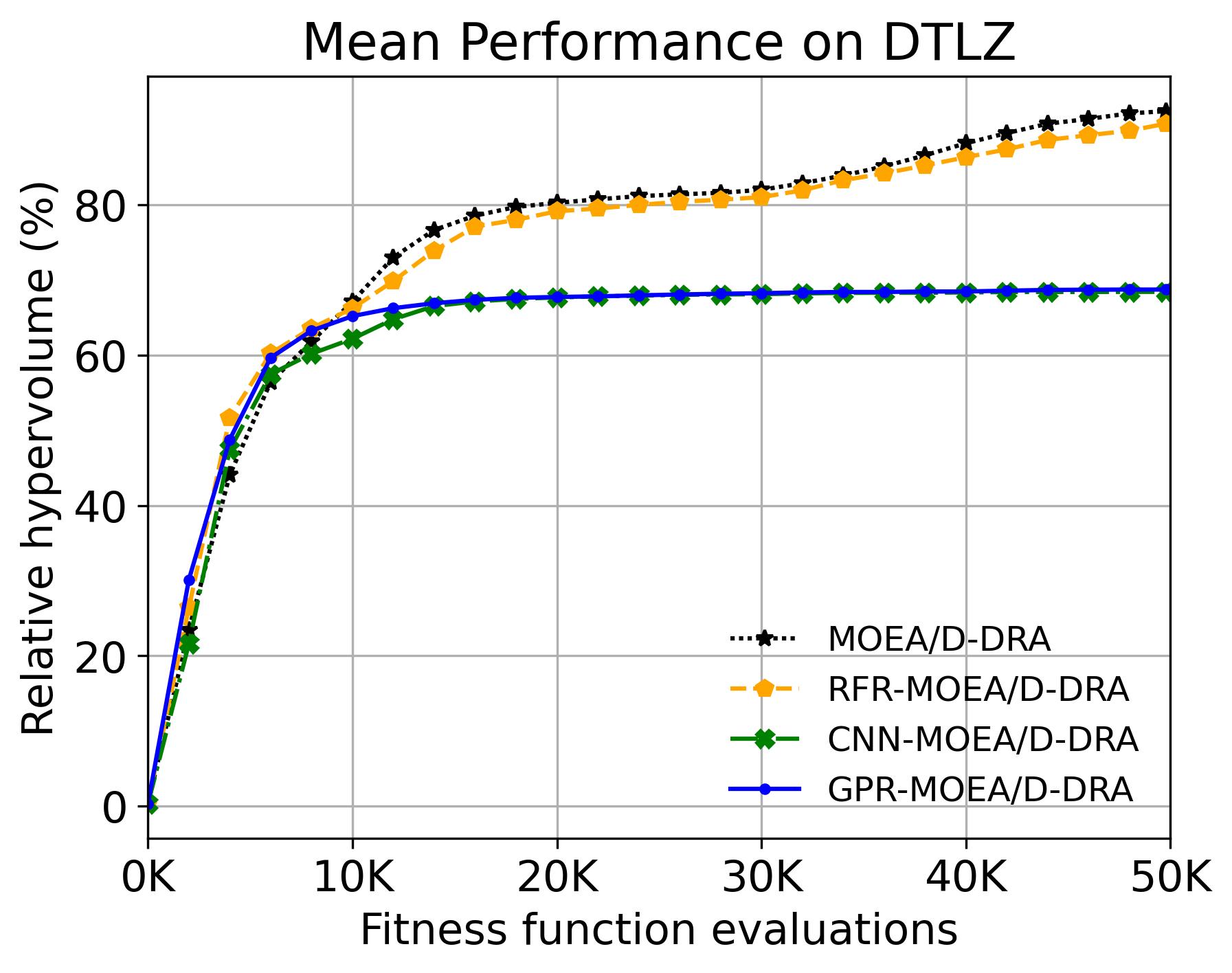}
        \caption{MOEA/D-DRA}
        \label{fig:dtlz_moead}
    \end{subfigure}

        \caption{Comparative mean performance of the surrogate-enhanced NSGA-II and MOEA/D-DRA variants and their baseline versions on the DTLZ problem suite (7 problems).}
        \label{fig:DTLZperf}
\end{figure*}

\begin{figure*}
    \centering
    \begin{subfigure}[t]{0.455\textwidth}
        \centering
        \includegraphics[width=\linewidth]{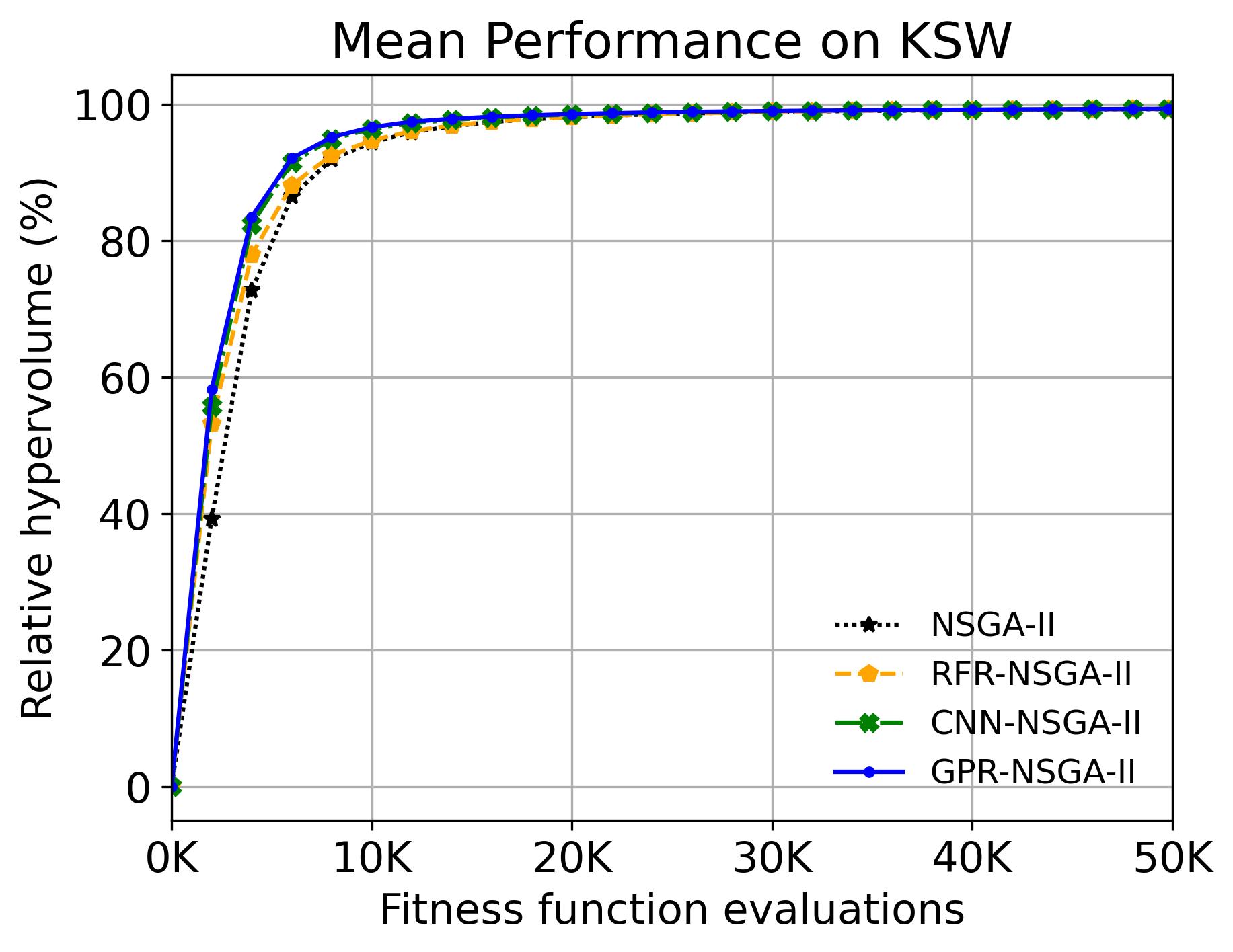}
        \caption{NSGA-II}
        \label{fig:ksw_nsgaii}
    \end{subfigure}
    \begin{subfigure}[t]{0.455\textwidth}
        \centering
        \includegraphics[width=\linewidth]{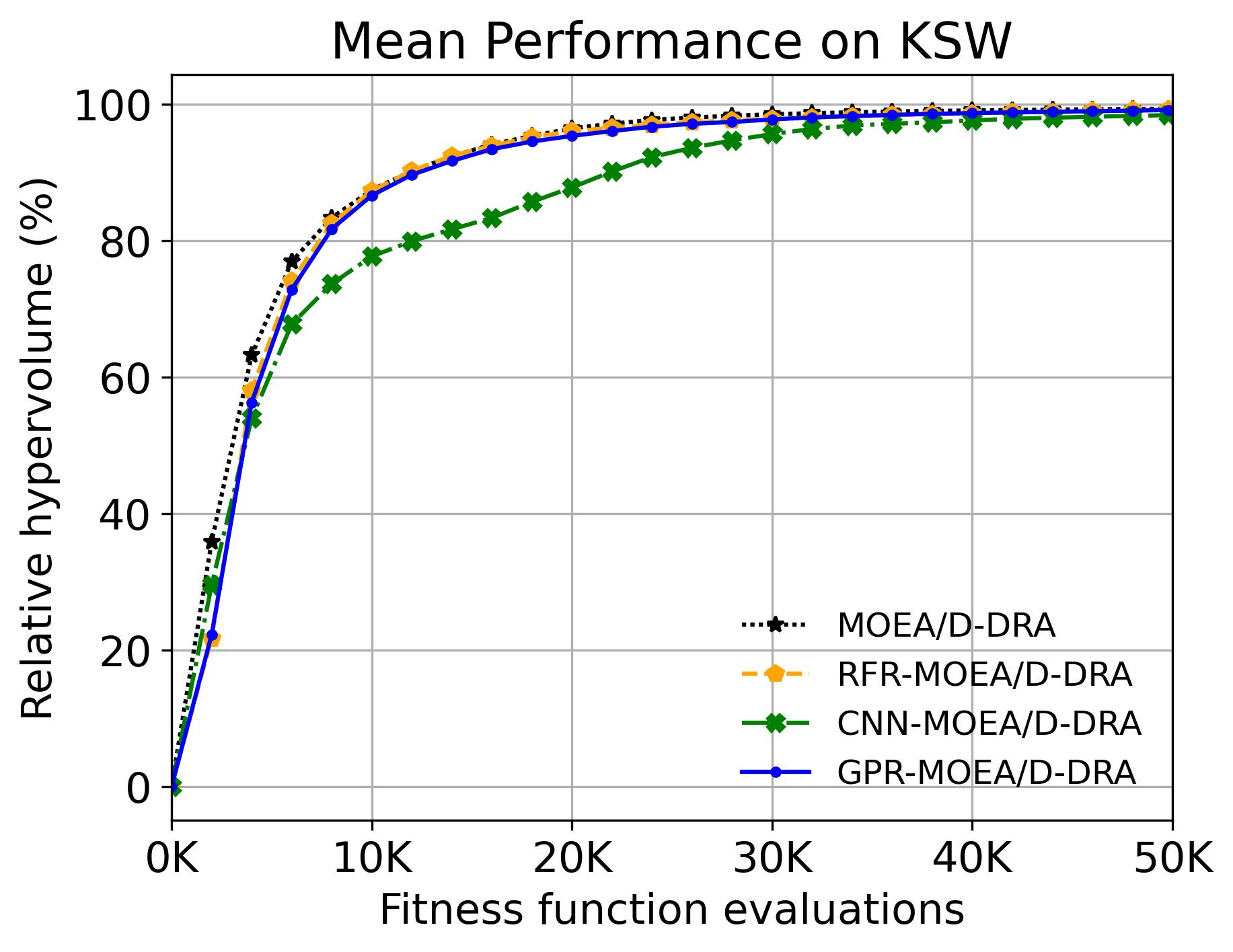}
        \caption{MOEA/D-DRA}
        \label{fig:ksw_moead}
    \end{subfigure}

        \caption{Comparative mean performance of the surrogate-enhanced NSGA-II and MOEA/D-DRA variants and their baseline versions on the KSW problem (1 problem).}
        \label{fig:KSWperf}
\end{figure*}

\begin{figure*}
    \centering
    \begin{subfigure}[t]{0.455\textwidth}
        \centering
        \includegraphics[width=\linewidth]{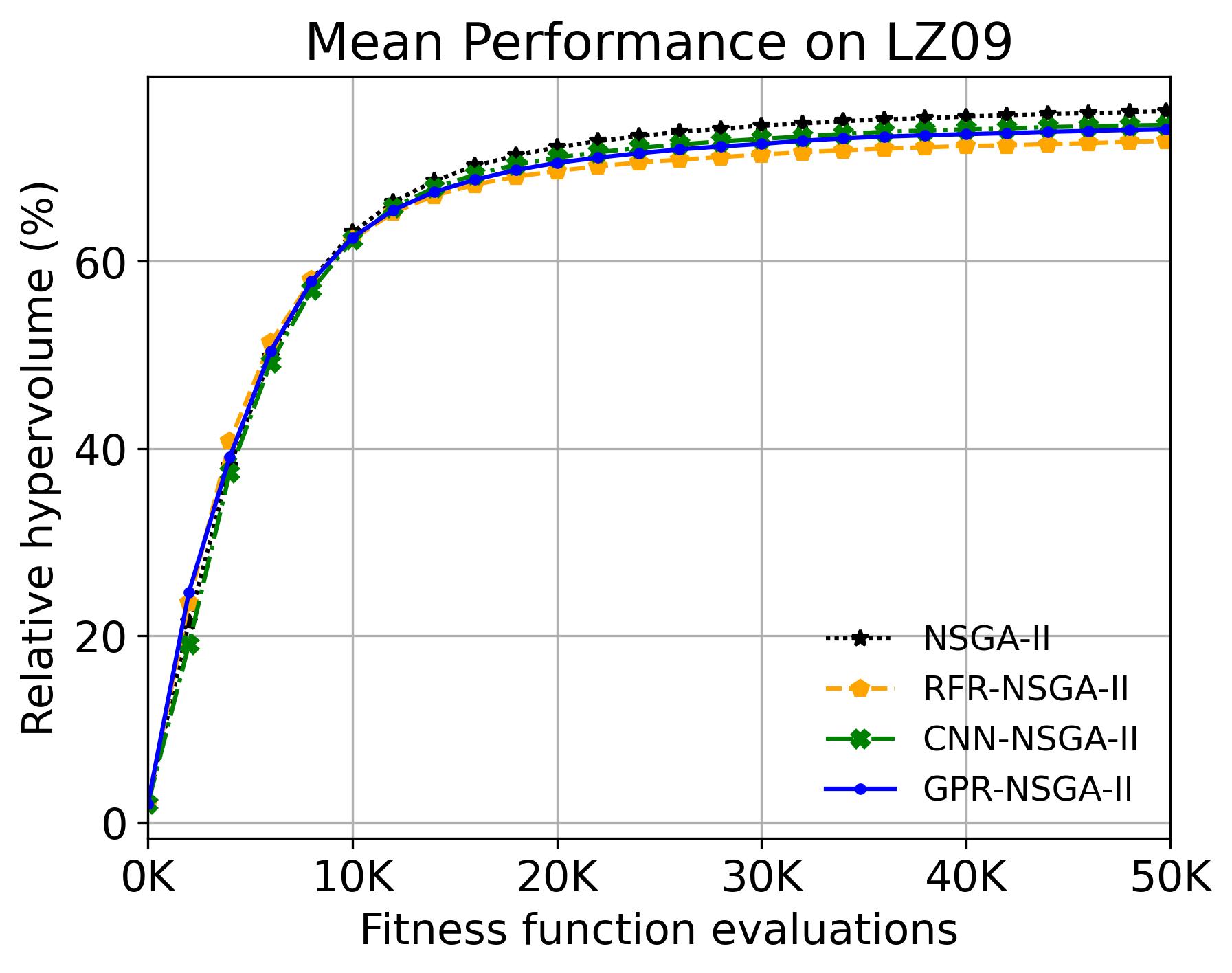}
        \caption{NSGA-II}
        \label{fig:lz09_nsgaii}
    \end{subfigure}
    \begin{subfigure}[t]{0.455\textwidth}
        \centering
        \includegraphics[width=\linewidth]{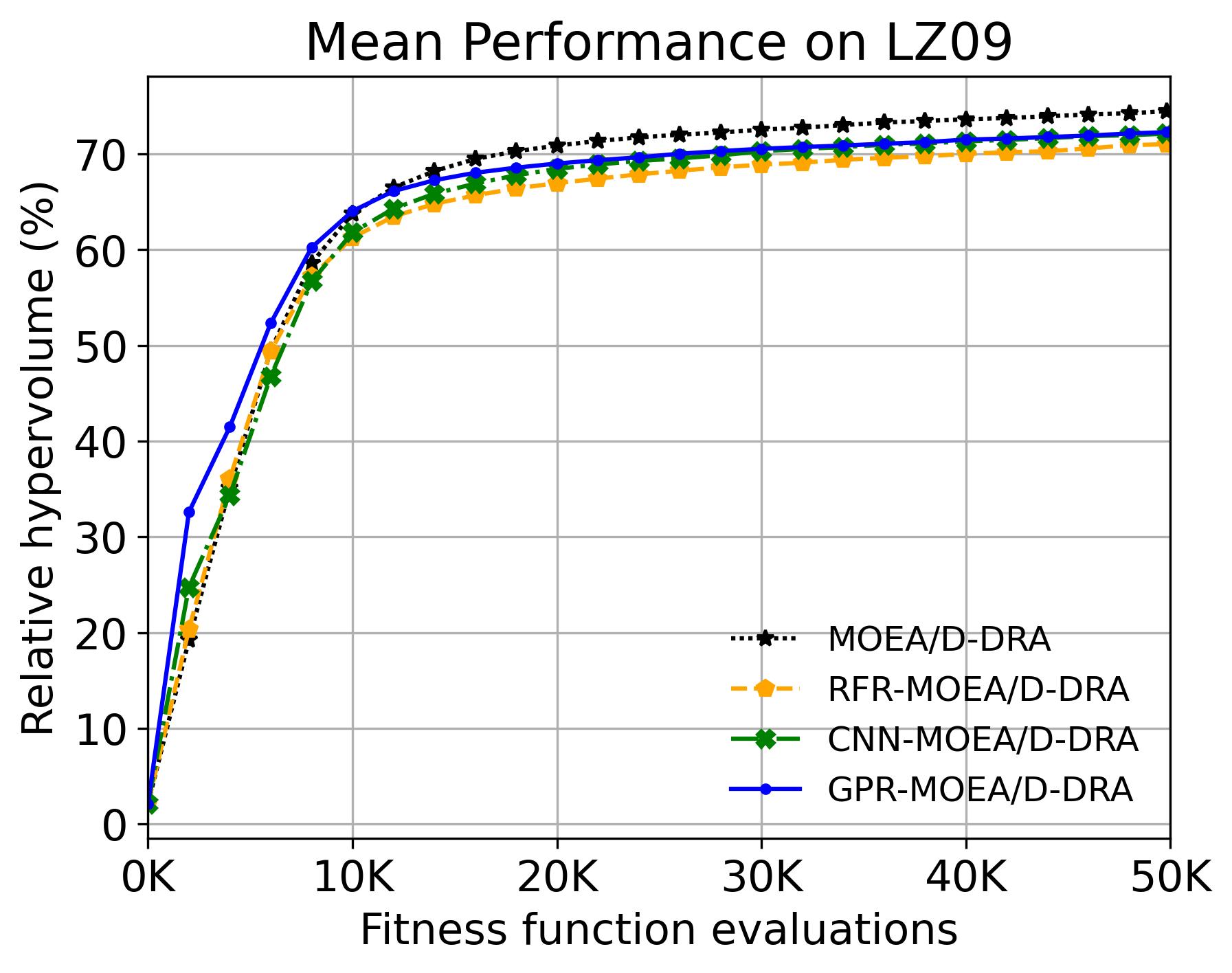}
        \caption{MOEA/D-DRA}
        \label{fig:lz09_moead}
    \end{subfigure}

        \caption{Comparative performance of the surrogate-enhanced NSGA-II and MOEA/D-DRA variants and their baseline versions on the LZ09 problem suite (9 problems).}
        \label{fig:LZ09perf}
\end{figure*}

\begin{figure*}
    \centering
    \begin{subfigure}[t]{0.455\textwidth}
        \centering
        \includegraphics[width=\linewidth]{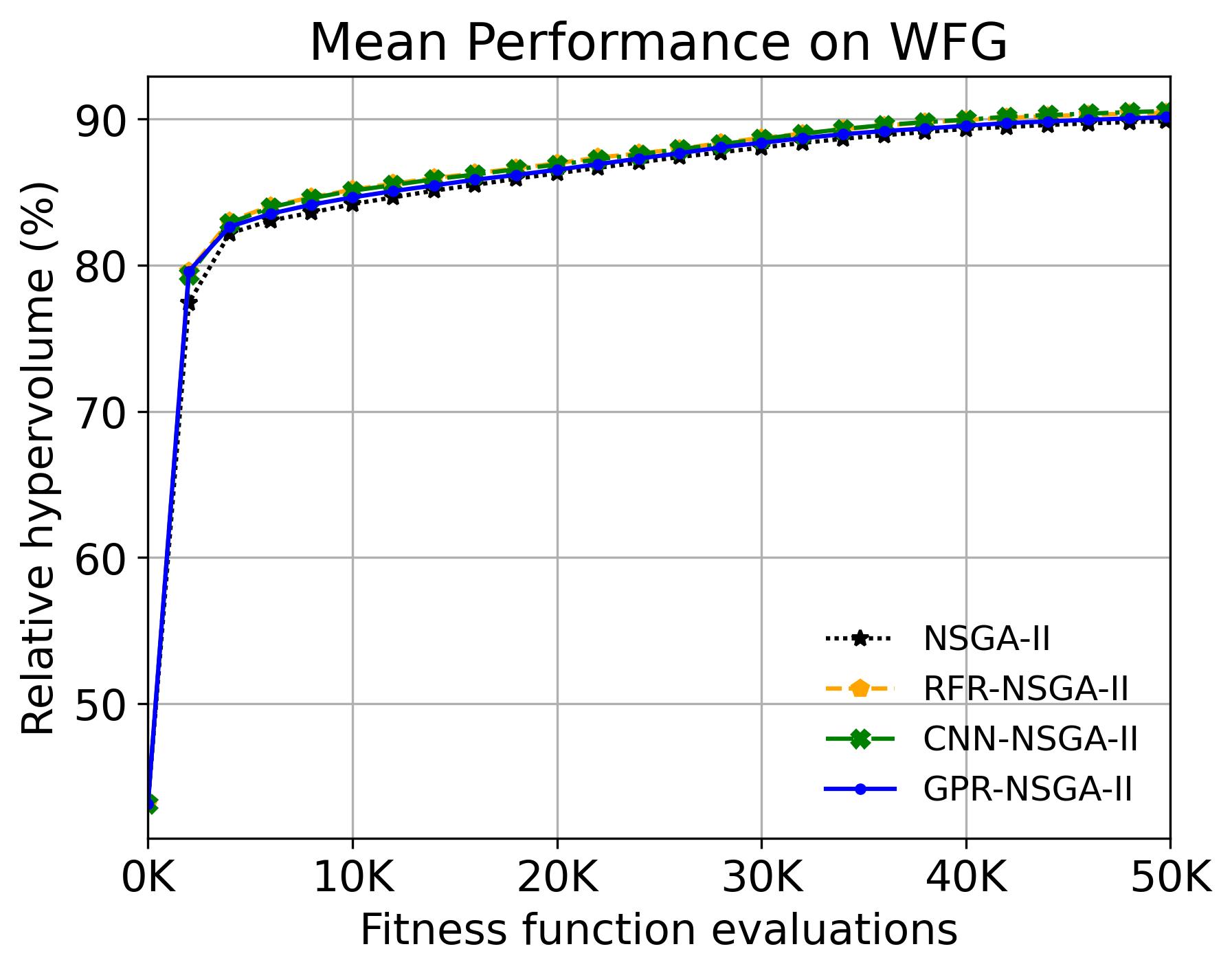}
        \caption{NSGA-II}
        \label{fig:wfg_nsgaii}
    \end{subfigure}
    \begin{subfigure}[t]{0.455\textwidth}
        \centering
        \includegraphics[width=\linewidth]{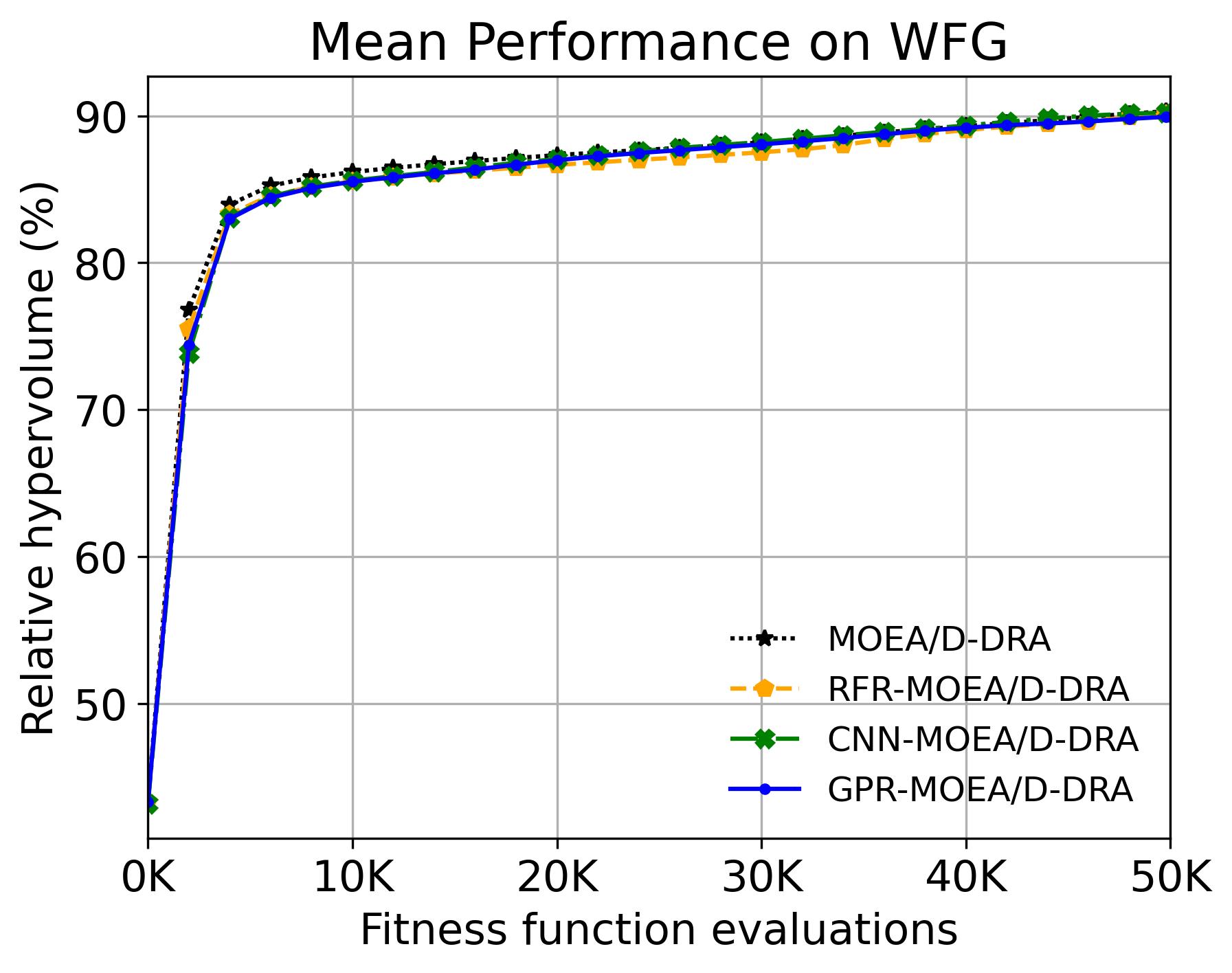}
        \caption{MOEA/D-DRA}
        \label{fig:wfg_moead}
    \end{subfigure}

        \caption{Comparative performance of the surrogate-enhanced NSGA-II and MOEA/D-DRA variants and their baseline versions on the WFG problem suite (9 problems).}
        \label{fig:WFGperf}
\end{figure*}

\begin{figure*}
    \centering
    \begin{subfigure}[t]{0.455\textwidth}
        \centering
        \includegraphics[width=\linewidth]{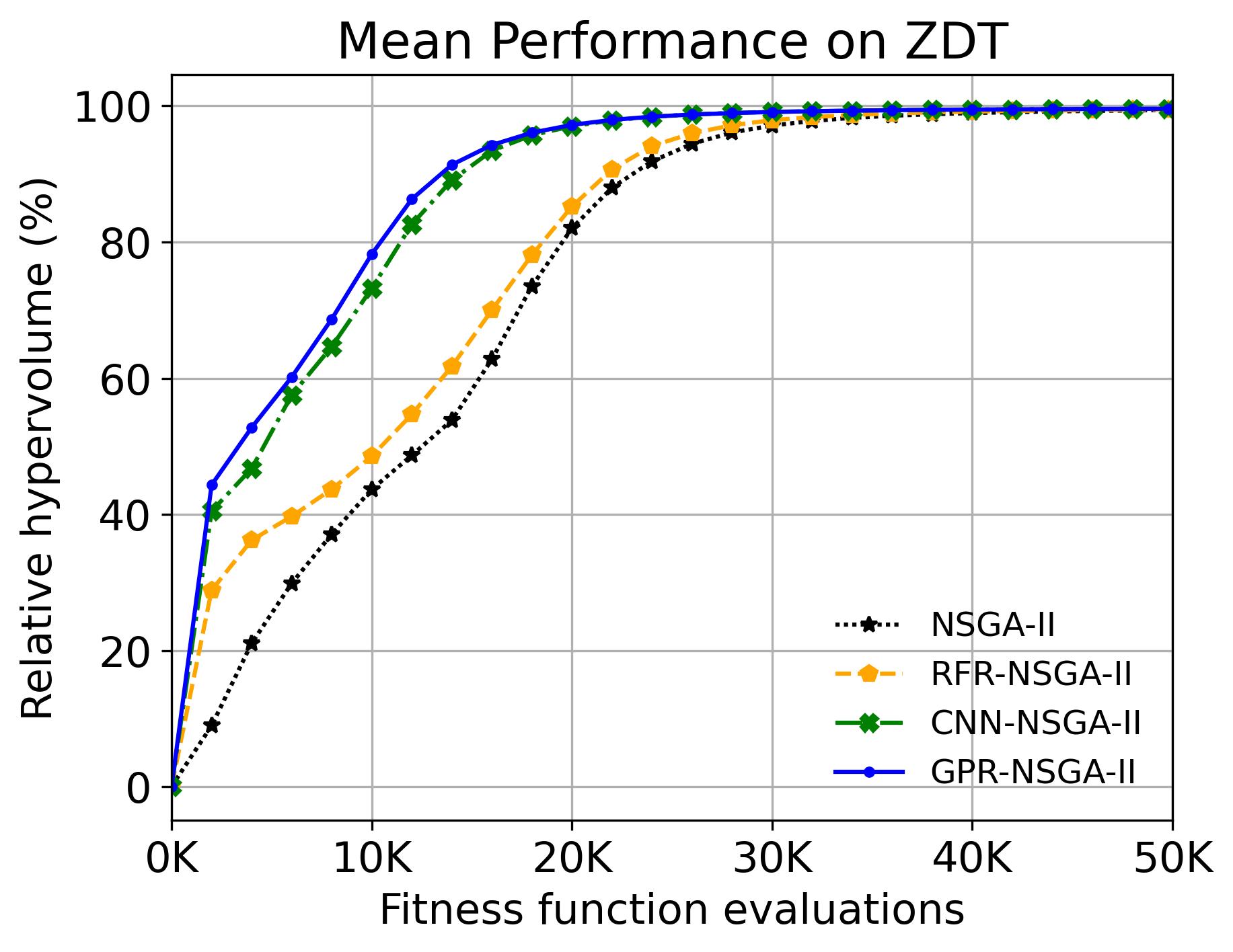}
        \caption{NSGA-II}
        \label{fig:zdt_nsgaii}
    \end{subfigure}
    \begin{subfigure}[t]{0.455\textwidth}
        \centering
        \includegraphics[width=\linewidth]{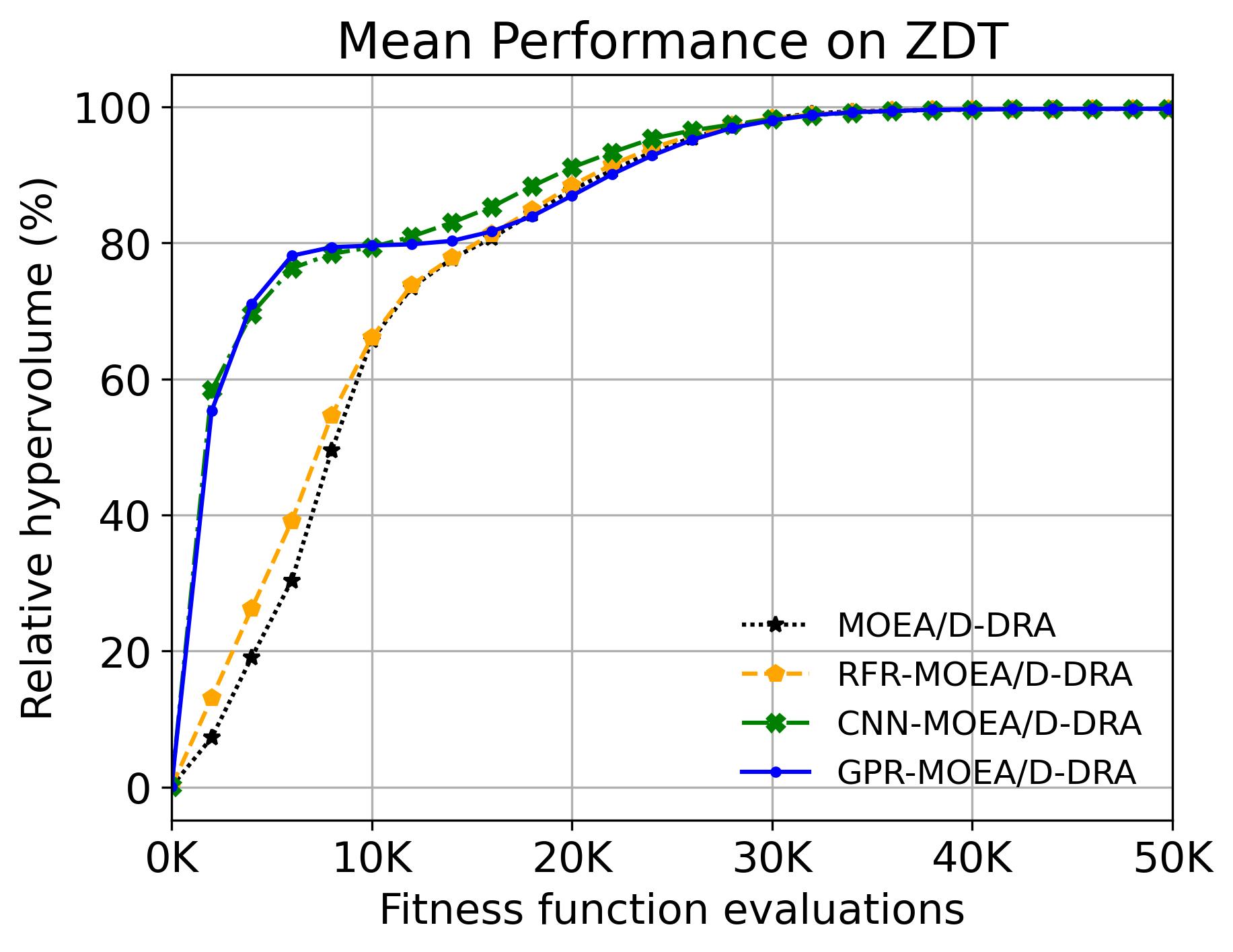}
        \caption{MOEA/D-DRA}
        \label{fig:zdt_moead}
    \end{subfigure}

        \caption{Comparative performance of the surrogate-enhanced NSGA-II and MOEA/D-DRA variants and their baseline versions on the ZDT problem suite (5 problems).}
        \label{fig:ZDTperf}
\end{figure*}

We present, in Figures \ref{fig:DTLZperf} to \ref{fig:ZDTperf}, the average comparative performance of the standard solvers against their surrogate-accelerated variants employing RFR, CNN and GPR on the 5 problem families we considered. To help with readability, whilst we do comment on individual problem performance, individual problem performance plots considering both $Hv(PF_c)$ and $IGD(PF_c)$ for all 31 benchmark problems are shown in the Appendix. 

\textbf{DTLZ Suite:} Figure \ref{fig:DTLZperf} shows the mean performance of the two groups of solvers on the DTLZ suite of problems (7 problems). The results show that NSGA-II convergence can be considerably improved by the Adaptive Accelerator using all three models (RFR, CNN and GPR) as the surrogate variants reach an average relative hypervolume of 50\% by generation 50, whereas, the standard NSGA-II only attains this $Hv(PF_c)$ at generation 80. The overall $Hv(PF_c)$ convergence profile of NSGA-II surrogate-accelerated variants on the DTLZ suite is influenced by their notable outperformance on the DTLZ7 problem where CNN and GPR achieve 80\%  $Hv(PF_c)$ at generation 10, whereas the baseline NSGA-II only achieves that performance at generation 95. In the case of MOEA/D-DRA, all surrogate-assisted variants and especially the CNN one perform well on DTLZ2, DTLZ5, DTLZ6 and DTLZ7. However, both CNN- and GPR-MOEA/D-DRA fail to reach meaningful $Hv(PF_c)$ values on DTLZ1 and DTLZ3, leading to the subpar performance over the entire suite that is shown in Figure \ref{fig:dtlz_moead}. It is very important to note that the Appendix $IGD(PF_c)$ plots confirm that all tested solvers are able to converge on any benchmark problem. However, for a few problems the best $PF_c$ some solvers discover within the allocated computation budget are not $Hv(PF_c)$ meaningful (i.e., yield a value close to 0) given the reference point used in the benchmark formulation to compute the $Hv(PF_c)$. In the Appendix (Figure \ref{fig:manyObjectivePerformance}), we also present the comparative performance on DTLZ5 and DTLZ7 versions with 4 objectives. The performance boosts delivered by all surrogate-assisted variants in these preliminary tests are on par, if not larger (in the case of RFR and NSGA-II) than those observed in the multi-objective versions of these two problems, indicating that the proposed Adaptive Accelerator strategy has potential to scale well on many-objective optimisation scenarios.           

\textbf{KSW:} The $Hv(PF_c)$ performance of the solvers on the KSW problem is shown in Figure \ref{fig:KSWperf}, where a trend similar to that exhibited on the DTLZ suite is observed. On the one hand, the three NSGA-II accelerators induced a consistently enhanced performance throughout the runtime. On the other hand, the MOEA/D-DRA surrogate-based variants failed to generate similar improvement, with CNN-MOEA/D-DRA determining a noticeable drop in convergence performance. 

\textbf{LZ09 Suite:} Comparative performance $Hv(PF_c)$ results on the LZ09 problem suite (9 problems) are shown in Figure \ref{fig:LZ09perf}. These results indicate that, over the problem suite, the  surrogates provide at best a minimal early convergence boost for NSGA-II. The early convergence acceleration is more evident for MOEA/D-DRA, with GPR-MOEA/D-DRA achieving 32.6\% suite average $Hv(PF_c)$ at generation 10, followed by CNN-MOEA/D-DRA achieving 24.7\%, RFR-MOEA/D-DRA achieving 20.3\%, and the baseline solver achieving 19.2\%. On individual benchmark problems, GPR-MOEA/D-DRA consistently achieves good performance across both $Hv(PF_c)$ and $IGD(PF_c)$ metrics in the first 50 generations, apart from LZ09\_F2 and LZ09\_F9. It is also worthwhile that the mean end-of-the-run $Hv(PF_c)$ attainment over the entire LZ09 suite is slightly lower for the surrogate-based variants. 

\textbf{WFG Suite:} Figure \ref{fig:WFGperf} shows the comparative $Hv(PF_c)$ results of the two groups of solvers on the WFG problem suite which comprises of 9 individual problems. Over the entire suite, surrogate-based variants have a slight convergence speed edge in the case of NSGA-II and show a slight disadvantage for MOEA/D-DRA. On individual problems, GPR-NSGA-II and CNN-NSGA-II surrogates perform very well on WFG1, WFG5, WFG6 and WFG8 as confirmed by both $Hv(PF_c)$ and $IGD(PF_c)$ metrics. GPR-MOEAD/D-DRA demonstrates some convergence speed improvements on WFG1 and WFG8, but all MOEA/D-DRA surrogate variants underperform on WFG9 on the $Hv(PF_c)$ metric.  

\textbf{ZDT Suite:} Comparative $Hv(PF_c)$ results on the ZDT problem suite (5 problems) are shown in Figure \ref{fig:ZDTperf} and they indicate that GPR- and CNN-based Adaptive Accelerators are able to drive notable convergence speed boosts for both NSGA-II and MOEA/D-DRA. The versions using RFR are also able to improve over baseline performance (especially in the case of NSGA-II), but the gains in convergence speed are far inferior to those demonstrated by the other two modelling strategies. Drilling down to individual problems, the CNN and GPR-based variants have a vastly superior convergence performance -- across both $Hv(PF_c)$ and $IGD(PF_c)$ -- on the entire ZDT suite for NSGA-II and on ZDT1, ZDT2 and ZDT6 for MOEA/D-DRA.

\subsection{Case Study: North-East Atlantic Fish Stock Assessment Surveys}
\label{sec:CaseStudy}
In this section, we describe a real-world computationally intensive problem that provided us with a relevant case study to further evaluate the efficacy of our surrogate-based MOEA convergence acceleration technique. The context is fish stock assessment in the North-East Atlantic, where the International Council for Exploration of the Seas (ICES) carries out bottom-trawl surveys. Protocols for conducting the surveys are detailed in the \textit{Manual for the North Sea International Bottom Trawl Surveys} \cite{ICES2019}. According to the manual, the surveys aim at providing consistent and standardised data for examining spatial and temporal changes in (a) the distribution and relative abundance of fish and fish assemblages; and (b) the biological parameters of commercial fish species for stock assessment purposes. The surveys make use of the segmentation of the North-East Atlantic into statistical grid rectangles measuring 1 degree in longitude by 0.5 degrees in latitude. Survey vessels typically conduct one or two trawling hauls within each grid cell, each lasting approximately 30 minutes. However, greater importance is placed on covering all grid cells rather than having two samples collected in a grid cell. Occasionally, coverage gaps occur due to adverse weather or technical failures of the vessels. 

Offshore wind energy plays a pivotal role in the UK's strategy to decarbonise its energy system and meet net-zero targets. The UK government has set bold offshore wind capacity goals: 50 GW by 2030 -- of which 5 GW is expected to come from floating wind -- and potentially scaling up to 140 GW by 2050 \cite{Desnz2023}. However, once offshore wind farms are established, the surrounding areas become hazardous for fishing and survey vessels to operate. This impacts two primary stakeholder groups:
\begin{itemize}
\item \textbf{Policy makers} must understand how the placement of offshore wind farms will influence fishing activities. Their challenge lies in identifying which grid cells should be excluded from development to minimise disruption.
\item \textbf{Fisheries scientists and survey teams} face reduced spatial coverage for the survey, raising concerns about the accuracy of survey-based models. The question becomes: How robust are fish abundance estimates if some areas are no longer surveyed?
\end{itemize}

To explore this issue, we modelled it as a binary optimisation problem with three objectives, focusing solely on Haddock populations for simplicity. Using DATRAS data from 2010 to 2023, we divided the North East Atlantic into 439 grid cells, with each candidate solution represented by a binary encoding indicating which cells were selected for inclusion in the abundance index calculations. We adopted a binary encoding scheme because each decision variable represents whether a grid cell is selected (1) or not (0). The first and second objectives measured the impact on the abundance index for quarters 1 and 3/4, respectively, while the third objective minimised the number of excluded (deselected) grid cells (i.e., cell value = 0). 
The impact was measured by comparing a solutions index with the true index, both produced using Delta -lognormal by age over time as follows. 

\begin{equation}
\text{Difference} = -\sum_{i=1}^{n} \left| \frac{y'_i}{y_i} - 1 \right|
\end{equation}

\noindent where $y$ is the true index value, $y'$ is the new index value based on a given solution candidate (i.e., with measurements from deselected cells excluded), and $i=1 \dots n$ is the period (i.e. from year 2010 to 2023).

Computing the Haddock index of abundance over a 14-year period is a very computationally intensive task. The delta-lognormal model generates indices for eight age groups in quarter 1 and nine age groups in quarter 3/4, for each year. Depending on the specific candidate solution, calculating these indices for both quarters takes approximately 15 to 30 minutes. As a result, evaluating the fitness of a single solution can require between 30 minutes and 1 hour.
Given these computational demands, running an optimisation with, for example, a budget of 50,000 fitness function evaluations becomes highly time-consuming. Evaluating 50k solutions sequentially -- each taking a minimum of 30 minutes -- would require over 1,042 days to complete (wall-clock time). Clearly, a sequential approach is impractical for this scale of problem.

To address this, we developed a custom population evaluator within the jMetalPy framework to enable parallel evaluation on an HTCondor-based computing cluster \cite{Thain2005}. The HTCondor cluster is composed of 4 machines, each with the following specifications: 24 CPUs, 64GB memory, NVIDIA RTX A5000 GPU,  Ubuntu 22.04.5 LTS operating system. This allowed us to evaluate
candidate solutions in parallel. We then compared the performance of standard NSGA-II against three NSGA-II-based surrogate-assisted variants on the fish abundance modelling problem, using the same parameters as in the benchmark problems and a computational budget of 40,000 fitness evaluations. With parallel execution, a single optimisation run was completed in approximately 11 days. We only managed to do a single run for the baseline NSGA-II solver and its three surrogate-based variants as it was impractical to repeat the experiments say 100 times as we did with the benchmarks. 

Furthermore, due to the complexity of the problem (binary encoding of size 439), and the desire to give all solvers a strong starting point, the initial population was generated using a combination of two approaches: 50\% of the solutions were generated randomly to introduce diversity, while the remaining 50\% were solutions derived through a "greedy" strategy that also aimed to create a baseline for the MOEA results. This greedy strategy involved a leave-one-out approach to first identify grid cells with the single highest total impact across the quarters of interest when excluded from the index. Afterwards, "greedy" solution candidates were iteratively created by adding individual exclusion cells in descending order of their leave-one-out total impact. Ultimately, 50\% of the MOEAs population was selected from the pool of "greedy" solution candidates, ensuring that we maintained a representative spread across the size of the excluded cliques of cell. 

\begin{figure}
    \centering
    \includegraphics[width=0.455\textwidth]{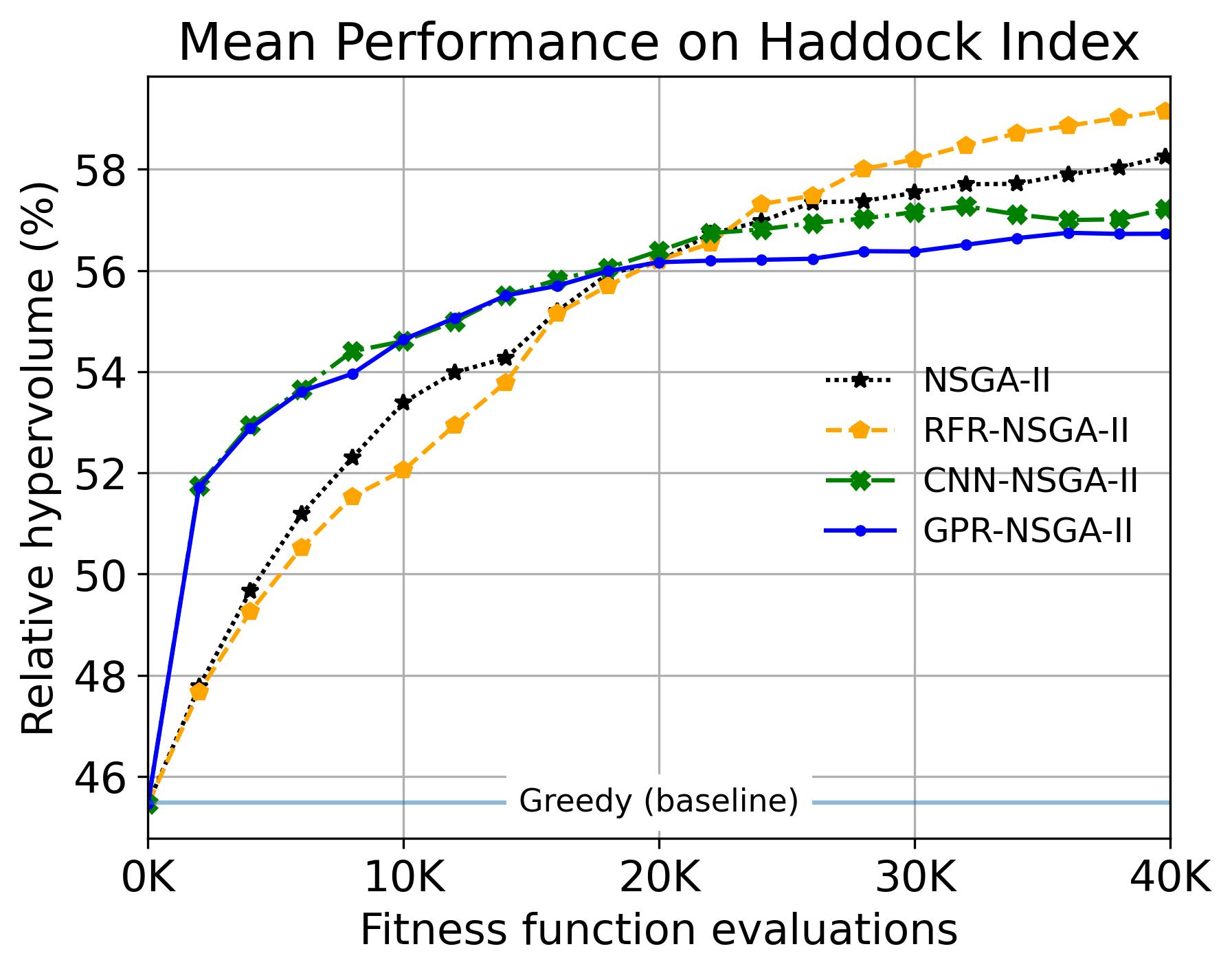}
    \caption{Comparative performance of surrogate-enhanced NSGA-II and its standard version on the Haddock Survey optimisation.}
    \label{fig:haddock}
\end{figure}

Comparative results are shown in Figure \ref{fig:haddock}, where the three surrogates accelerate the convergence speed of NSGA-II in the early generations. As seen with several benchmark problems, the early-stage acceleration delivered by GPR-NSGA-II and CNN-NSGA-II is noticeable. When compared to the greedy baseline, the Adaptive Accelerators using CNN and GPR models are able to maintain a lead of 1 to 2 days over the standard solver in the first third of the run. Towards the end of the optimisation run, the performance of the GPR and CNN variants is matched end even slightly surpassed by the standard NSGA-II run. RFR-NSGA-II does not provide any early stage convergence improvement, but outperforms all solvers on end-of-the-run relative hypervolume attainment. In terms of surrogate exit points, all three surrogates exited at generation 6, having run for 5 generations. Given that the average training time for a single surrogate model was 3 minutes and 46 seconds (RFR), 3 minutes and 49 seconds (CNN) and 1 minute and 52 seconds (GPR), the total computational burden associated with surrogate model training was negligible. However, the performance of all three surrogate-based variants on the case study scenario indicate their general usefulness (speed and quality attainment) within a computational approach designed to offer fisheries scientists a means to stress test abundance indices when aiming to assess and ultimately improve their robustness.  

\section{Conclusions}
\label{sec:Conclusion}

In this paper we have demonstrated the efficacy of an on-the-fly surrogate modelling technique designed to accelerate the early stage convergence speed of multi-objective evolutionary algorithms. Despite being data-driven , the proposed technique is not resource-intensive as it only uses the most recently evaluated population to train models and also features an adaptive surrogate deactivation strategy. Using a streamlined configuration (with fixed parameter settings), the tested surrogate-enhanced solvers proved to accelerate the mean early convergence speed of NSGA-II and MOEA/D-DRA on a test aggregating 31 widely known benchmark problems. In particular, the convergence speed gains demonstrated by our NSGA-II Adaptive Accelerator variants using GPR and CNN models over this benchmark set clearly surpassed those achieved by recently proposed interpolation-based approaches. To demonstrate performance on multi-objective scenarios with a wide range of characteristics, we also applied our acceleration technique on a real-world computationally-intensive fisheries modelling problem in the North-East Atlantic that features a fairly large binary encoding and a specialised initialisation. The case study results show valuable convergence improvements when compared with the standard MOEA approach. 

Overall, numerical results indicate that our streamlined Adaptive Acceleration approach can demonstrably enhance the early-stage convergence of MOEAs. Consequently, when addressing computationally intensive optimisation problems, practitioners can obtain a high-quality set of PN solutions with substantially fewer fitness function evaluations. This, in turn, resulting in significant savings in both time and computational resources.

Even though it is generally successful in accelerating the convergence of both tested MOEAs over the benchmark set on average, our streamlined strategy also displays limitations as some surrogate-MOEA pairs underperform on specific test problems and even problem families (e.g., CNN and GPR variants of MOEA/D-DRA on DTLZ1 and DTLZ3). At the same time, RFR variants that tend to deliver a more luckluster performance on average perform very well on DTLZ3 and DTLZ6 (for NSGA-II) and much better than other MOEA/D-DRA surrogates on DTLZ1, DTLZ3, DTLZ4 and LZ09\_F6. Whilst tempting to simply interpret this as a corollary of the No Free Lunch Theorems \cite{Wolpert1997}, the differential behaviour of the RFR variants also offers significant potential for a further future refinement of our strategy based on ensemble ML models. This future refinement could also explore a potential update for the fixed parameterisation we propose in Section \ref{sec:Parameters}. The current "rule of 1/2" fixed setting (i.e., half-life surrogate deactivation criterion, 50\% surrogate solutions to be re-integrated, max surrogate generations set to 1/2 of host solver setting) and limitation to a single generation for training produces stable results and reduces the complexity of our proposed approach, but a thorough sensitivity analysis might explore if tailoring these setting can improve the general performance of  Adaptive Accelerators to a level that warrants the associated increase in complexity.        

Furthermore, future research will investigate the effect of offspring population size on the performance of the Adaptive Accelerators. We will also compare the performance of the Adaptive Accelerators with other similar surrogate modelling approaches such as the Kriging assisted Reference Vector Guided Evolutionary Algorithm (K-RVEA) \cite{Chugh2018} the Classification-based Surrogate-assisted Evolutionary Algorithm (CSEA) \cite{Pan2019} This will give more opportunity for a comprehensive comparison of algorithmic behaviour and run-time performance.

\section*{Acknowledgments}
This work has been supported by the COMET-K2 ''Center for Symbiotic Mechatronics'' of the Linz Center of Mechatronics (LCM) funded by the Austrian federal government and the federal state of Upper Austria. We also would like to acknowledge 
<name withheld> for providing the case study and associated code for running the delta-lognormal models.

\printcredits

\bibliographystyle{cas-model2-names}
\bibliography{Tiwonge_Main_References}

\appendix
\section{Appendix}
In this Appendix, we plot the $Hv(PF_c)$ and $IGD(PF_c)$ comparative performance of the NSGA-II and MOEA/D based solvers on the individual 31 benchmark problems in Figures \ref{fig:annexNSGAII} to \ref{fig:annexMOEAD_igd}.

\begin{figure*}[t]
    \centering
    \begin{subfigure}[t]{0.455\textwidth}
        \centering
        \includegraphics[width=\textwidth]{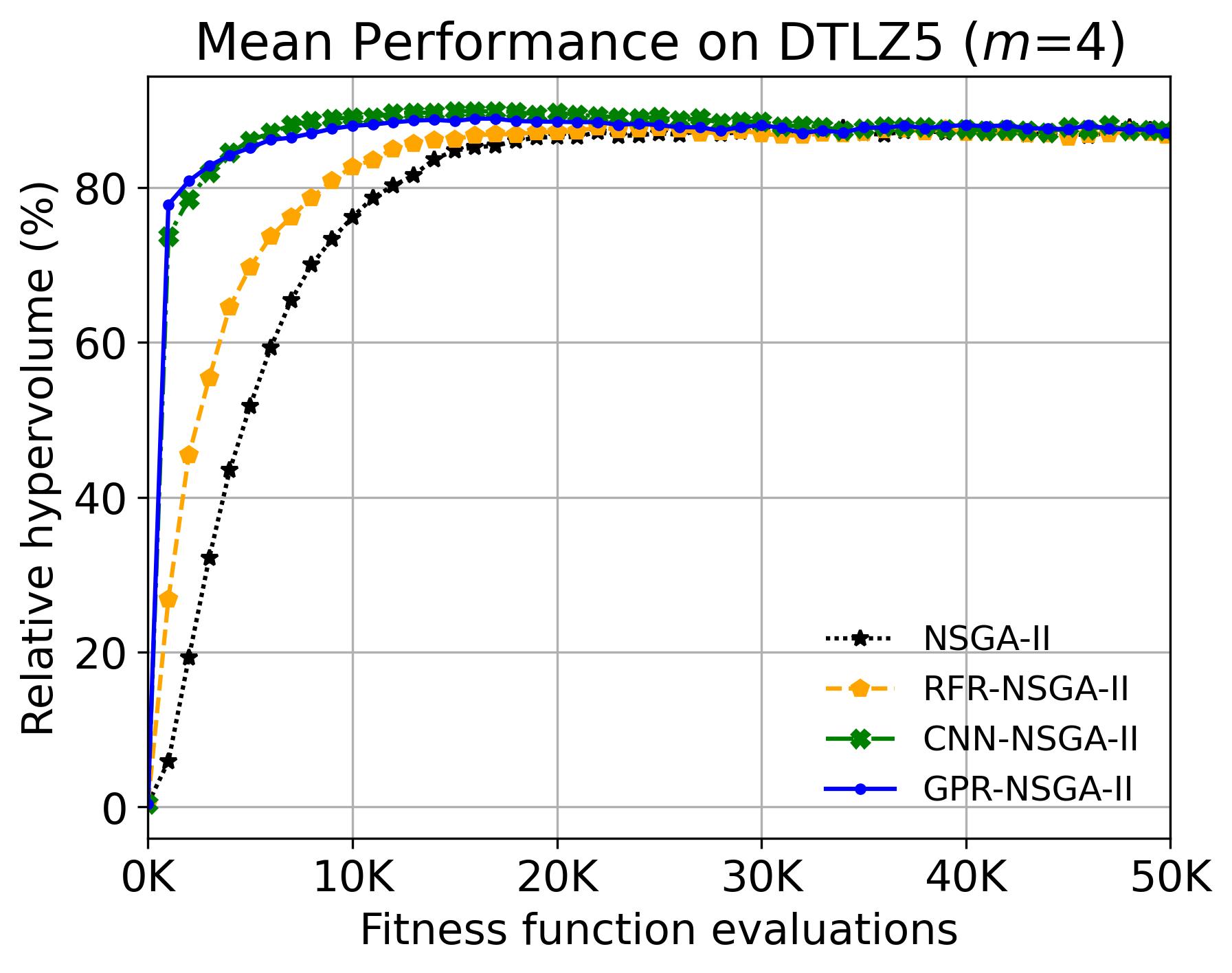}
        \caption{NSGA-II}
    \end{subfigure}
    \begin{subfigure}[t]{0.455\textwidth}
        \centering
        \includegraphics[width=\textwidth]{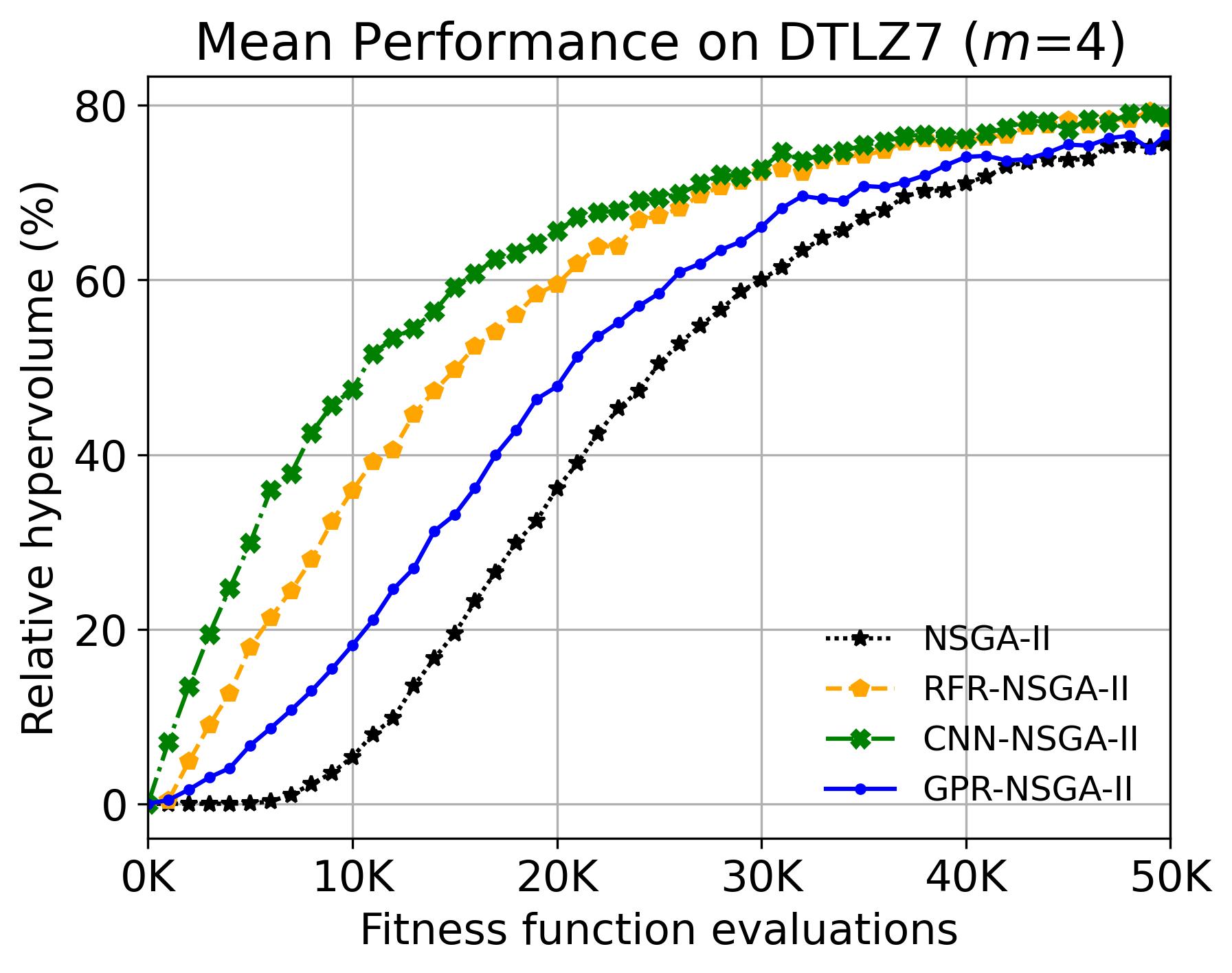}
        \caption{NSGA-II}
    \end{subfigure}
    \begin{subfigure}[b]{0.455\textwidth}
        \centering
        \includegraphics[width=\textwidth]{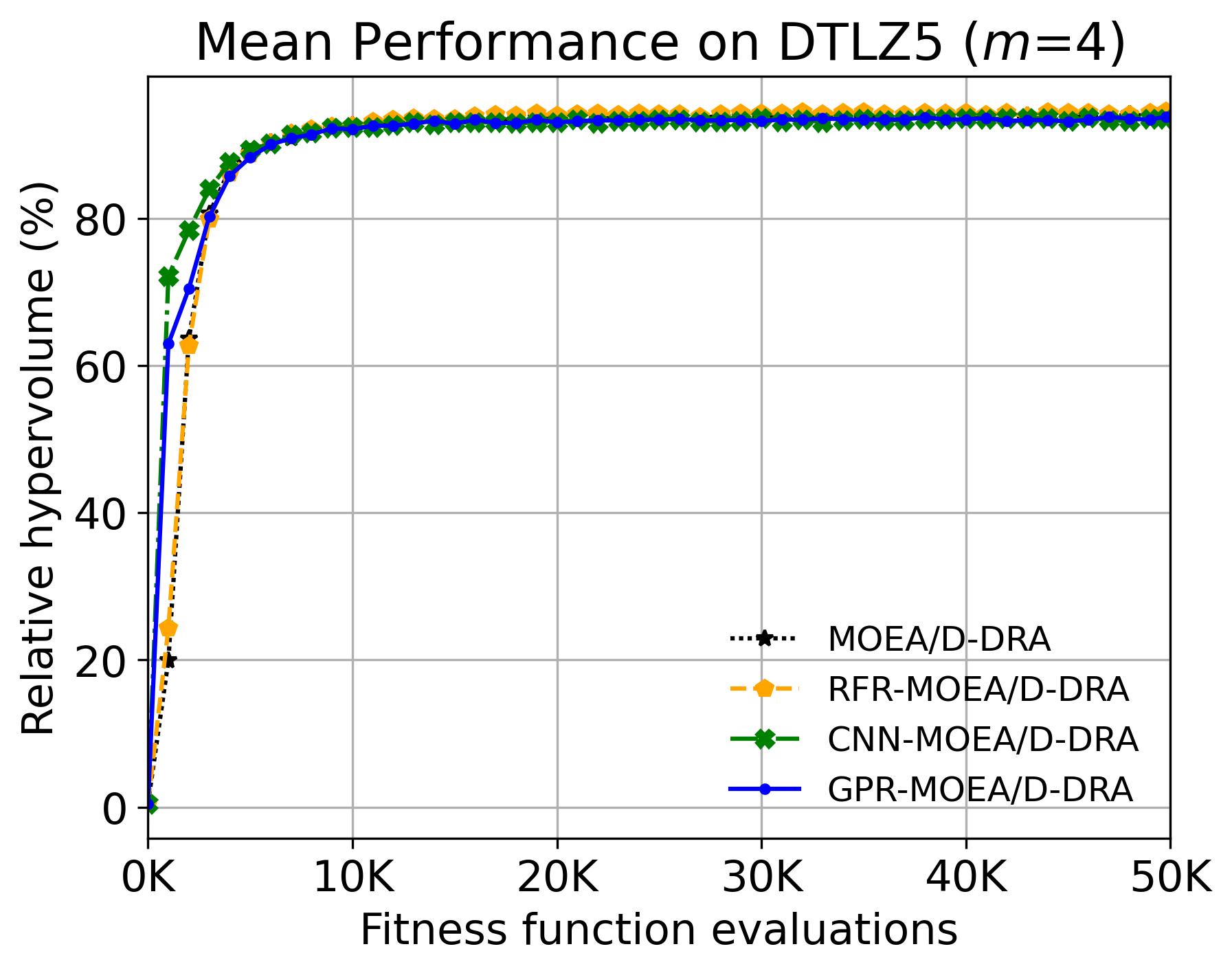}
        \caption{MOEA/D-DRA}
    \end{subfigure}
    \begin{subfigure}[b]{0.455\textwidth}
        \centering
        \includegraphics[width=\textwidth]{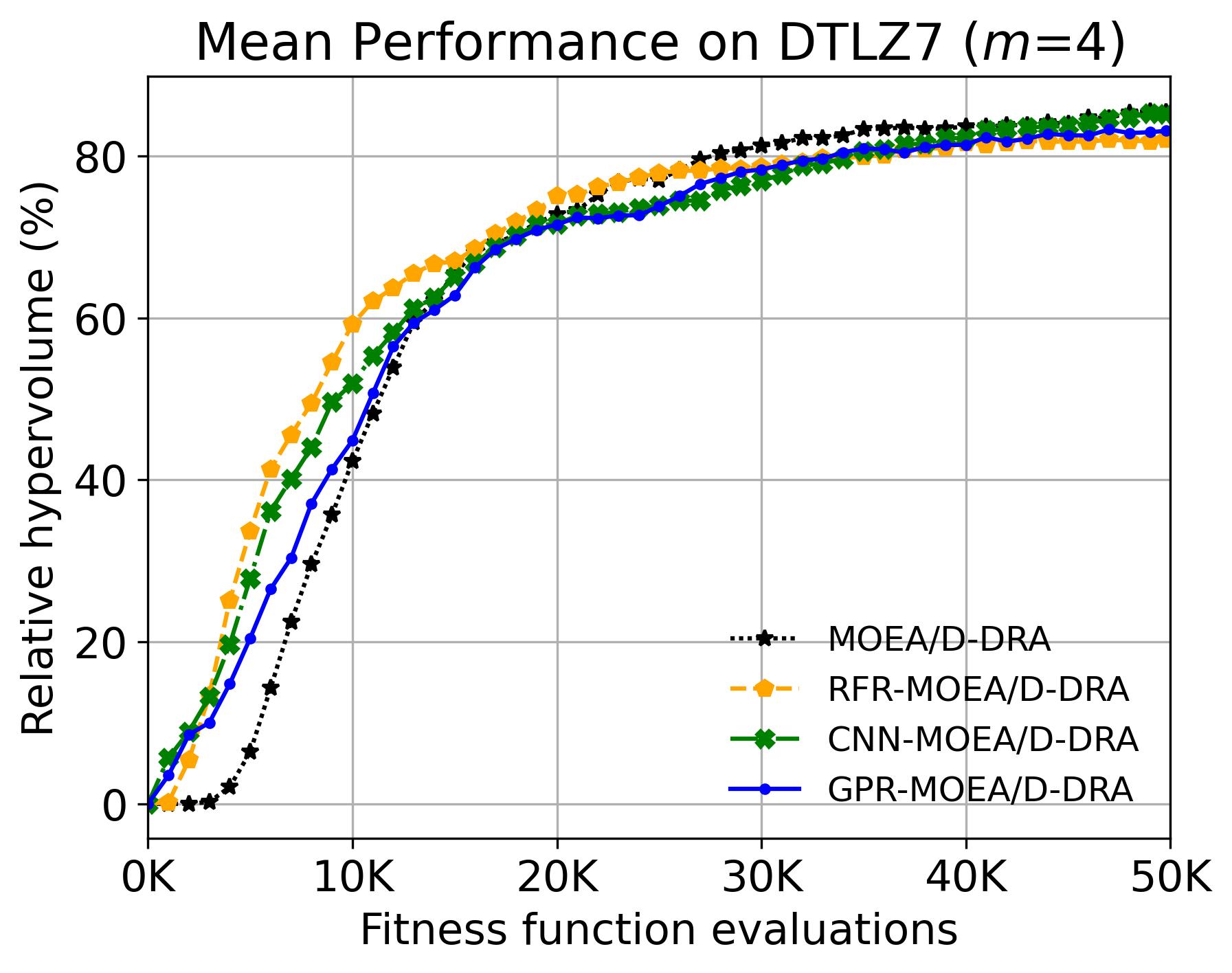}
        \caption{MOEA/D-DRA}
    \end{subfigure}
     \caption{$Hv(PF_c)$-measured mean comparative  performance of NSGA-II and MOEA/D-DRA based solvers on DTLZ5 and DTLZ7 versions with 4 objectives.}
    \label{fig:manyObjectivePerformance}
\end{figure*}

In Figure \ref{fig:manyObjectivePerformance} we show the comparative performance of the solvers on two DTLZ problem versions with 4 objectives (i.e., many objective optimisation problem extensions).

\newcommand{\figwid}{0.20}
\begin{figure*}[h]
    \centering
    \begin{subfigure}[t]{\figwid\textwidth}
        \centering
        \includegraphics[width=\linewidth]{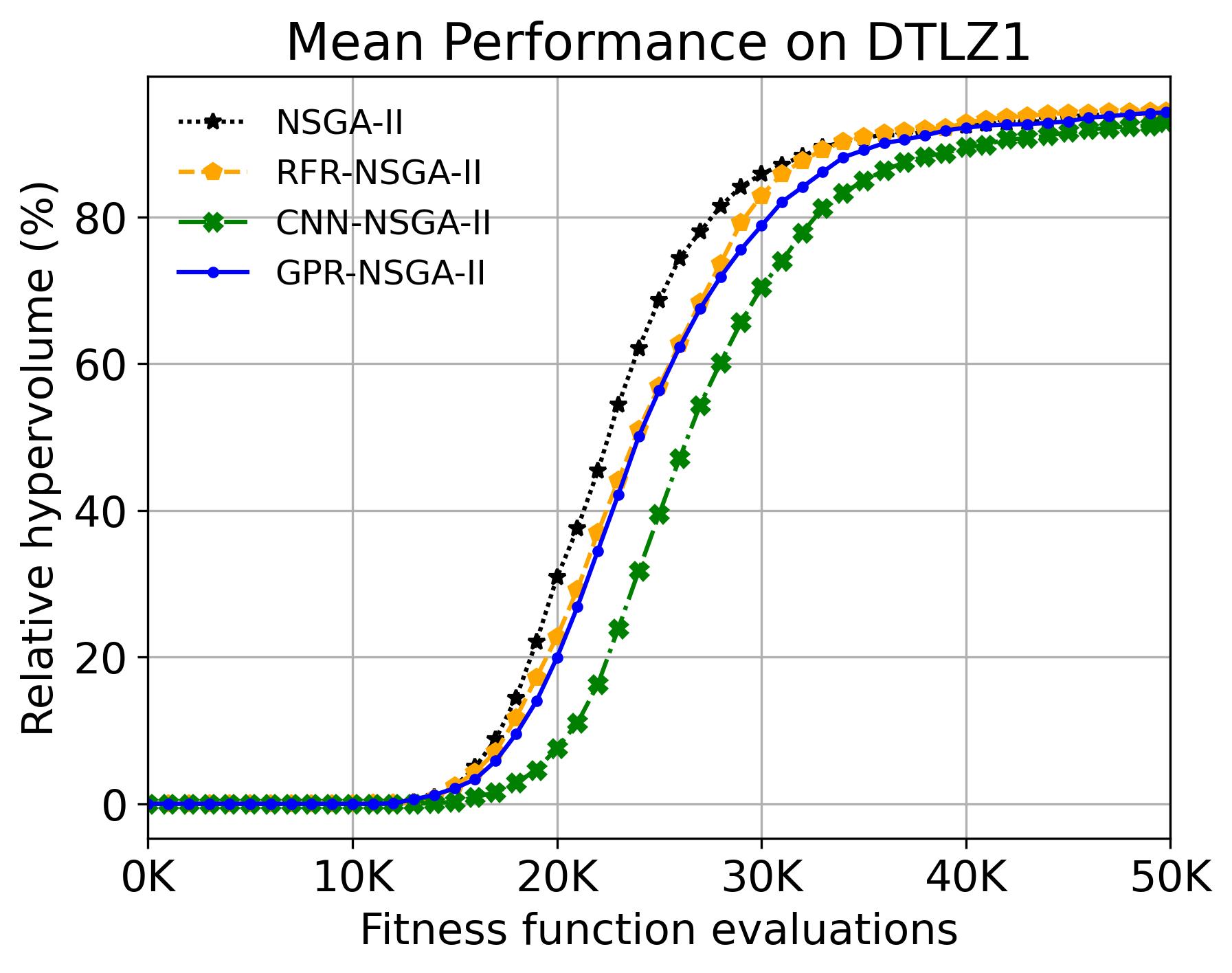}
    \end{subfigure}
    \begin{subfigure}[t]{\figwid\textwidth}
        \centering
        \includegraphics[width=\linewidth]{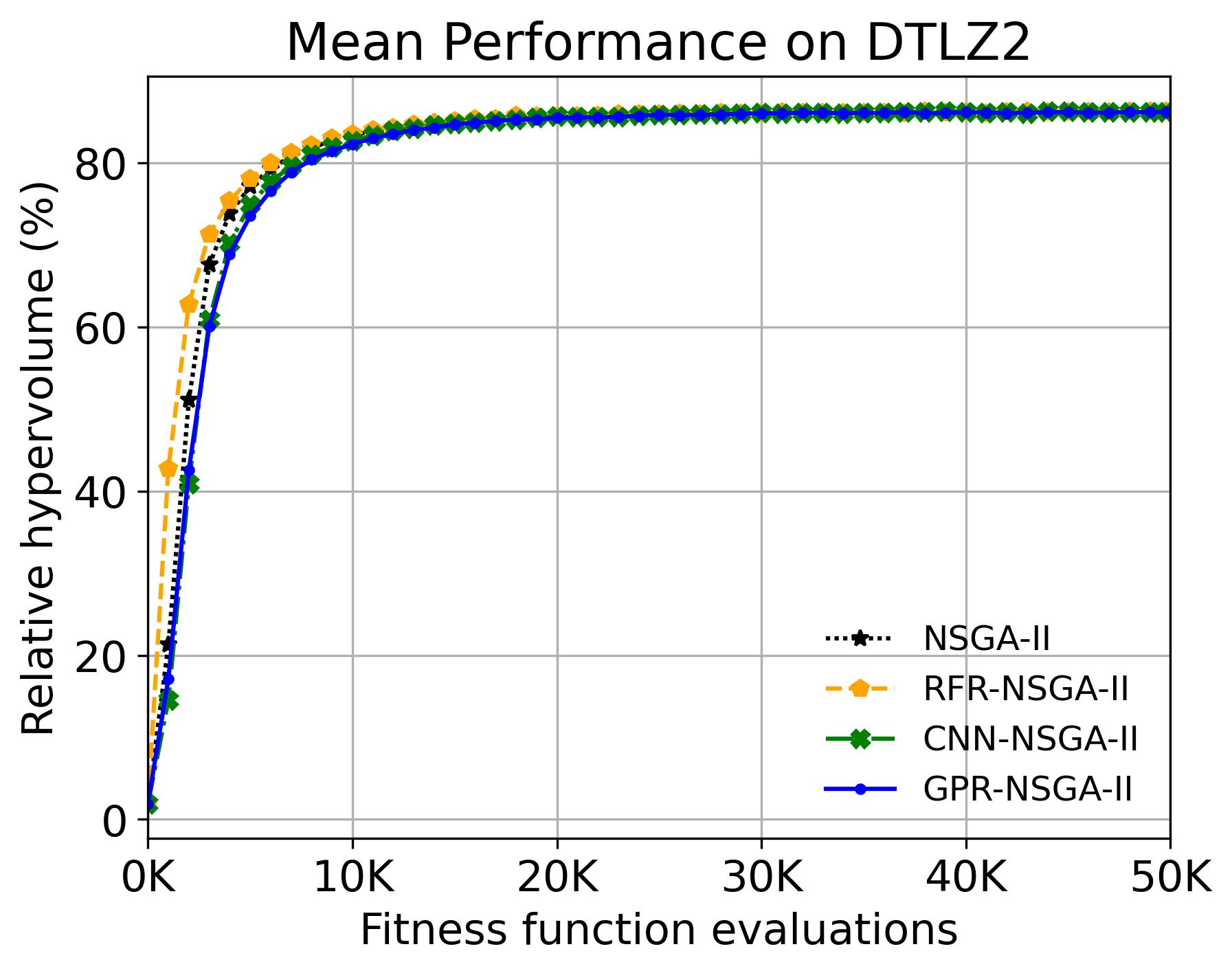}
    \end{subfigure}
    \begin{subfigure}[t]{\figwid\textwidth}
        \centering
        \includegraphics[width=\linewidth]{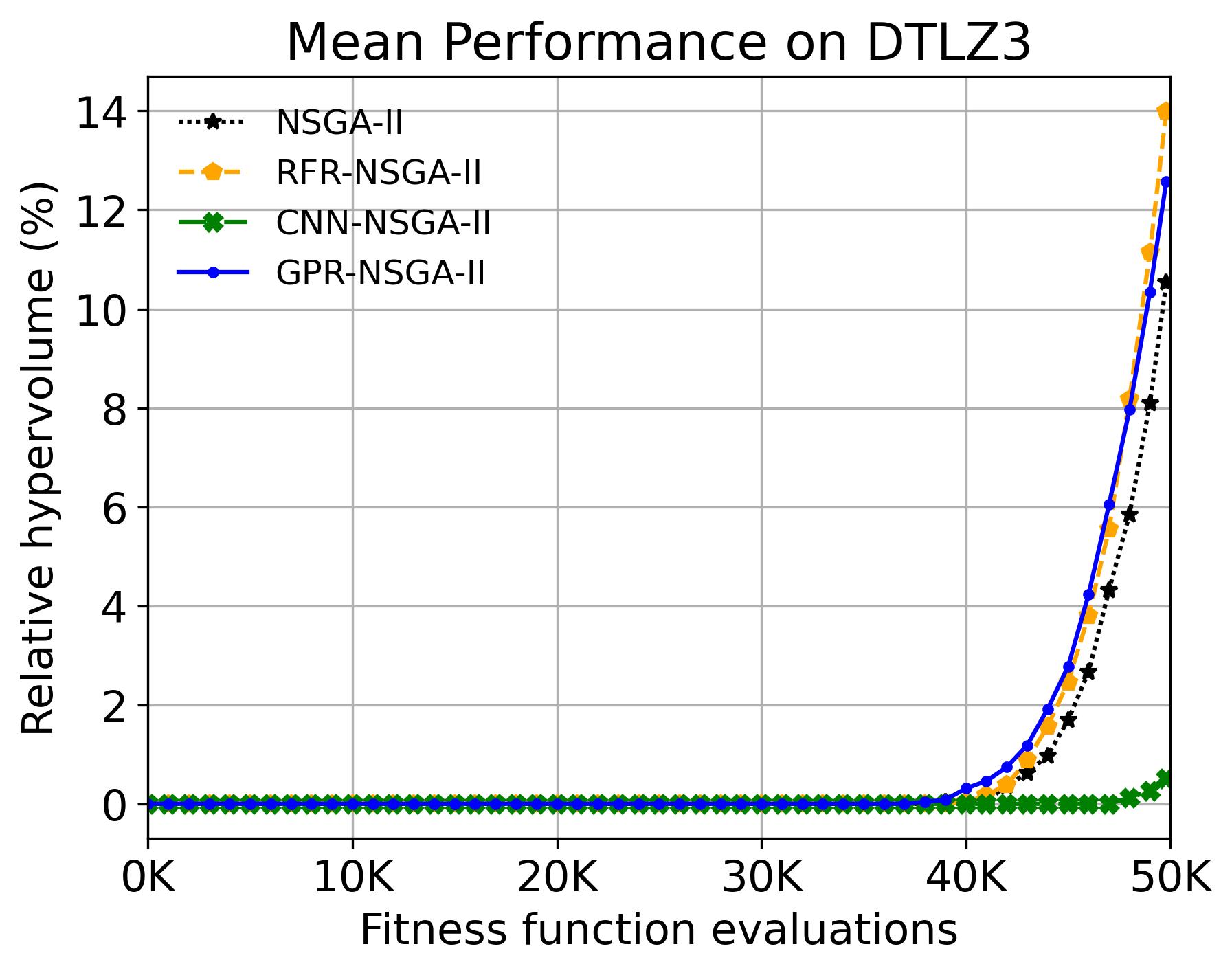}
    \end{subfigure}    
    \begin{subfigure}[t]{\figwid\textwidth}
        \centering
        \includegraphics[width=\linewidth]{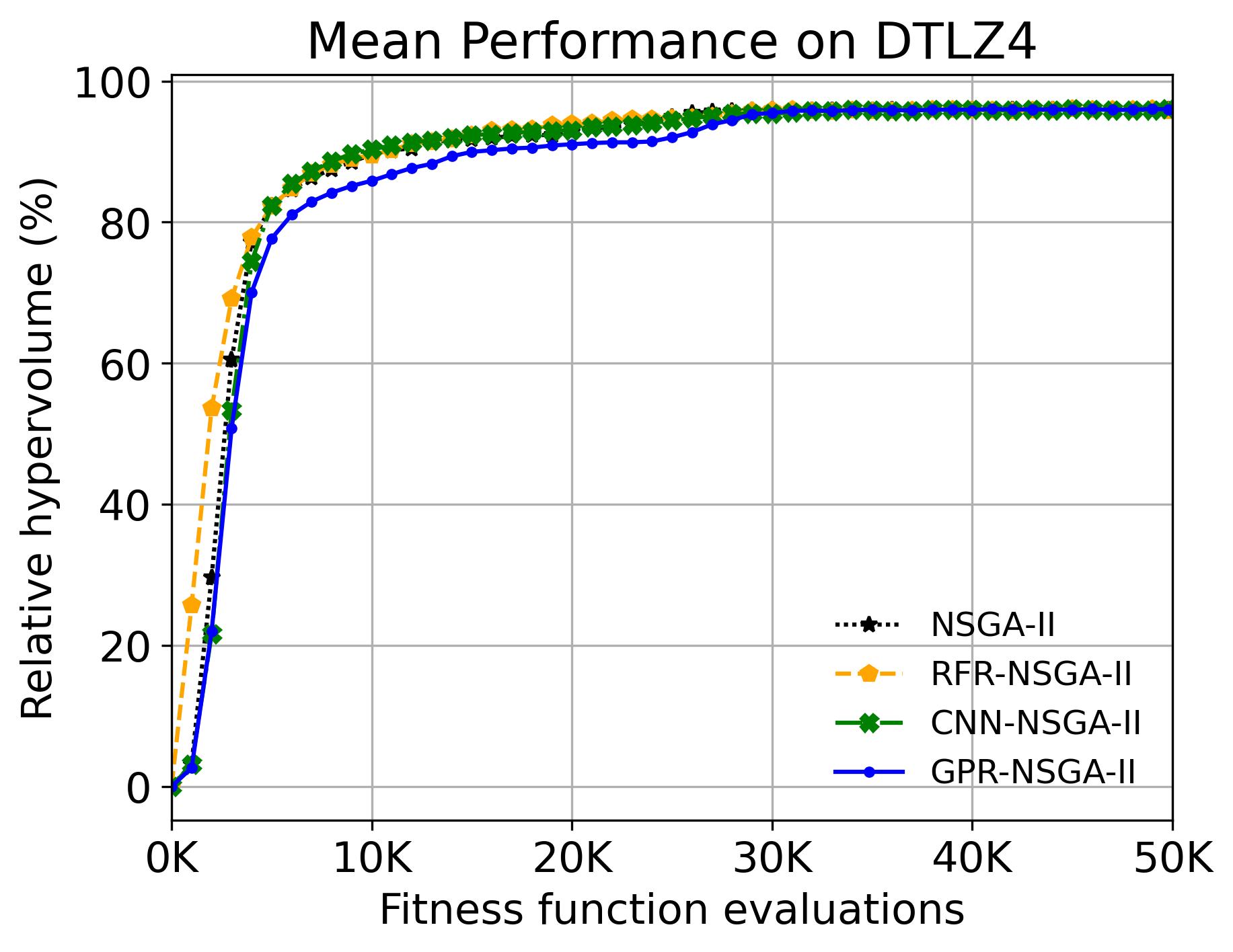}
    \end{subfigure}
    \begin{subfigure}[t]{\figwid\textwidth}
        \centering
        \includegraphics[width=\linewidth]{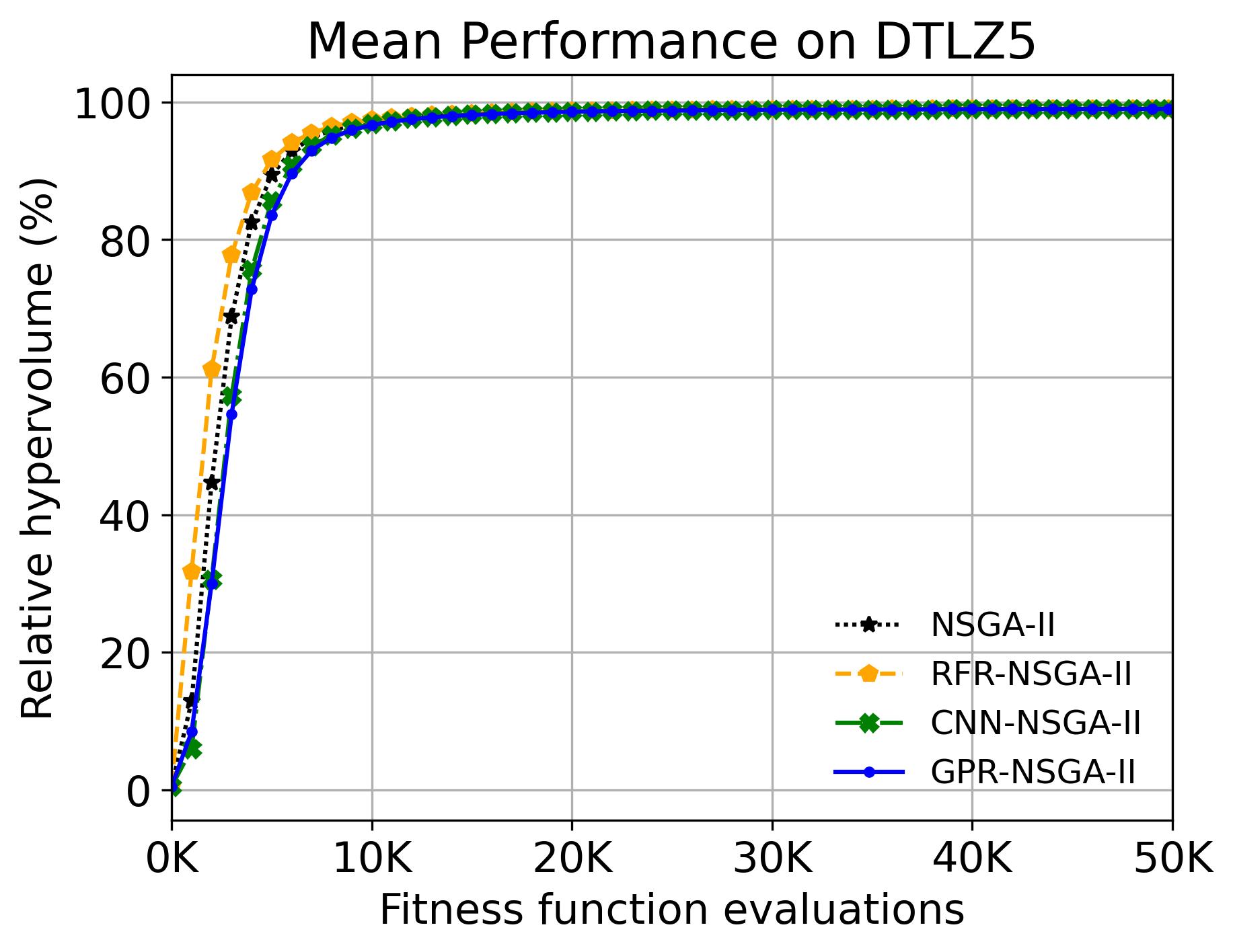}
    \end{subfigure}
    \begin{subfigure}[t]{\figwid\textwidth}
        \centering
        \includegraphics[width=\linewidth]{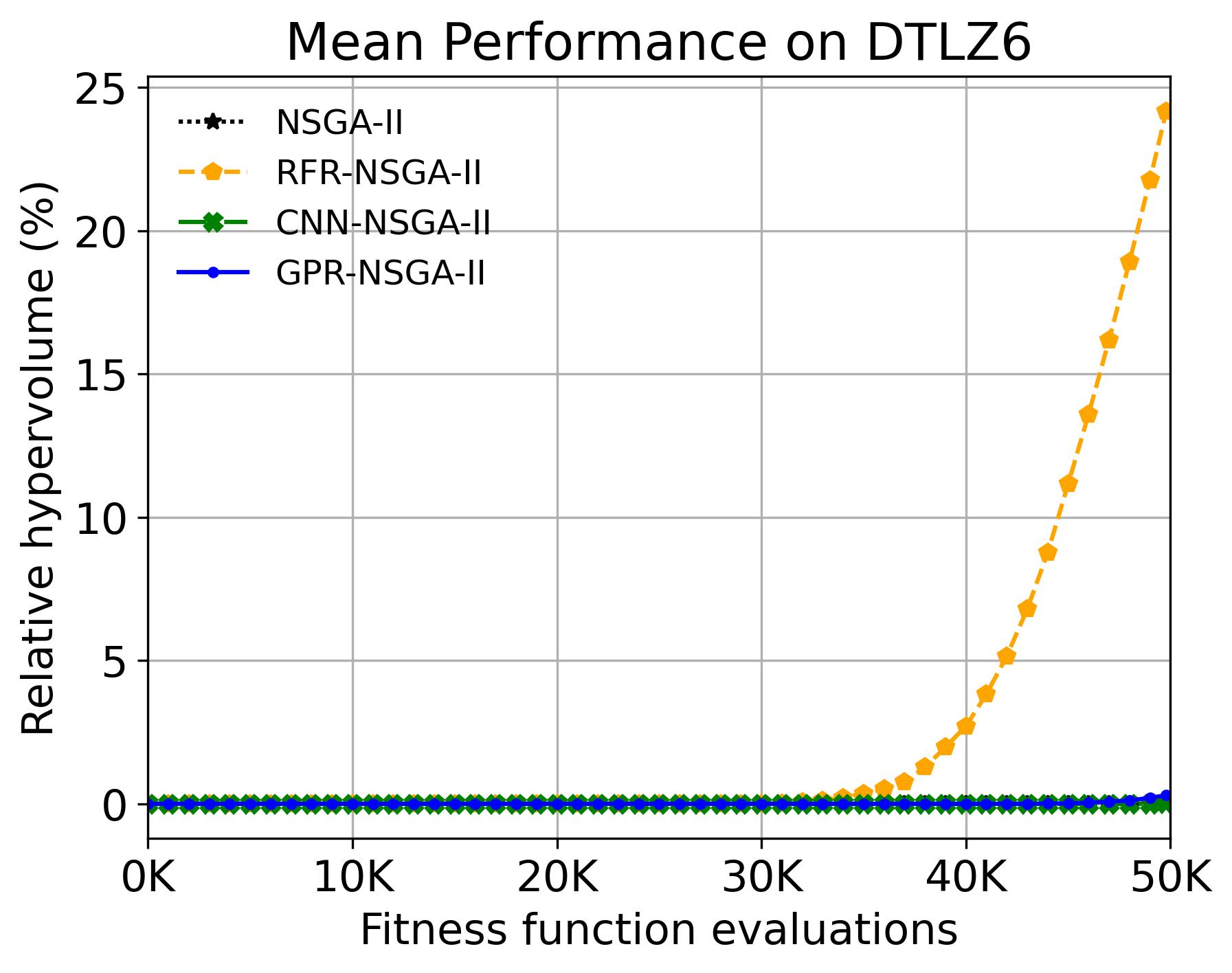}
    \end{subfigure}
    \begin{subfigure}[t]{\figwid\textwidth}
        \centering
        \includegraphics[width=\linewidth]{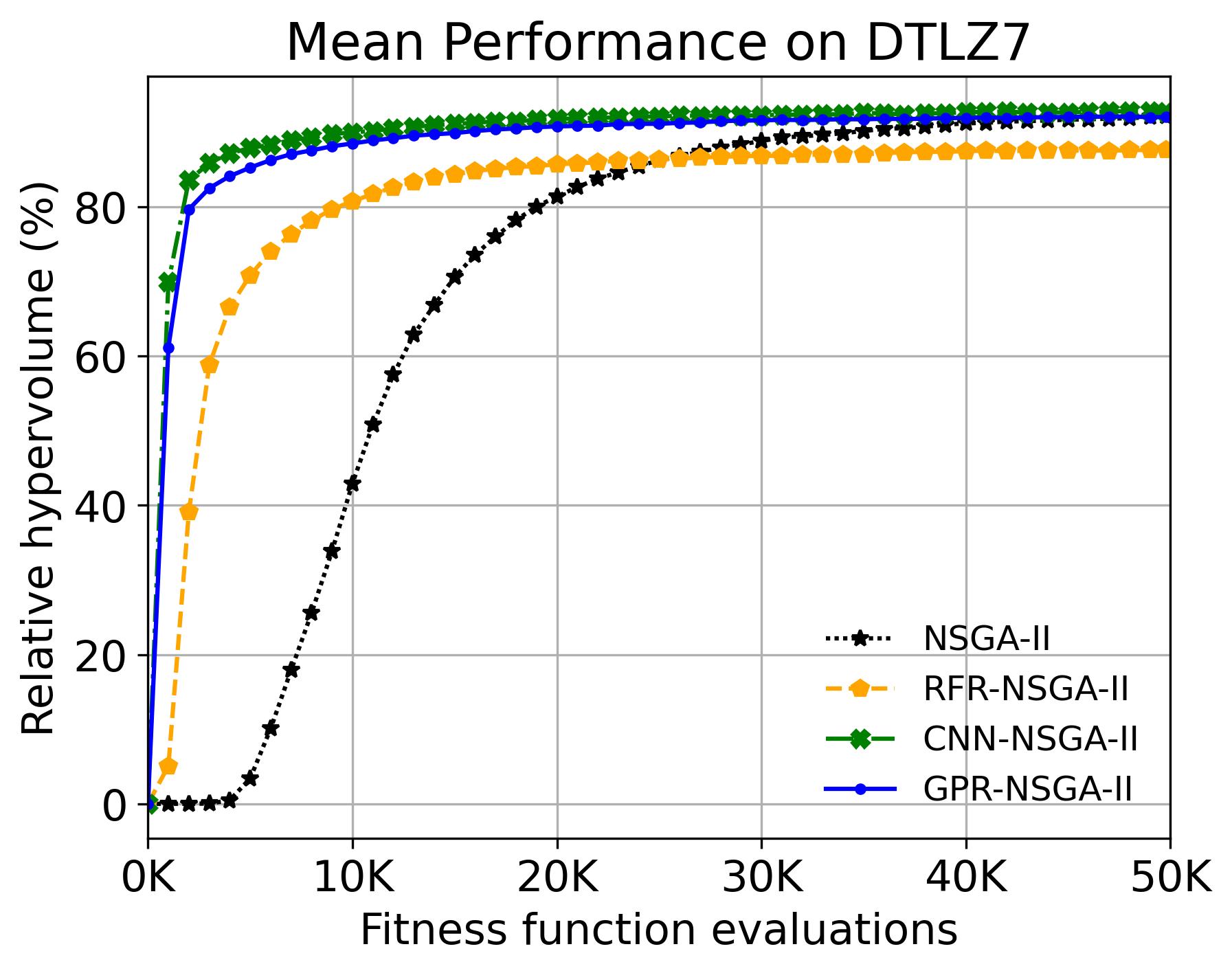}
    \end{subfigure}
    \begin{subfigure}[t]{\figwid\textwidth}
        \centering
        \includegraphics[width=\linewidth]{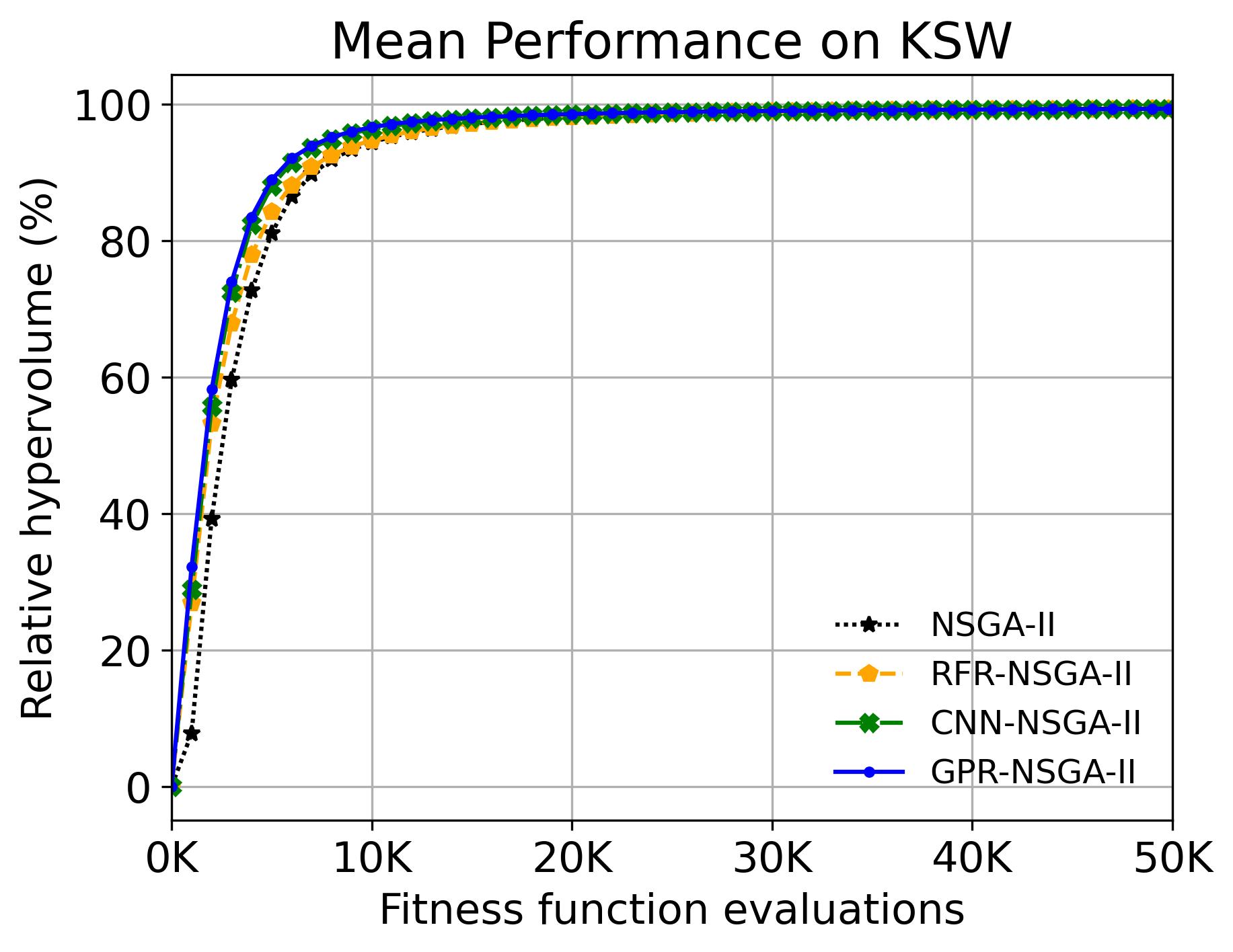}
    \end{subfigure}
    \begin{subfigure}[t]{\figwid\textwidth}
        \centering
        \includegraphics[width=\linewidth]{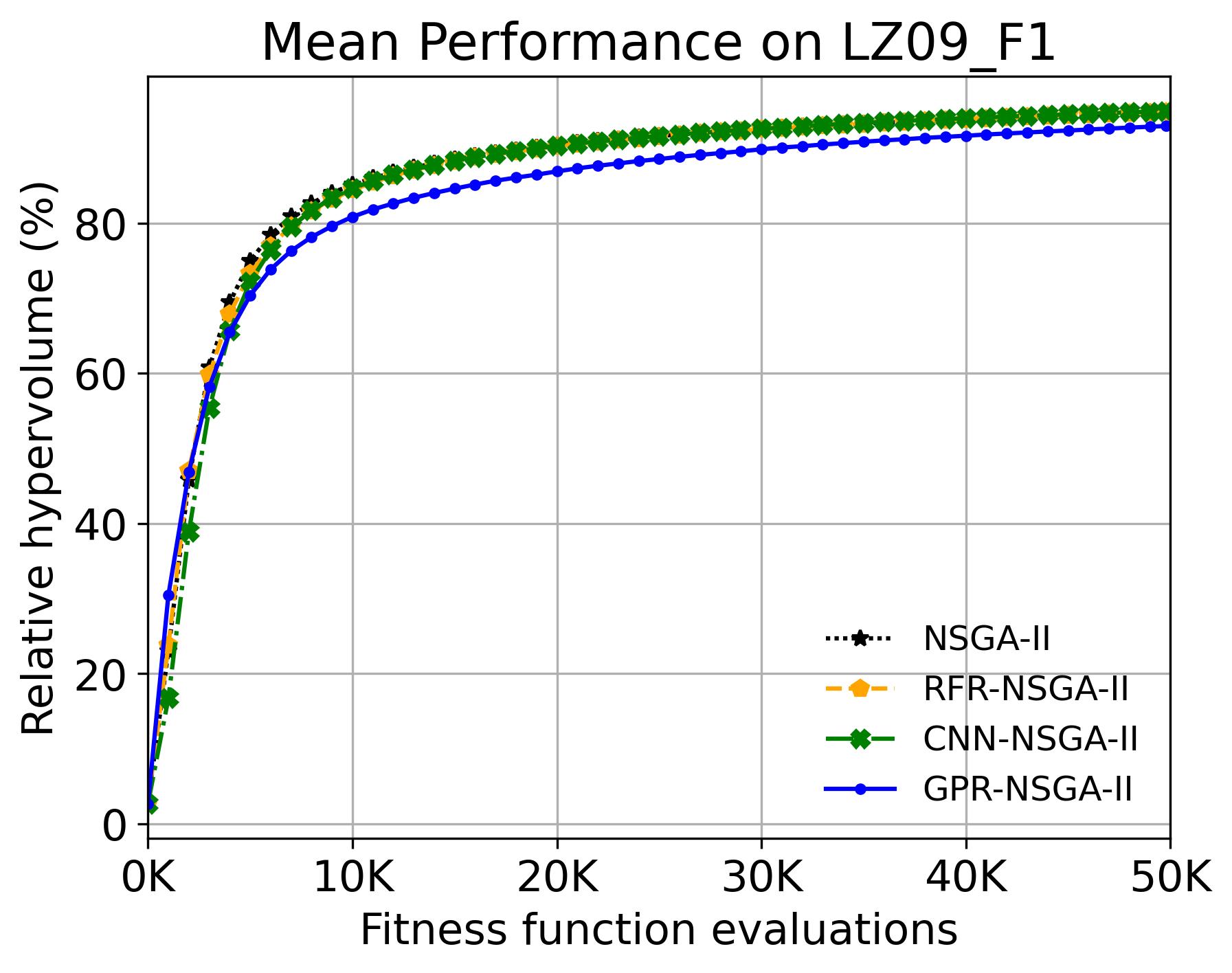}
    \end{subfigure}
    \begin{subfigure}[t]{\figwid\textwidth}
        \centering
        \includegraphics[width=\linewidth]{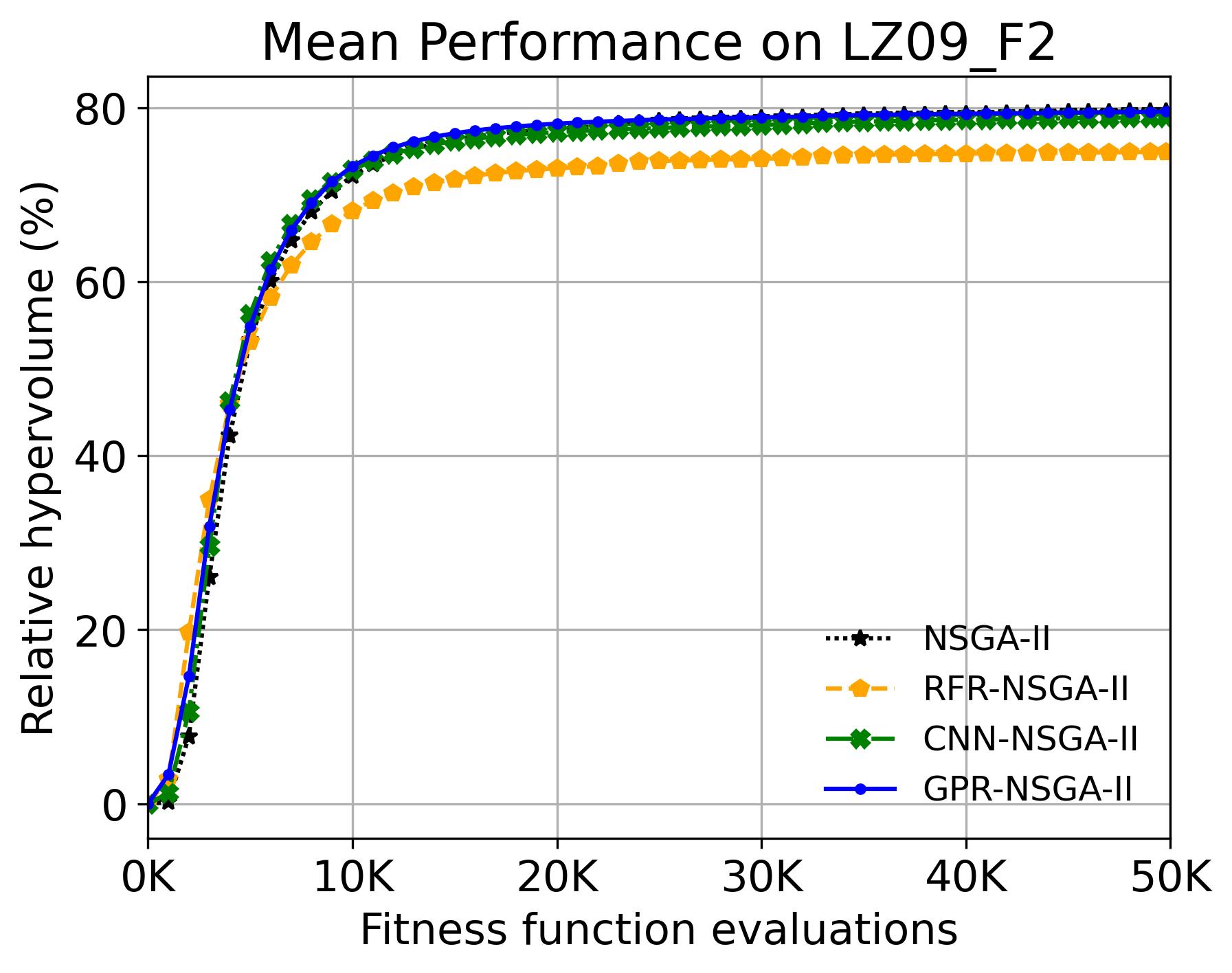}
    \end{subfigure}
    \begin{subfigure}[t]{\figwid\textwidth}
        \centering
        \includegraphics[width=\linewidth]{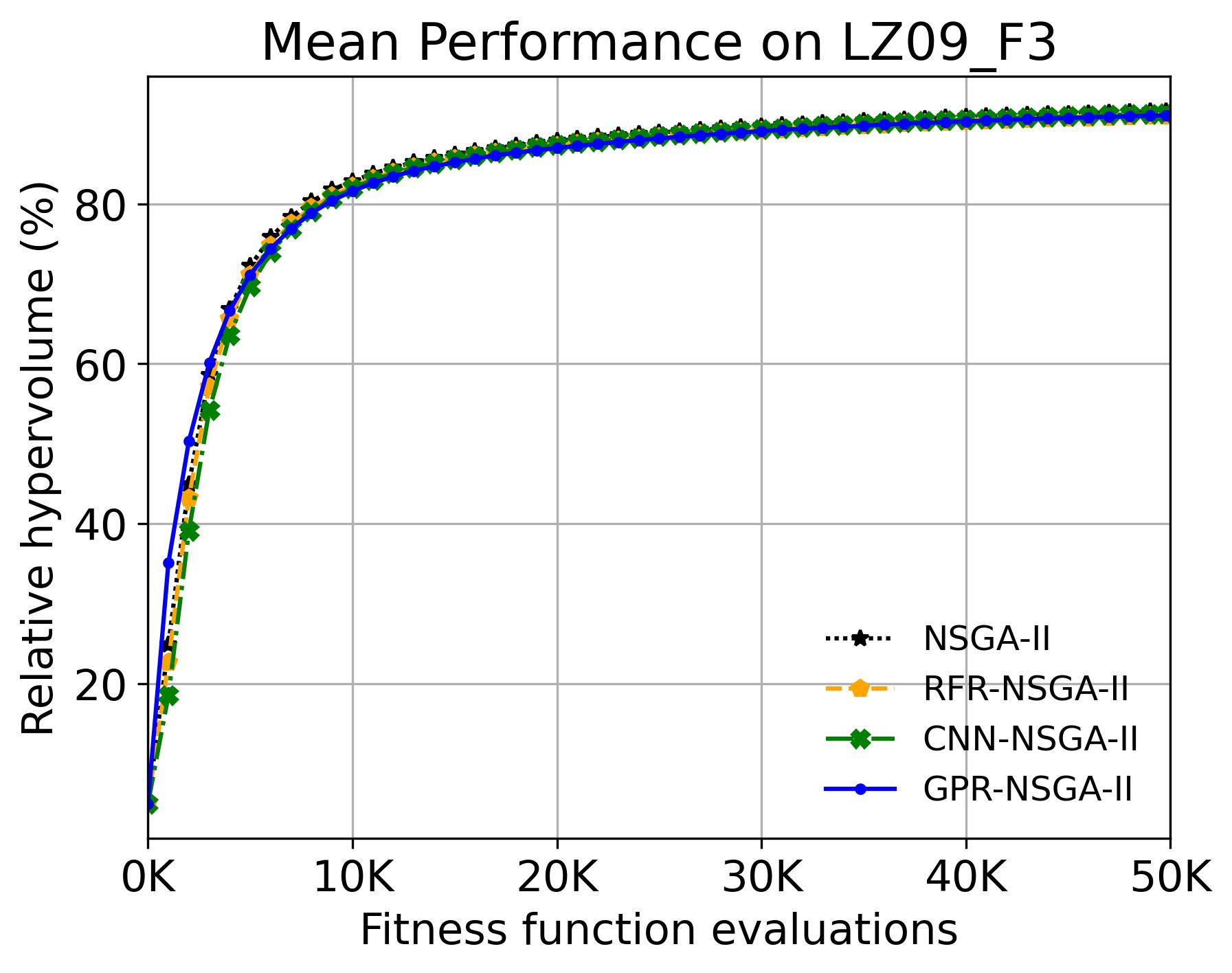}
    \end{subfigure}
    \begin{subfigure}[t]{\figwid\textwidth}
        \centering
        \includegraphics[width=\linewidth]{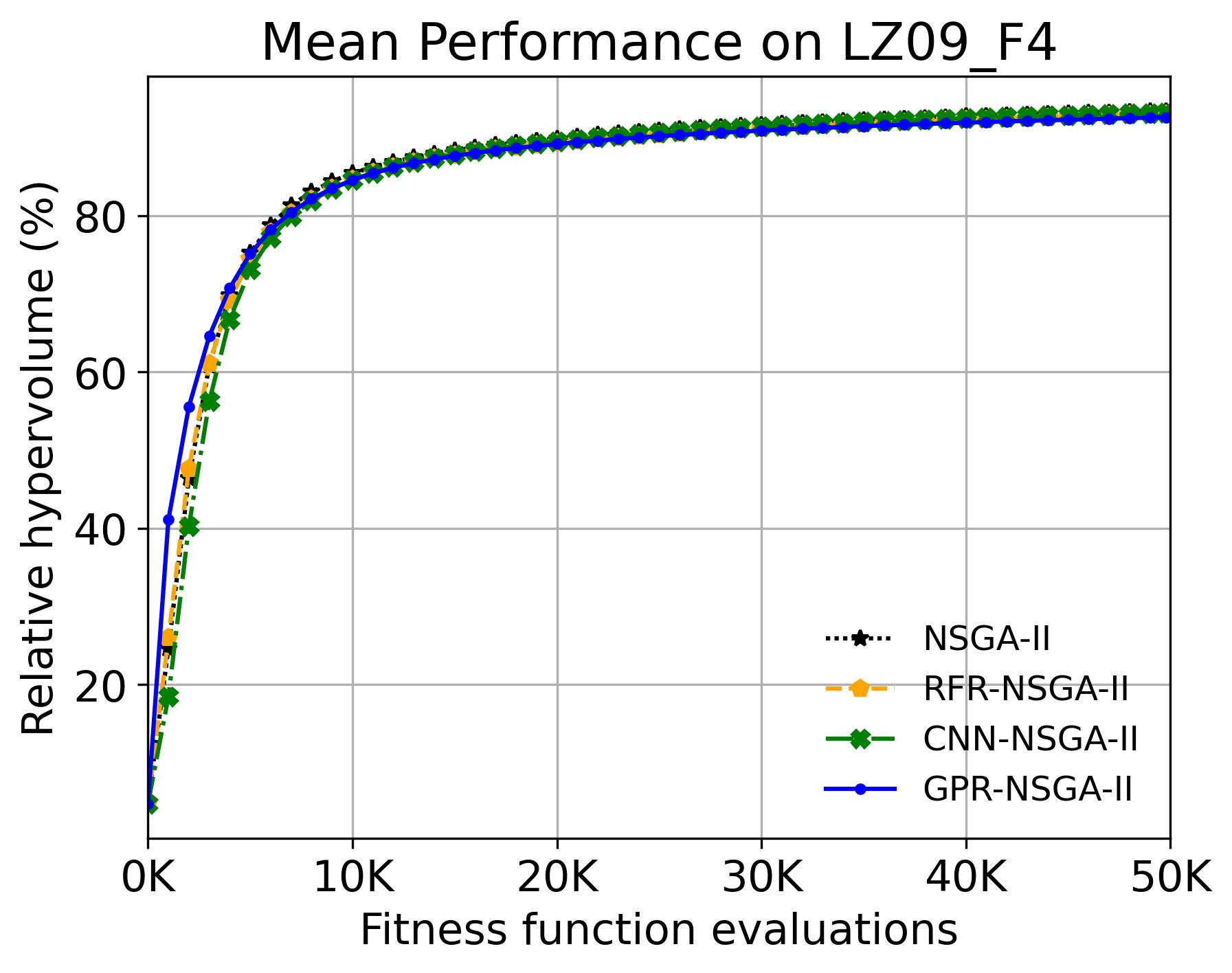}
    \end{subfigure}
    \begin{subfigure}[t]{\figwid\textwidth}
        \centering
        \includegraphics[width=\linewidth]{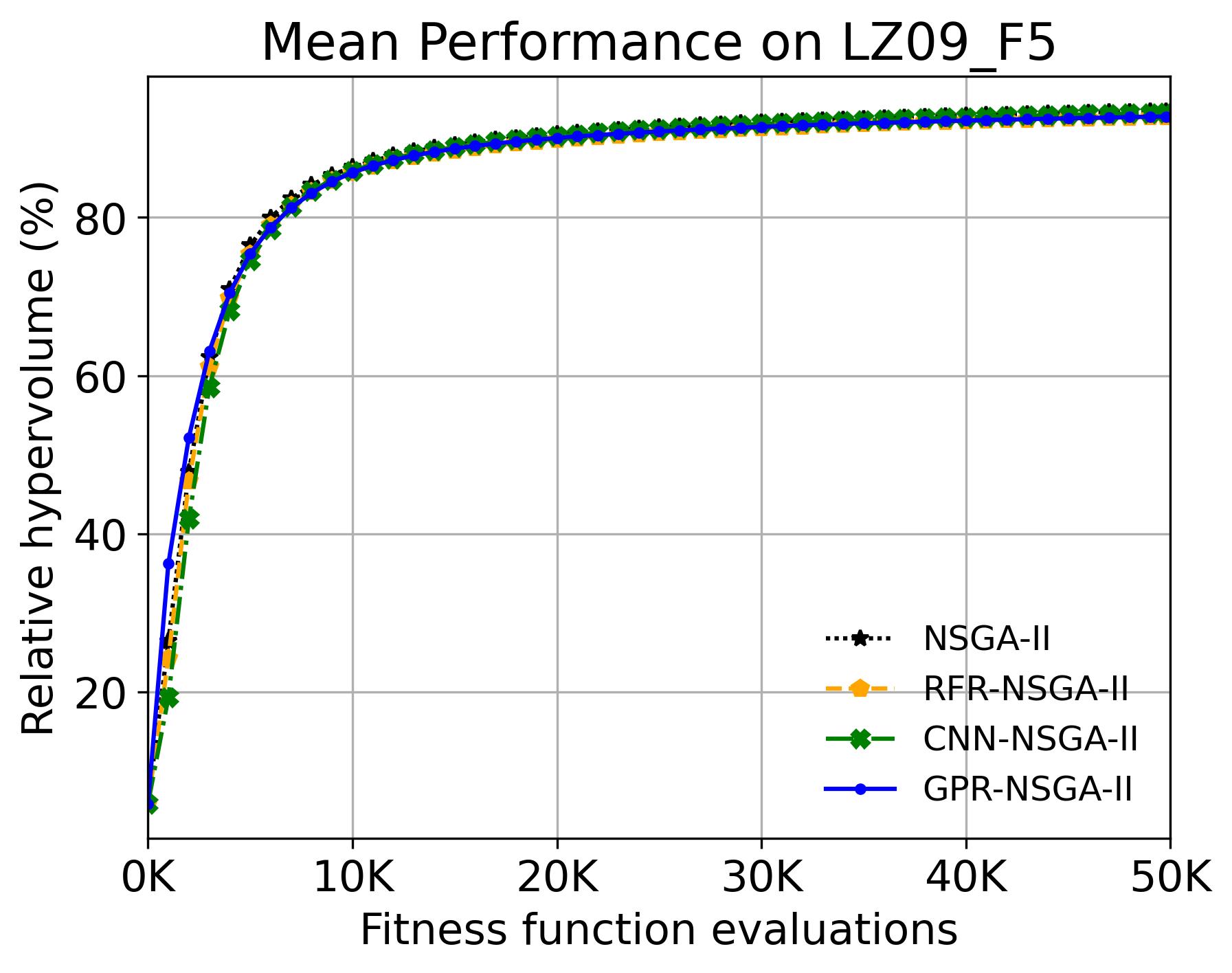}
    \end{subfigure}
    \begin{subfigure}[t]{\figwid\textwidth}
        \centering
        \includegraphics[width=\linewidth]{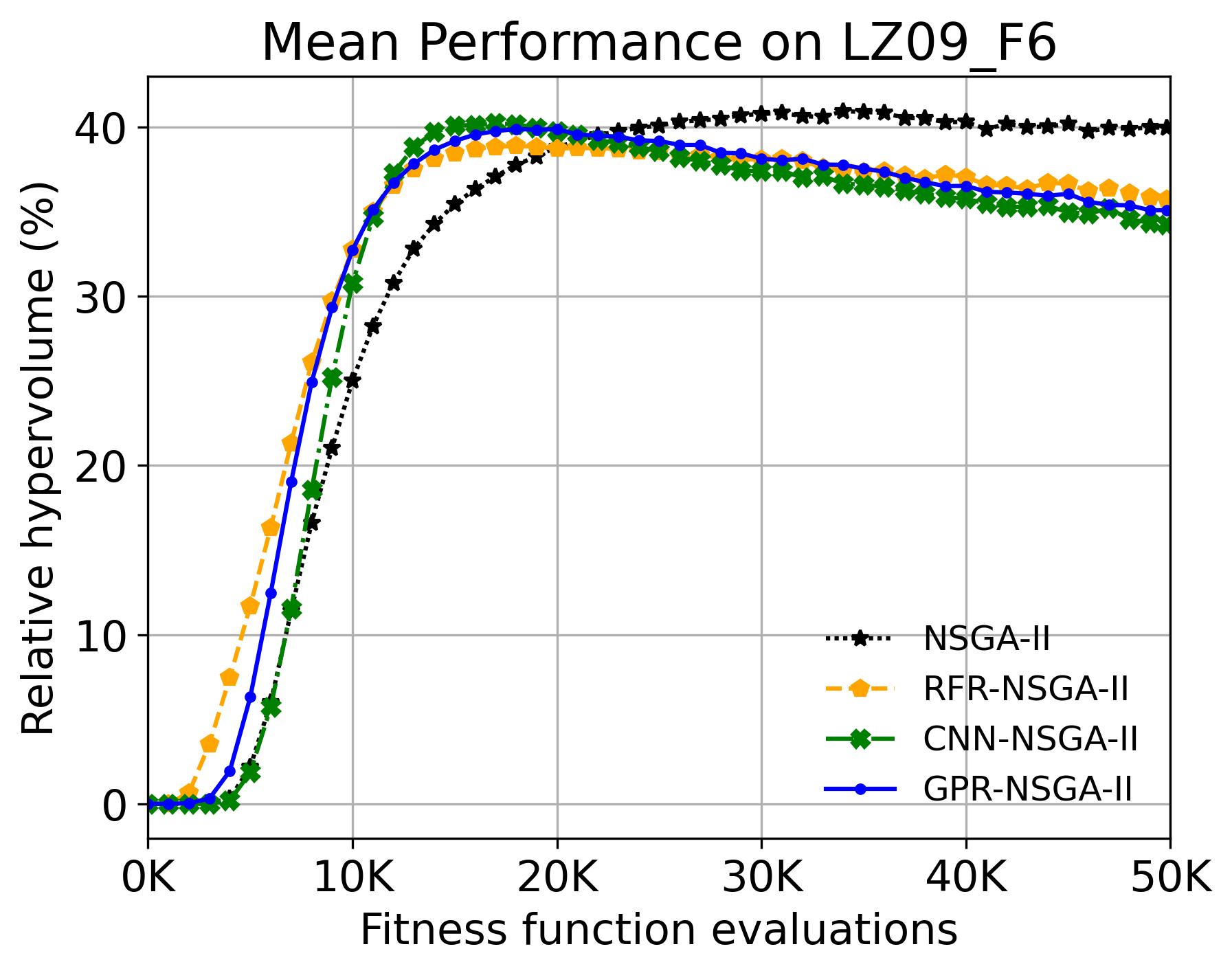}
    \end{subfigure}
    \begin{subfigure}[t]{\figwid\textwidth}
        \centering
        \includegraphics[width=\linewidth]{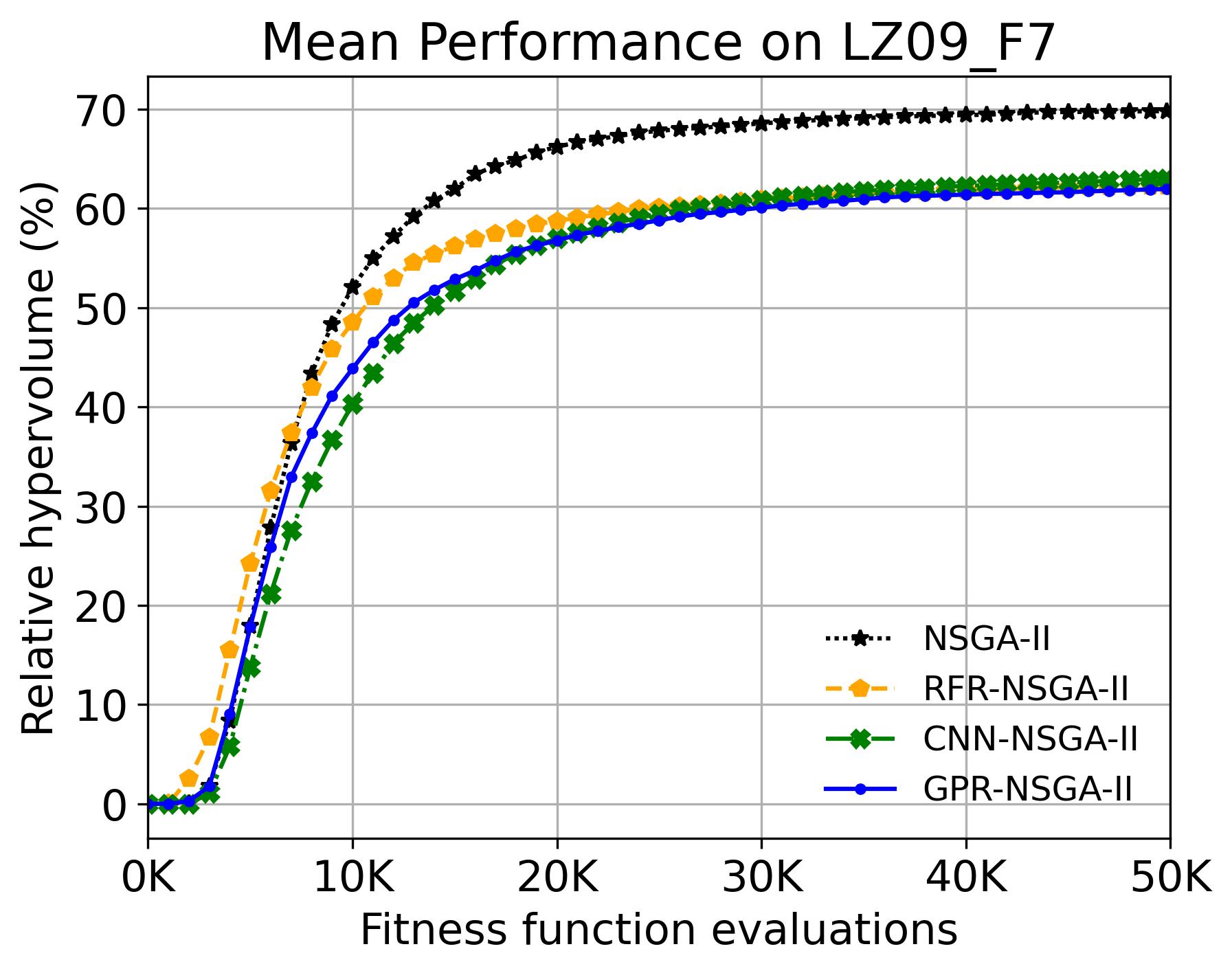}
    \end{subfigure}
    \begin{subfigure}[t]{\figwid\textwidth}
        \centering
        \includegraphics[width=\linewidth]{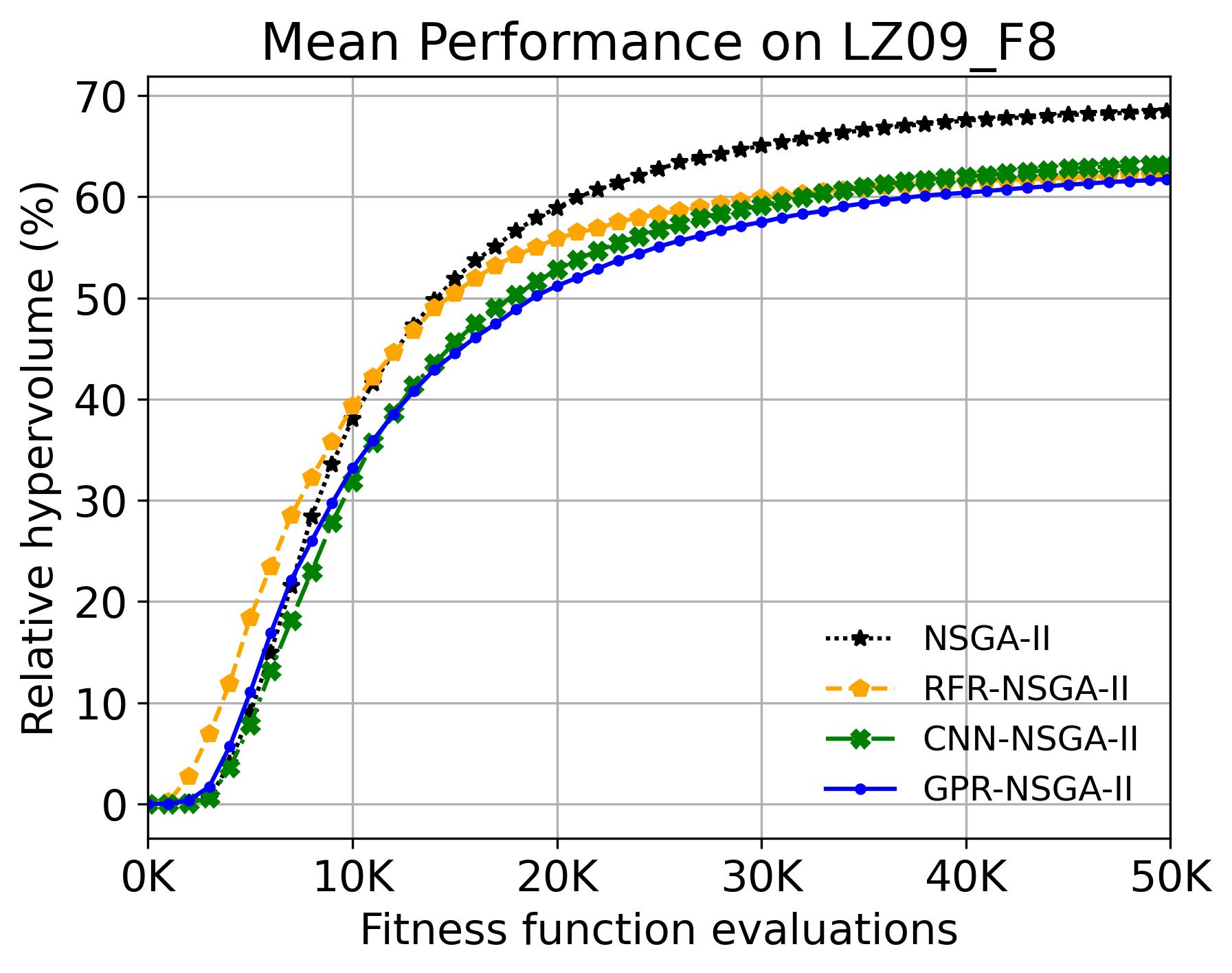}
    \end{subfigure}
    \begin{subfigure}[t]{\figwid\textwidth}
        \centering
        \includegraphics[width=\linewidth]{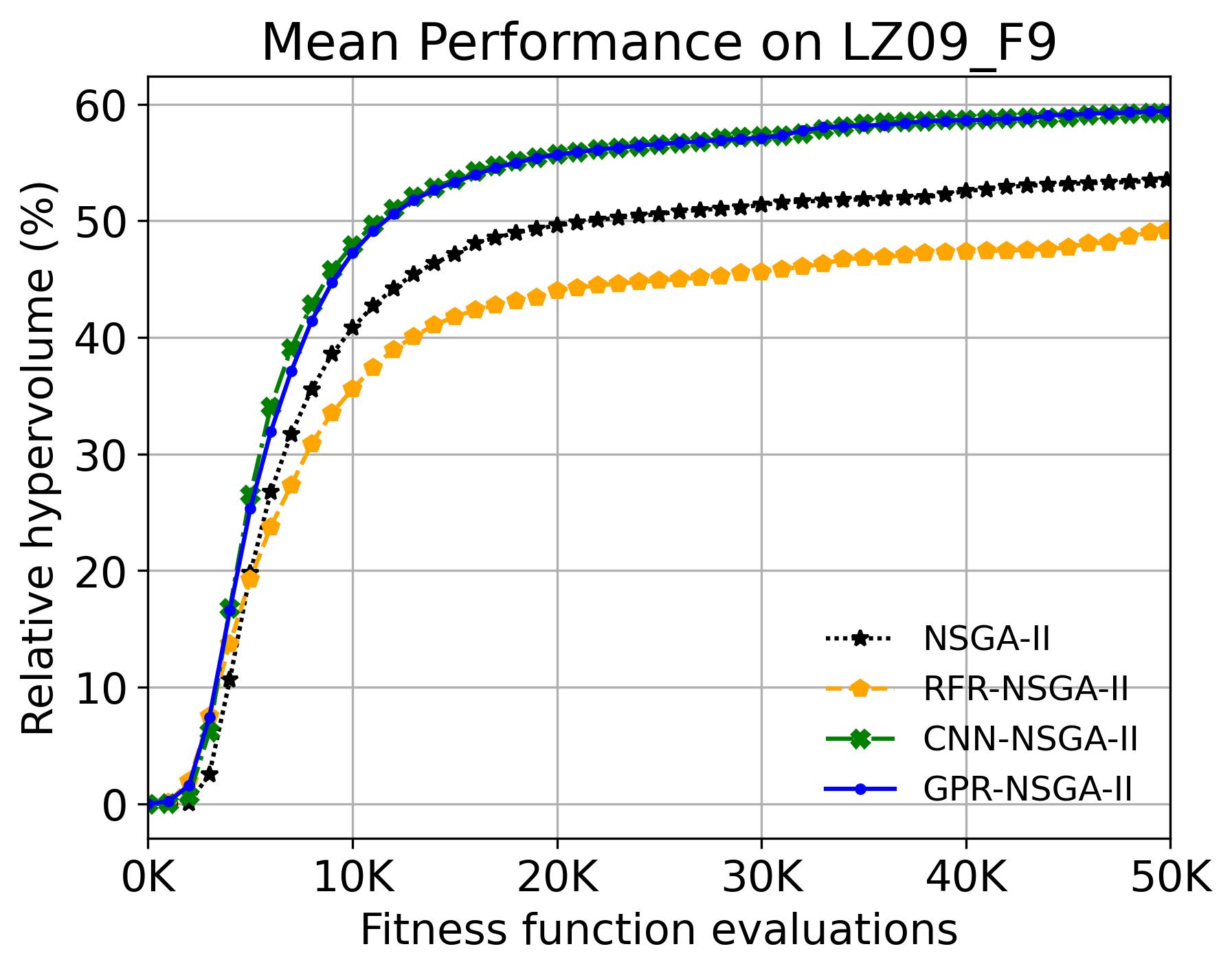}
    \end{subfigure}
    \begin{subfigure}[t]{\figwid\textwidth}
        \centering
        \includegraphics[width=\linewidth]{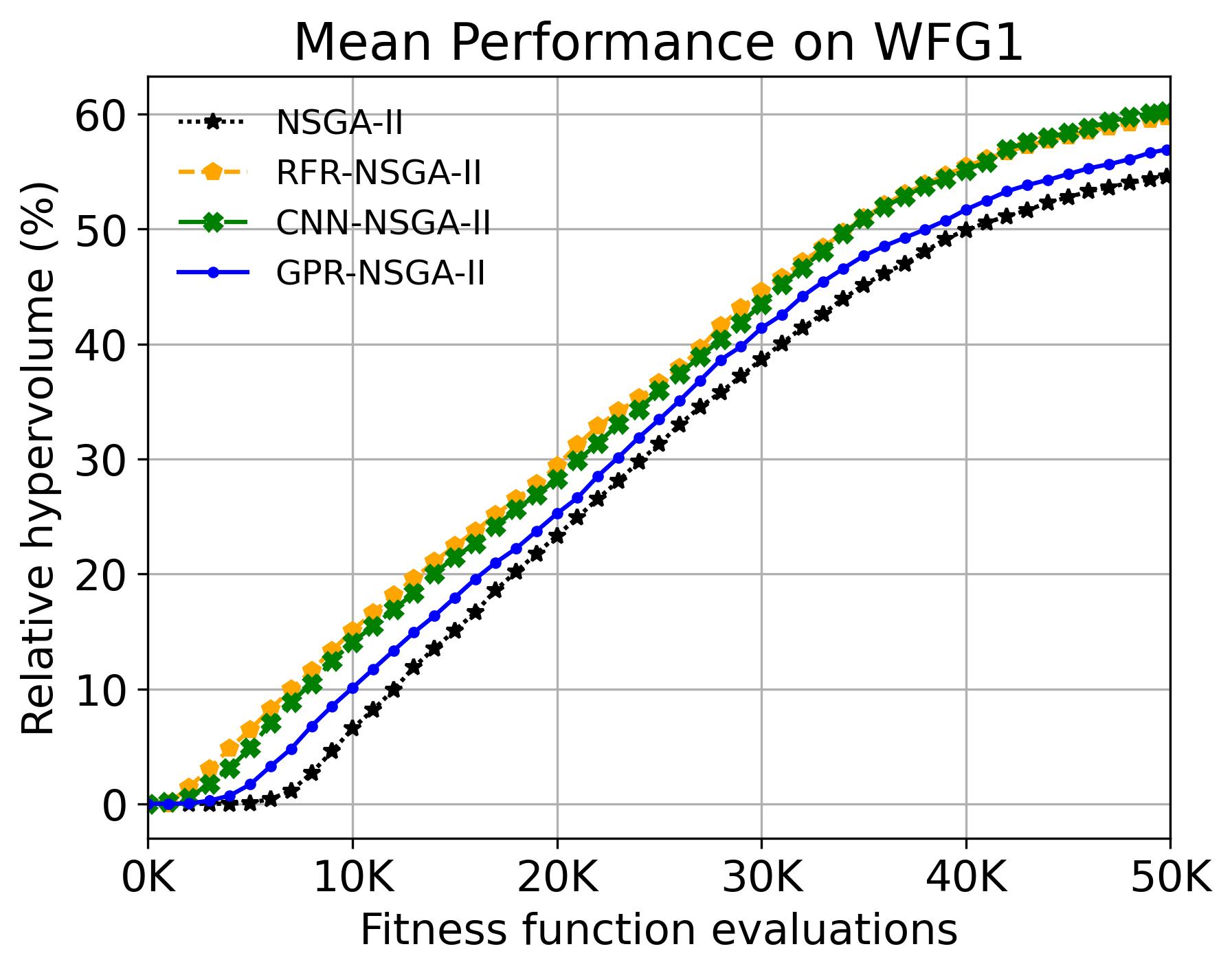}
    \end{subfigure}
    \begin{subfigure}[t]{\figwid\textwidth}
        \centering
        \includegraphics[width=\linewidth]{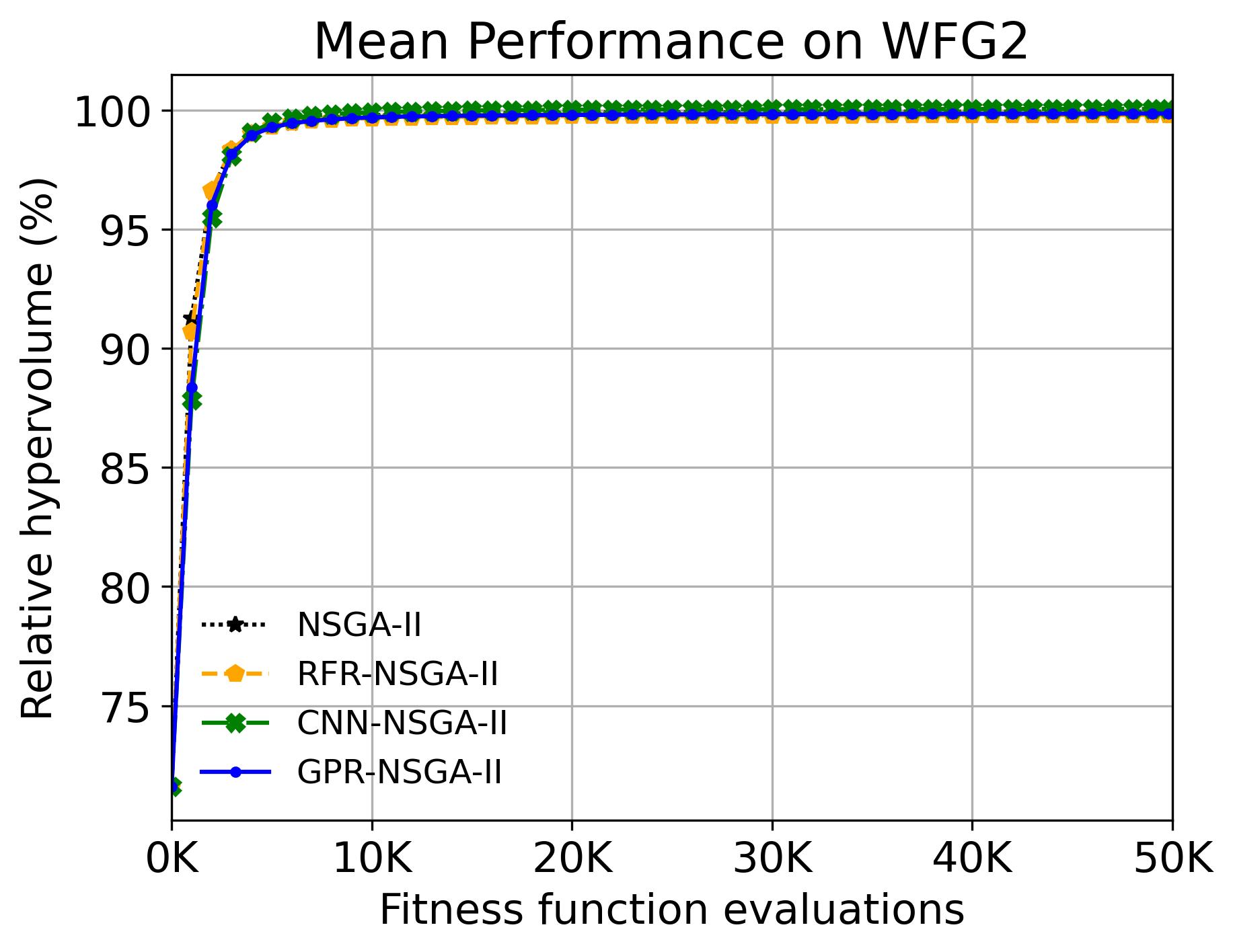}
    \end{subfigure}
    \begin{subfigure}[t]{\figwid\textwidth}
        \centering
        \includegraphics[width=\linewidth]{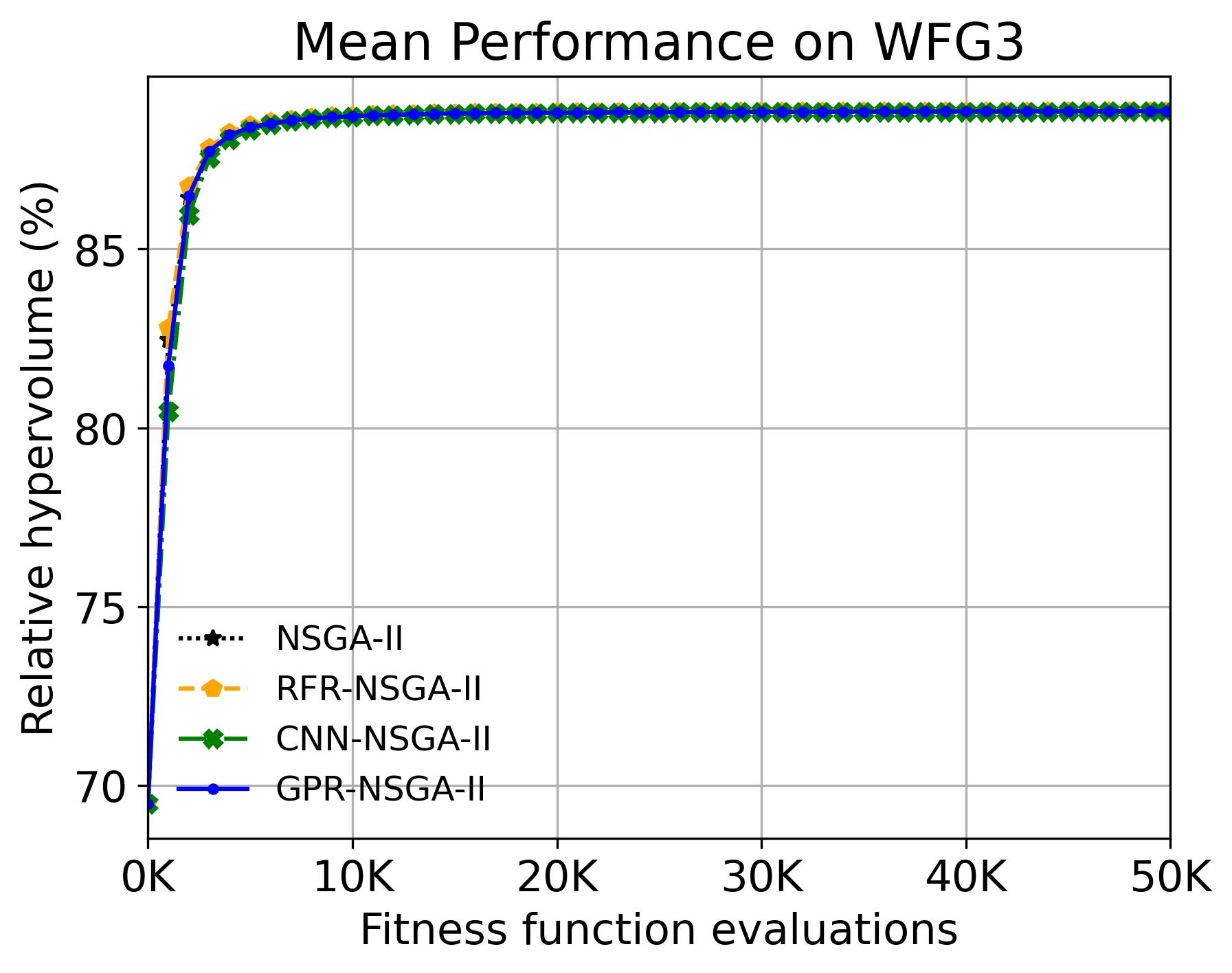}
    \end{subfigure}
    \begin{subfigure}[t]{\figwid\textwidth}
        \centering
        \includegraphics[width=\linewidth]{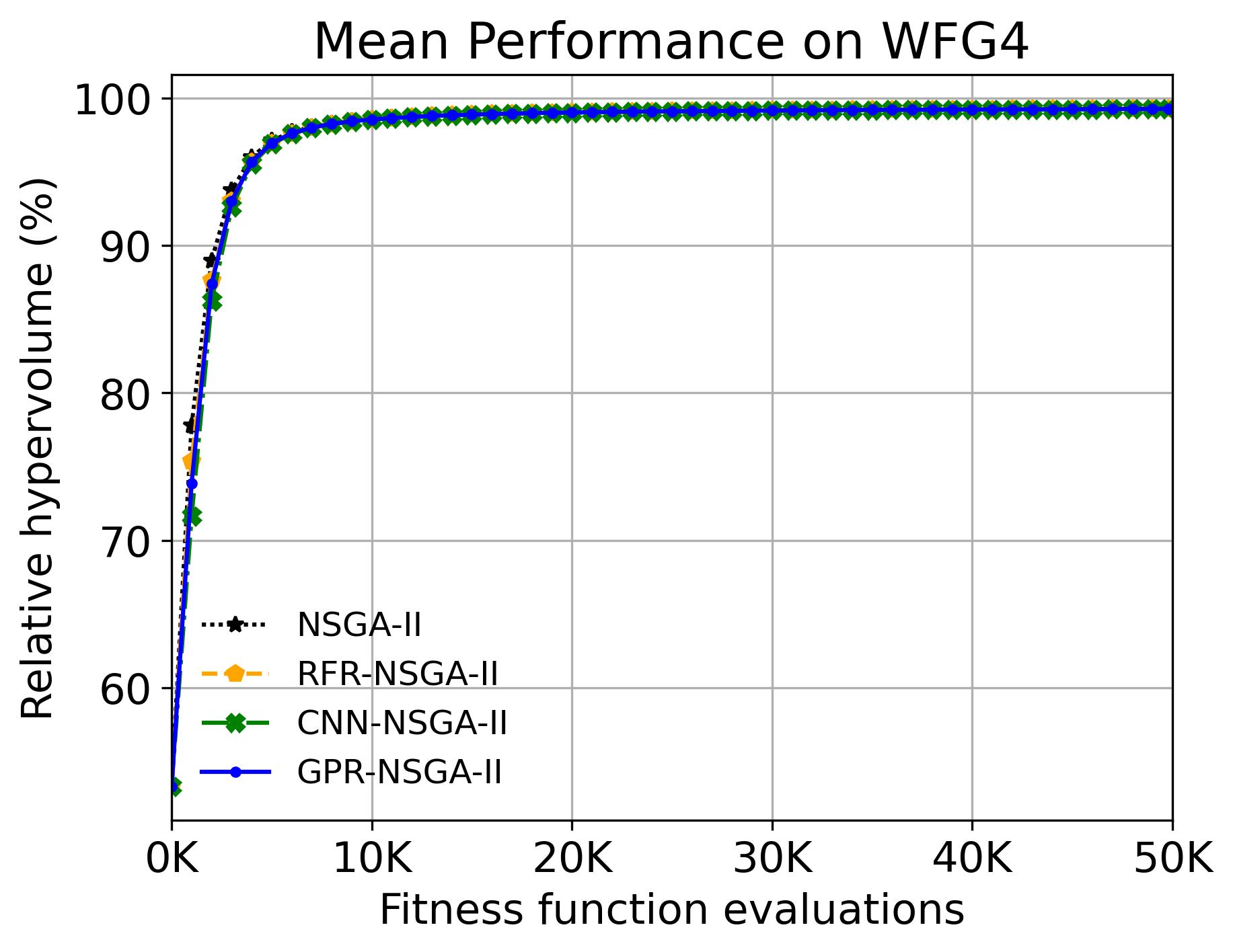}
    \end{subfigure}
    \begin{subfigure}[t]{\figwid\textwidth}
        \centering
        \includegraphics[width=\linewidth]{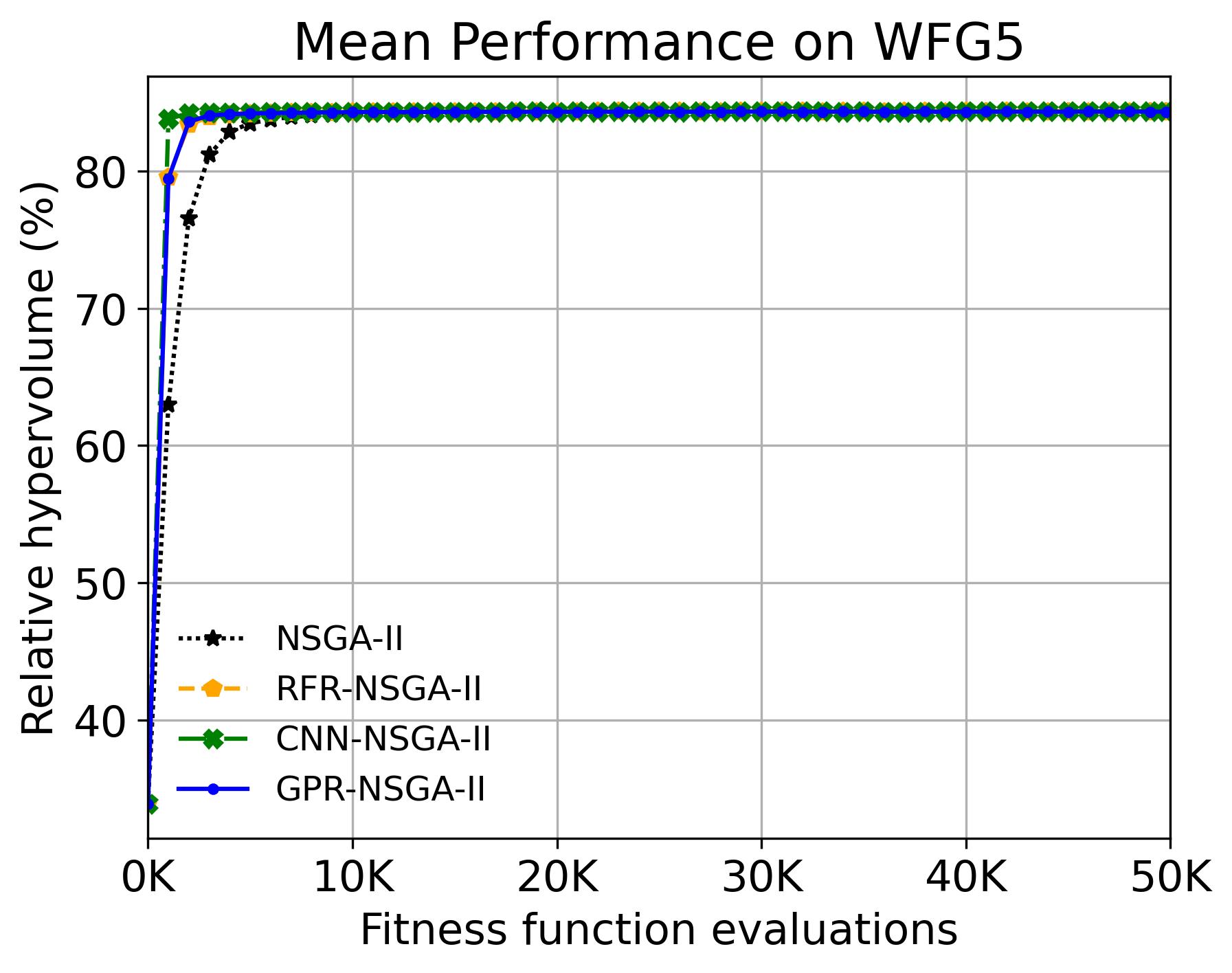}
    \end{subfigure}
    \begin{subfigure}[t]{\figwid\textwidth}
        \centering
        \includegraphics[width=\linewidth]{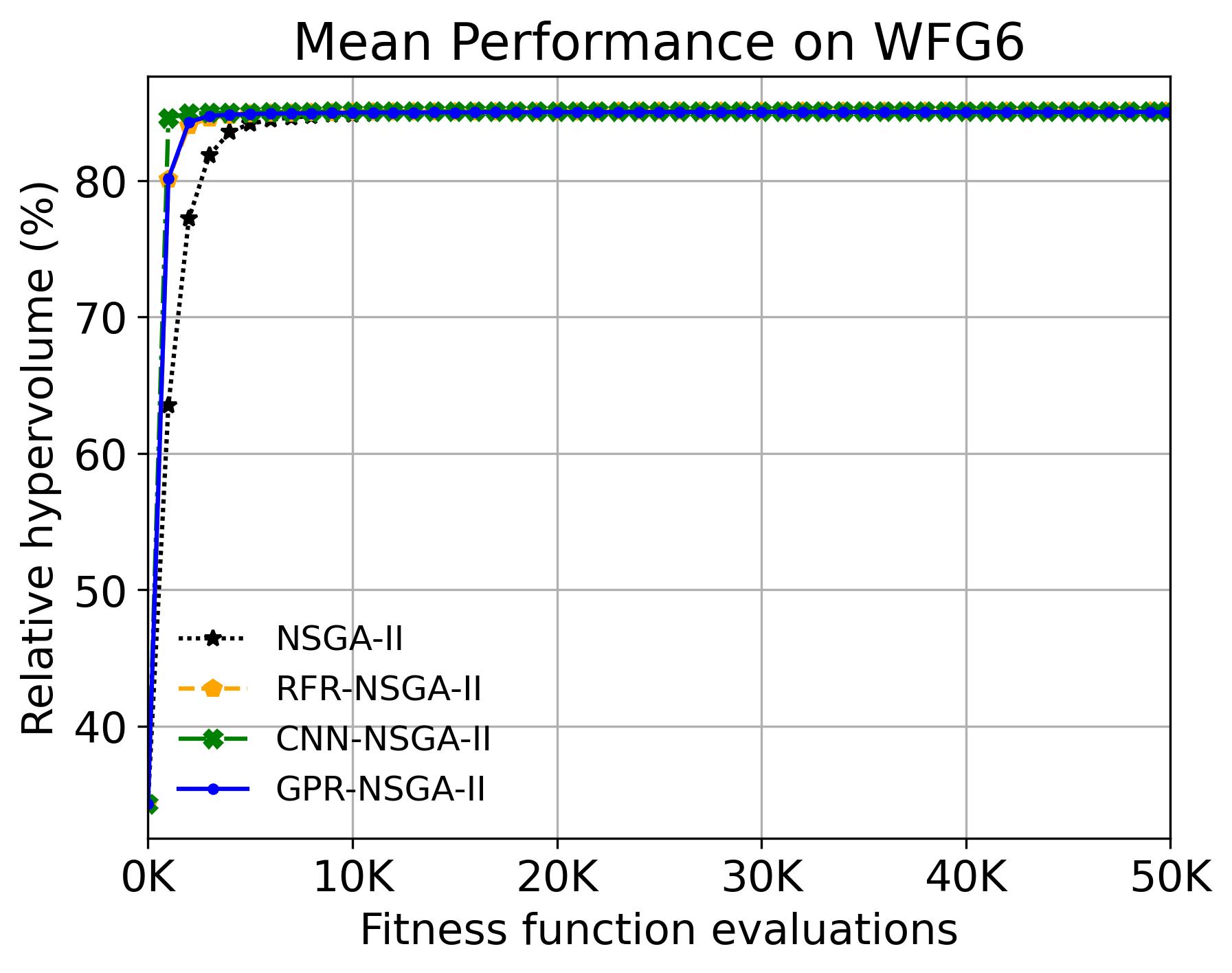}
    \end{subfigure}
    \begin{subfigure}[t]{\figwid\textwidth}
        \centering
        \includegraphics[width=\linewidth]{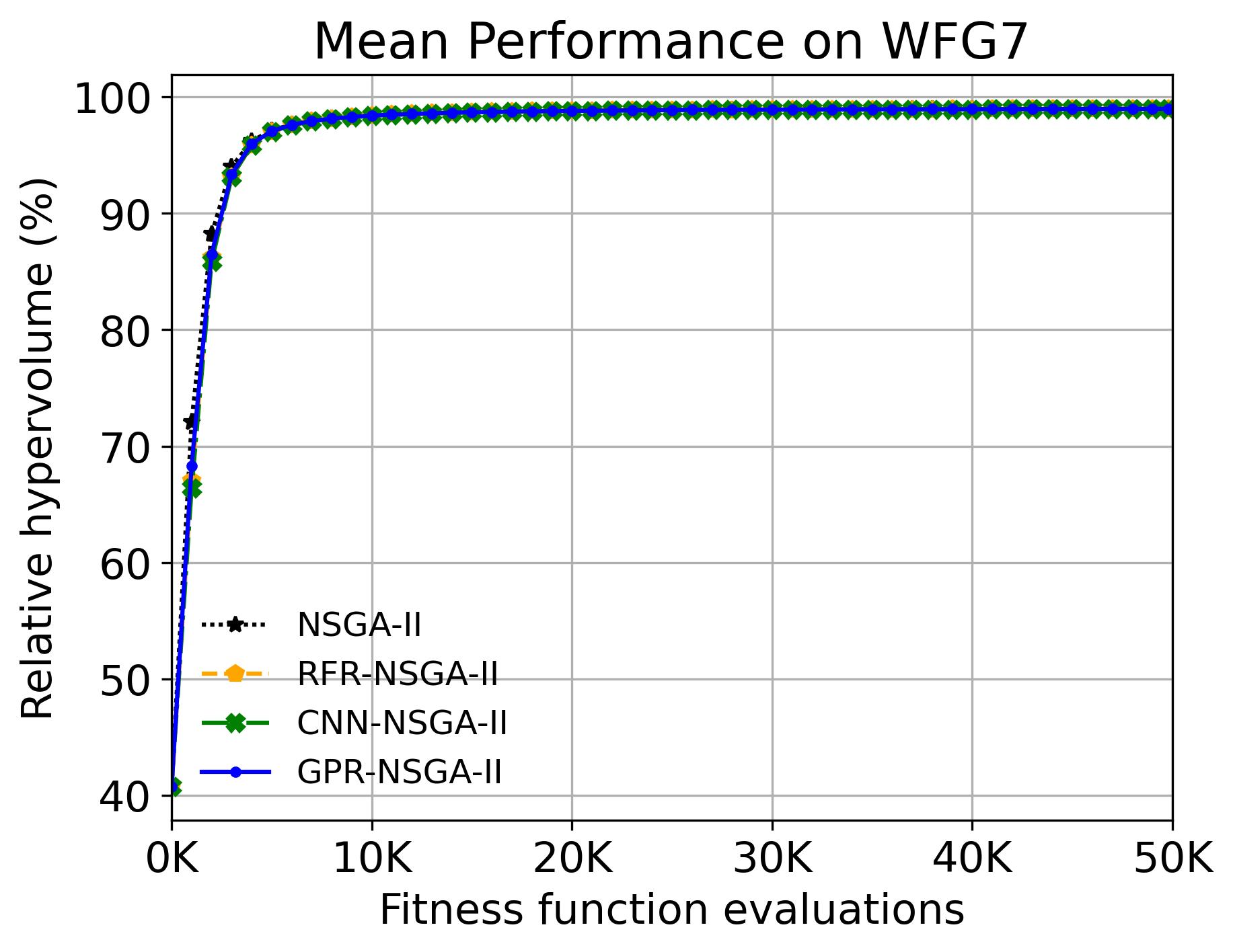}
    \end{subfigure}
    \begin{subfigure}[t]{\figwid\textwidth}
        \centering
        \includegraphics[width=\linewidth]{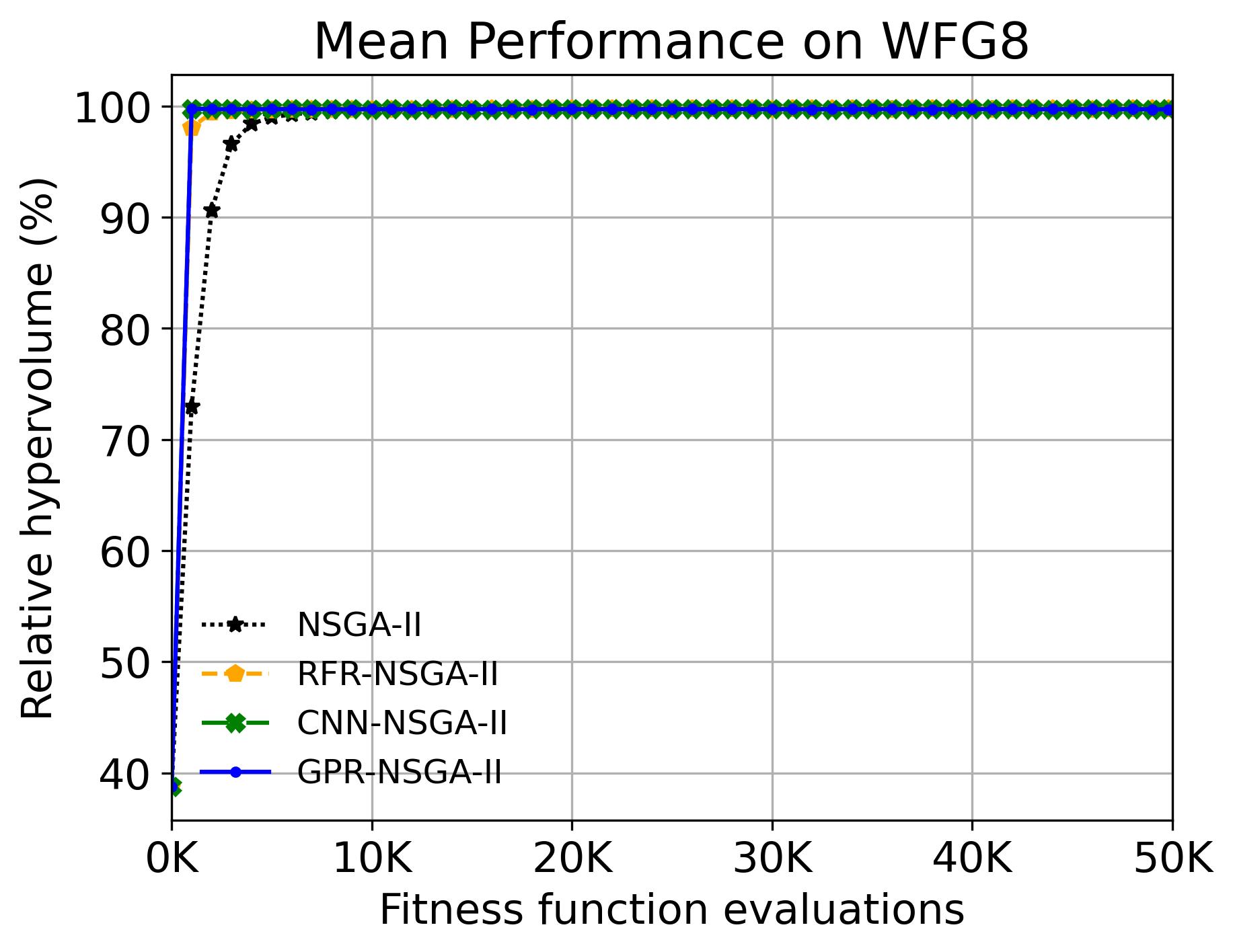}
    \end{subfigure}
    \begin{subfigure}[t]{\figwid\textwidth}
        \centering
        \includegraphics[width=\linewidth]{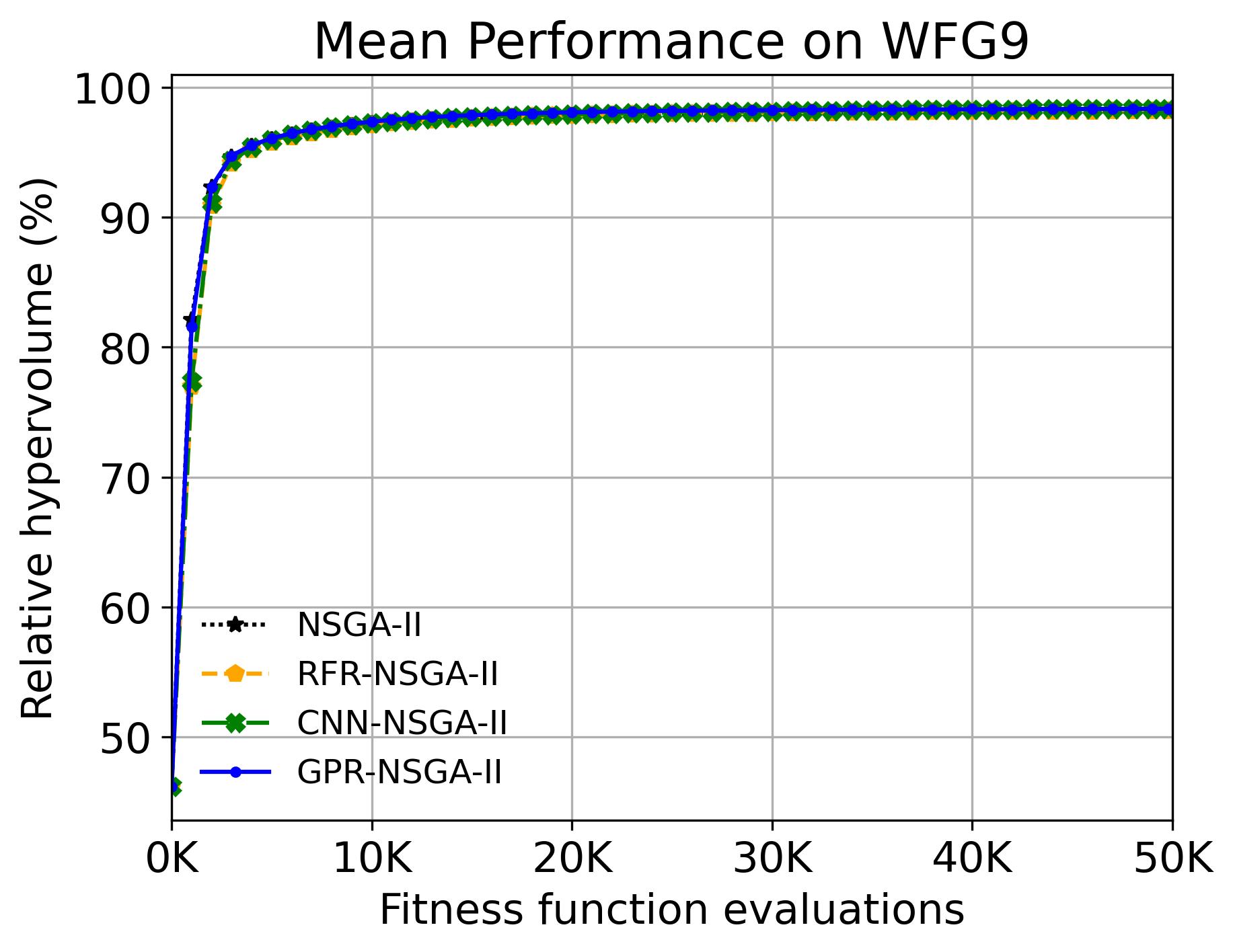}
    \end{subfigure}
    \begin{subfigure}[t]{\figwid\textwidth}
        \centering
        \includegraphics[width=\linewidth]{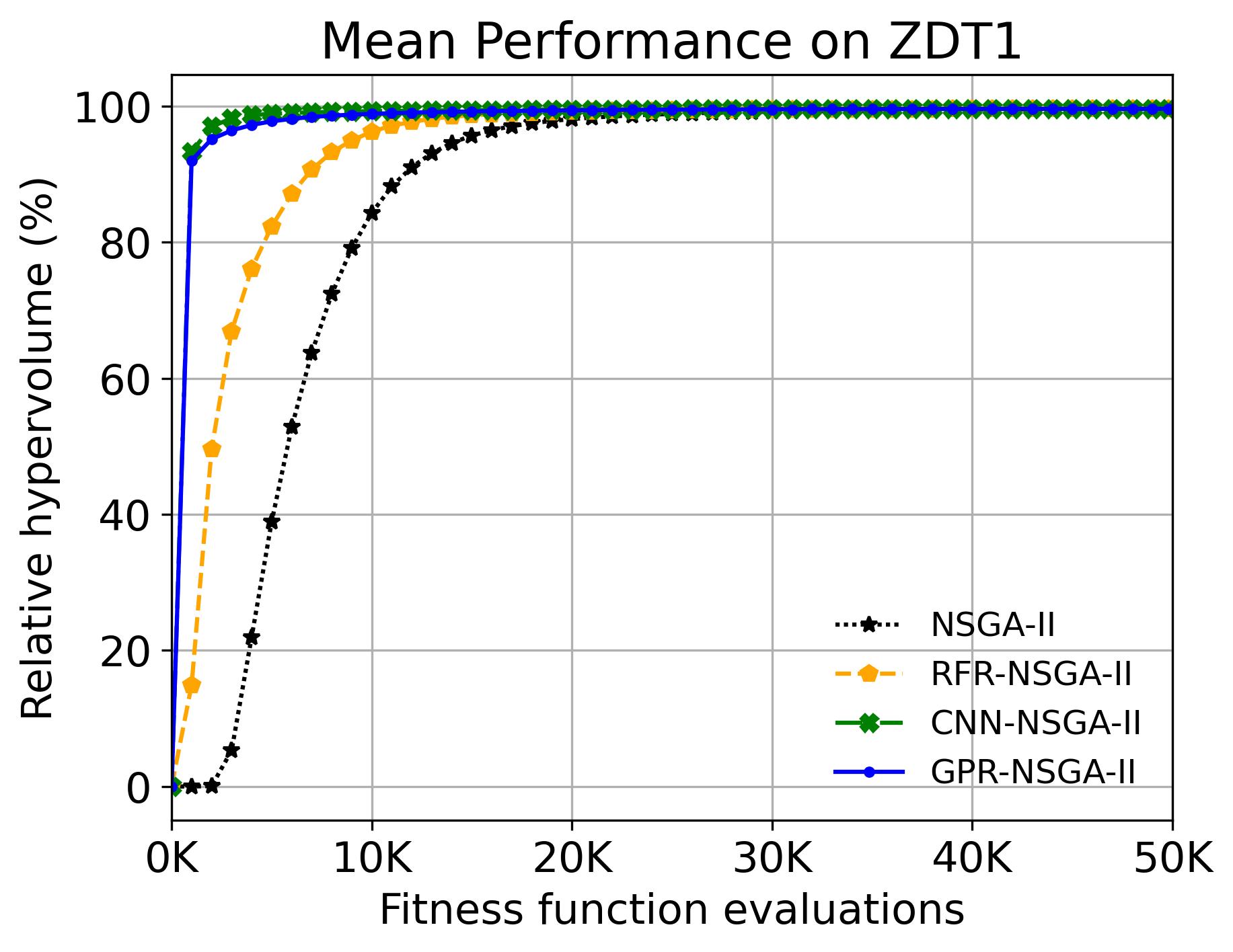}
    \end{subfigure}
    \begin{subfigure}[t]{\figwid\textwidth}
        \centering
        \includegraphics[width=\linewidth]{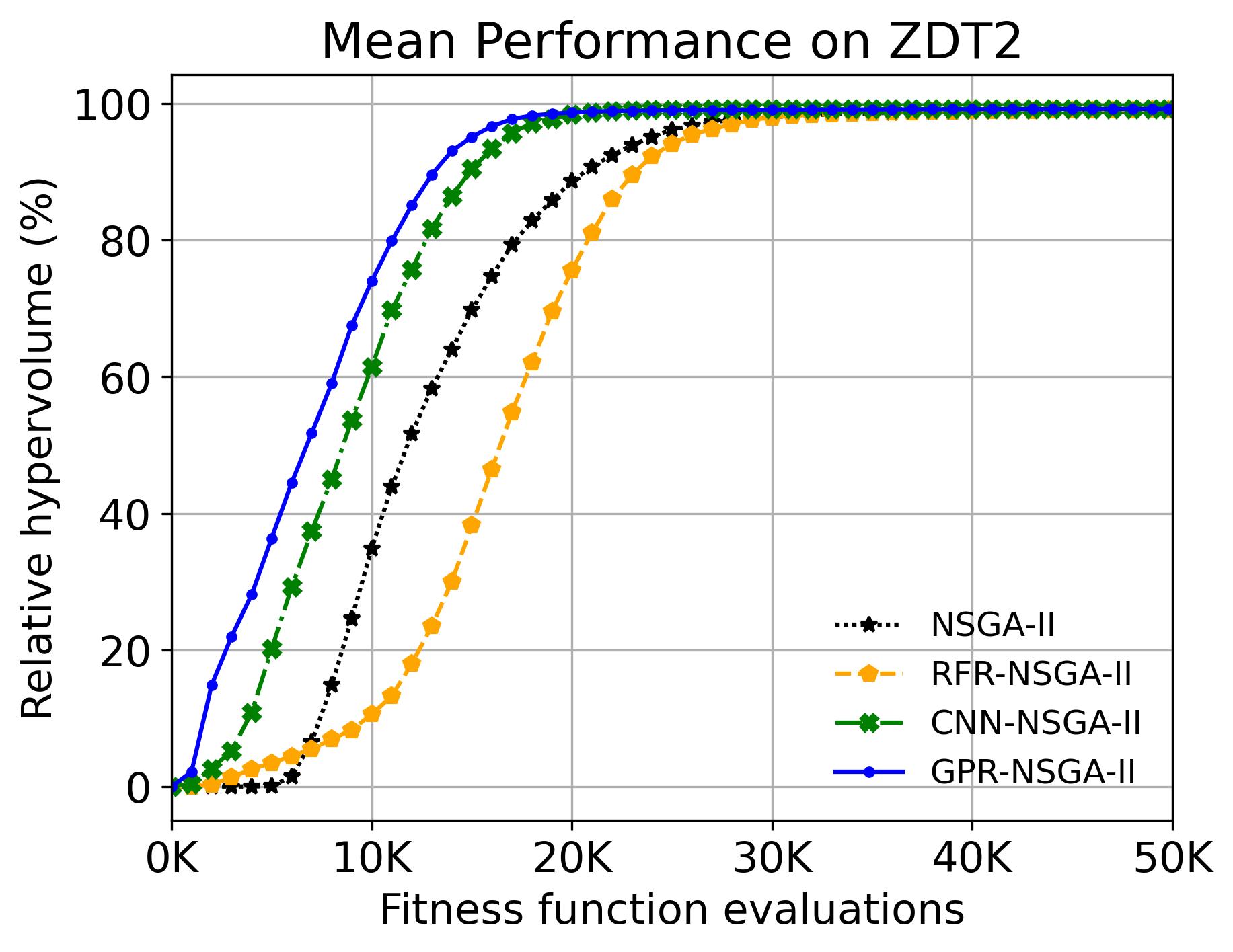}
    \end{subfigure}
    \begin{subfigure}[t]{\figwid\textwidth}
        \centering
        \includegraphics[width=\linewidth]{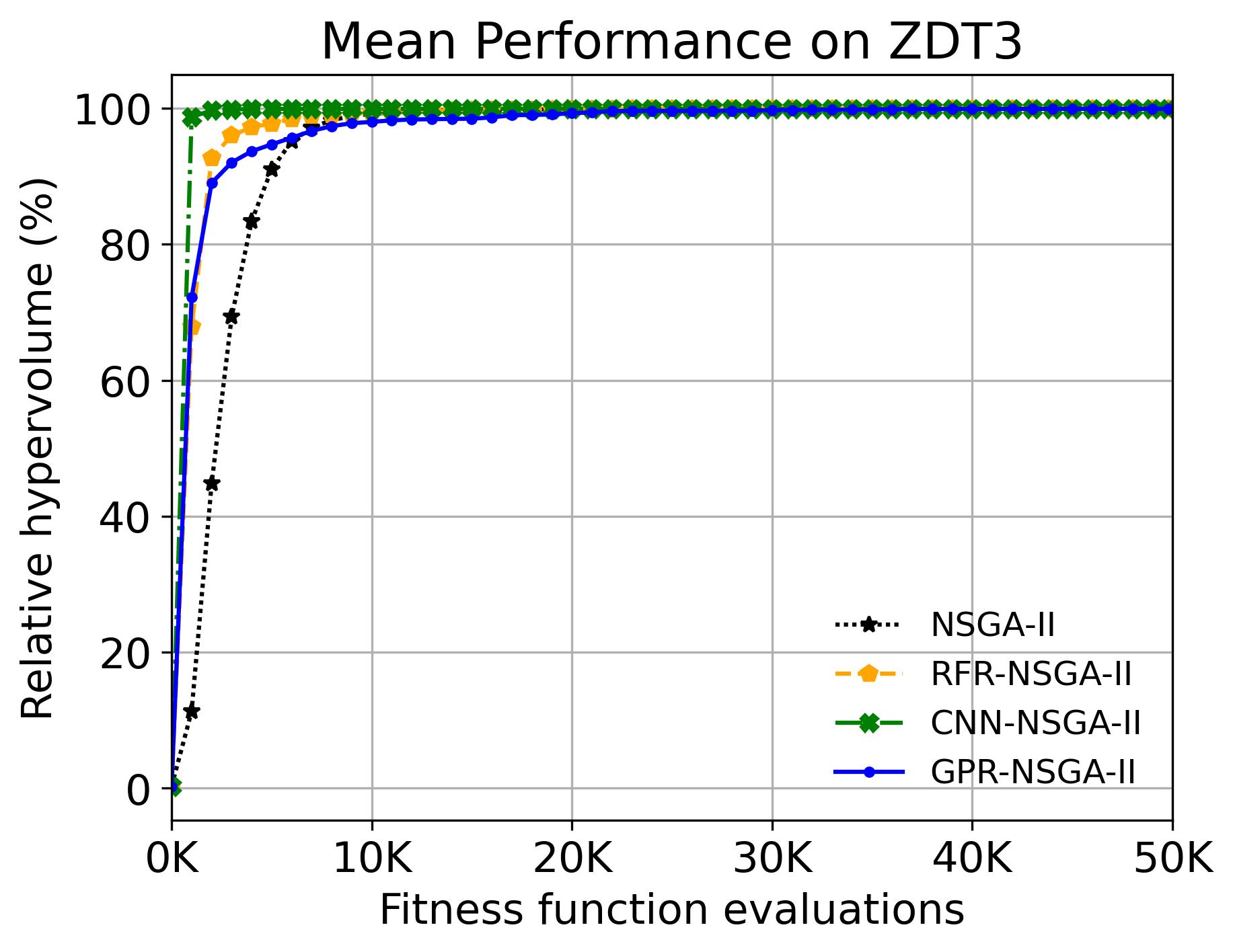}
    \end{subfigure}
    \begin{subfigure}[t]{\figwid\textwidth}
        \centering
        \includegraphics[width=\linewidth]{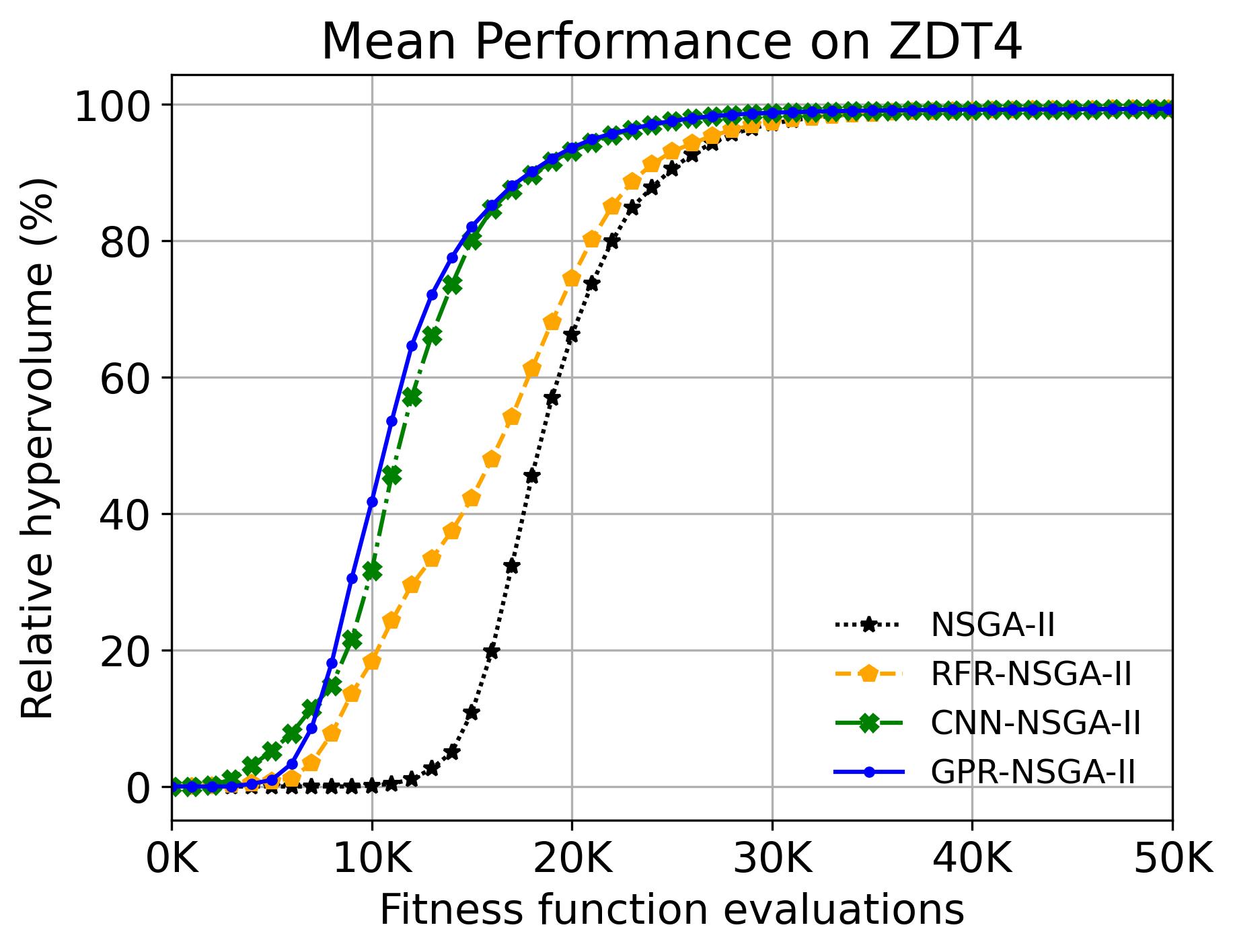}
    \end{subfigure}
    \begin{subfigure}[t]{\figwid\textwidth}
        \centering
        \includegraphics[width=\linewidth]{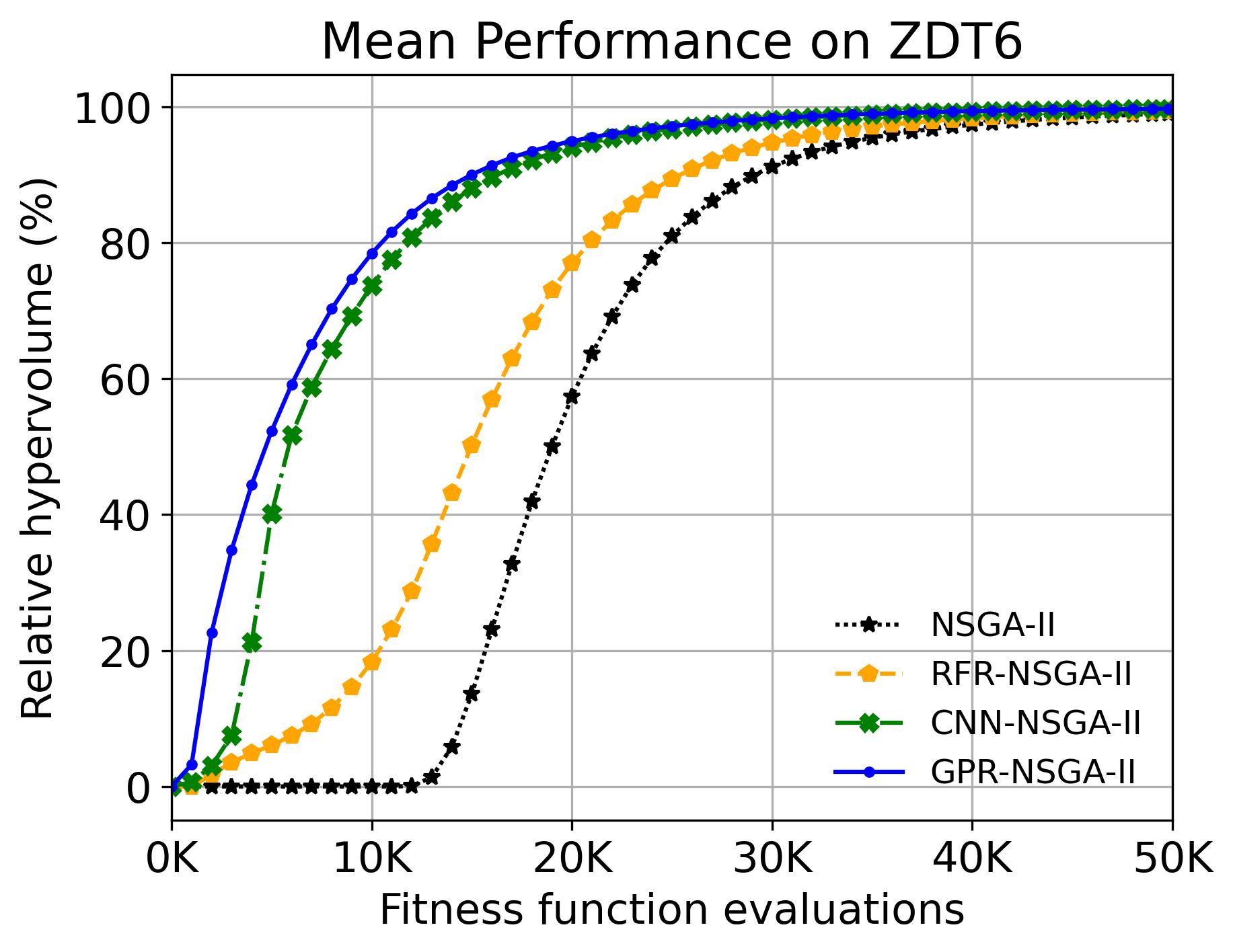}
    \end{subfigure}

        \caption{Comparison of NSGA-II and its associated surrogate-enhanced solvers on individual benchmark problems using $Hv(PF_c)$ -- i.e., the relative hypervolume -- as a performance indicator..}
        \label{fig:annexNSGAII}
\end{figure*}

\begin{figure*}[h]
    \centering
    \begin{subfigure}[t]{\figwid\textwidth}
        \centering
        \includegraphics[width=\linewidth]{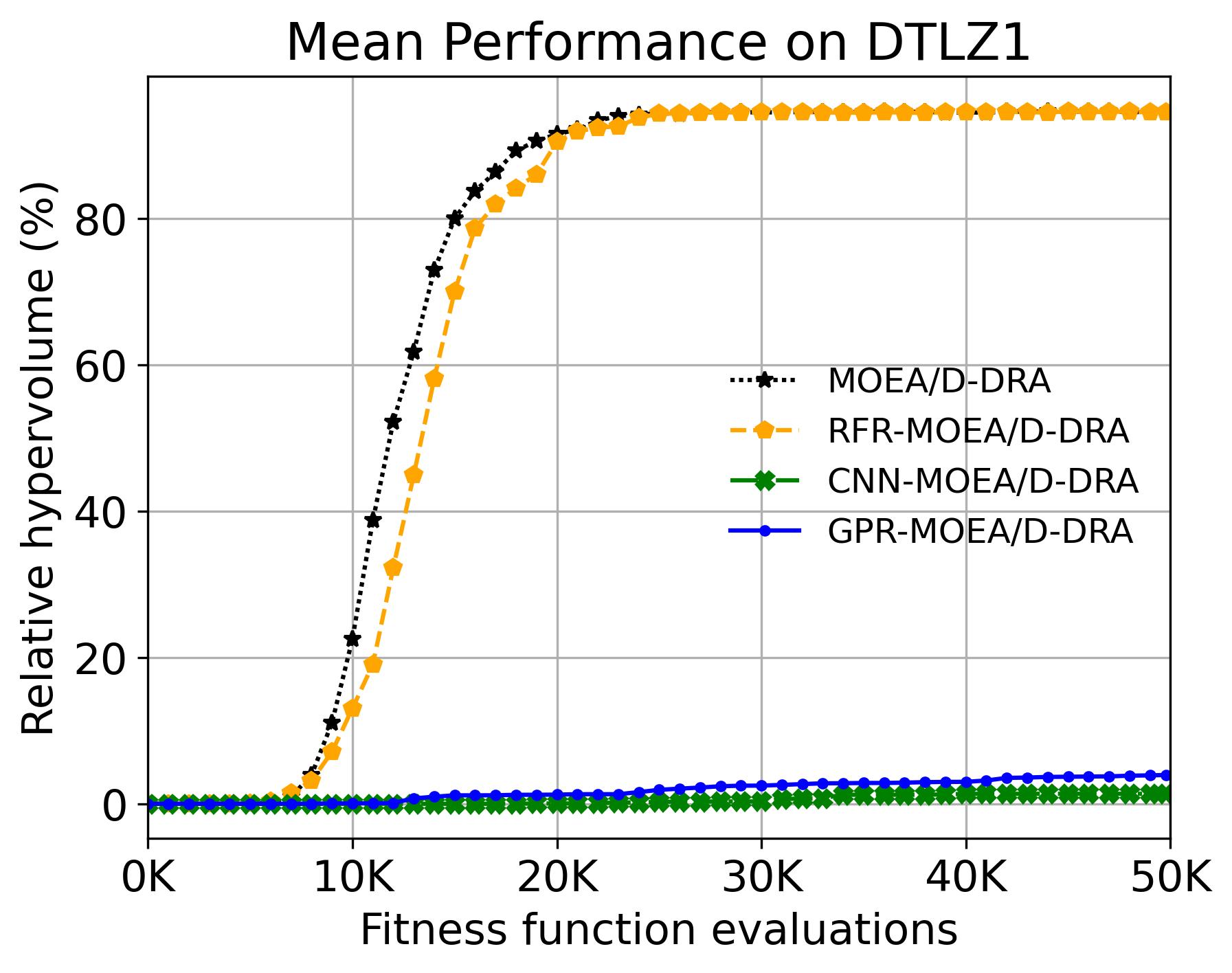}
    \end{subfigure}
    \begin{subfigure}[t]{\figwid\textwidth}
        \centering
        \includegraphics[width=\linewidth]{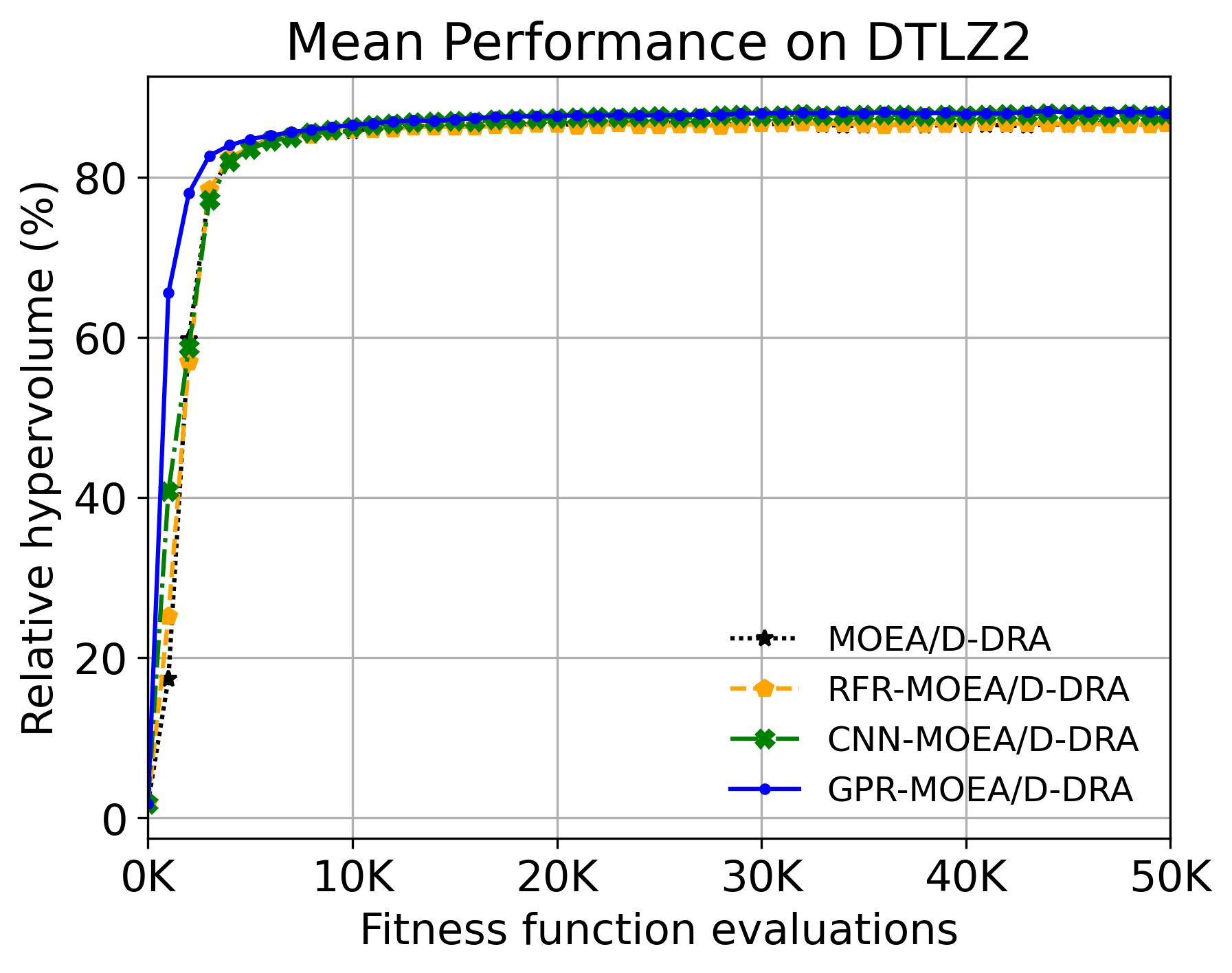}
    \end{subfigure}
    \begin{subfigure}[t]{\figwid\textwidth}
        \centering
        \includegraphics[width=\linewidth]{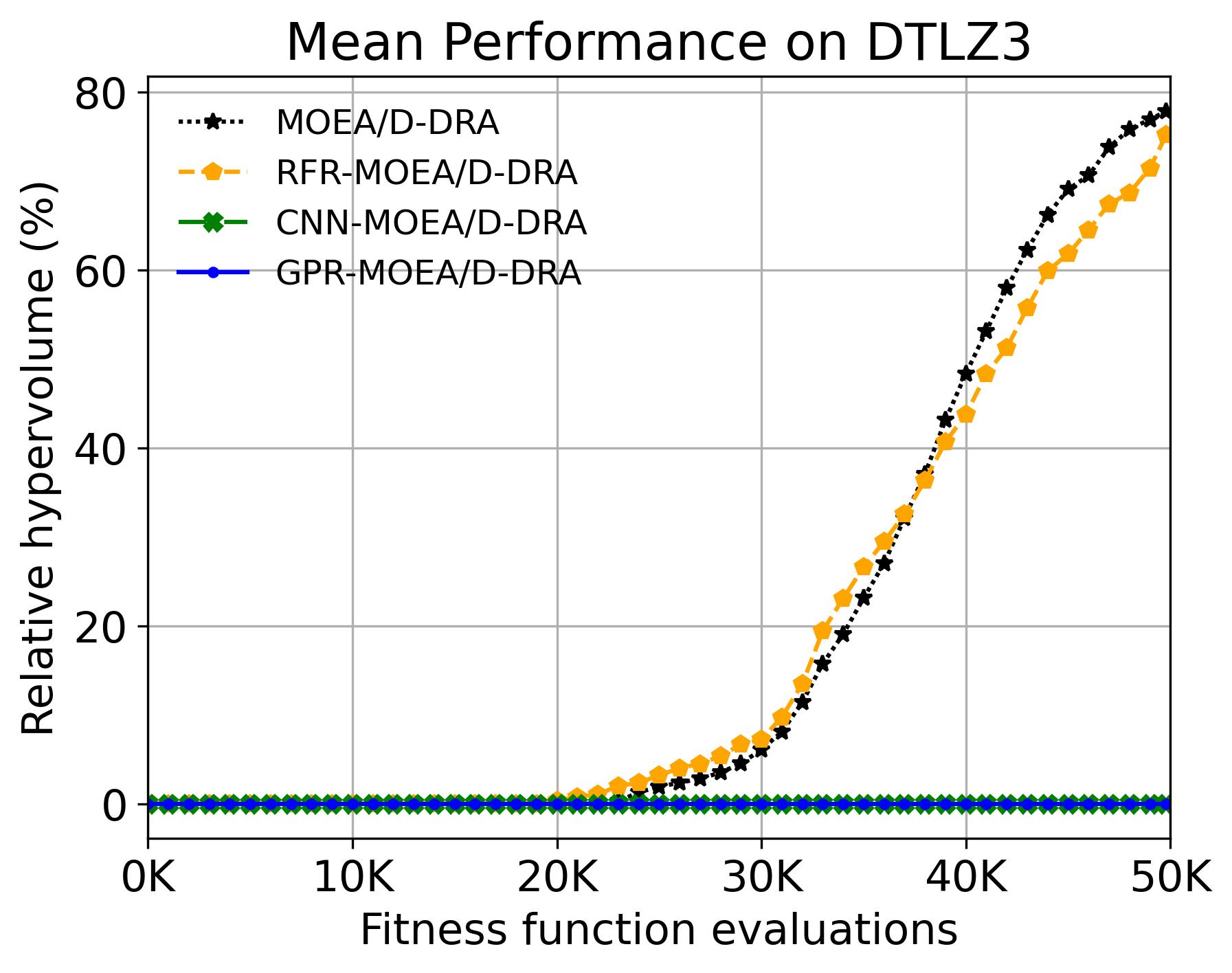}
    \end{subfigure}    
    \begin{subfigure}[t]{\figwid\textwidth}
        \centering
        \includegraphics[width=\linewidth]{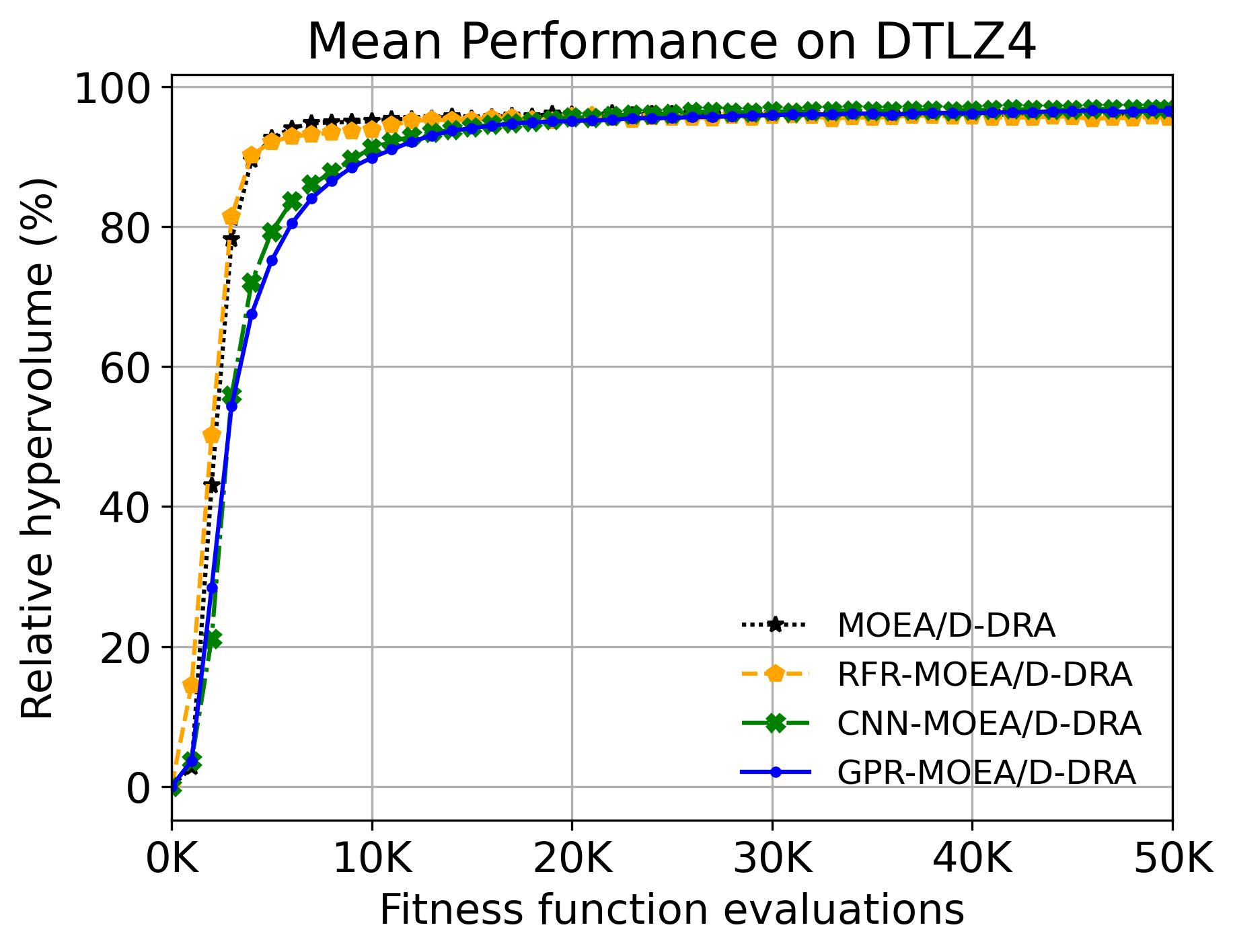}
    \end{subfigure}
    \begin{subfigure}[t]{\figwid\textwidth}
        \centering
        \includegraphics[width=\linewidth]{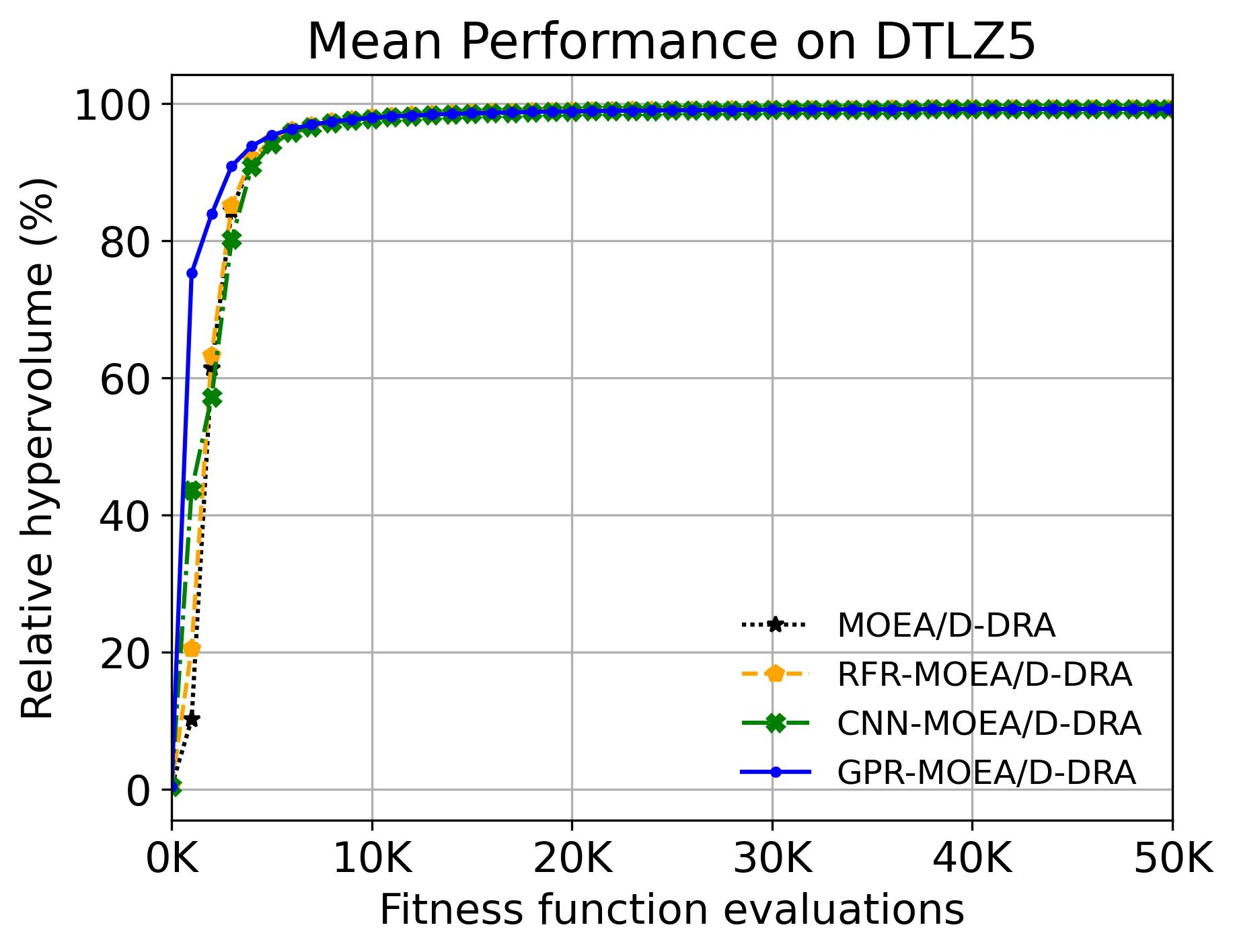}
    \end{subfigure}
    \begin{subfigure}[t]{\figwid\textwidth}
        \centering
        \includegraphics[width=\linewidth]{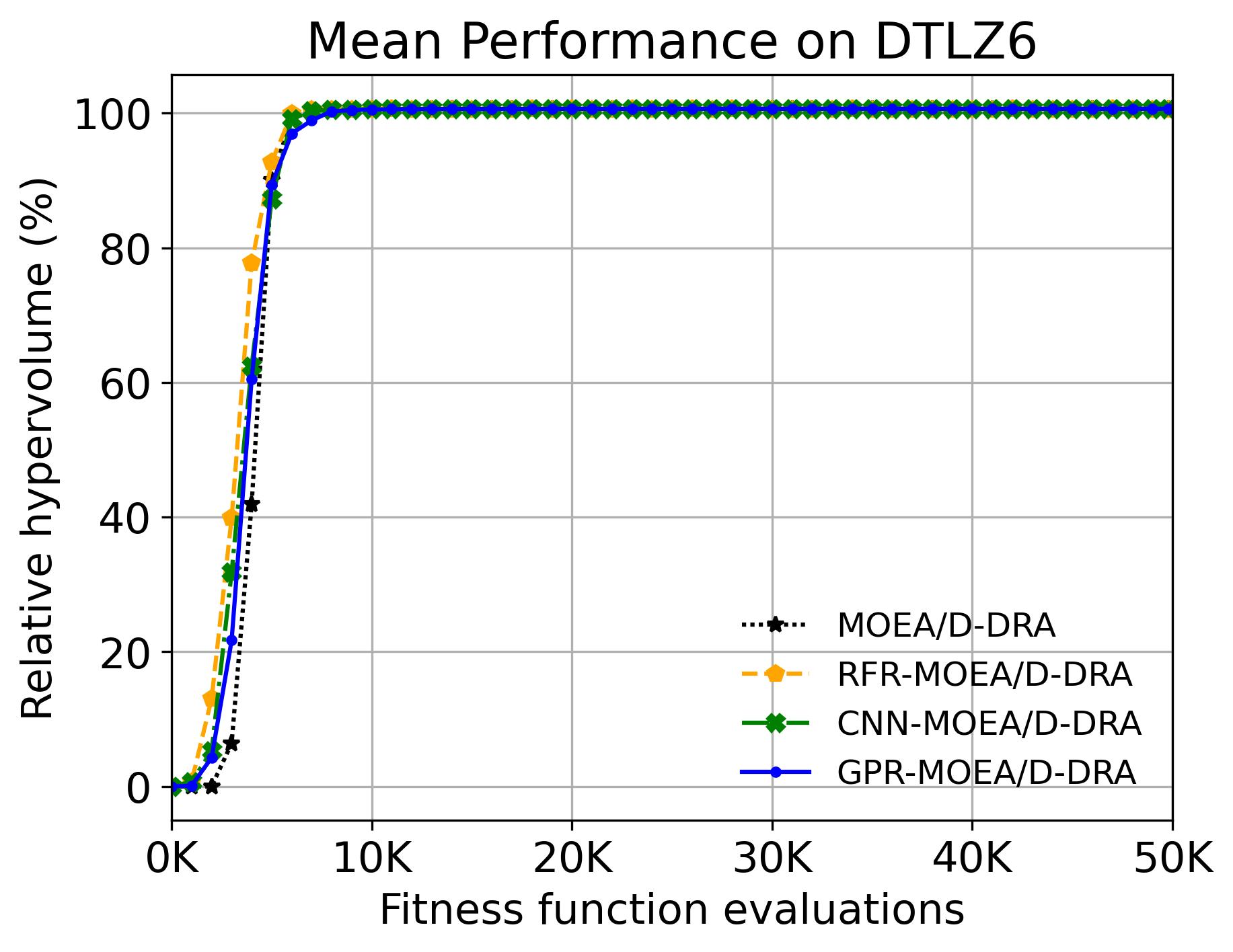}
    \end{subfigure}
    \begin{subfigure}[t]{\figwid\textwidth}
        \centering
        \includegraphics[width=\linewidth]{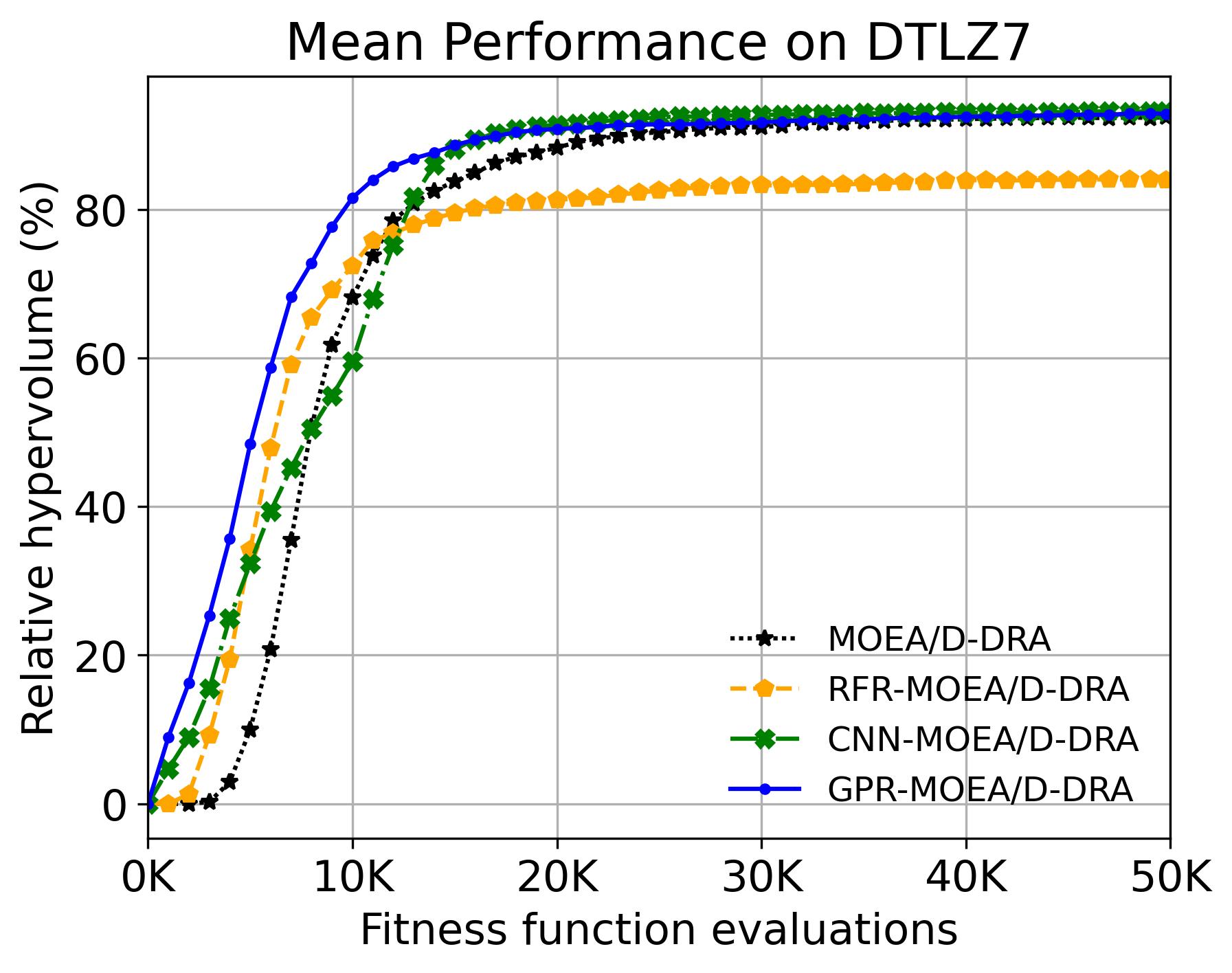}
    \end{subfigure}
    \begin{subfigure}[t]{\figwid\textwidth}
        \centering
        \includegraphics[width=\linewidth]{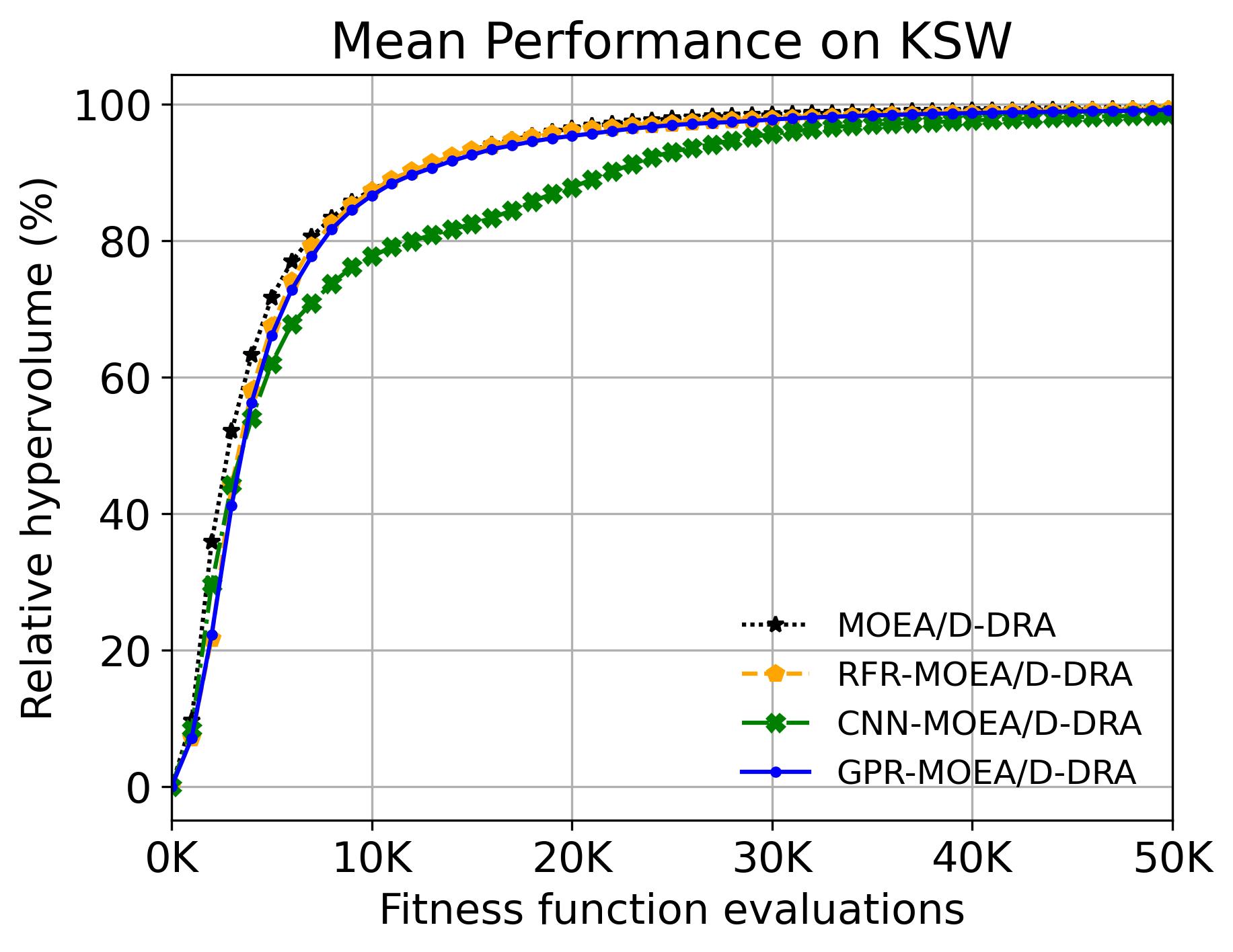}
    \end{subfigure}
    \begin{subfigure}[t]{\figwid\textwidth}
        \centering
        \includegraphics[width=\linewidth]{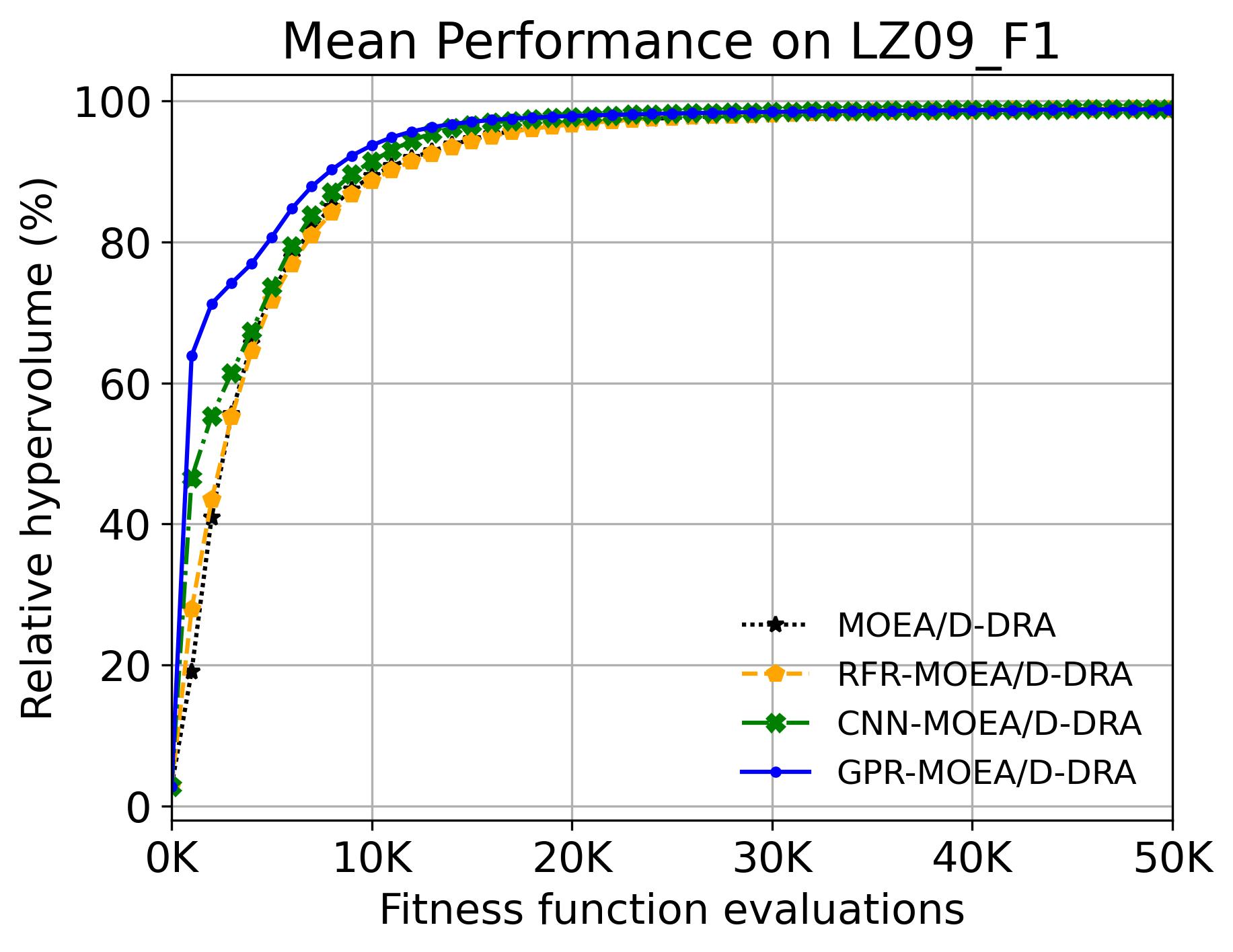}
    \end{subfigure}
    \begin{subfigure}[t]{\figwid\textwidth}
        \centering
        \includegraphics[width=\linewidth]{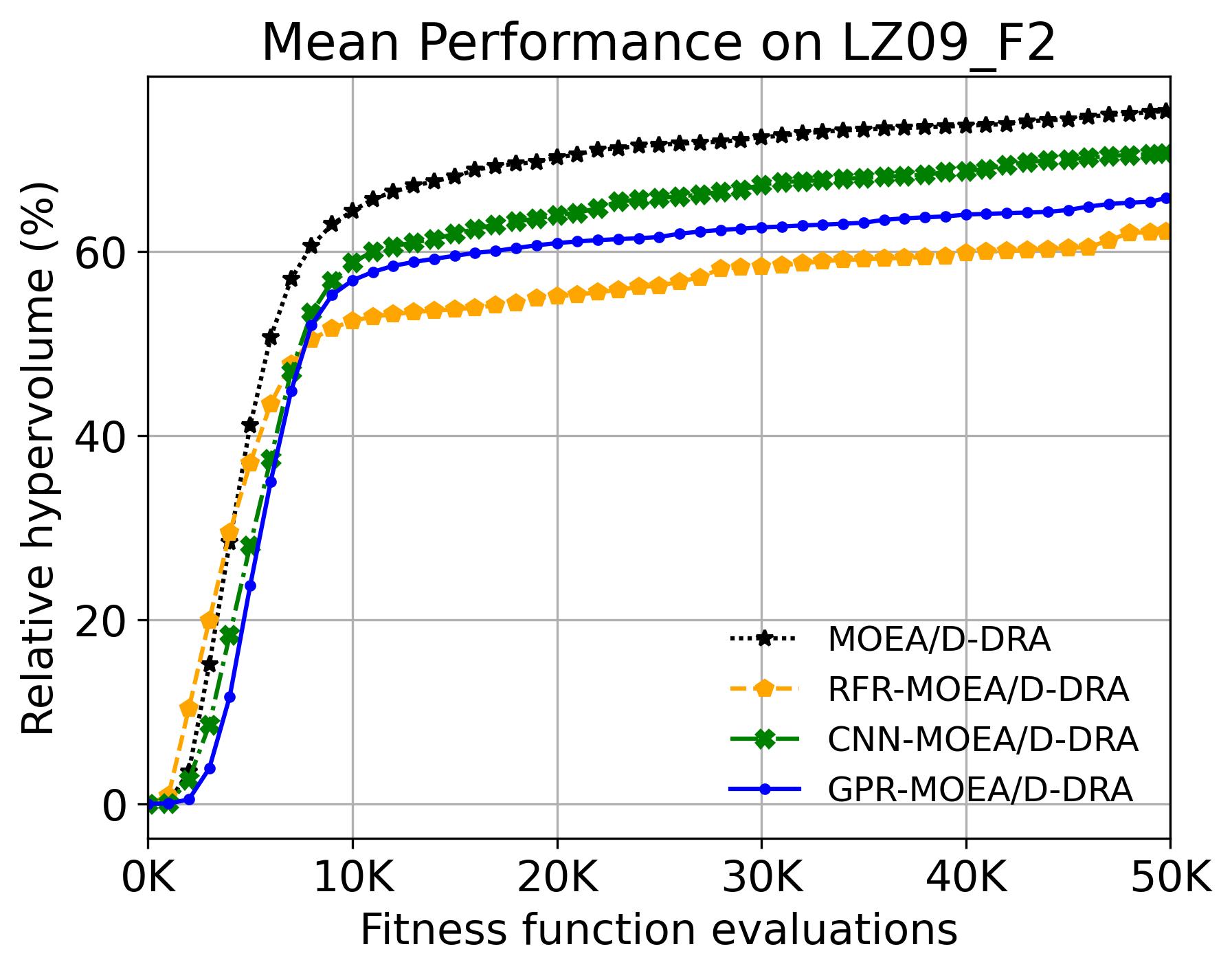}
    \end{subfigure}
    \begin{subfigure}[t]{\figwid\textwidth}
        \centering
        \includegraphics[width=\linewidth]{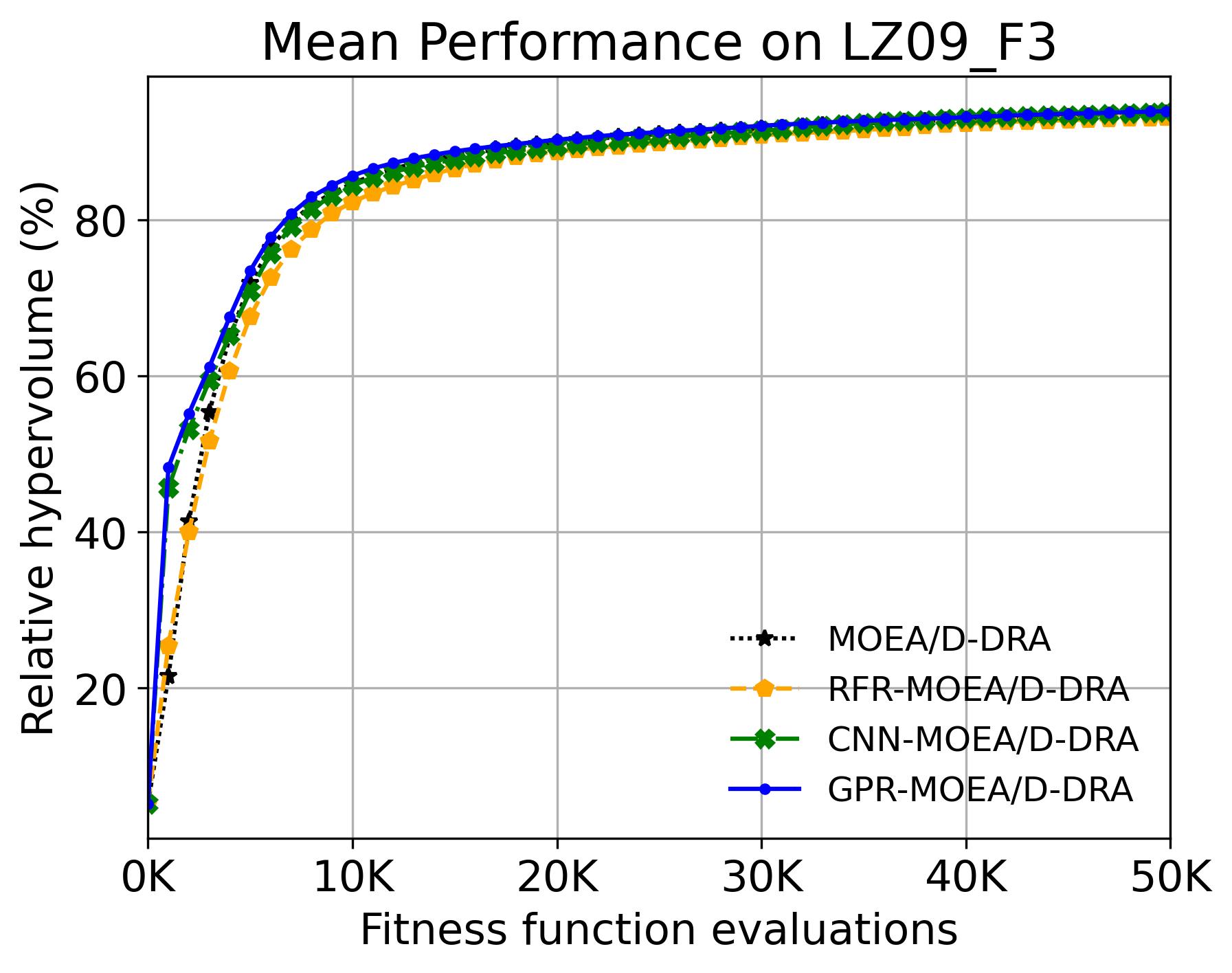}
    \end{subfigure}
    \begin{subfigure}[t]{\figwid\textwidth}
        \centering
        \includegraphics[width=\linewidth]{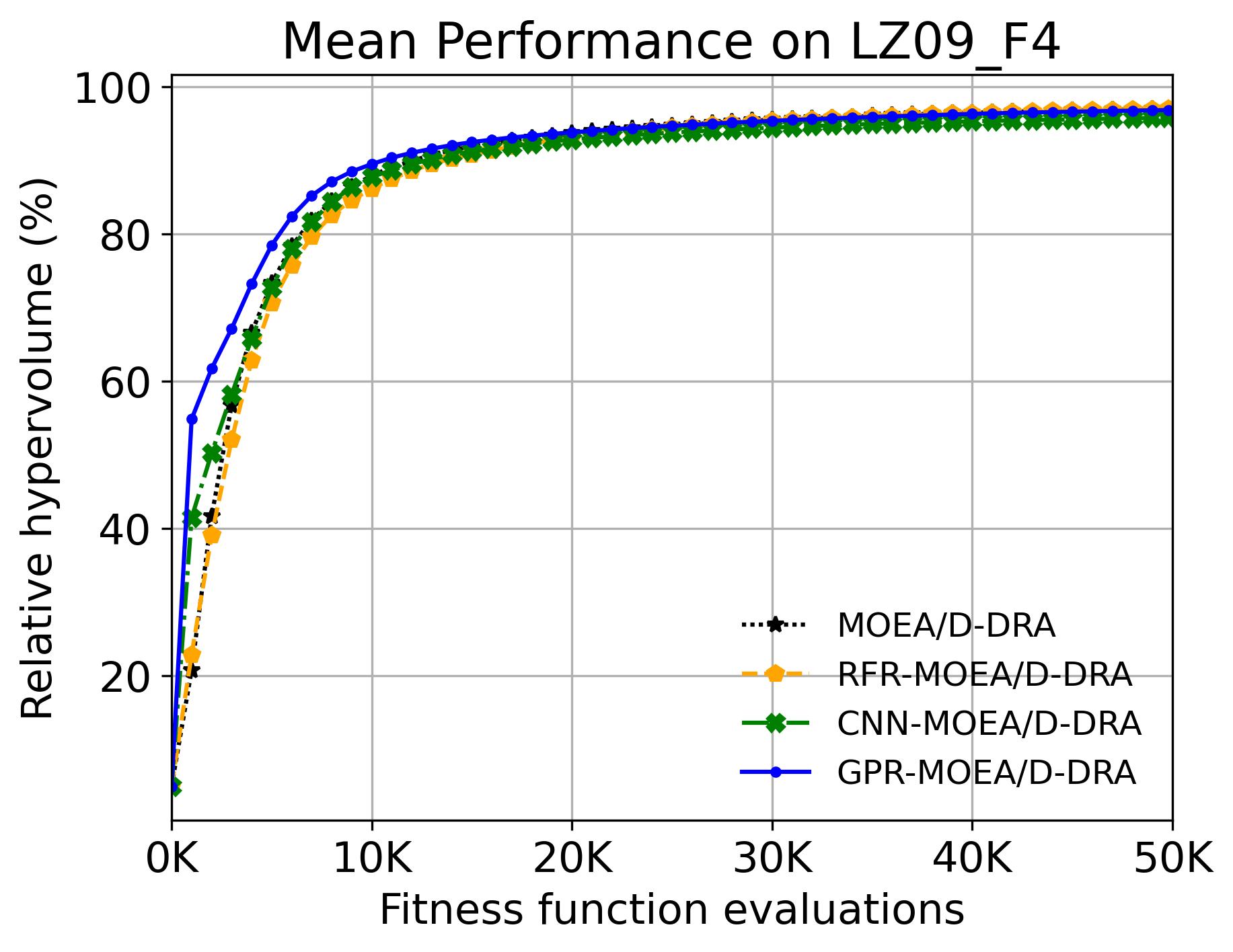}
    \end{subfigure}
    \begin{subfigure}[t]{\figwid\textwidth}
        \centering
        \includegraphics[width=\linewidth]{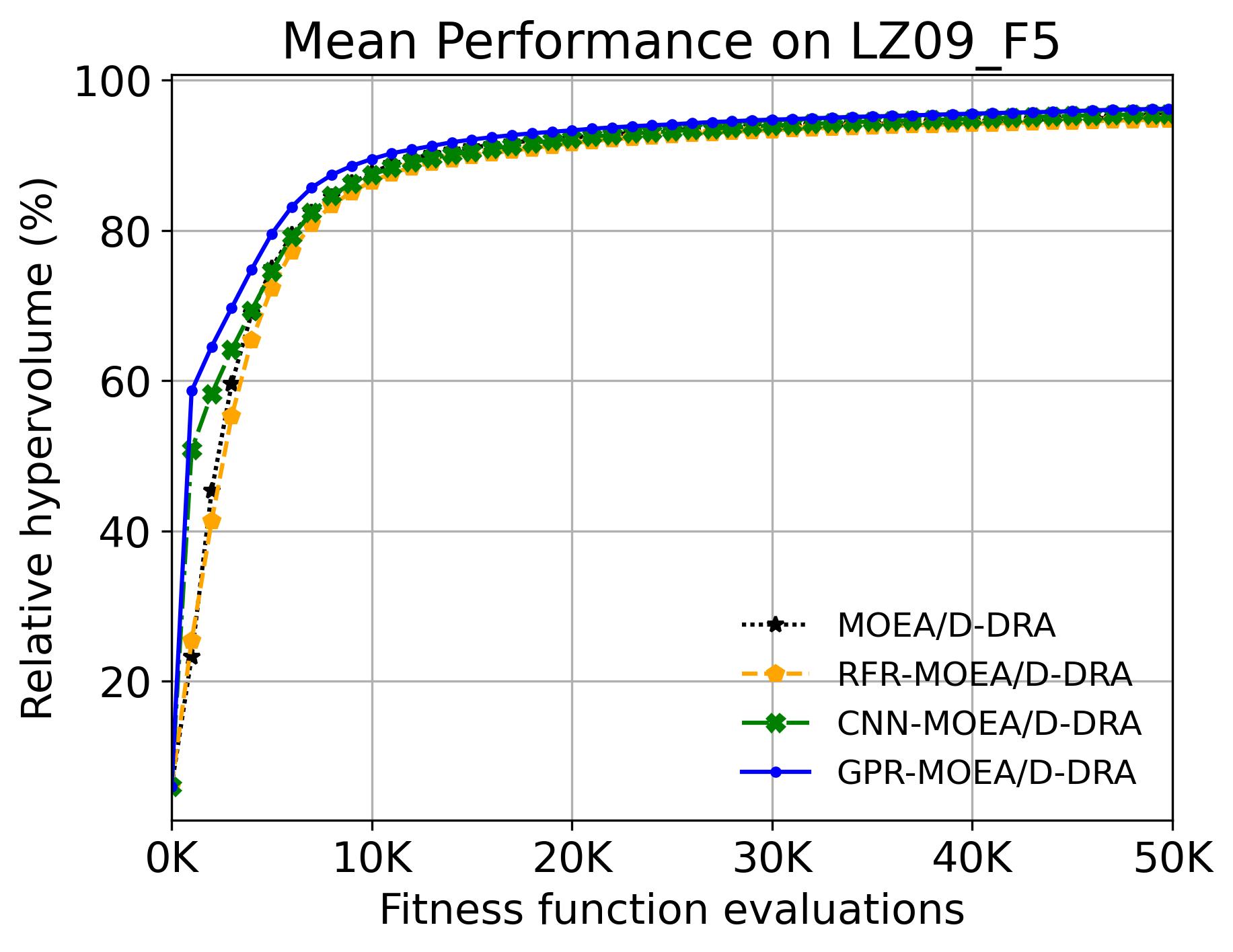}
    \end{subfigure}
    \begin{subfigure}[t]{\figwid\textwidth}
        \centering
        \includegraphics[width=\linewidth]{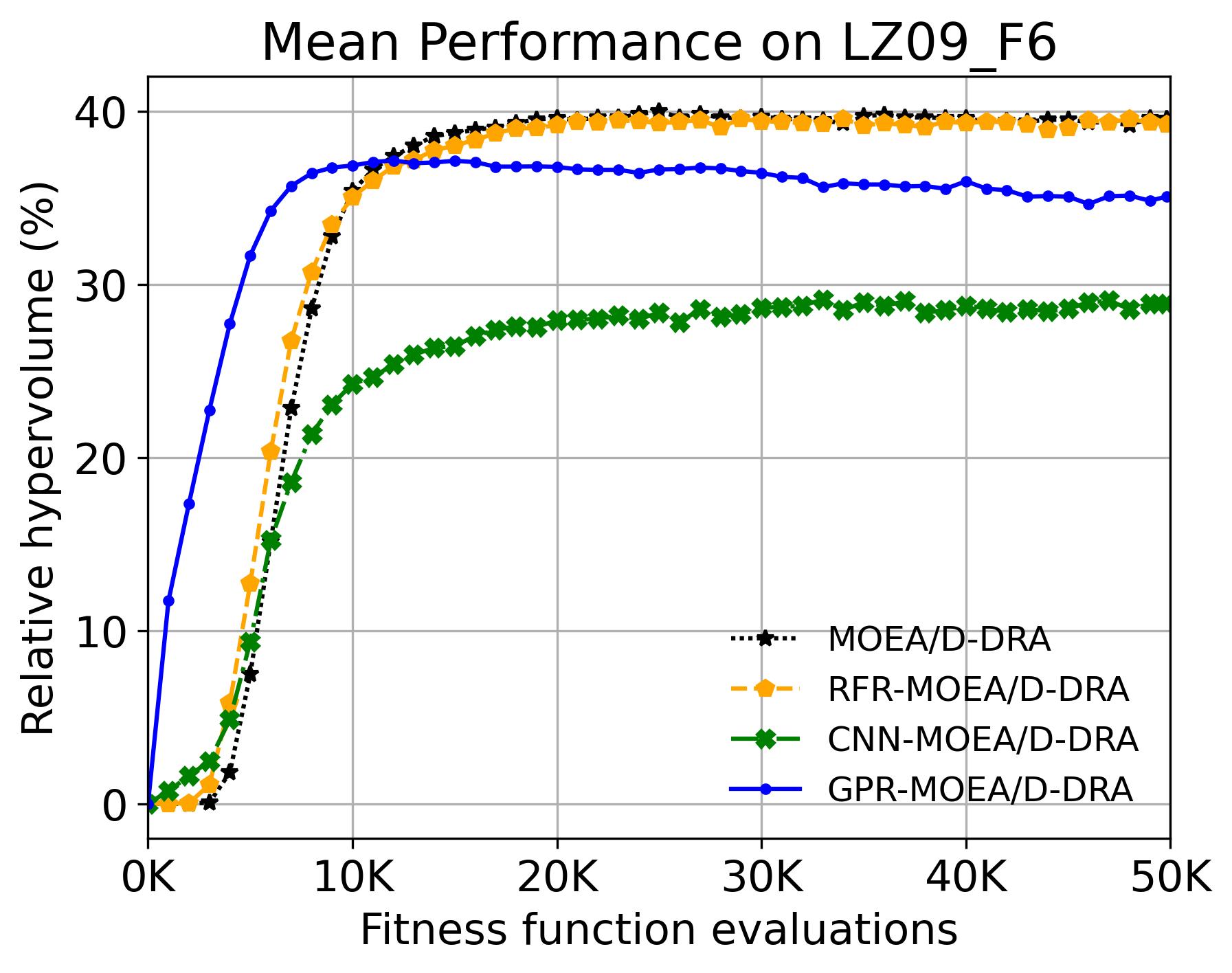}
    \end{subfigure}
    \begin{subfigure}[t]{\figwid\textwidth}
        \centering
        \includegraphics[width=\linewidth]{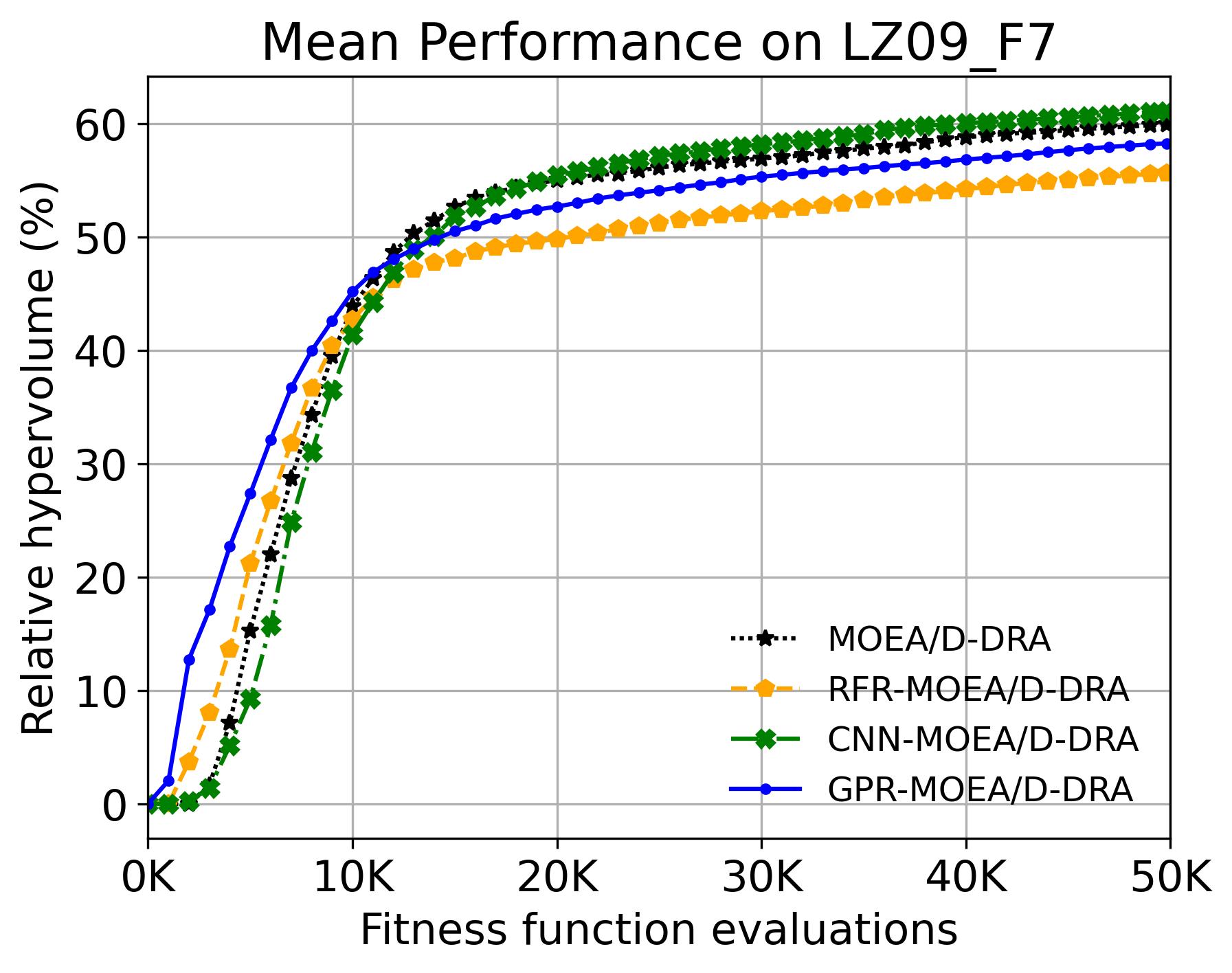}
    \end{subfigure}
    \begin{subfigure}[t]{\figwid\textwidth}
        \centering
        \includegraphics[width=\linewidth]{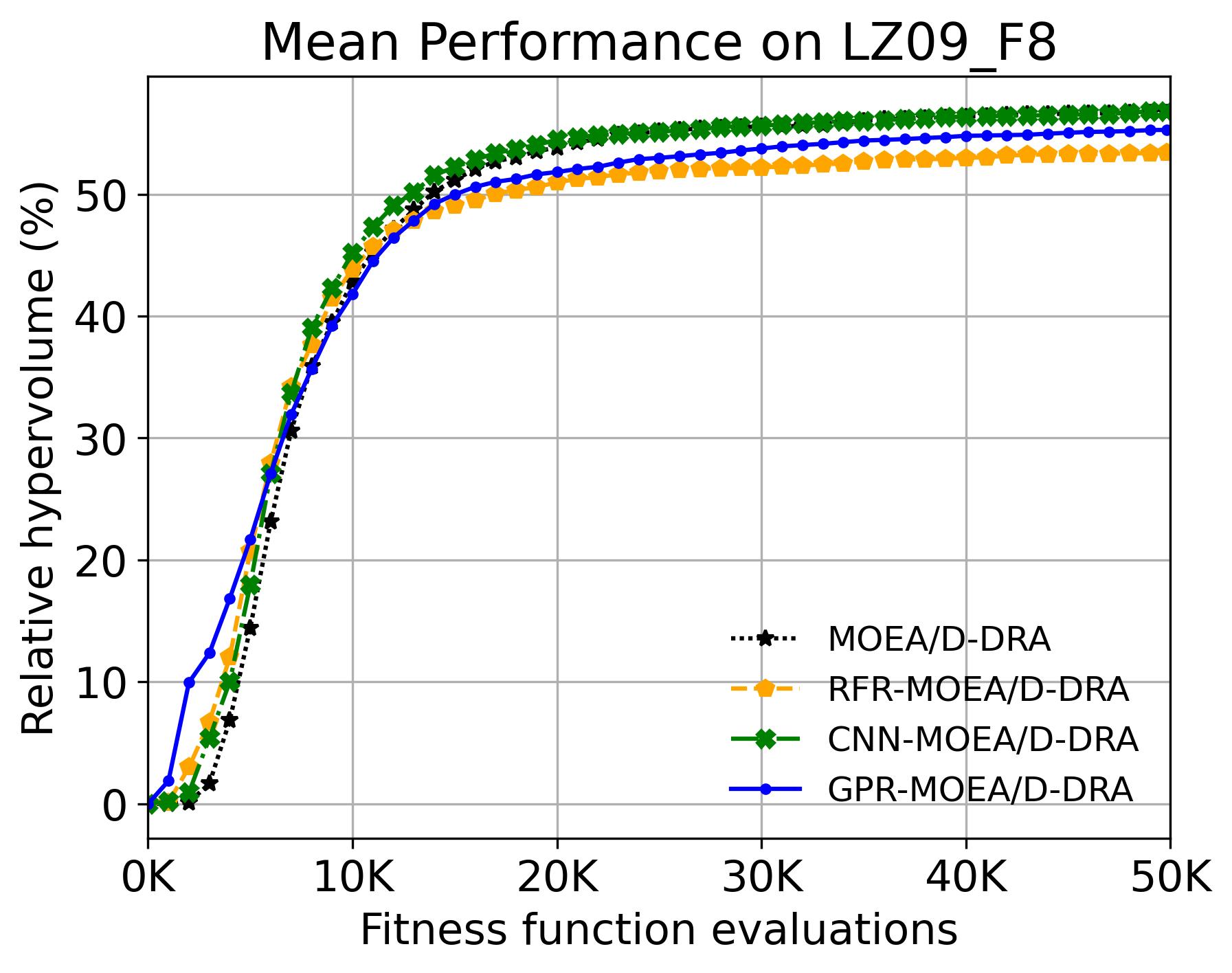}
    \end{subfigure}
    \begin{subfigure}[t]{\figwid\textwidth}
        \centering
        \includegraphics[width=\linewidth]{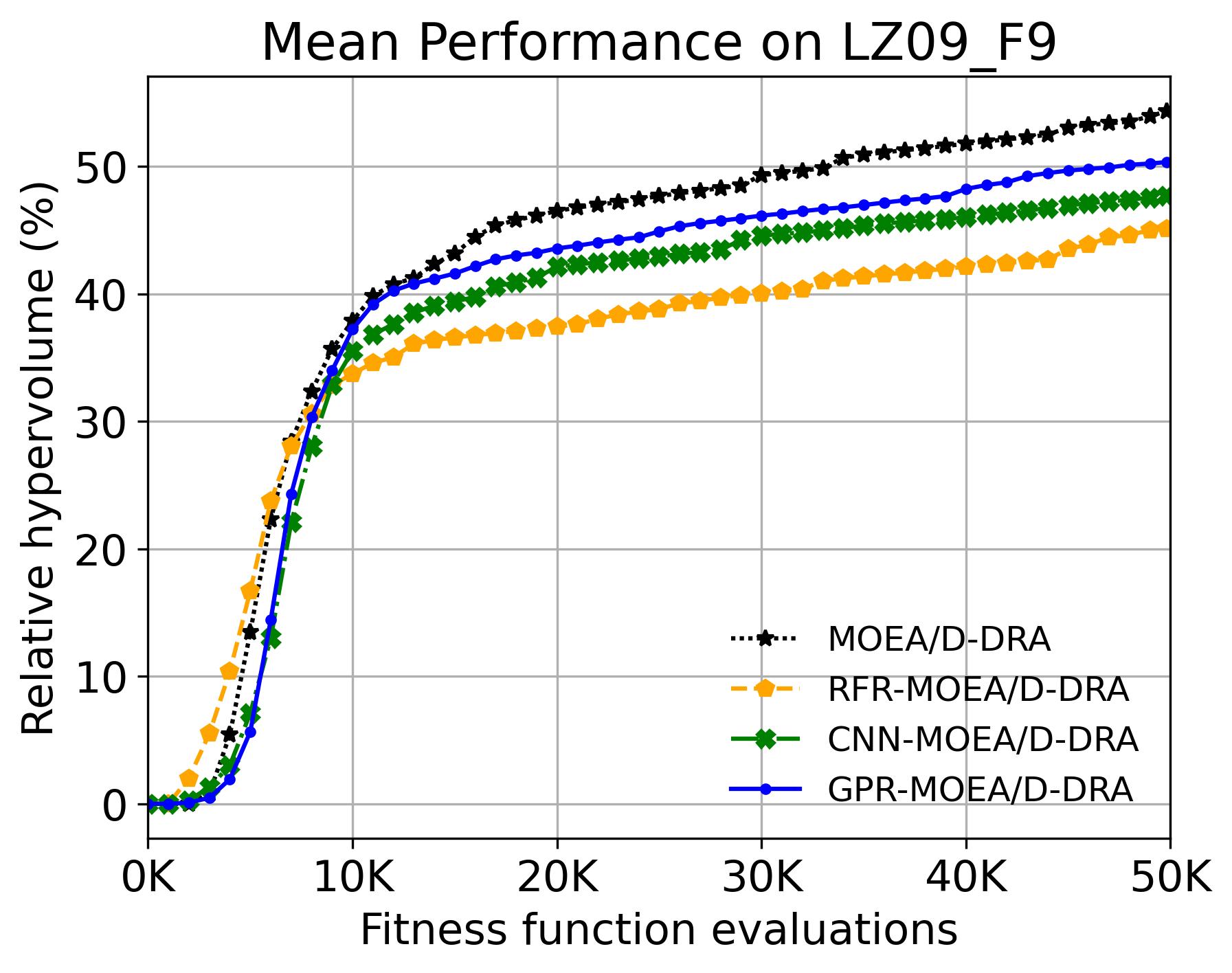}
    \end{subfigure}
    \begin{subfigure}[t]{\figwid\textwidth}
        \centering
        \includegraphics[width=\linewidth]{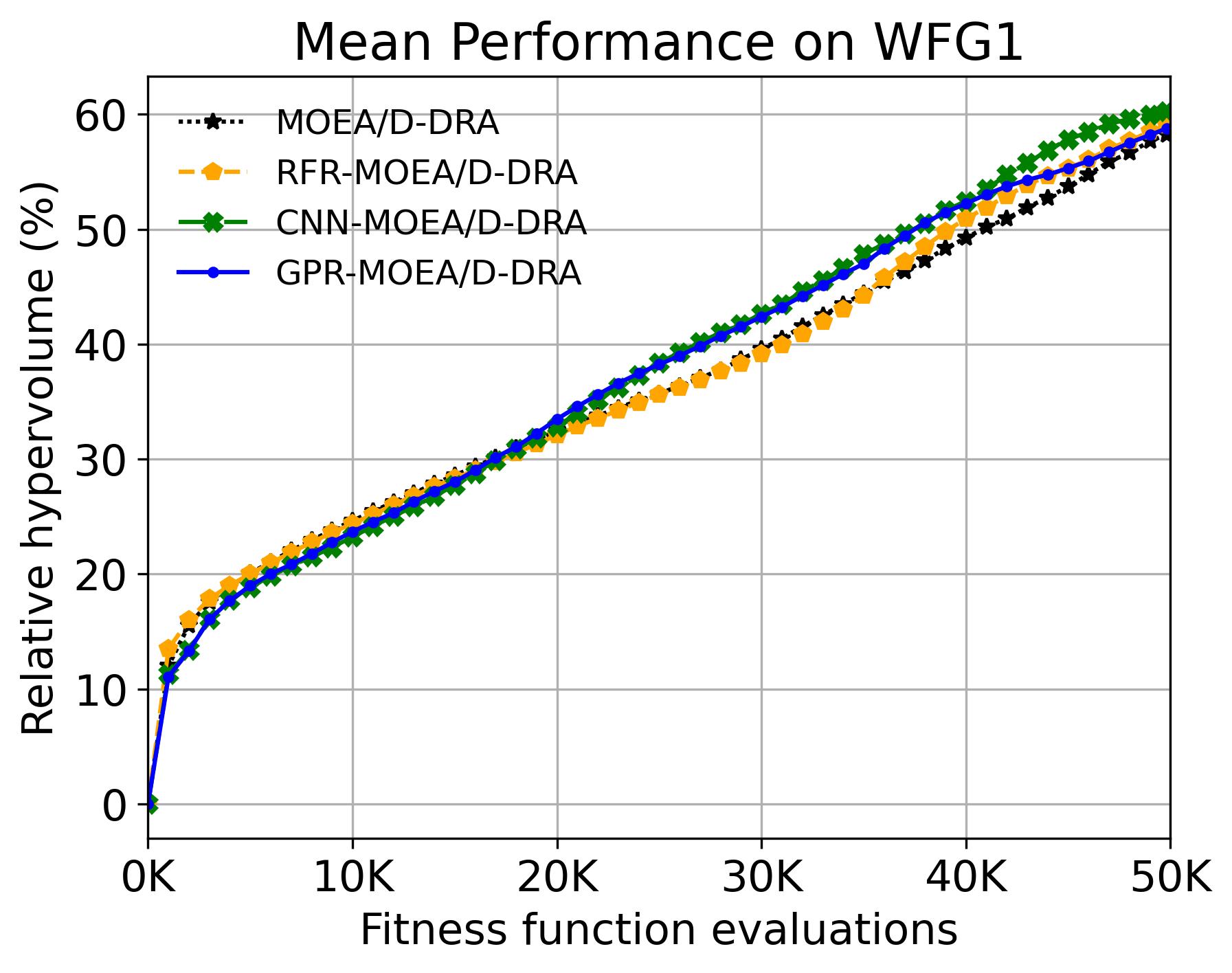}
    \end{subfigure}
    \begin{subfigure}[t]{\figwid\textwidth}
        \centering
        \includegraphics[width=\linewidth]{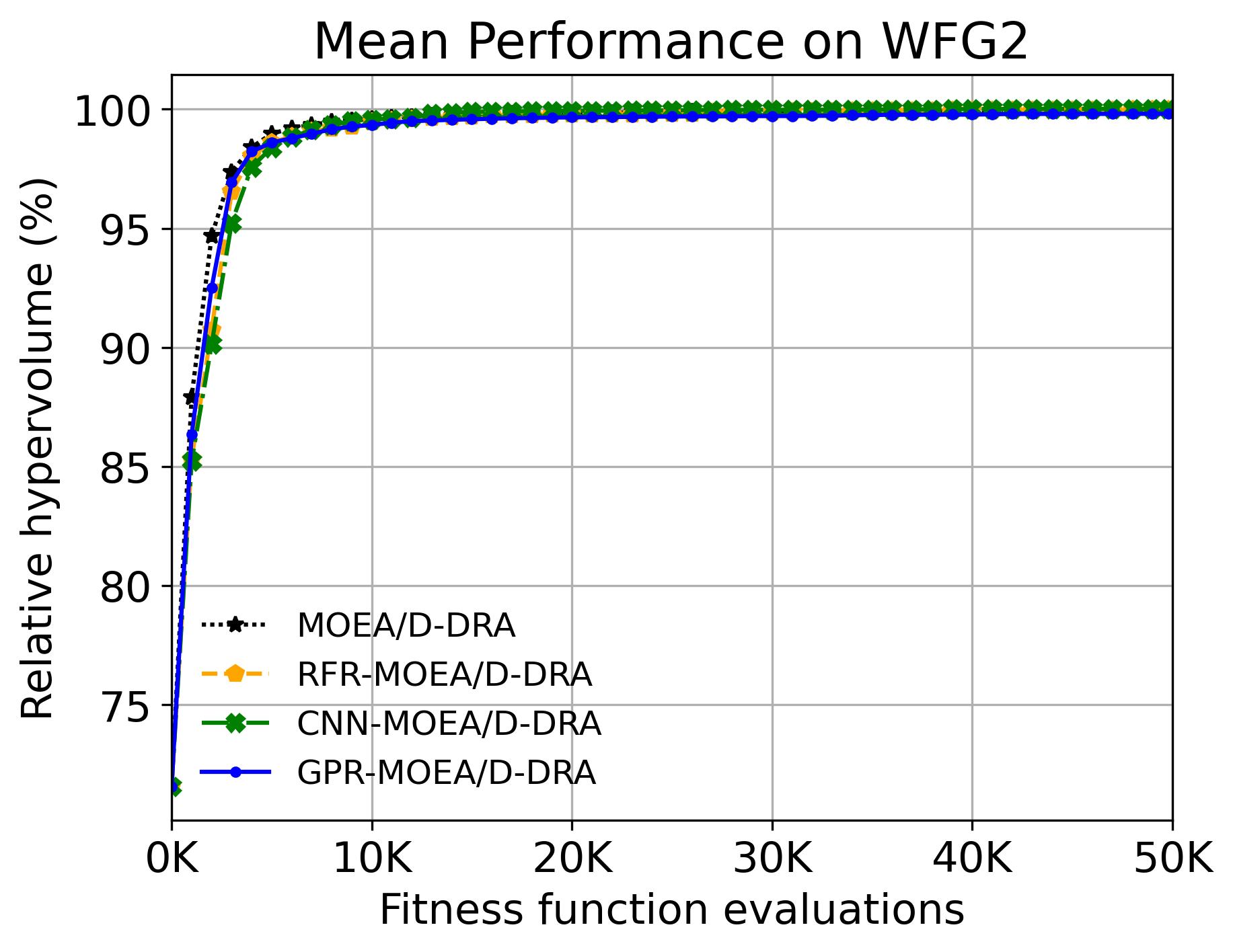}
    \end{subfigure}
    \begin{subfigure}[t]{\figwid\textwidth}
        \centering
        \includegraphics[width=\linewidth]{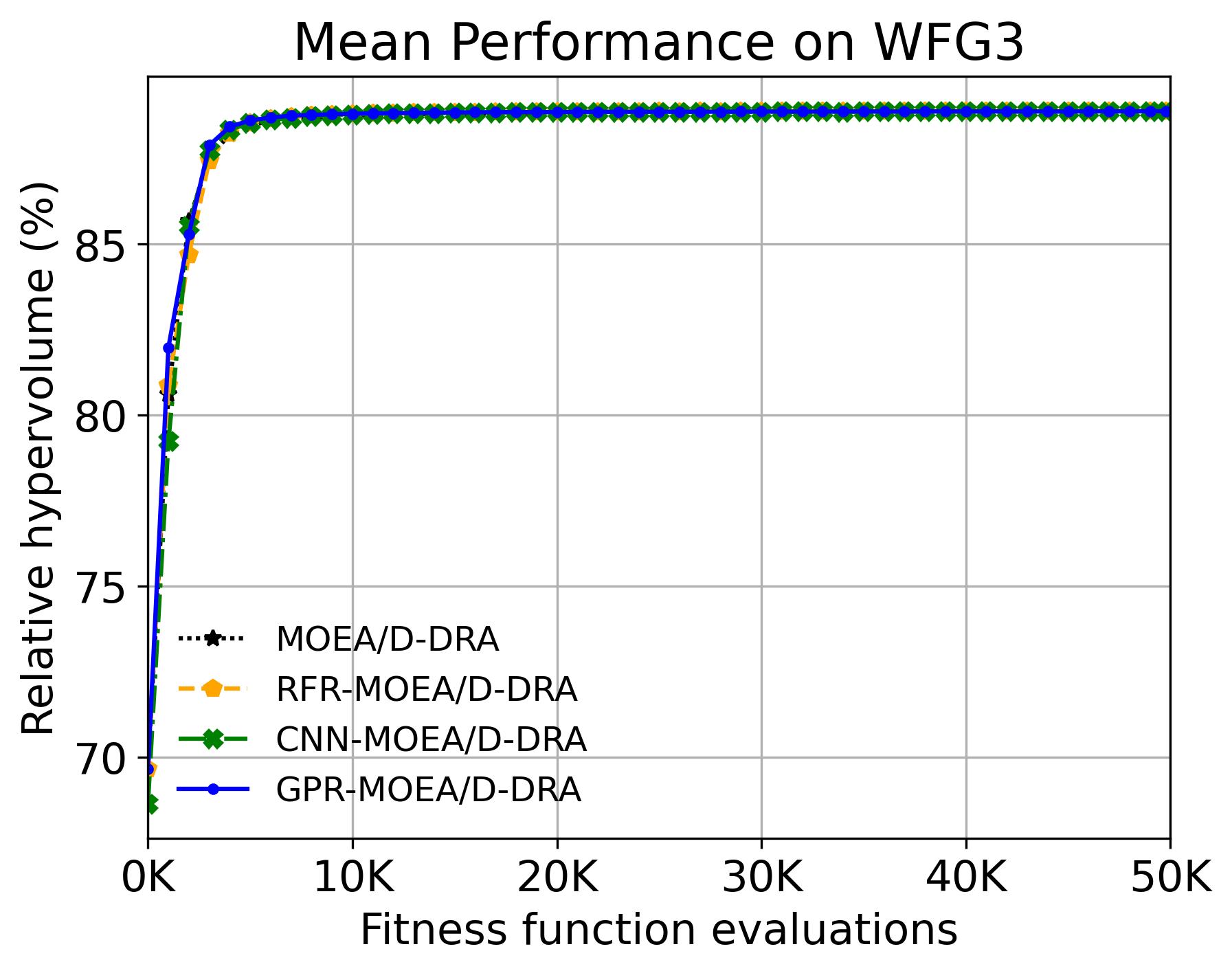}
    \end{subfigure}
    \begin{subfigure}[t]{\figwid\textwidth}
        \centering
        \includegraphics[width=\linewidth]{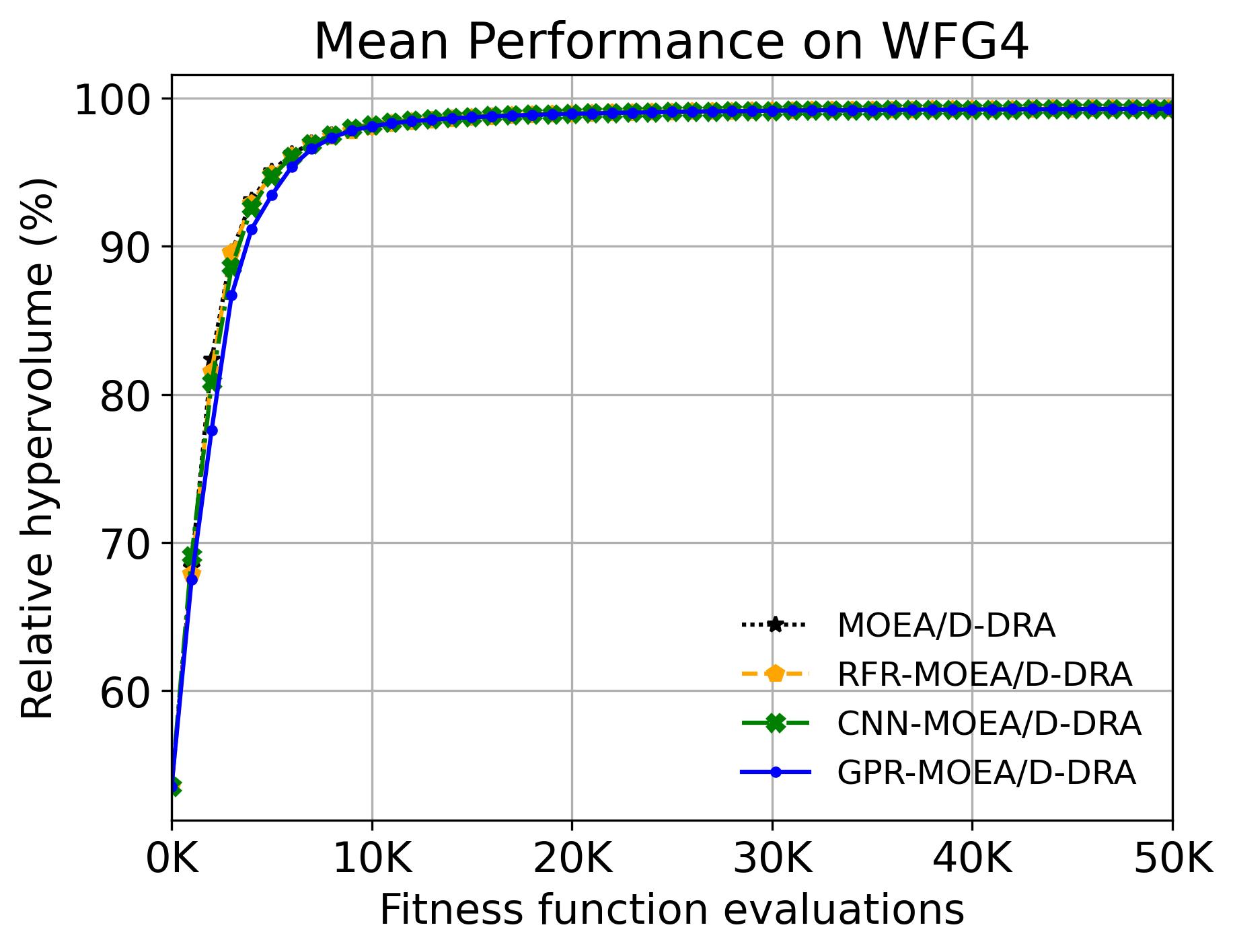}
    \end{subfigure}
    \begin{subfigure}[t]{\figwid\textwidth}
        \centering
        \includegraphics[width=\linewidth]{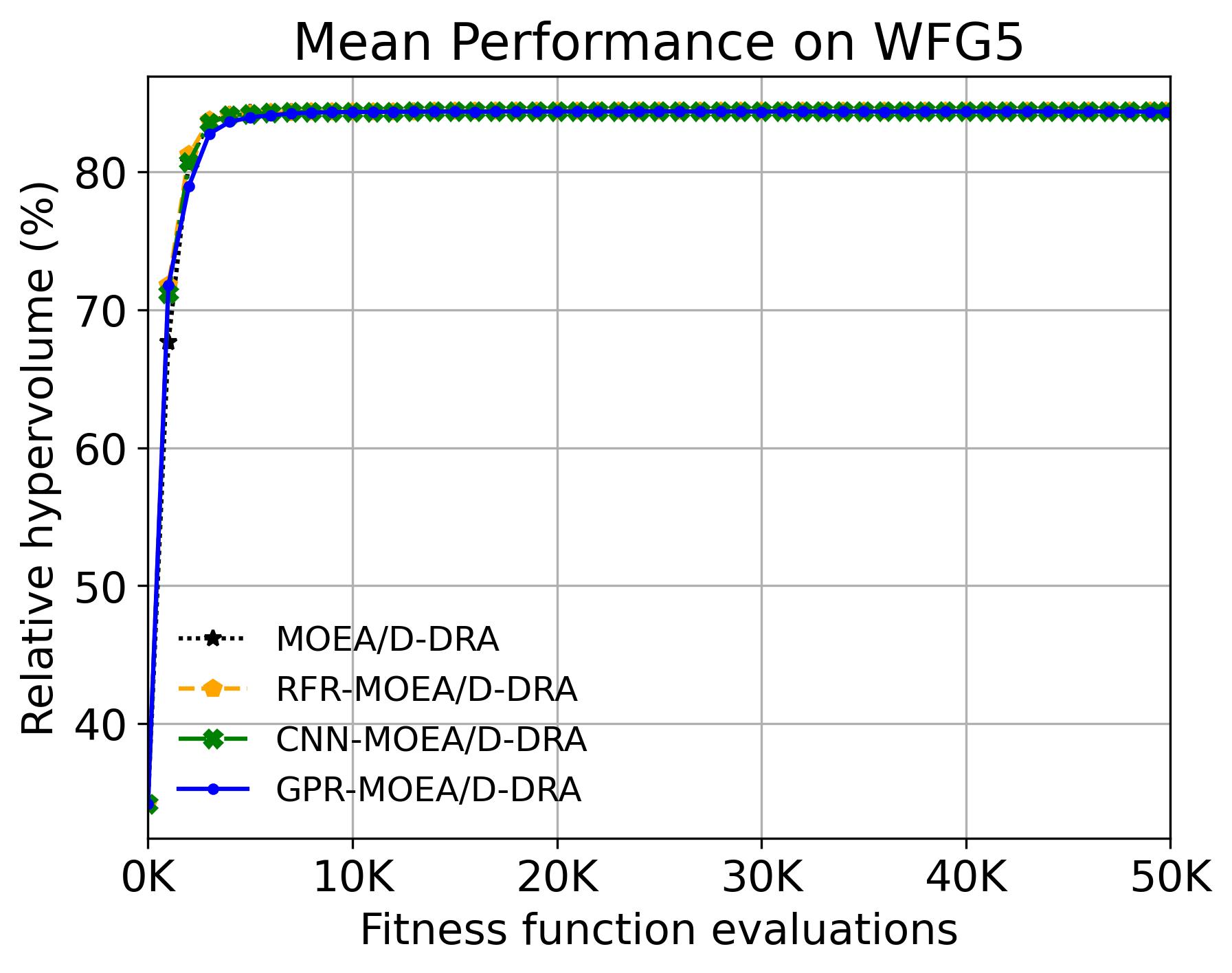}
    \end{subfigure}
    \begin{subfigure}[t]{\figwid\textwidth}
        \centering
        \includegraphics[width=\linewidth]{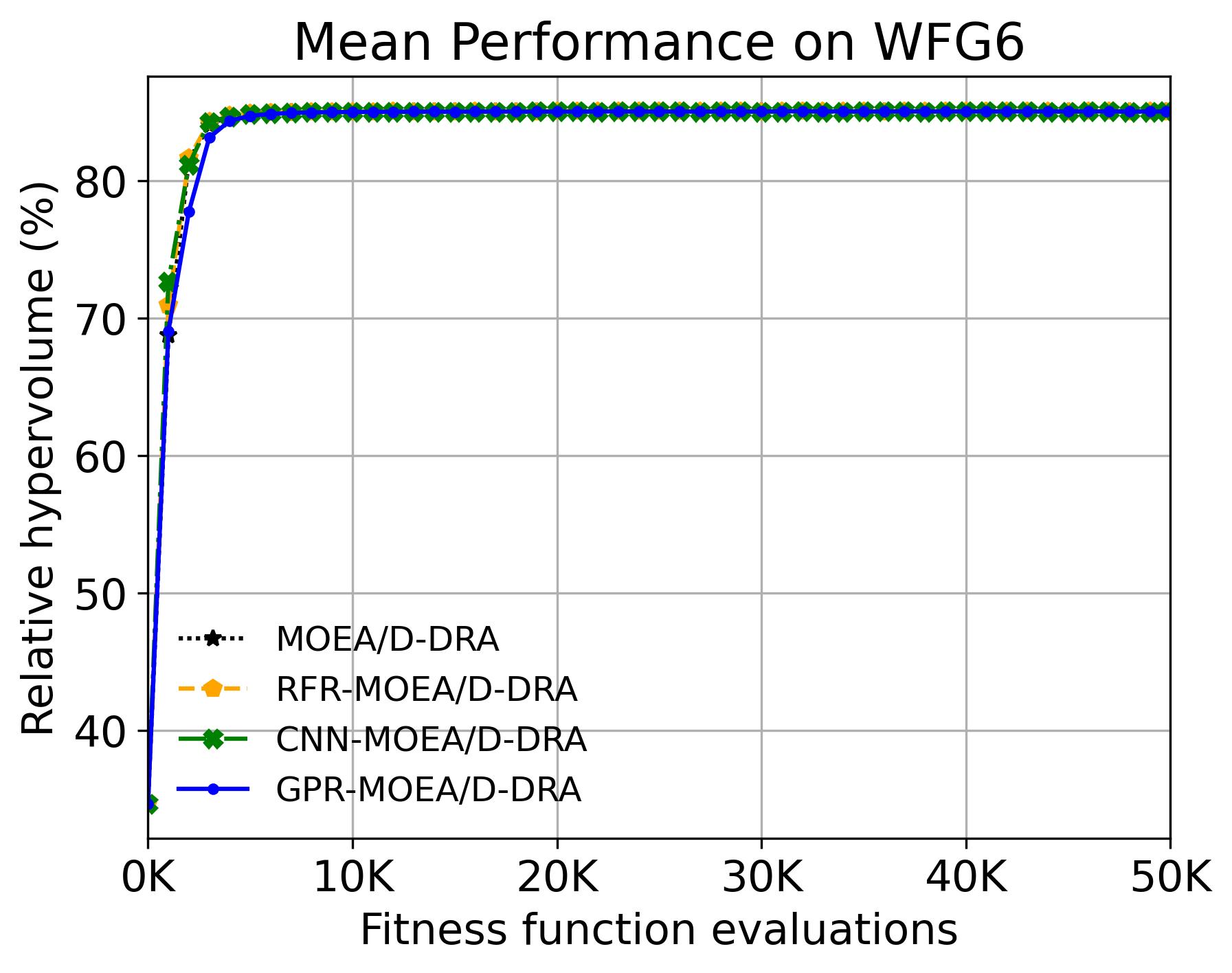}
    \end{subfigure}
    \begin{subfigure}[t]{\figwid\textwidth}
        \centering
        \includegraphics[width=\linewidth]{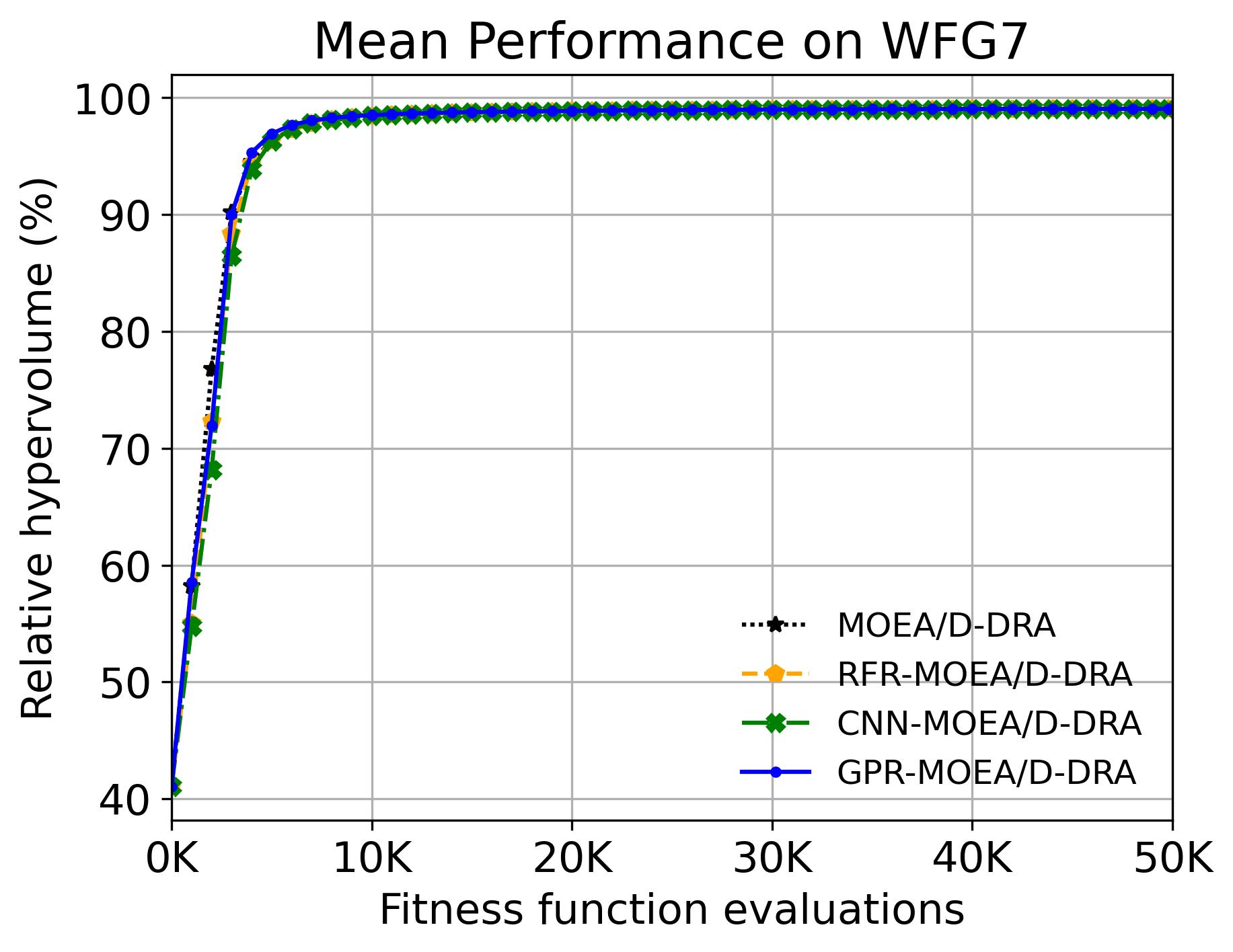}
    \end{subfigure}
    \begin{subfigure}[t]{\figwid\textwidth}
        \centering
        \includegraphics[width=\linewidth]{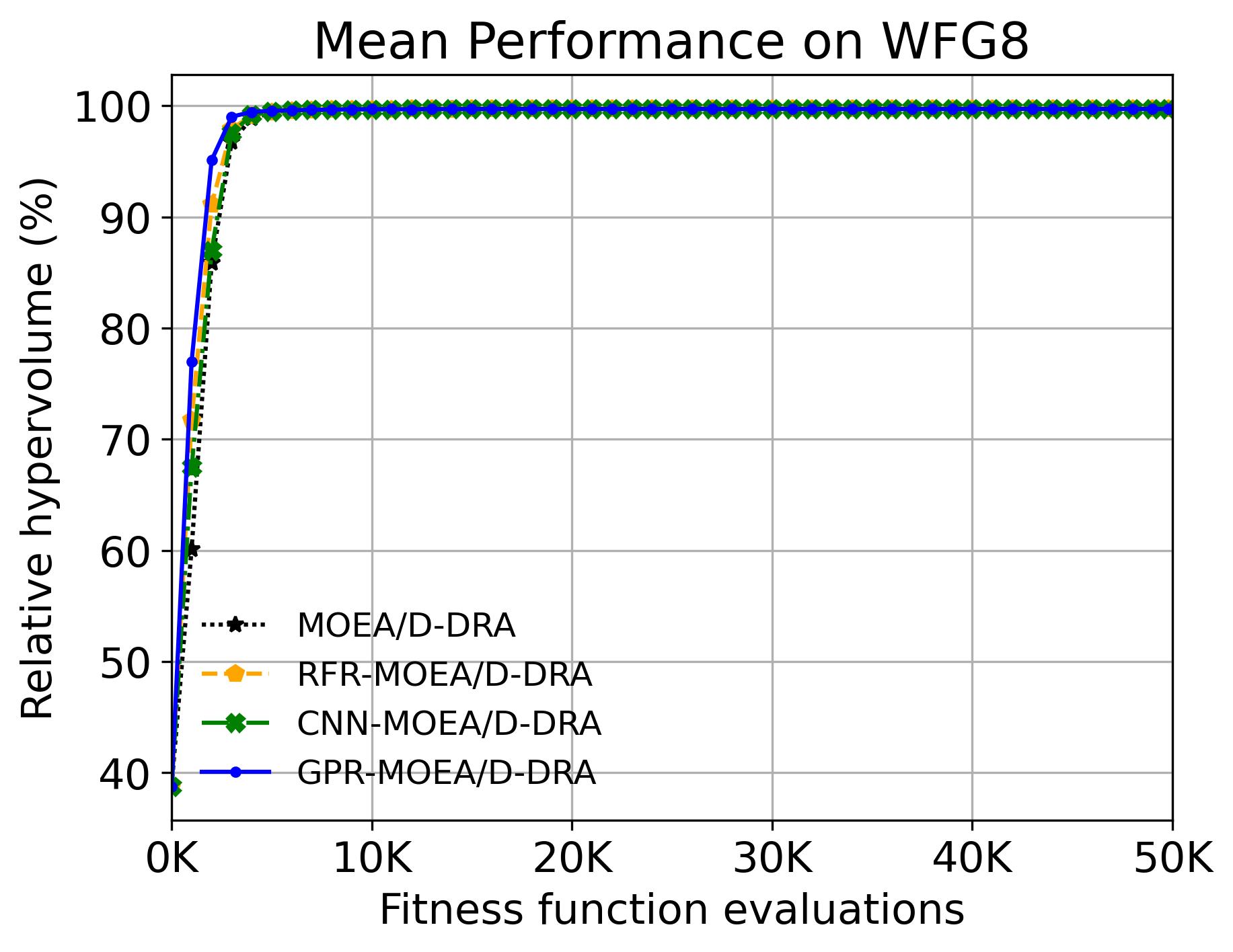}
    \end{subfigure}
    \begin{subfigure}[t]{\figwid\textwidth}
        \centering
        \includegraphics[width=\linewidth]{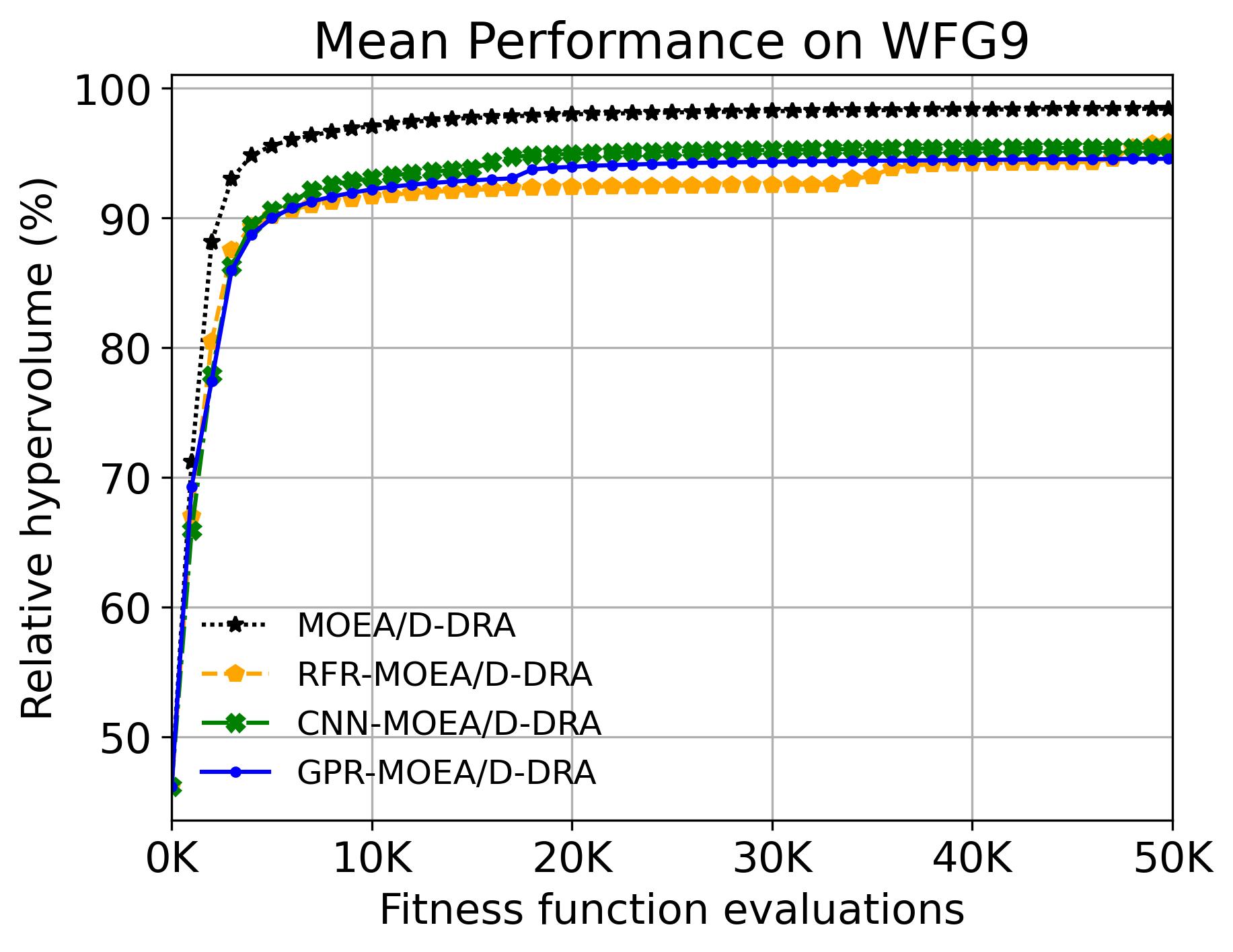}
    \end{subfigure}
    \begin{subfigure}[t]{\figwid\textwidth}
        \centering
        \includegraphics[width=\linewidth]{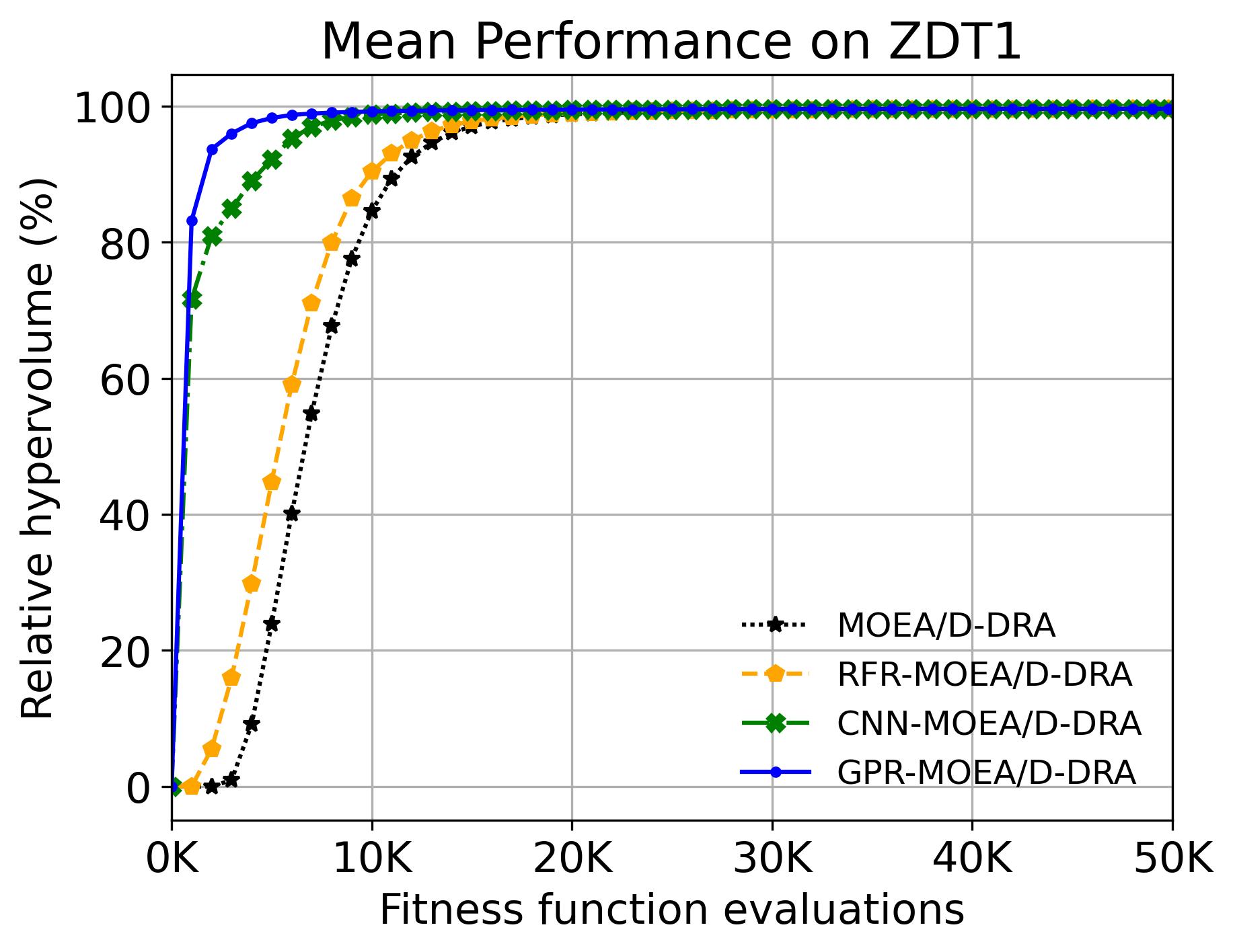}
    \end{subfigure}
    \begin{subfigure}[t]{\figwid\textwidth}
        \centering
        \includegraphics[width=\linewidth]{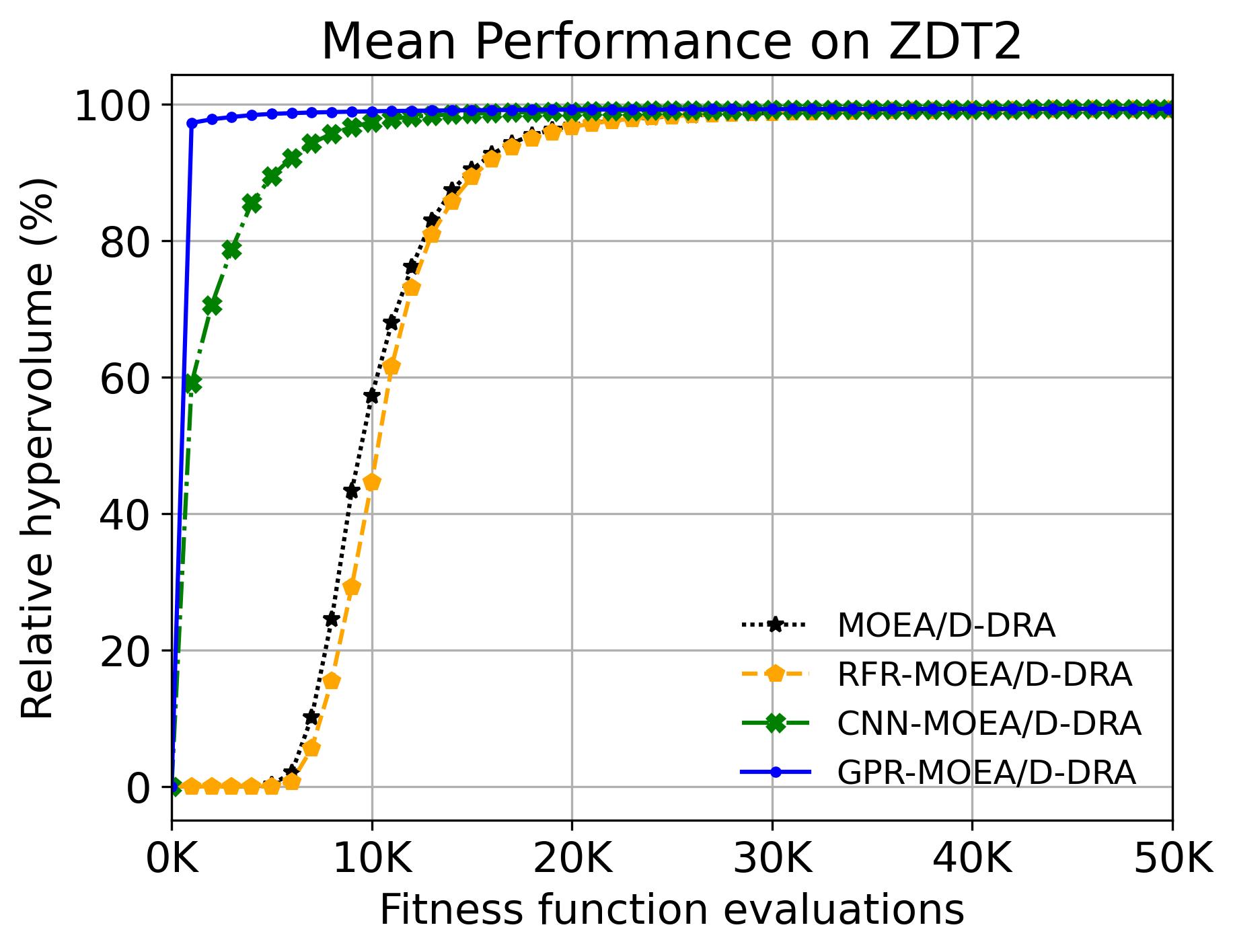}
    \end{subfigure}
    \begin{subfigure}[t]{\figwid\textwidth}
        \centering
        \includegraphics[width=\linewidth]{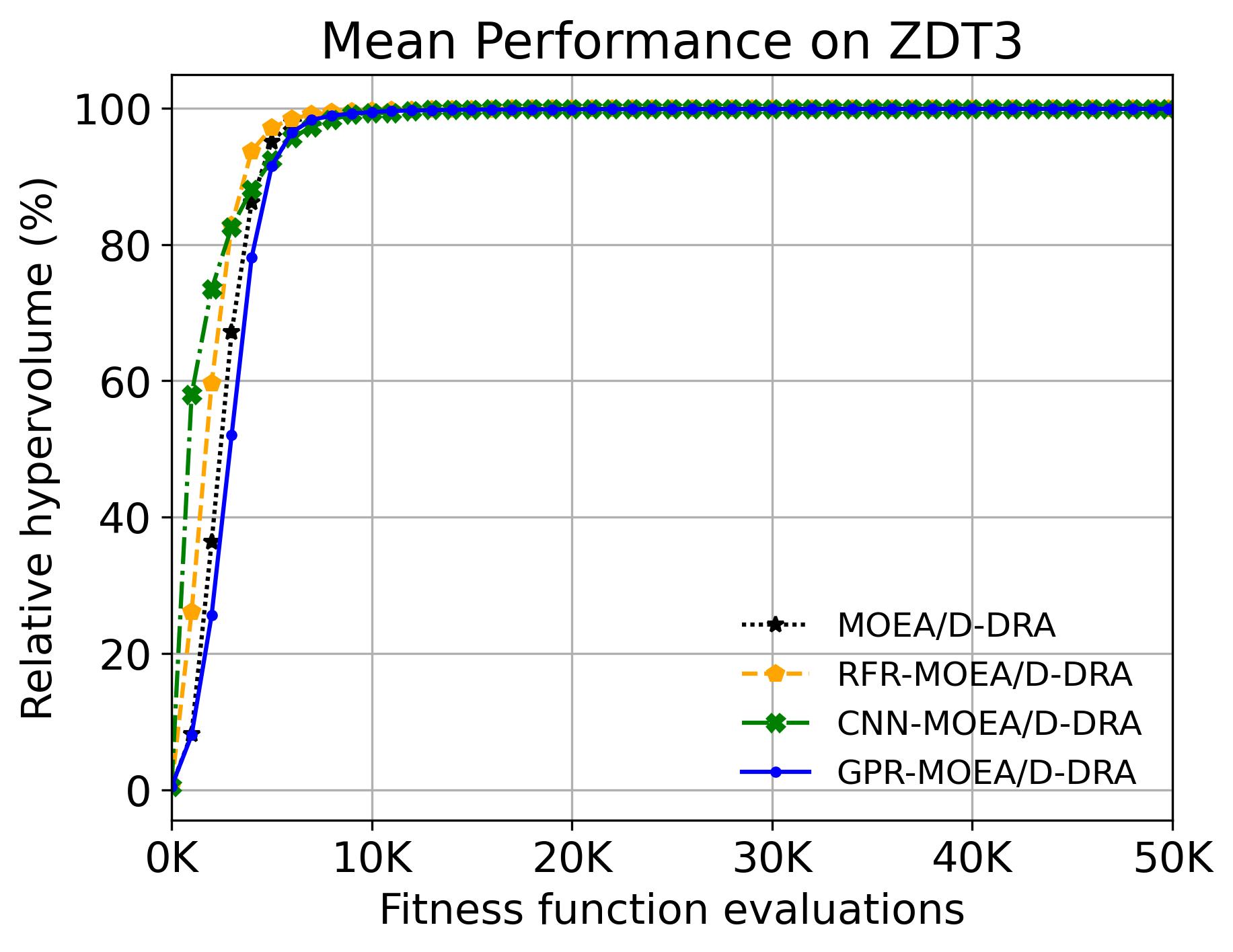}
    \end{subfigure}
    \begin{subfigure}[t]{\figwid\textwidth}
        \centering
        \includegraphics[width=\linewidth]{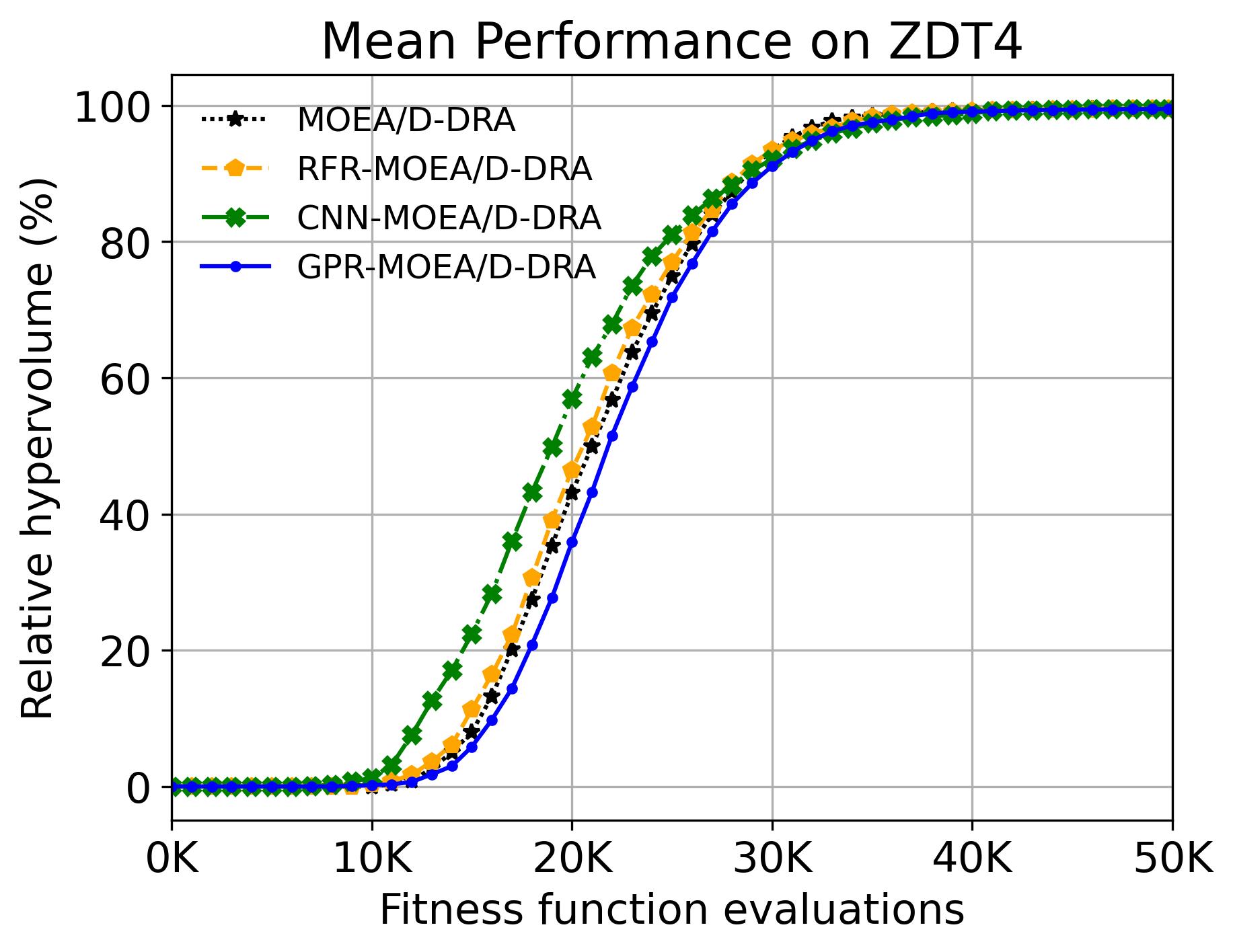}
    \end{subfigure}
    \begin{subfigure}[t]{\figwid\textwidth}
        \centering
        \includegraphics[width=\linewidth]{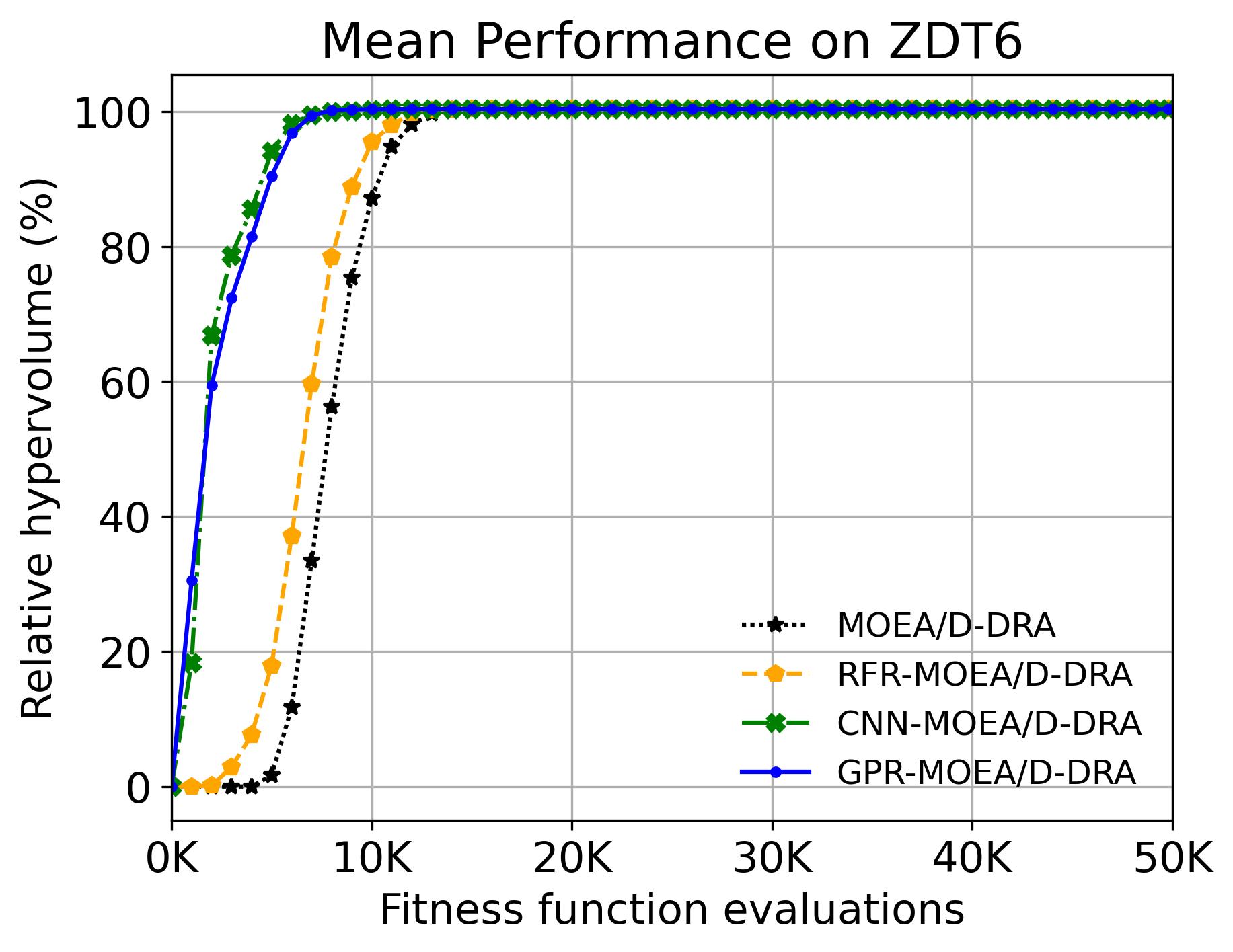}
    \end{subfigure}

        \caption{Comparison of MOEA/D and its associated surrogate-enhanced solvers on individual benchmark problems using $Hv(PF_c)$ -- i.e., the relative hypervolume -- as a performance indicator.}
        \label{fig:annexMOEAD}
\end{figure*}

\begin{figure*}[h]
    \centering
    \begin{subfigure}[t]{\figwid\textwidth}
        \centering
        \includegraphics[width=\linewidth]{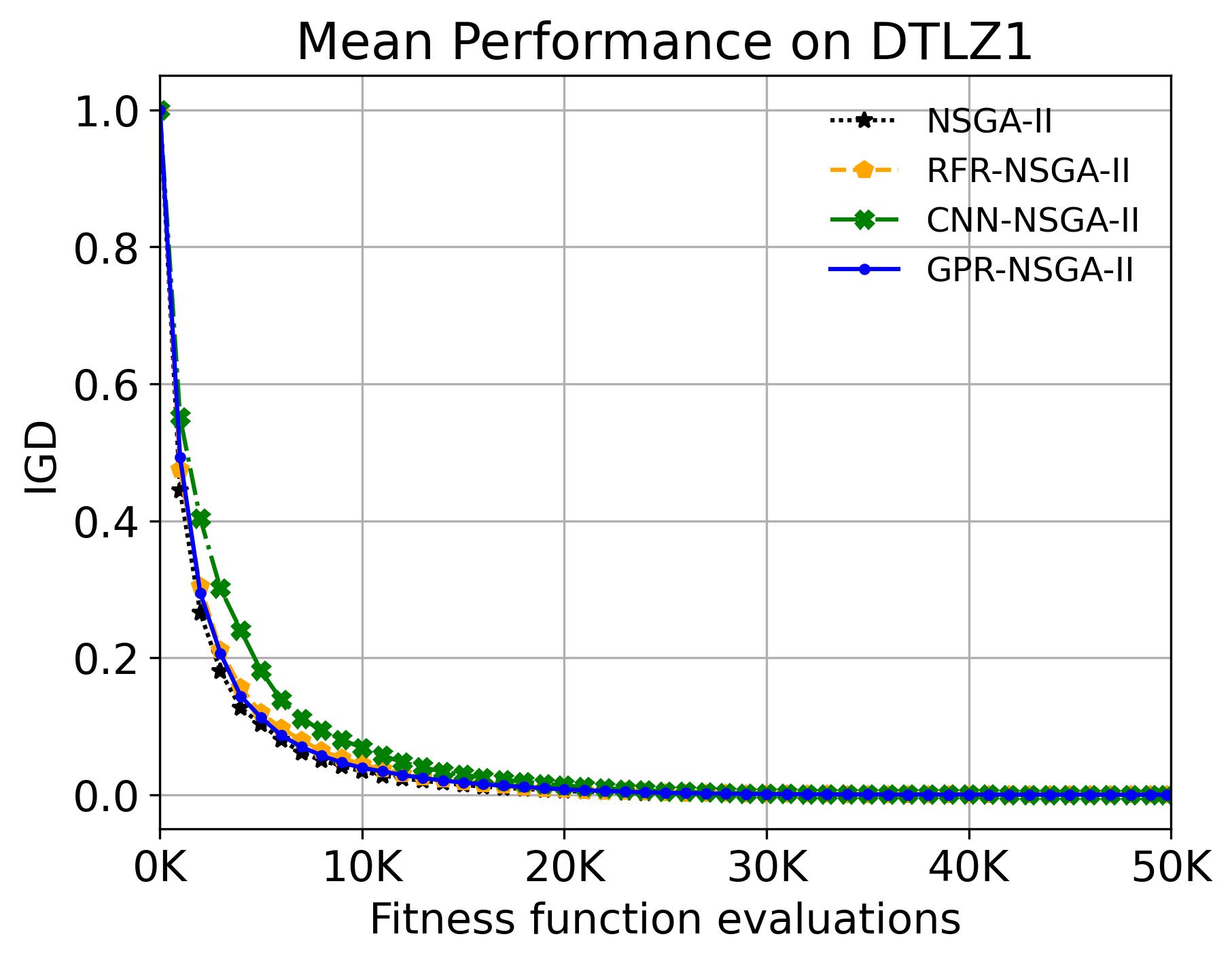}
    \end{subfigure}
    \begin{subfigure}[t]{\figwid\textwidth}
        \centering
        \includegraphics[width=\linewidth]{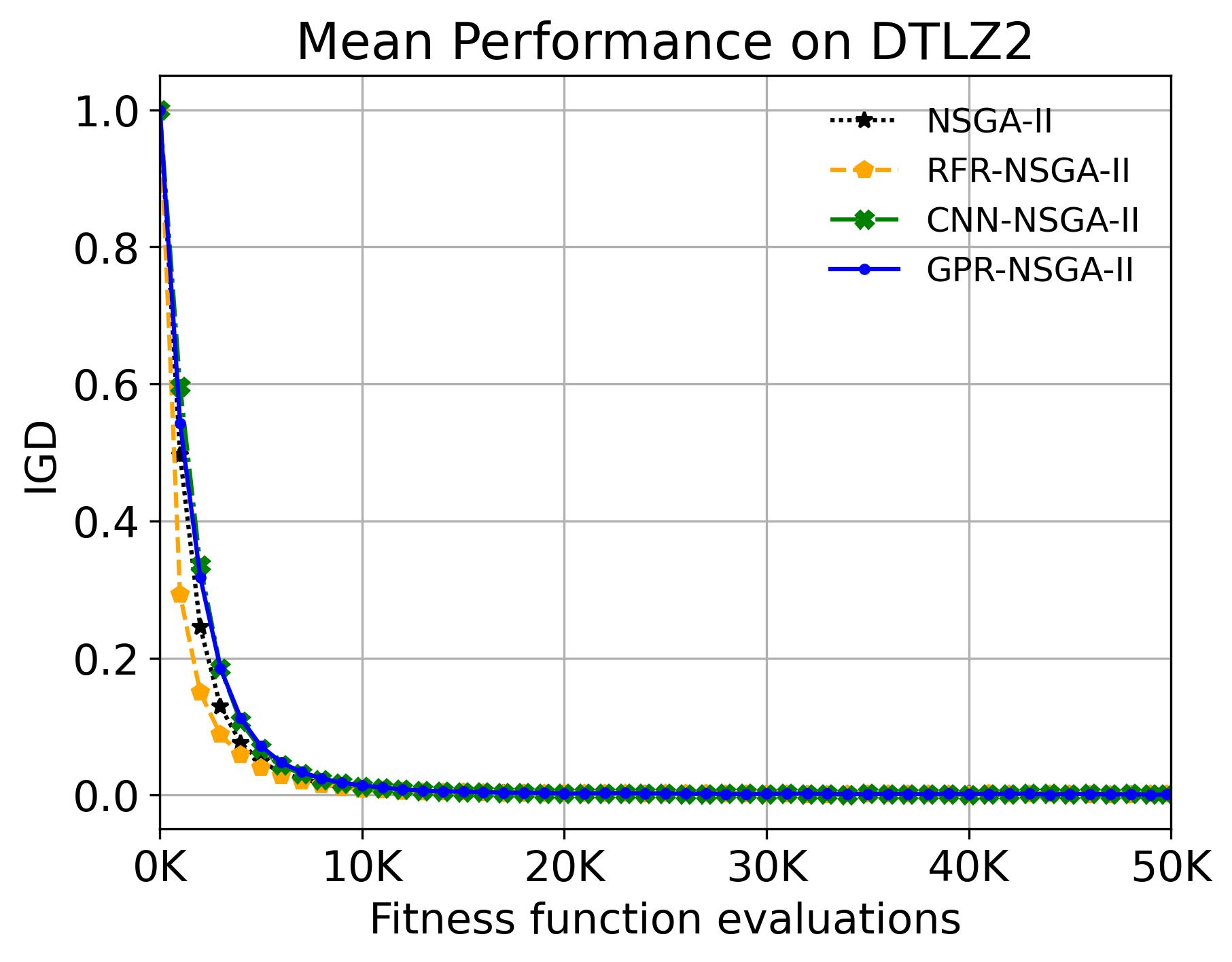}
    \end{subfigure}
    \begin{subfigure}[t]{\figwid\textwidth}
        \centering
        \includegraphics[width=\linewidth]{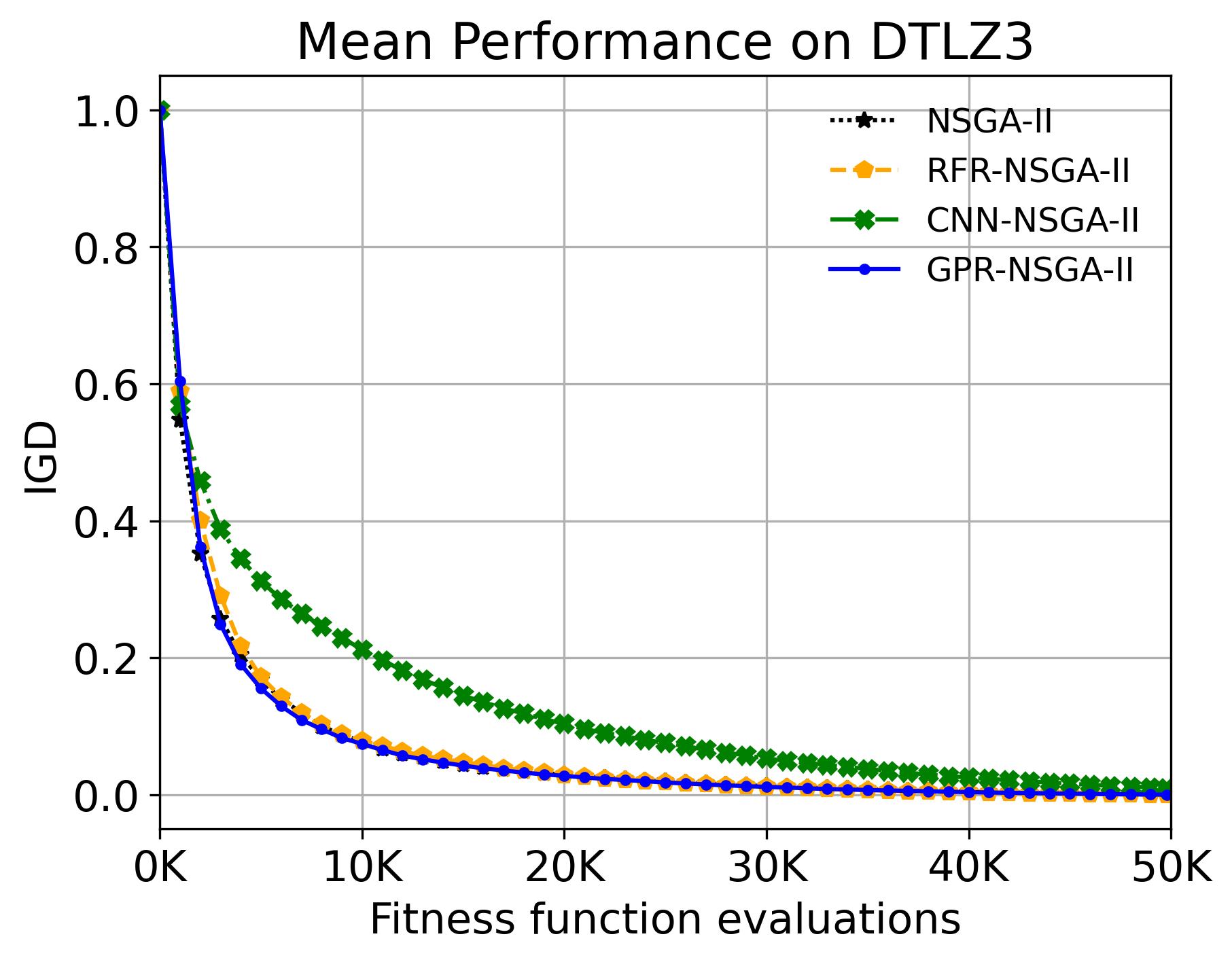}
    \end{subfigure}    
    \begin{subfigure}[t]{\figwid\textwidth}
        \centering
        \includegraphics[width=\linewidth]{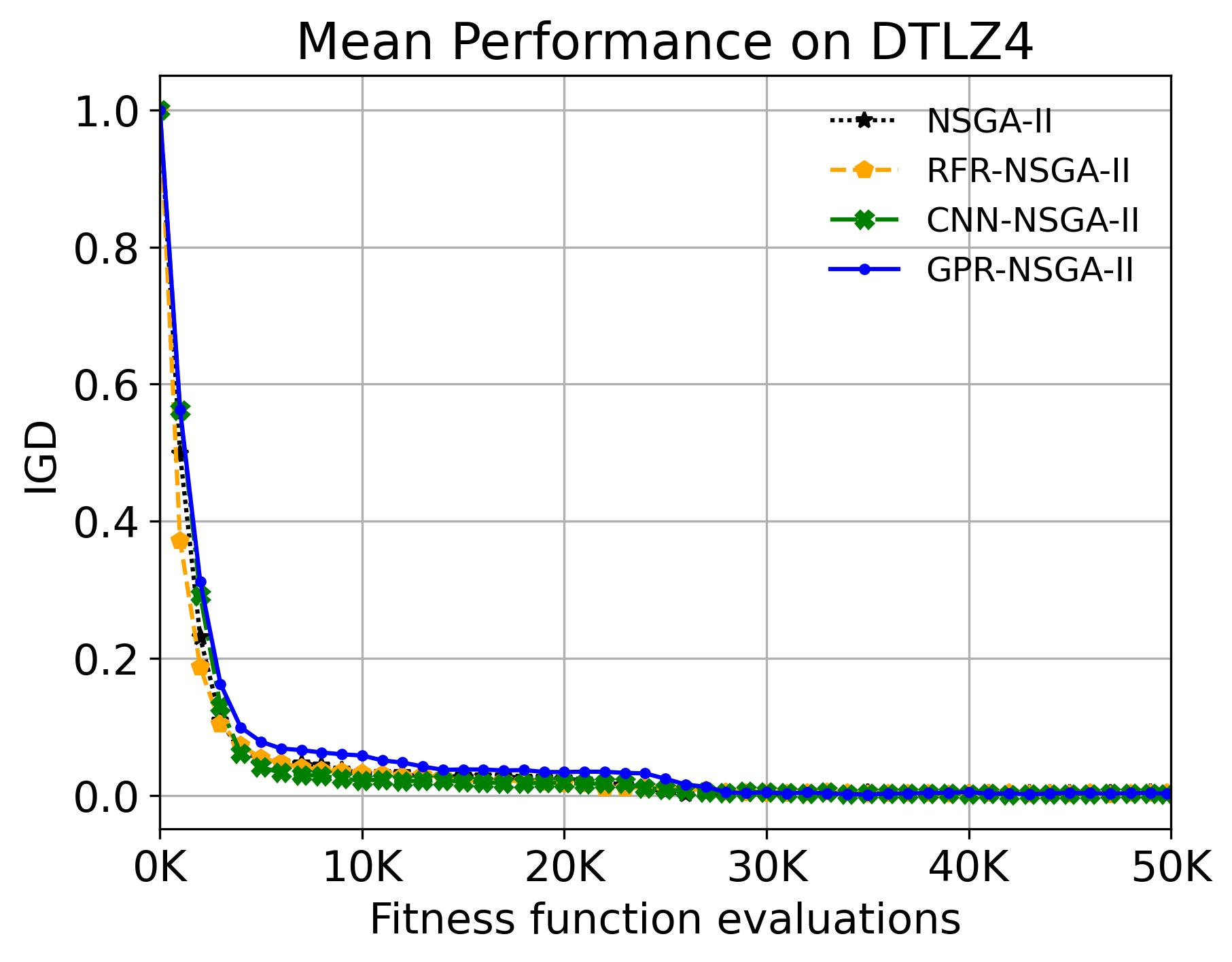}
    \end{subfigure}
    \begin{subfigure}[t]{\figwid\textwidth}
        \centering
        \includegraphics[width=\linewidth]{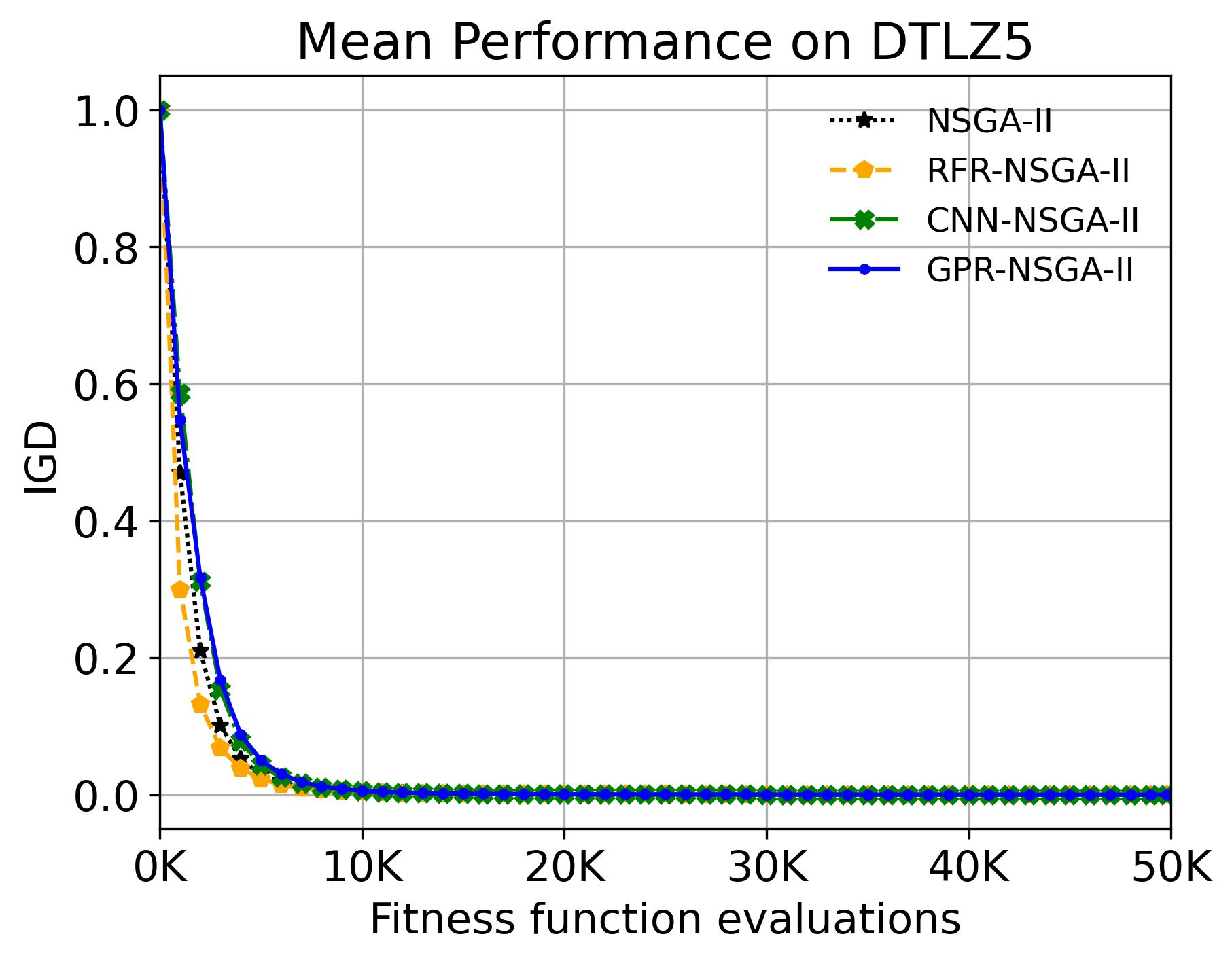}
    \end{subfigure}
    \begin{subfigure}[t]{\figwid\textwidth}
        \centering
        \includegraphics[width=\linewidth]{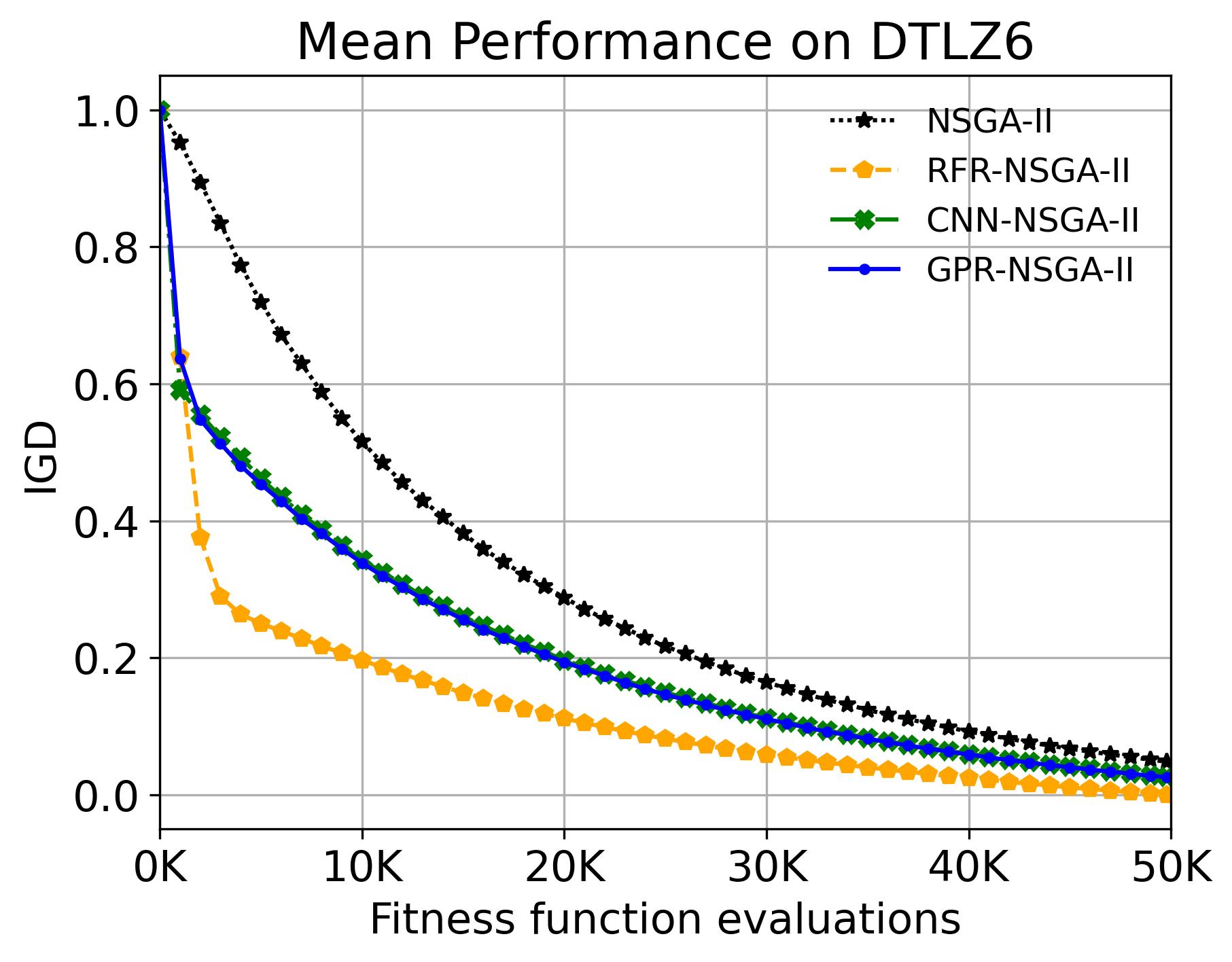}
    \end{subfigure}
    \begin{subfigure}[t]{\figwid\textwidth}
        \centering
        \includegraphics[width=\linewidth]{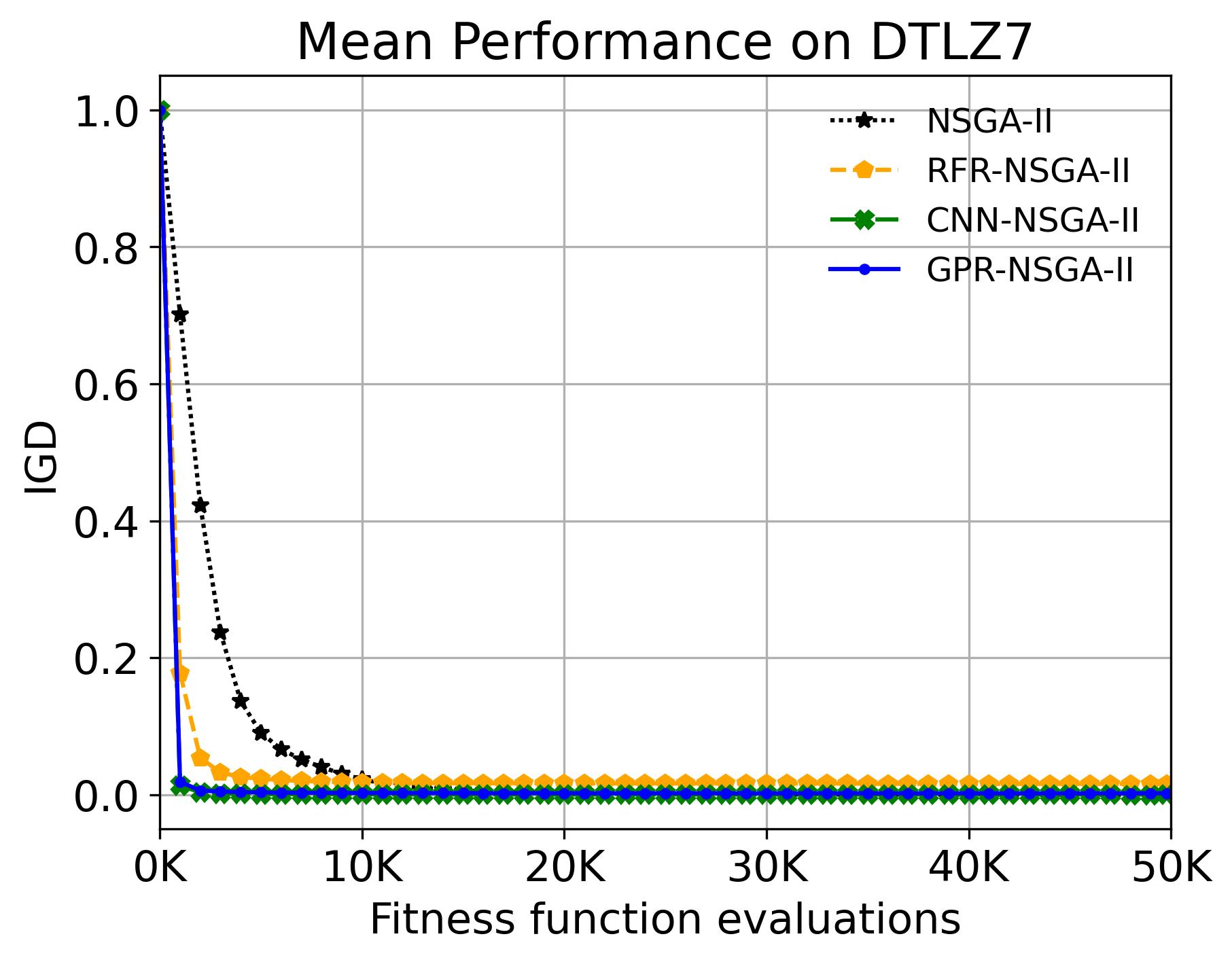}
    \end{subfigure}
    \begin{subfigure}[t]{\figwid\textwidth}
        \centering
        \includegraphics[width=\linewidth]{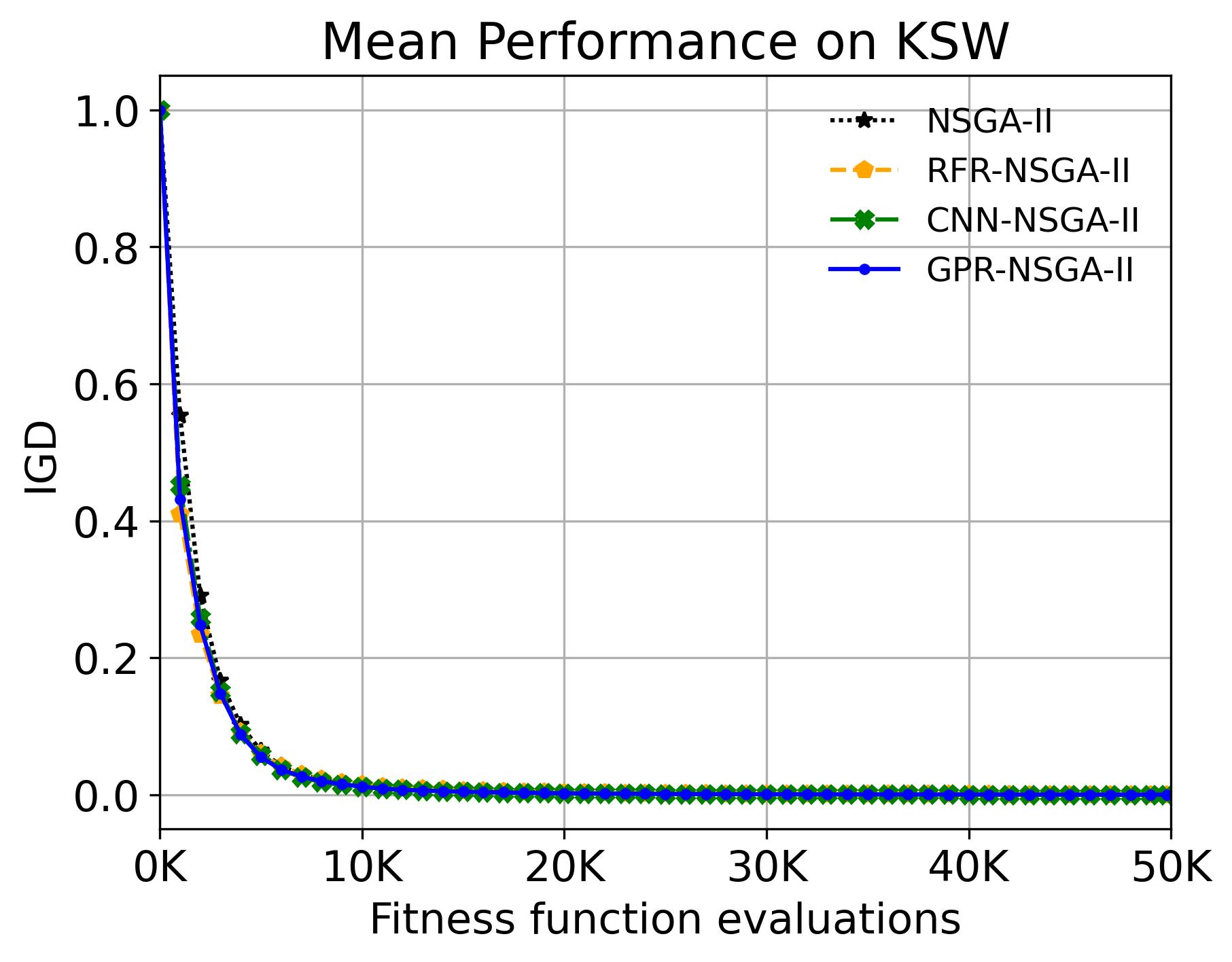}
    \end{subfigure}
    \begin{subfigure}[t]{\figwid\textwidth}
        \centering
        \includegraphics[width=\linewidth]{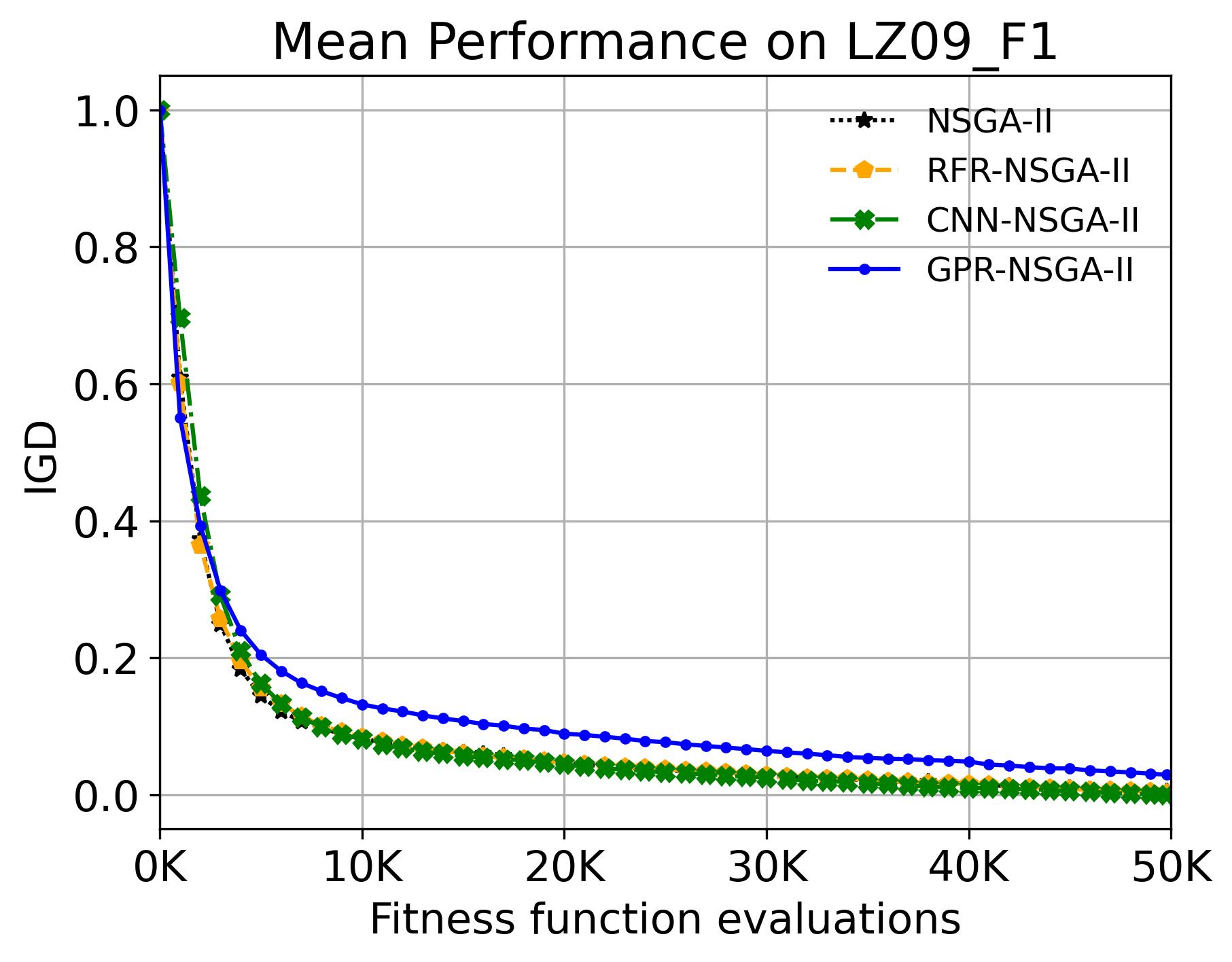}
    \end{subfigure}
    \begin{subfigure}[t]{\figwid\textwidth}
        \centering
        \includegraphics[width=\linewidth]{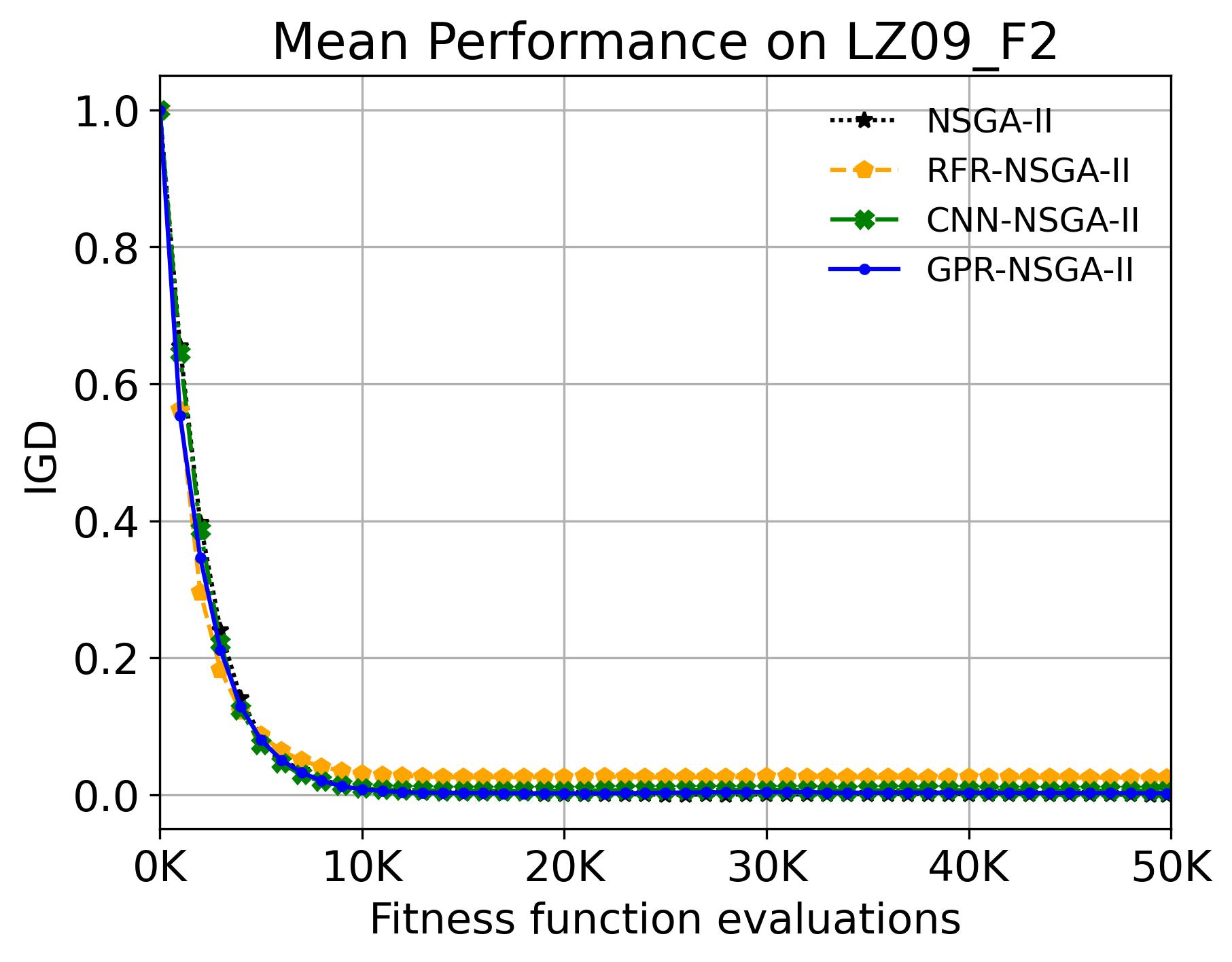}
    \end{subfigure}
    \begin{subfigure}[t]{\figwid\textwidth}
        \centering
        \includegraphics[width=\linewidth]{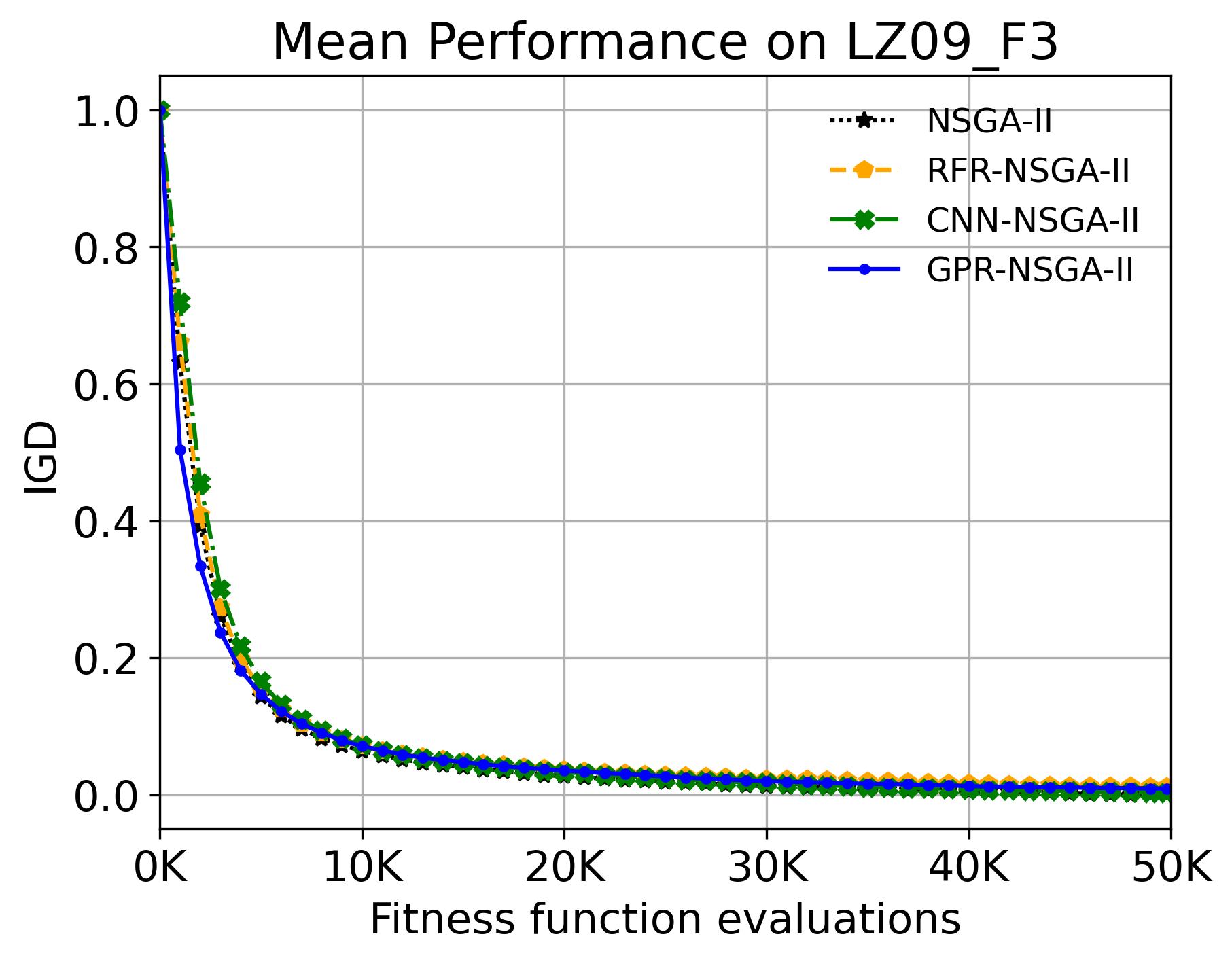}
    \end{subfigure}
    \begin{subfigure}[t]{\figwid\textwidth}
        \centering
        \includegraphics[width=\linewidth]{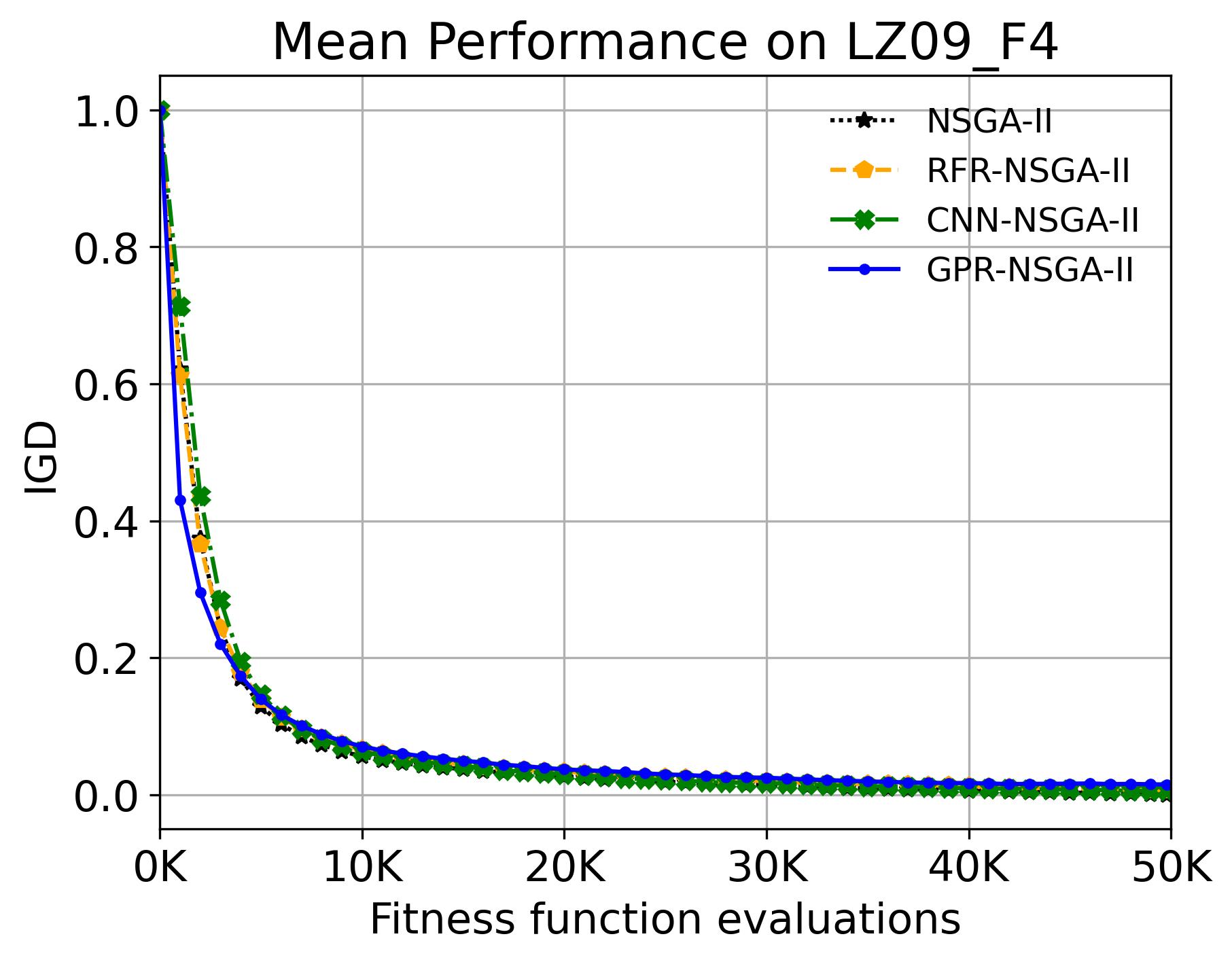}
    \end{subfigure}
    \begin{subfigure}[t]{\figwid\textwidth}
        \centering
        \includegraphics[width=\linewidth]{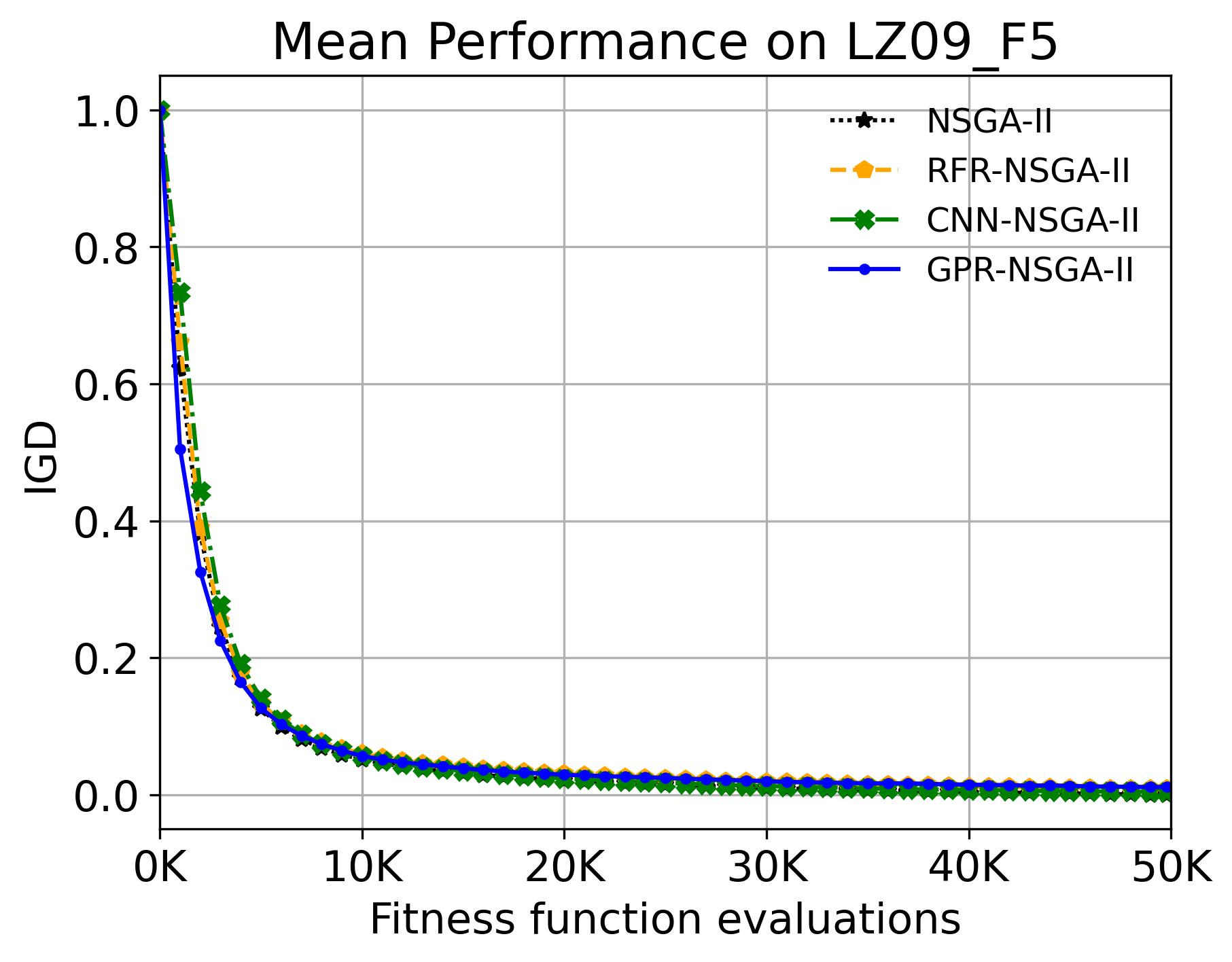}
    \end{subfigure}
    \begin{subfigure}[t]{\figwid\textwidth}
        \centering
        \includegraphics[width=\linewidth]{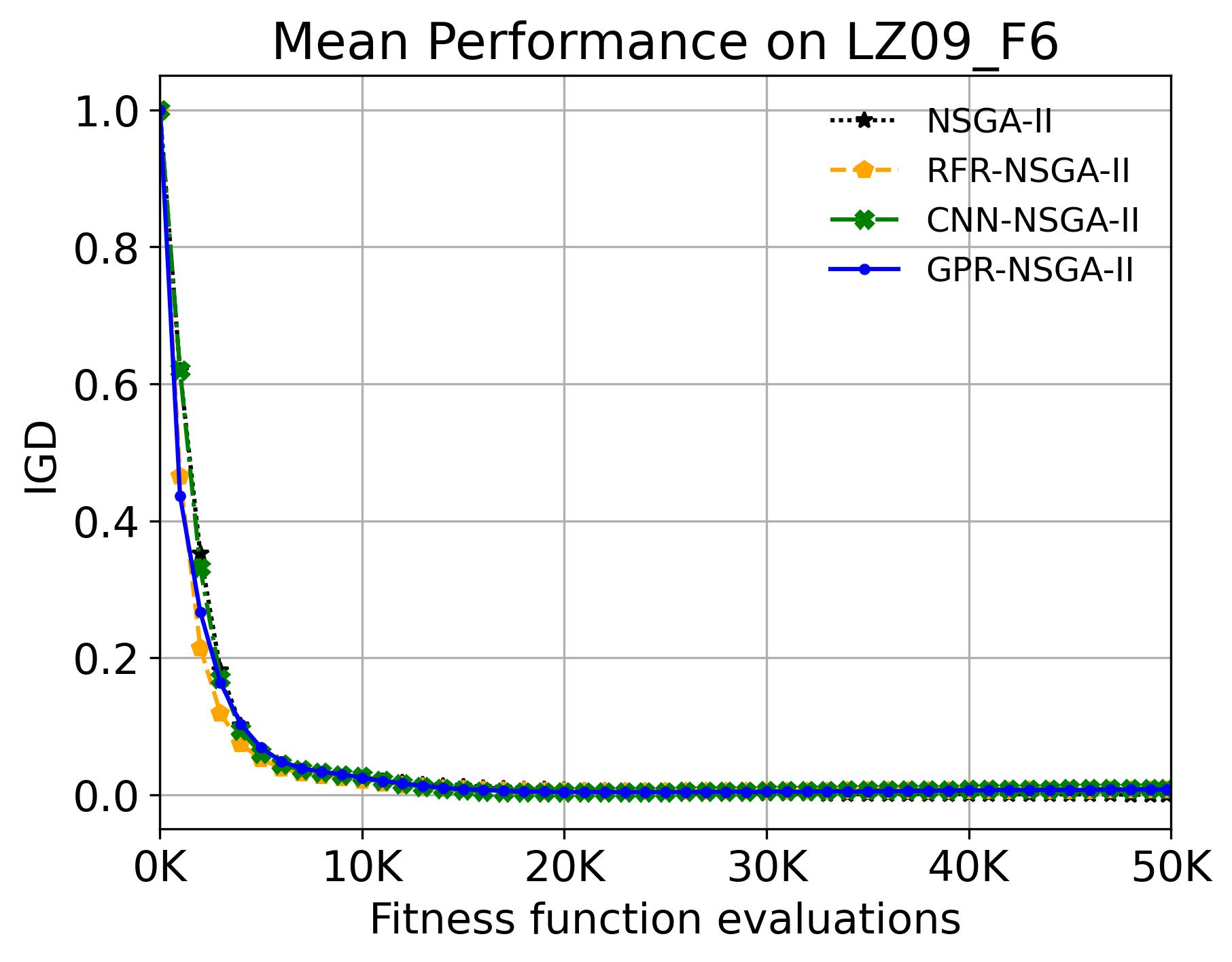}
    \end{subfigure}
    \begin{subfigure}[t]{\figwid\textwidth}
        \centering
        \includegraphics[width=\linewidth]{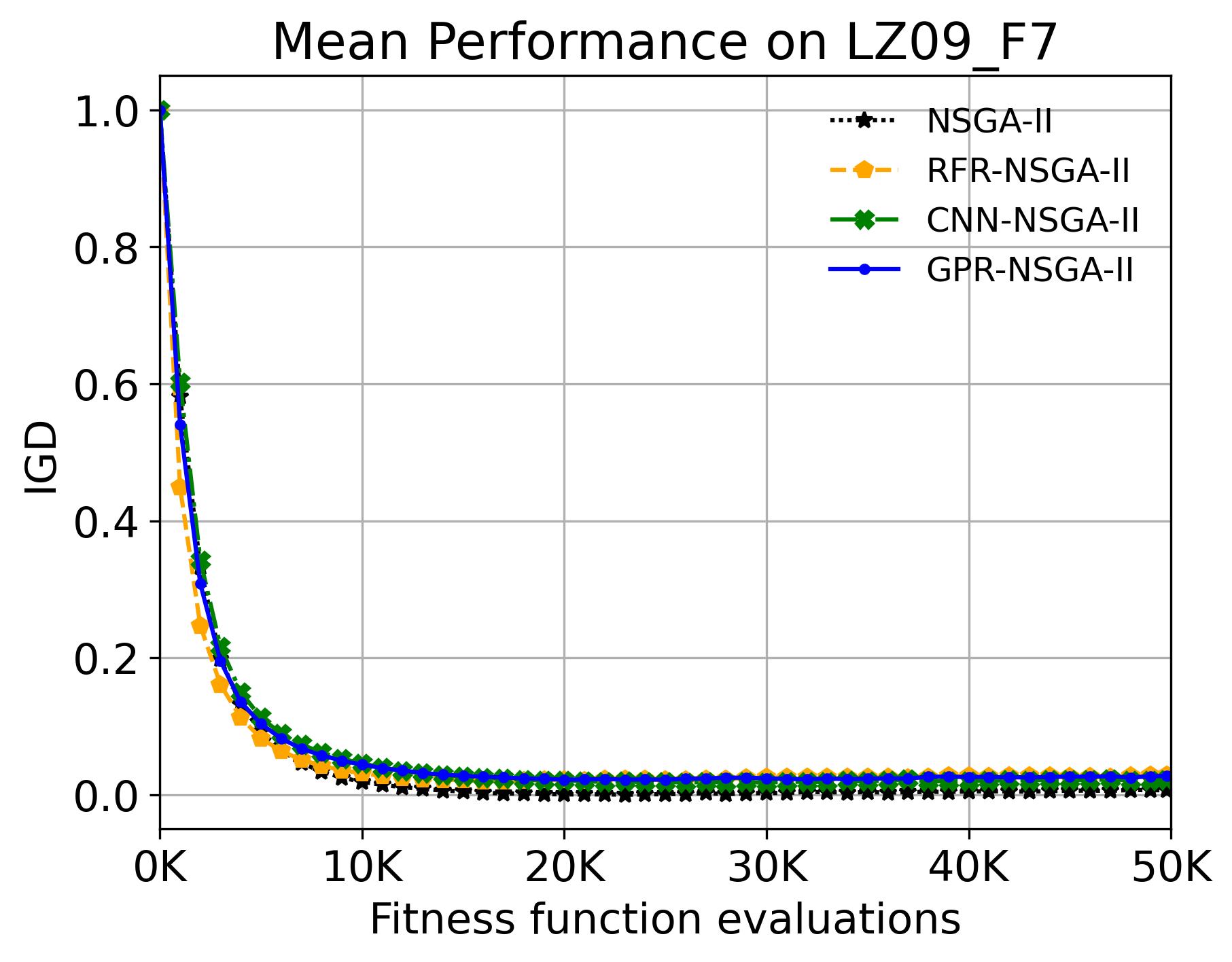}
    \end{subfigure}
    \begin{subfigure}[t]{\figwid\textwidth}
        \centering
        \includegraphics[width=\linewidth]{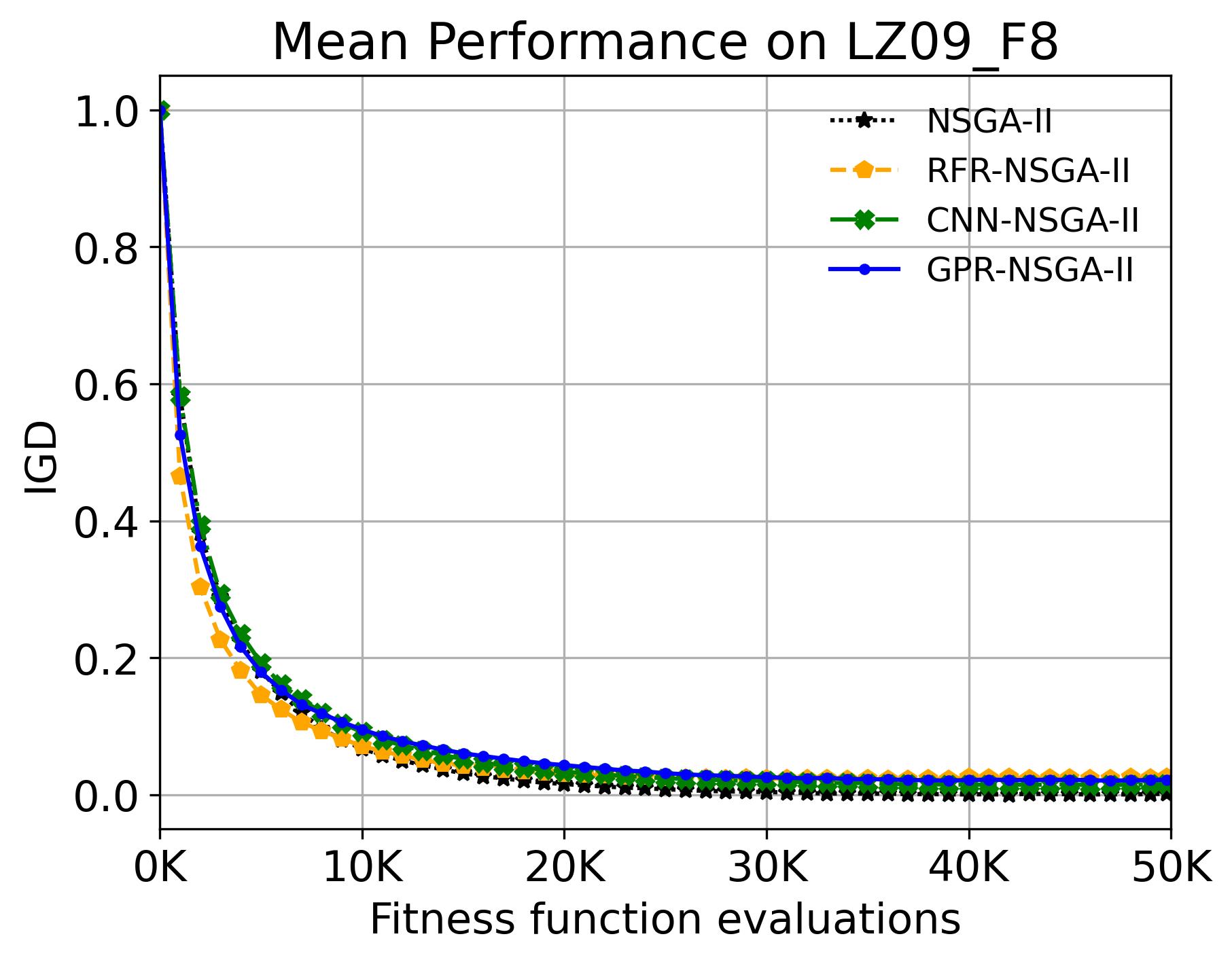}
    \end{subfigure}
    \begin{subfigure}[t]{\figwid\textwidth}
        \centering
        \includegraphics[width=\linewidth]{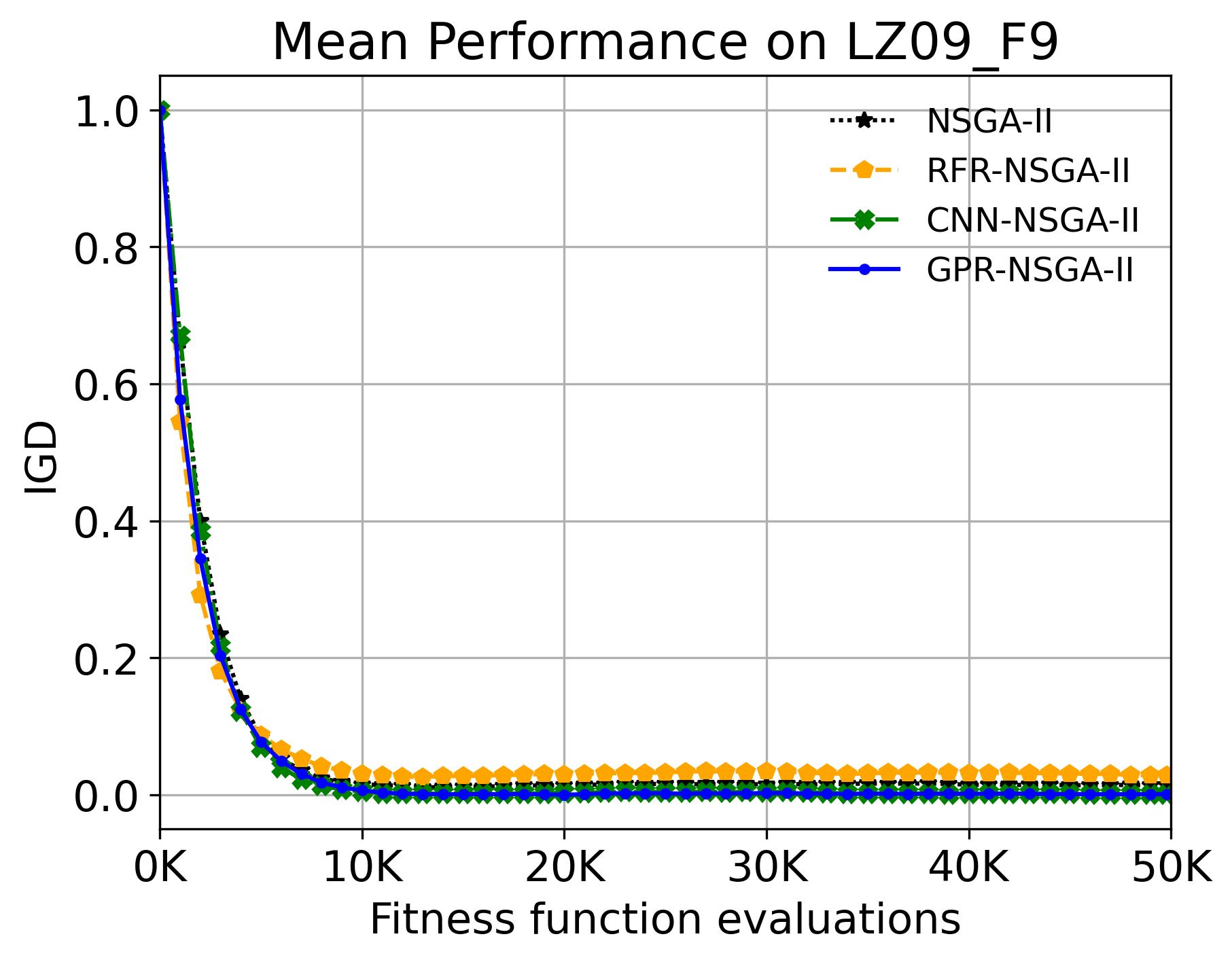}
    \end{subfigure}
    \begin{subfigure}[t]{\figwid\textwidth}
        \centering
        \includegraphics[width=\linewidth]{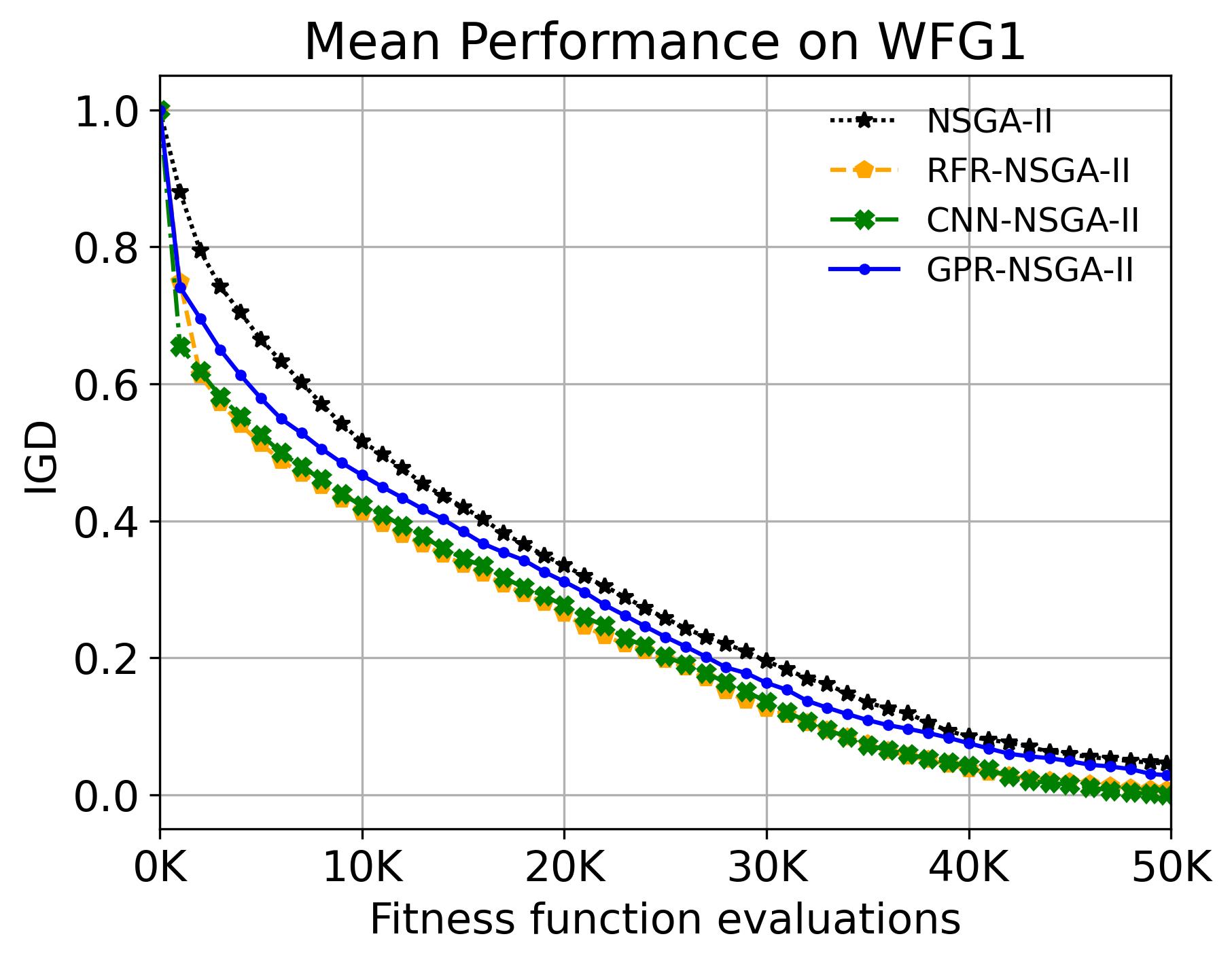}
    \end{subfigure}
    \begin{subfigure}[t]{\figwid\textwidth}
        \centering
        \includegraphics[width=\linewidth]{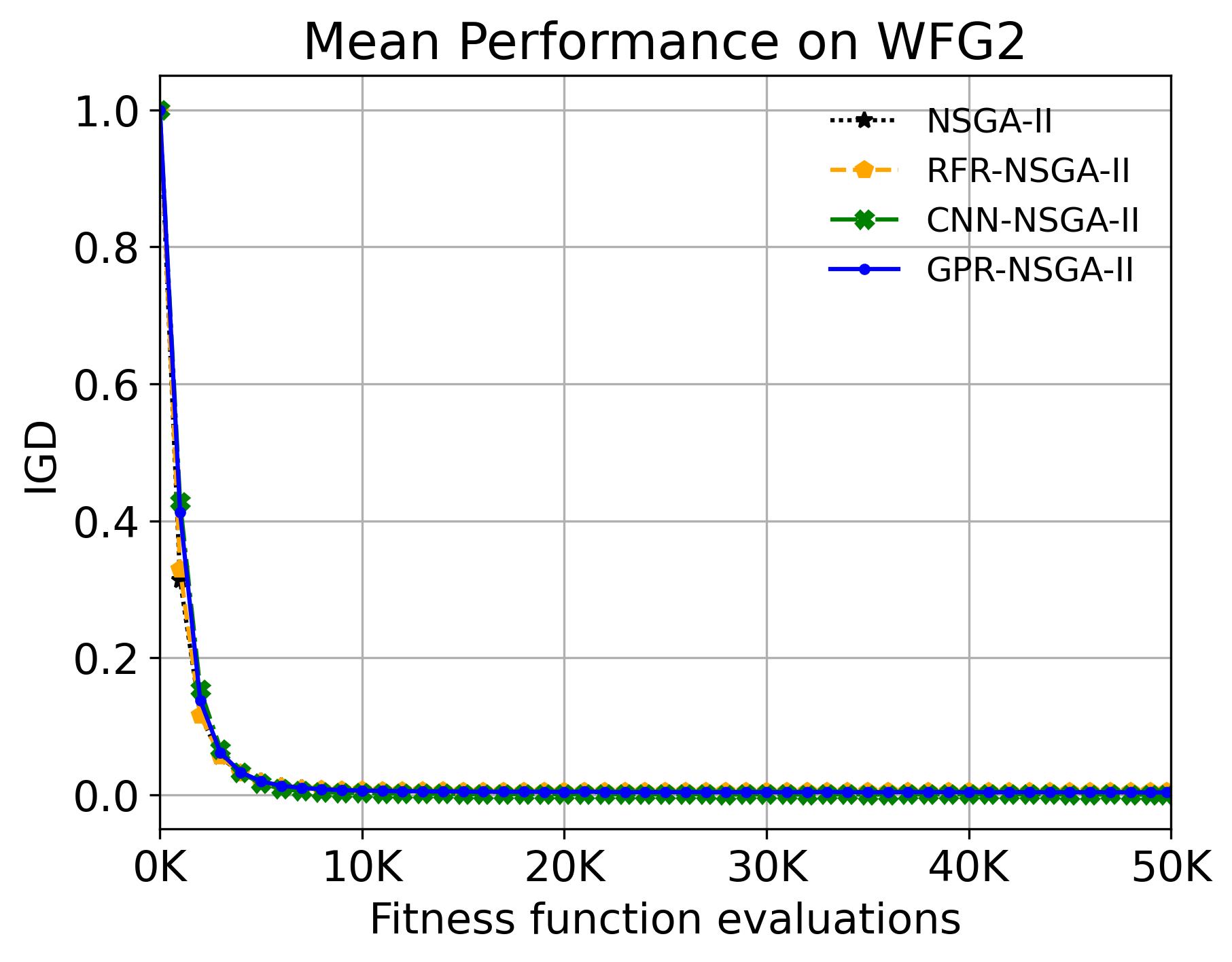}
    \end{subfigure}
    \begin{subfigure}[t]{\figwid\textwidth}
        \centering
        \includegraphics[width=\linewidth]{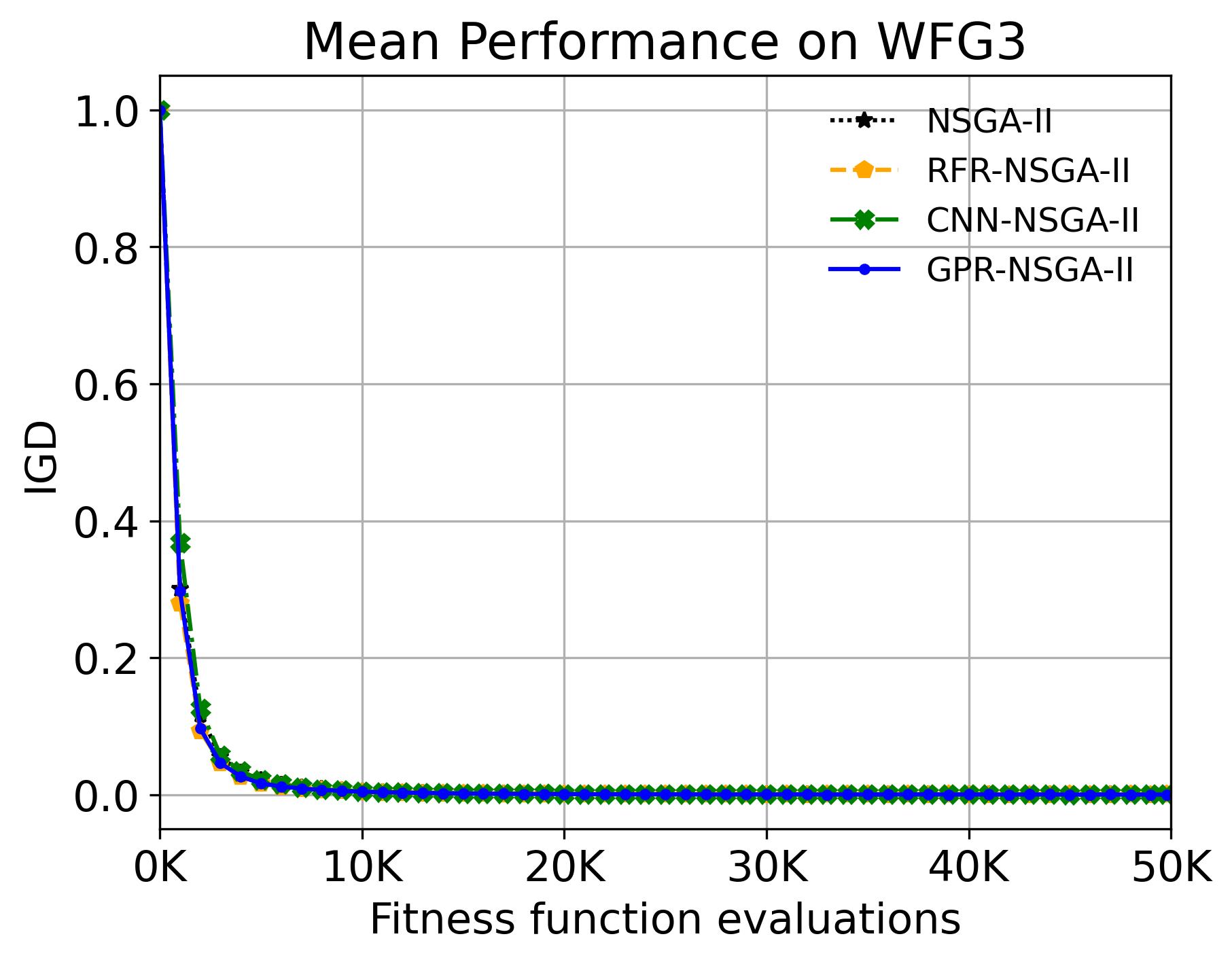}
    \end{subfigure}
    \begin{subfigure}[t]{\figwid\textwidth}
        \centering
        \includegraphics[width=\linewidth]{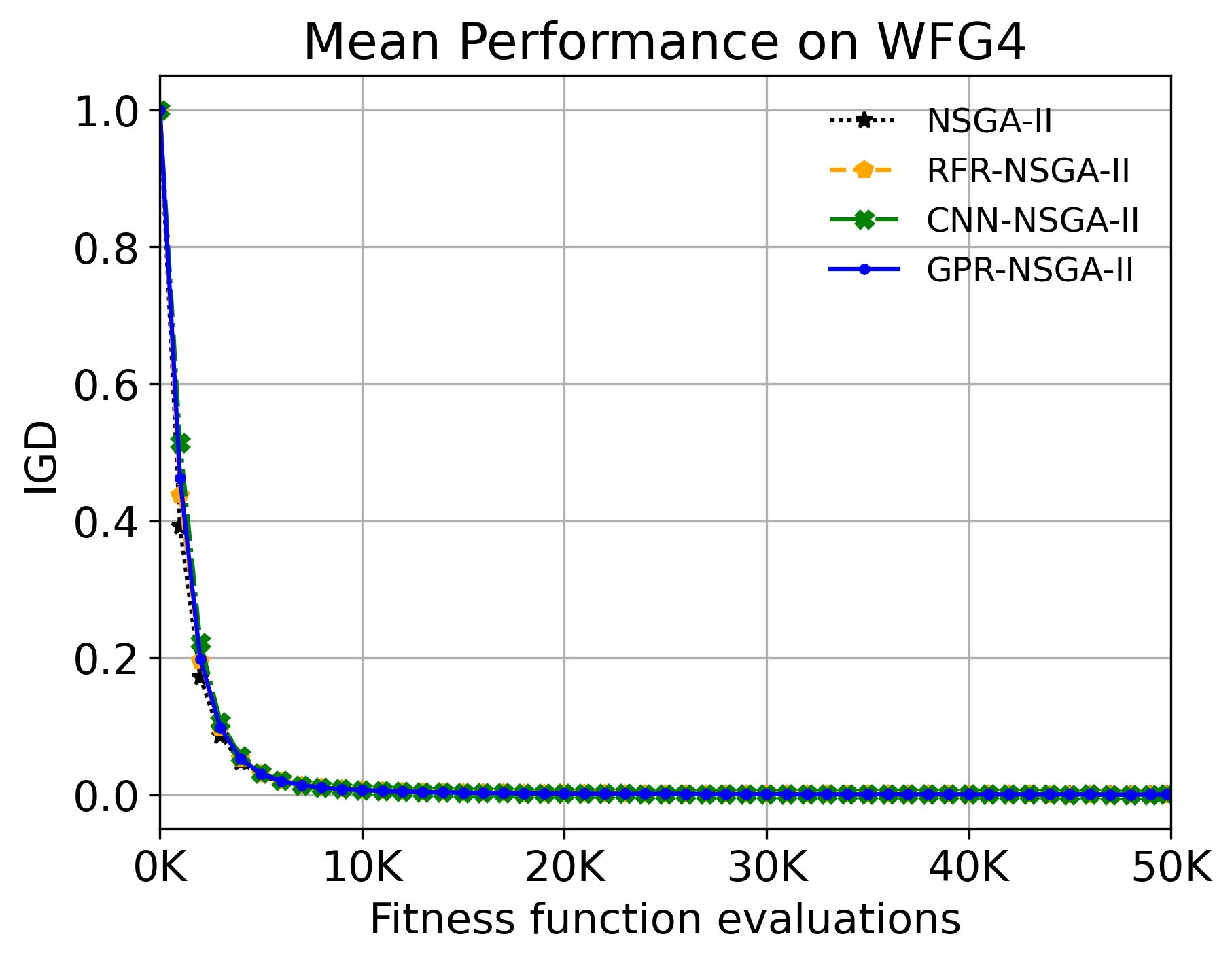}
    \end{subfigure}
    \begin{subfigure}[t]{\figwid\textwidth}
        \centering
        \includegraphics[width=\linewidth]{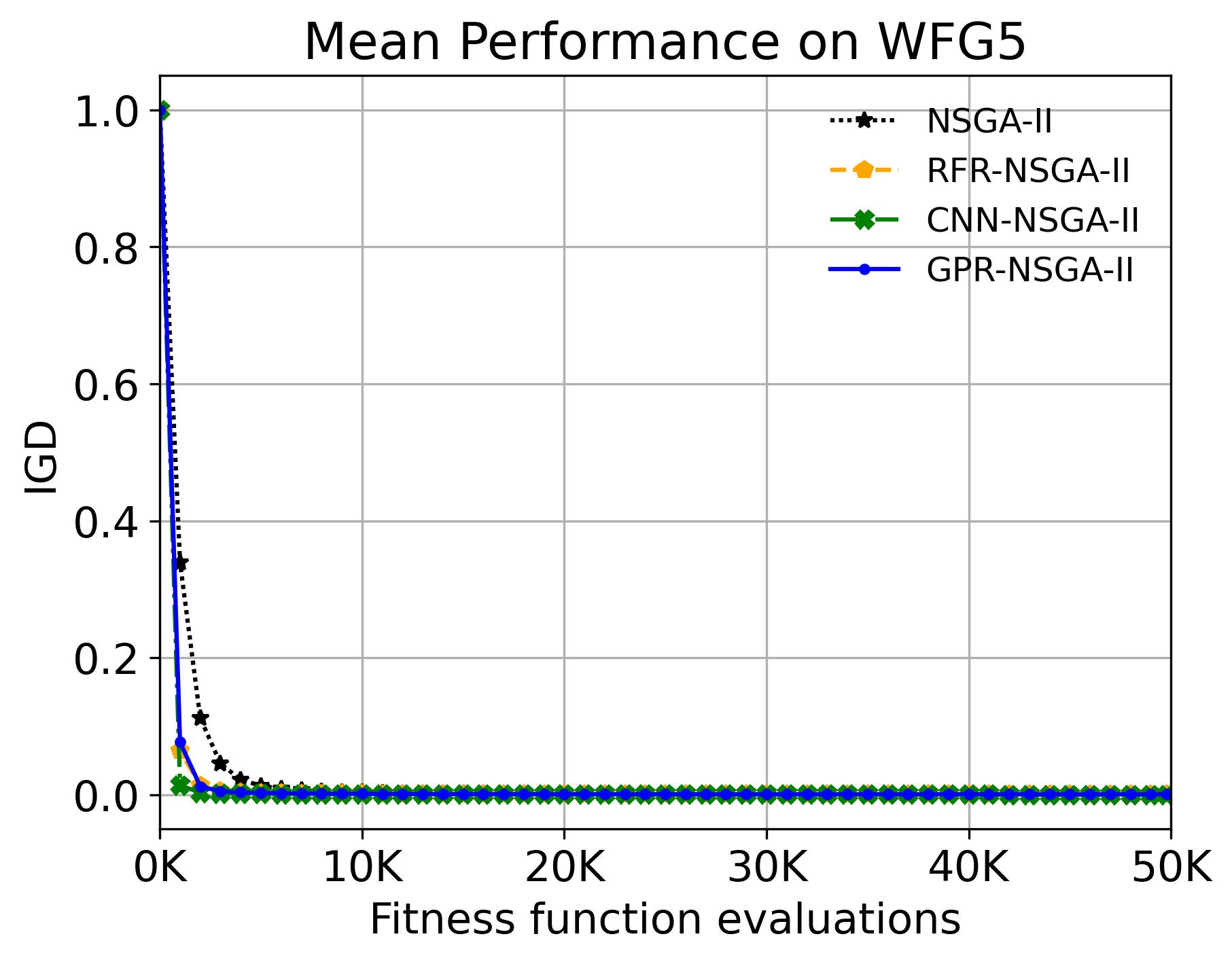}
    \end{subfigure}
    \begin{subfigure}[t]{\figwid\textwidth}
        \centering
        \includegraphics[width=\linewidth]{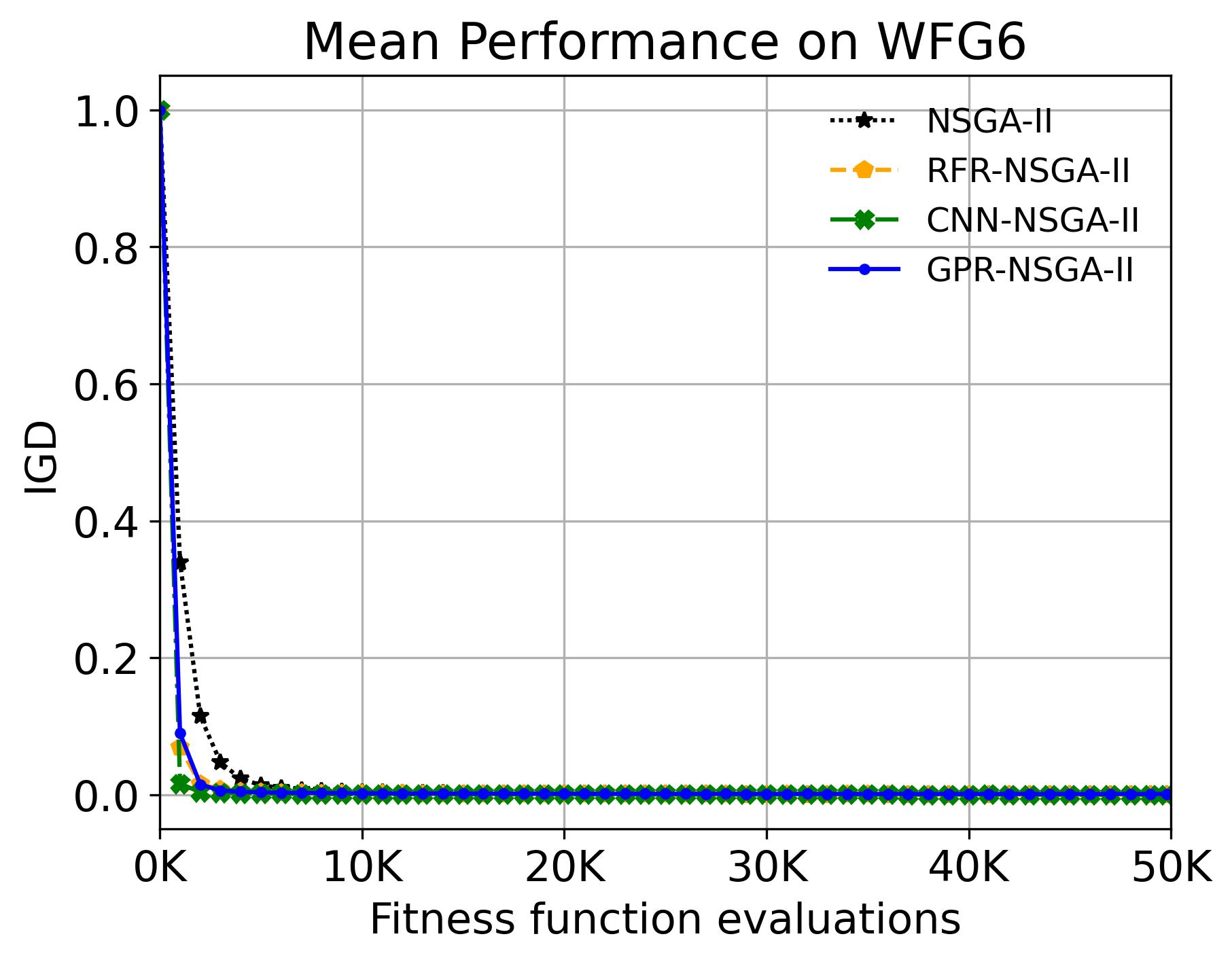}
    \end{subfigure}
    \begin{subfigure}[t]{\figwid\textwidth}
        \centering
        \includegraphics[width=\linewidth]{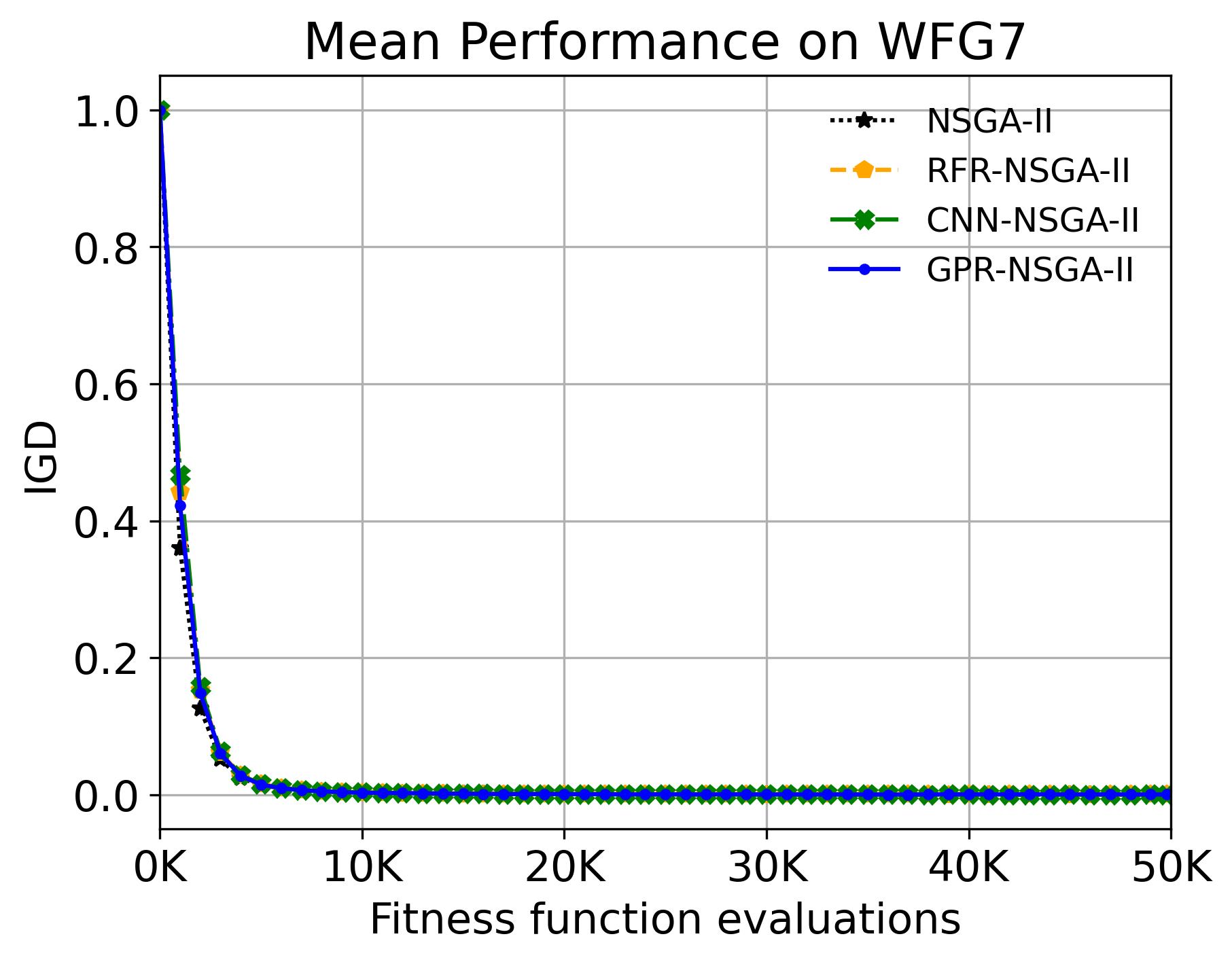}
    \end{subfigure}
    \begin{subfigure}[t]{\figwid\textwidth}
        \centering
        \includegraphics[width=\linewidth]{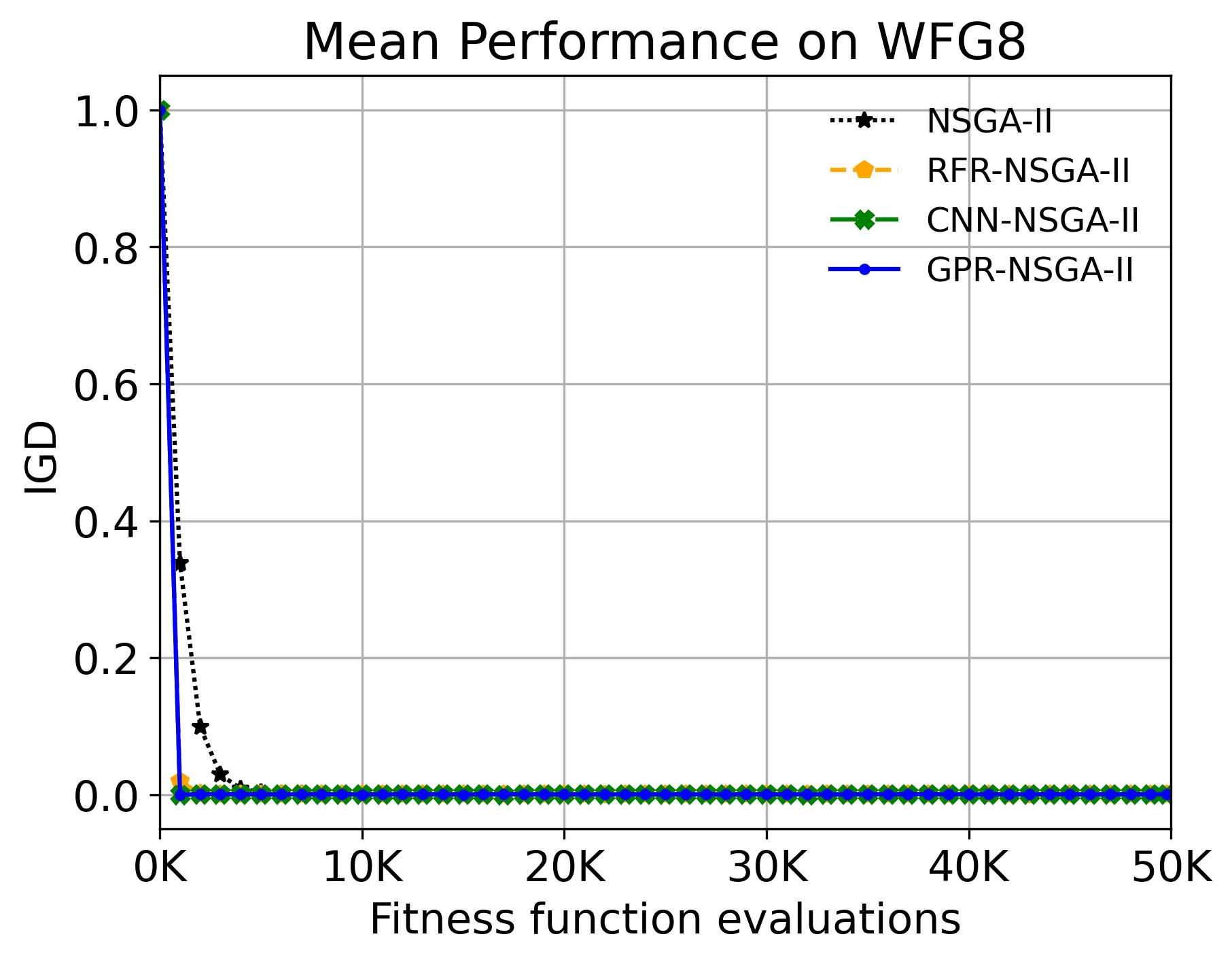}
    \end{subfigure}
    \begin{subfigure}[t]{\figwid\textwidth}
        \centering
        \includegraphics[width=\linewidth]{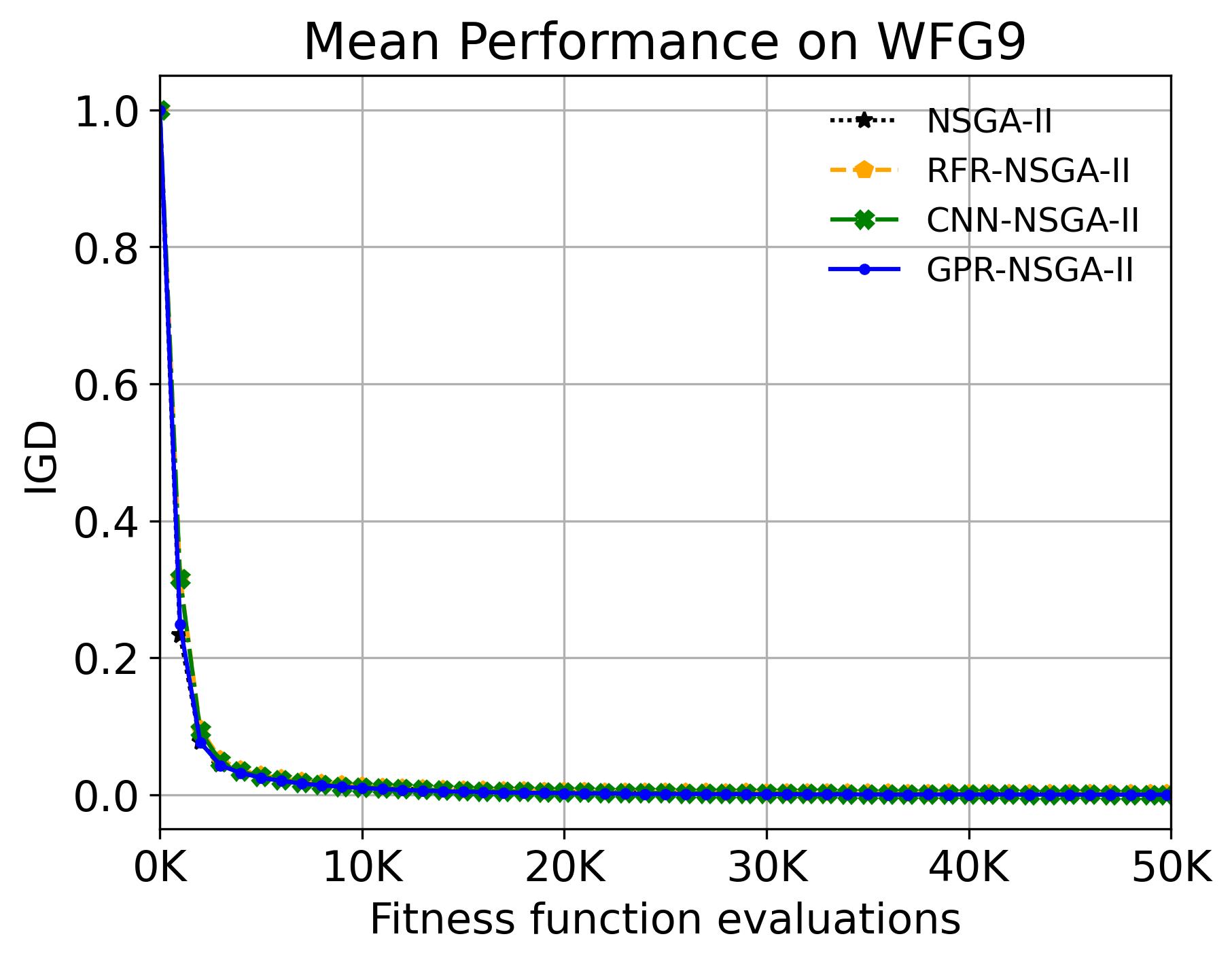}
    \end{subfigure}
    \begin{subfigure}[t]{\figwid\textwidth}
        \centering
        \includegraphics[width=\linewidth]{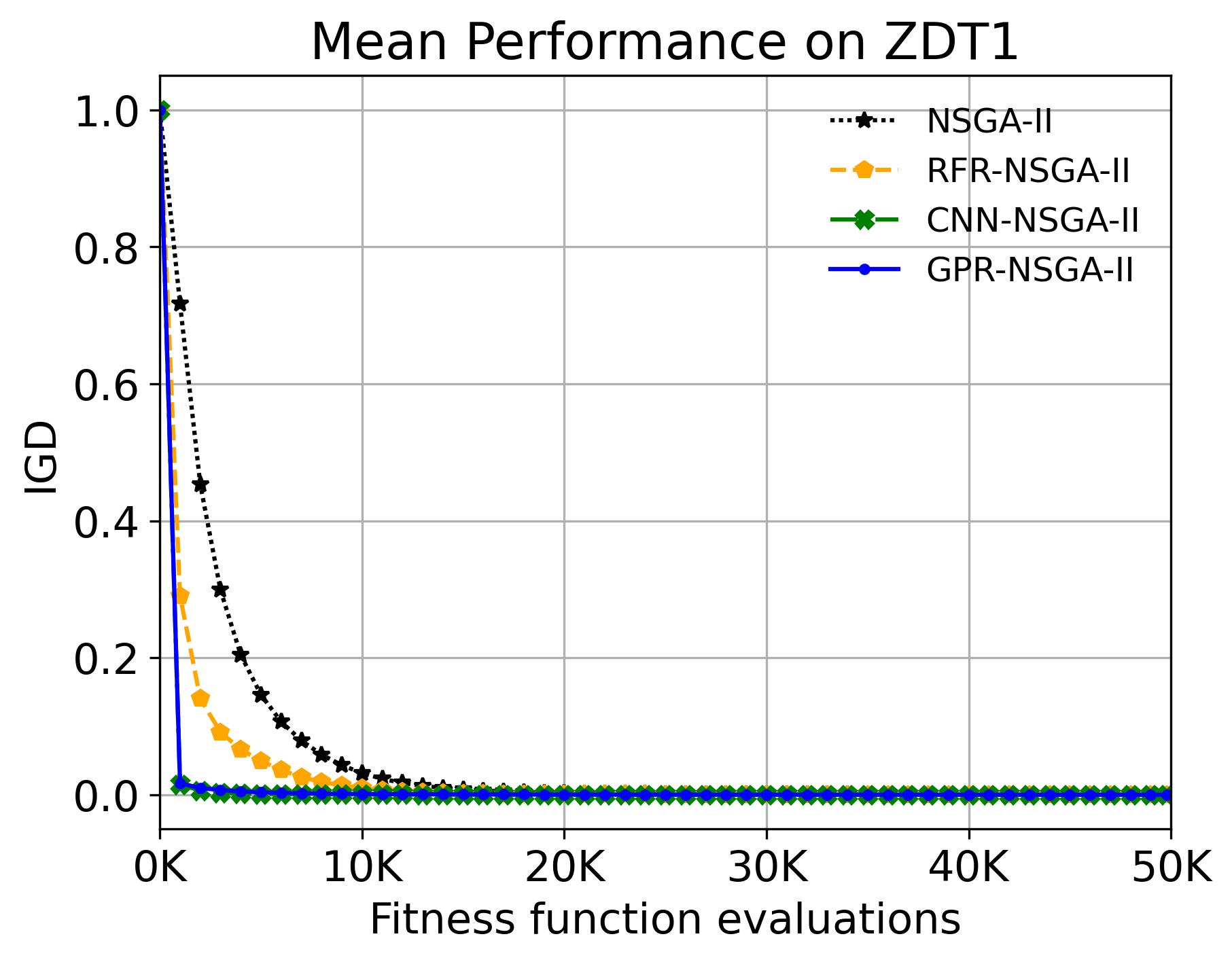}
    \end{subfigure}
    \begin{subfigure}[t]{\figwid\textwidth}
        \centering
        \includegraphics[width=\linewidth]{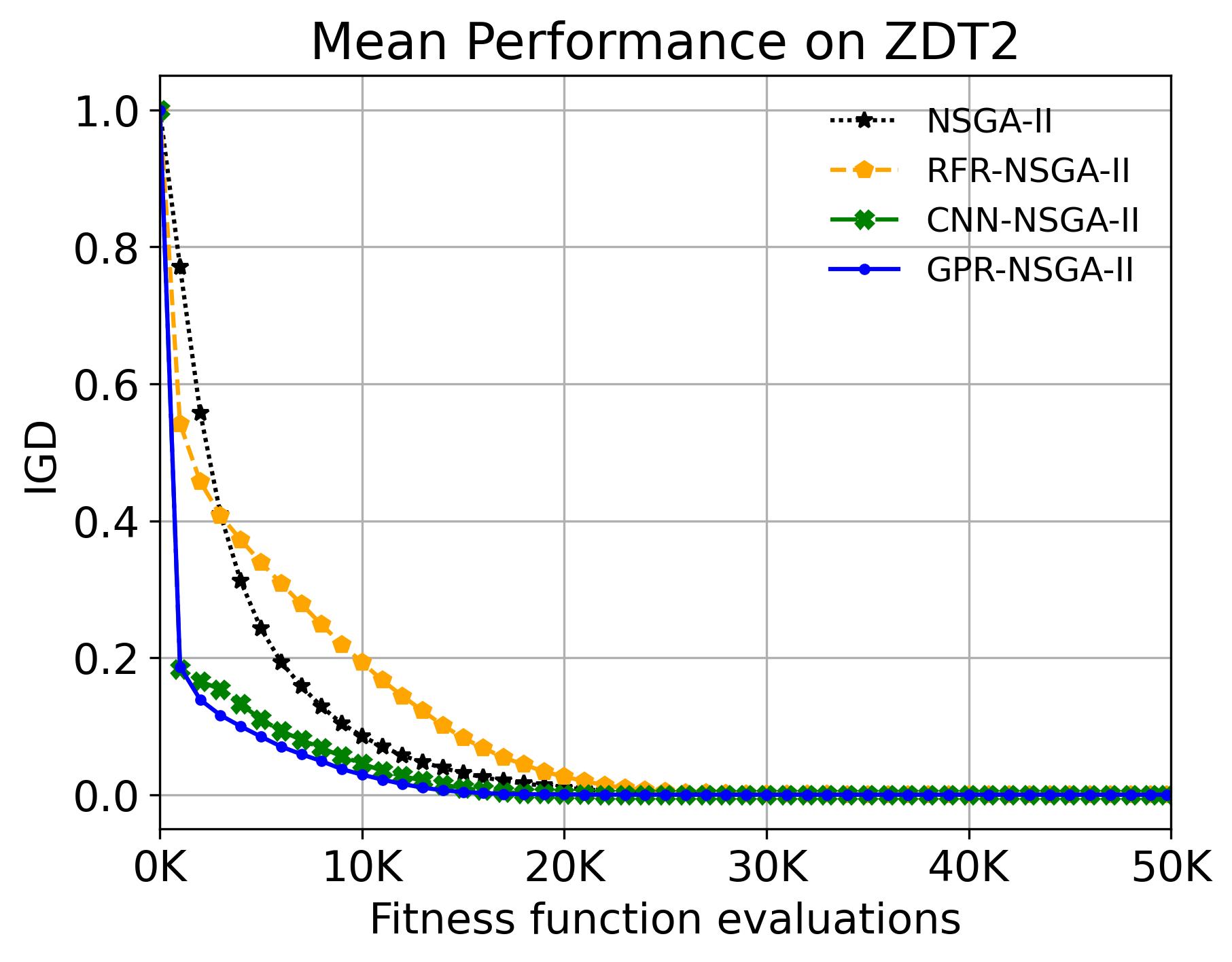}
    \end{subfigure}
    \begin{subfigure}[t]{\figwid\textwidth}
        \centering
        \includegraphics[width=\linewidth]{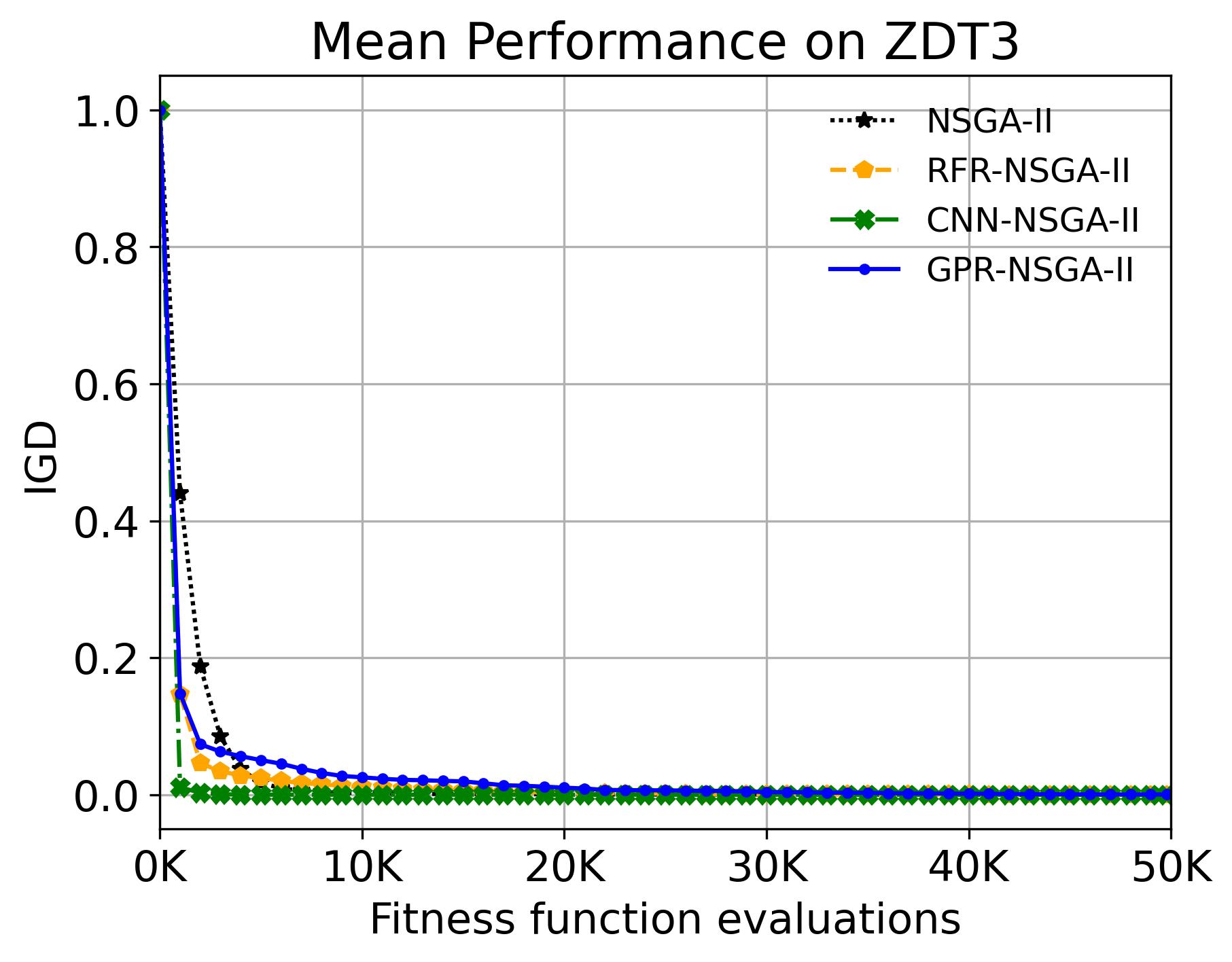}
    \end{subfigure}
    \begin{subfigure}[t]{\figwid\textwidth}
        \centering
        \includegraphics[width=\linewidth]{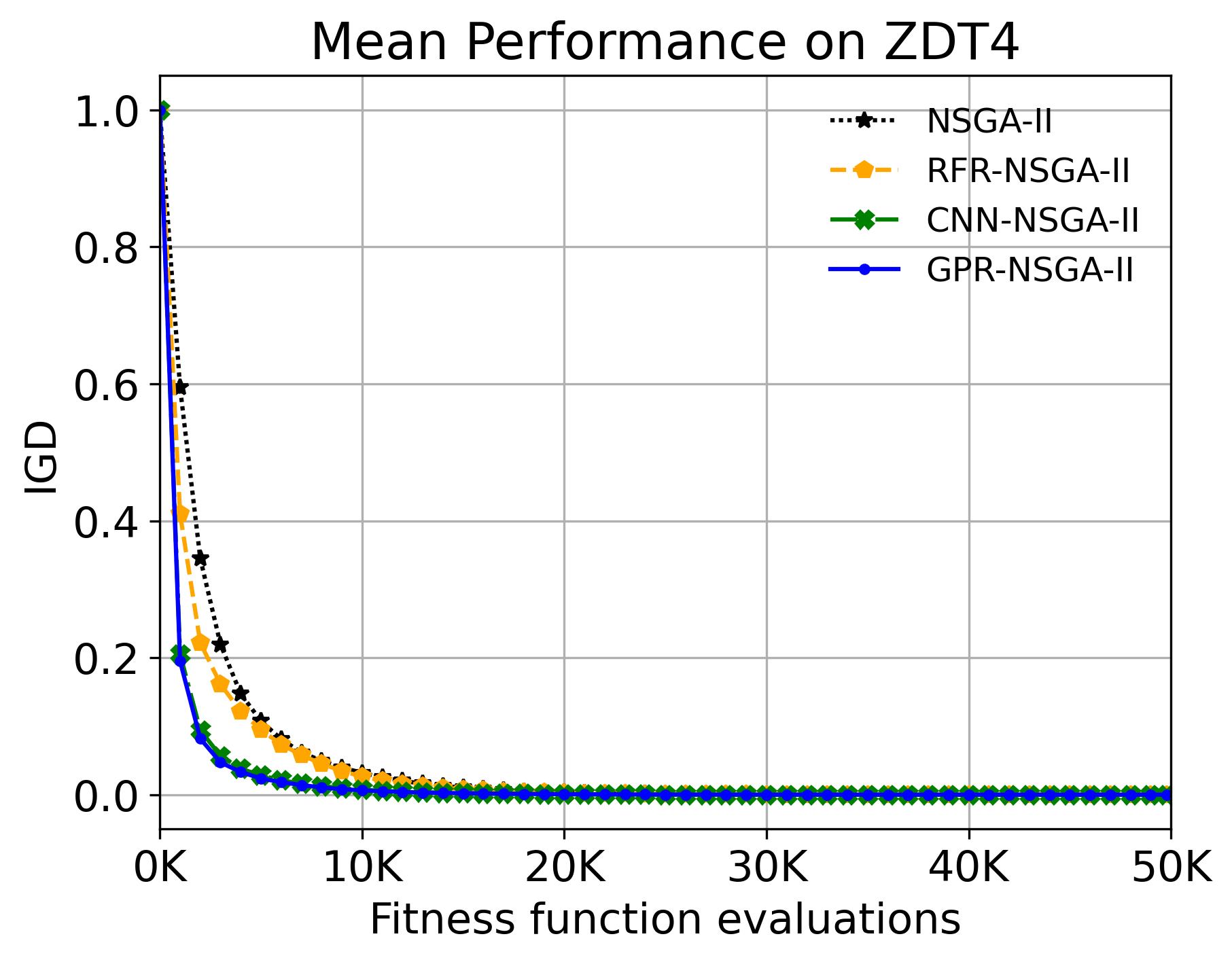}
    \end{subfigure}
    \begin{subfigure}[t]{\figwid\textwidth}
        \centering
        \includegraphics[width=\linewidth]{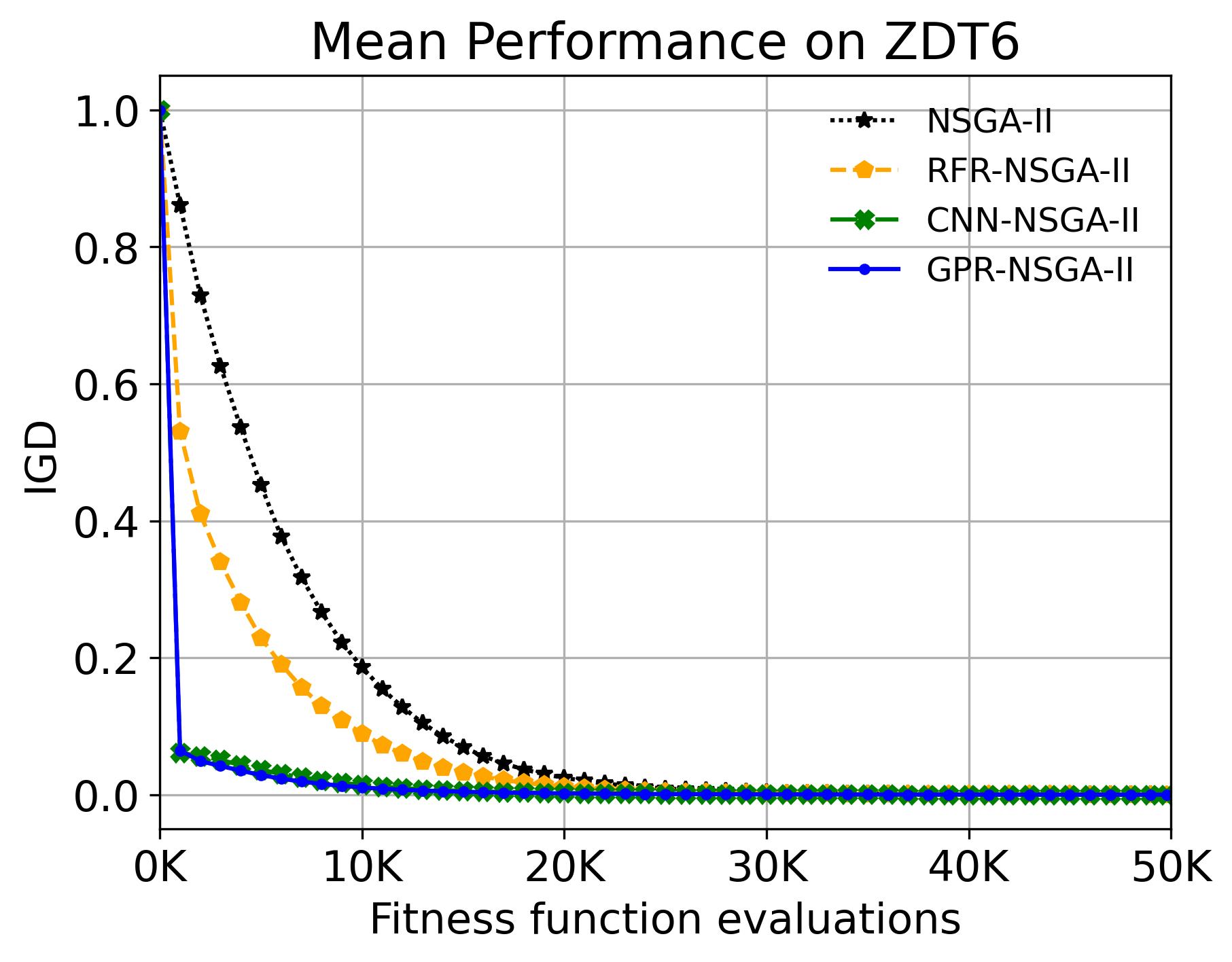}
    \end{subfigure}

        \caption{Comparison of NSGA-II and its associated surrogate-enhanced solvers on individual benchmark problems using  $IGD(PF_c)$ -- i.e., the normalised inverse generational distance -- as a performance indicator.}
        \label{fig:annexNSGAII_igd}
\end{figure*}

\begin{figure*}[h]
    \centering
    \begin{subfigure}[t]{\figwid\textwidth}
        \centering
        \includegraphics[width=\linewidth]{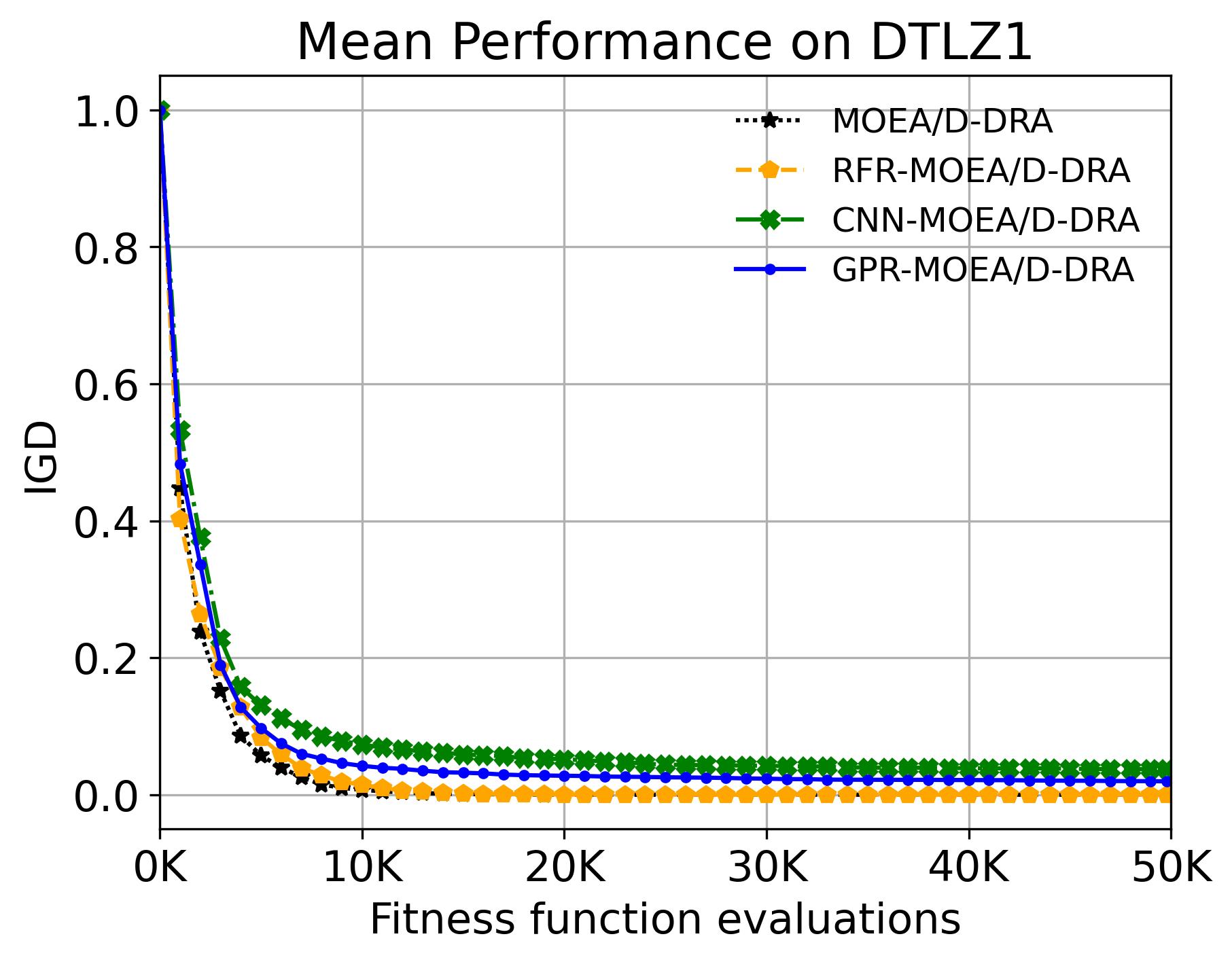}
    \end{subfigure}
    \begin{subfigure}[t]{\figwid\textwidth}
        \centering
        \includegraphics[width=\linewidth]{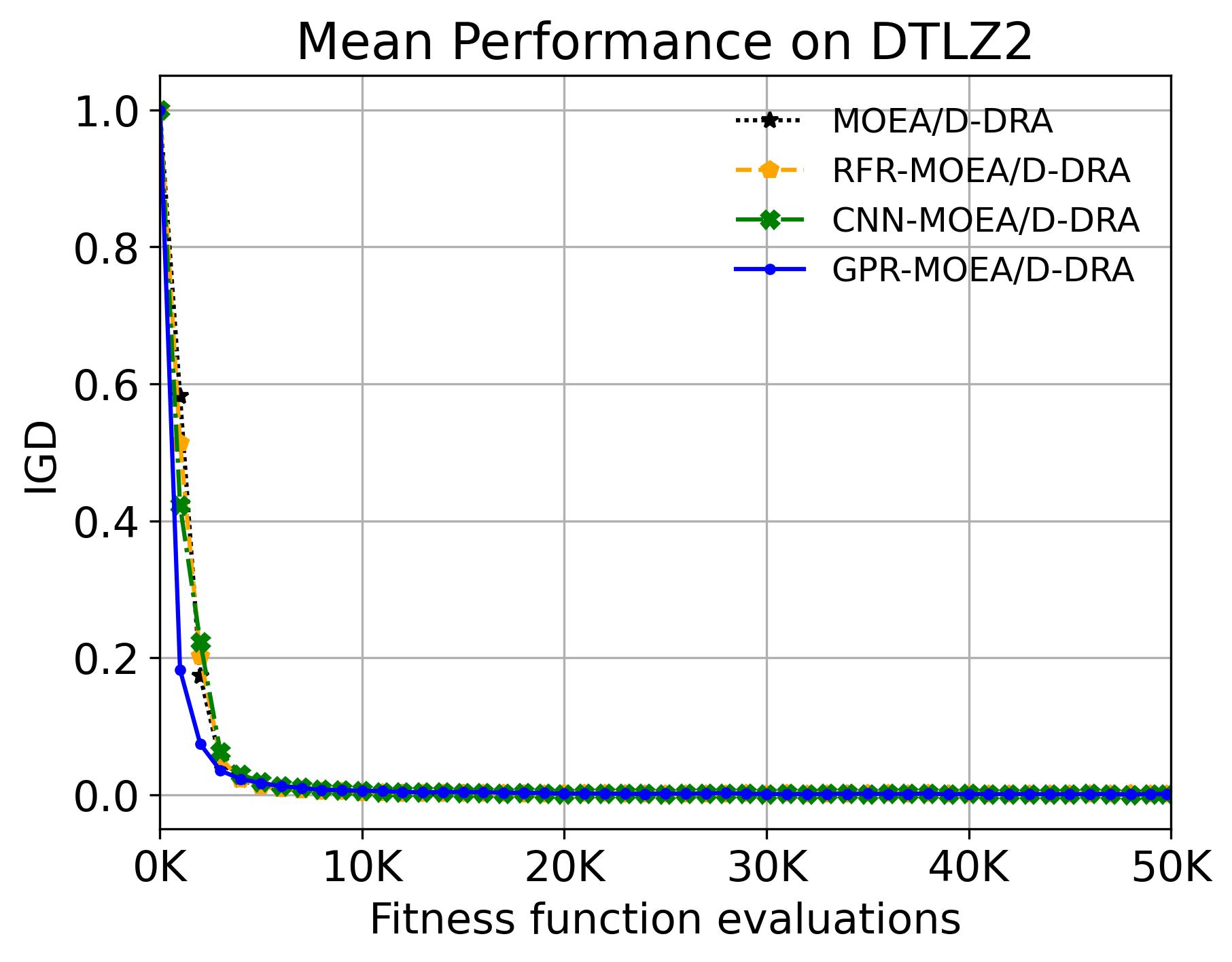}
    \end{subfigure}
    \begin{subfigure}[t]{\figwid\textwidth}
        \centering
        \includegraphics[width=\linewidth]{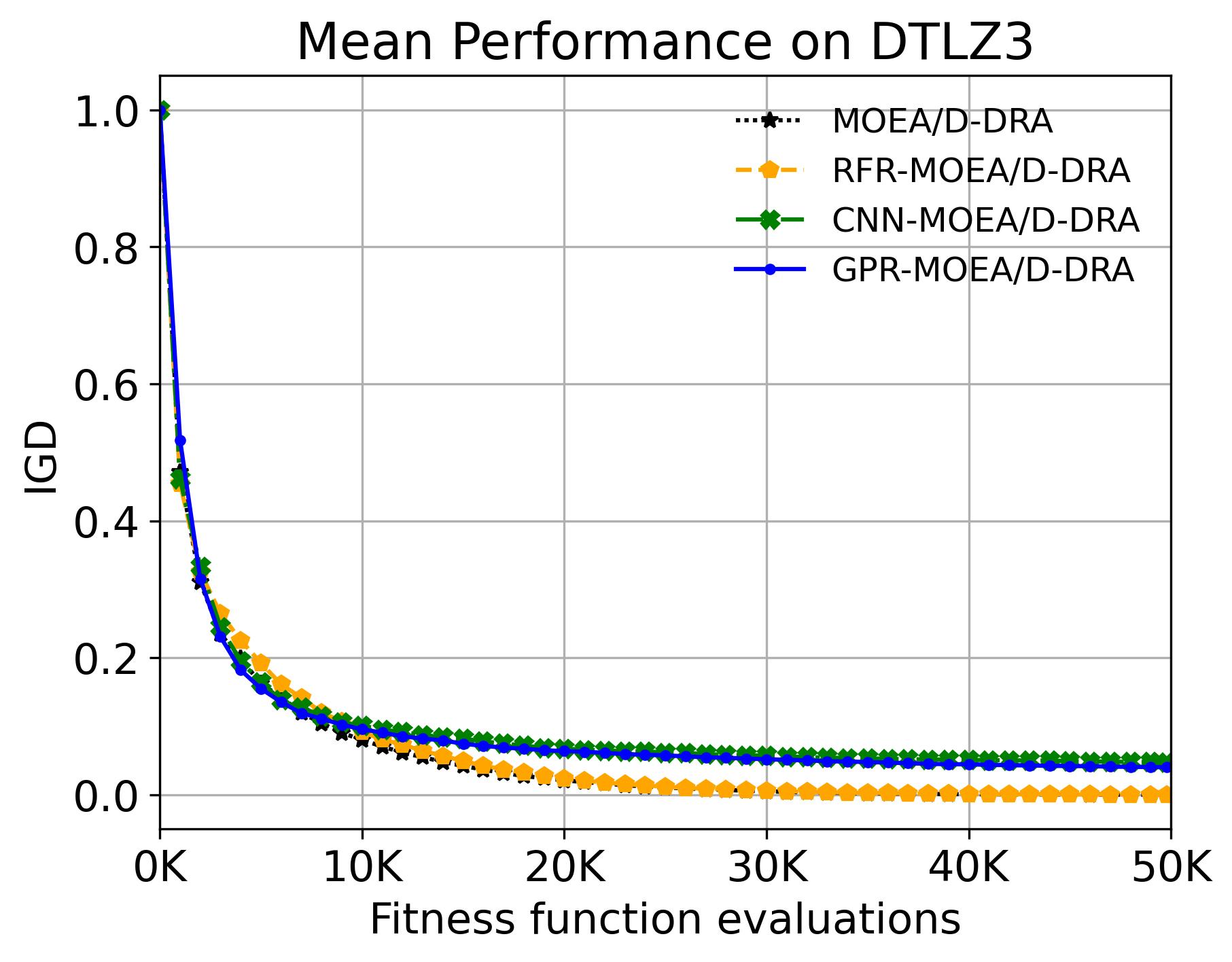}
    \end{subfigure}    
    \begin{subfigure}[t]{\figwid\textwidth}
        \centering
        \includegraphics[width=\linewidth]{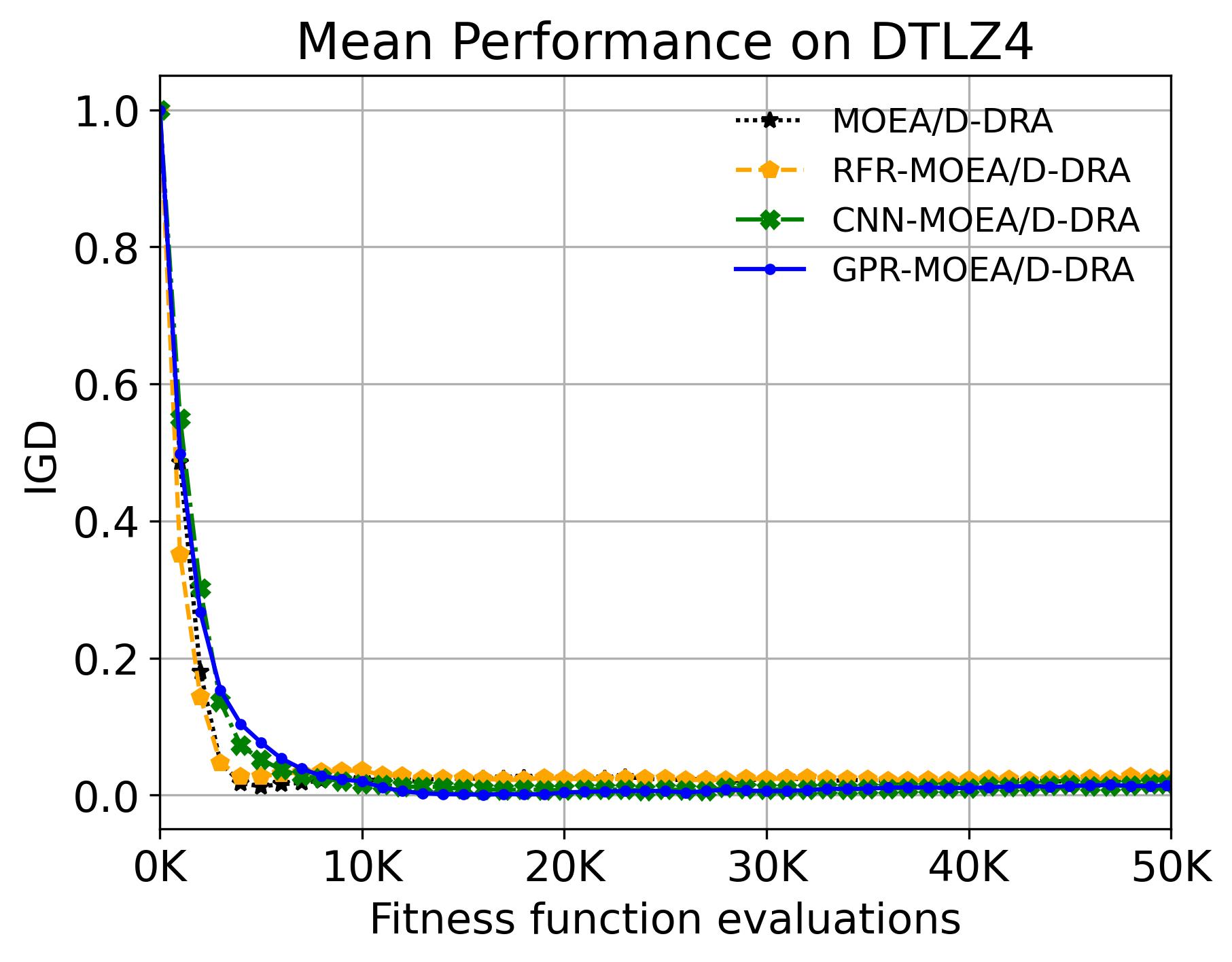}
    \end{subfigure}
    \begin{subfigure}[t]{\figwid\textwidth}
        \centering
        \includegraphics[width=\linewidth]{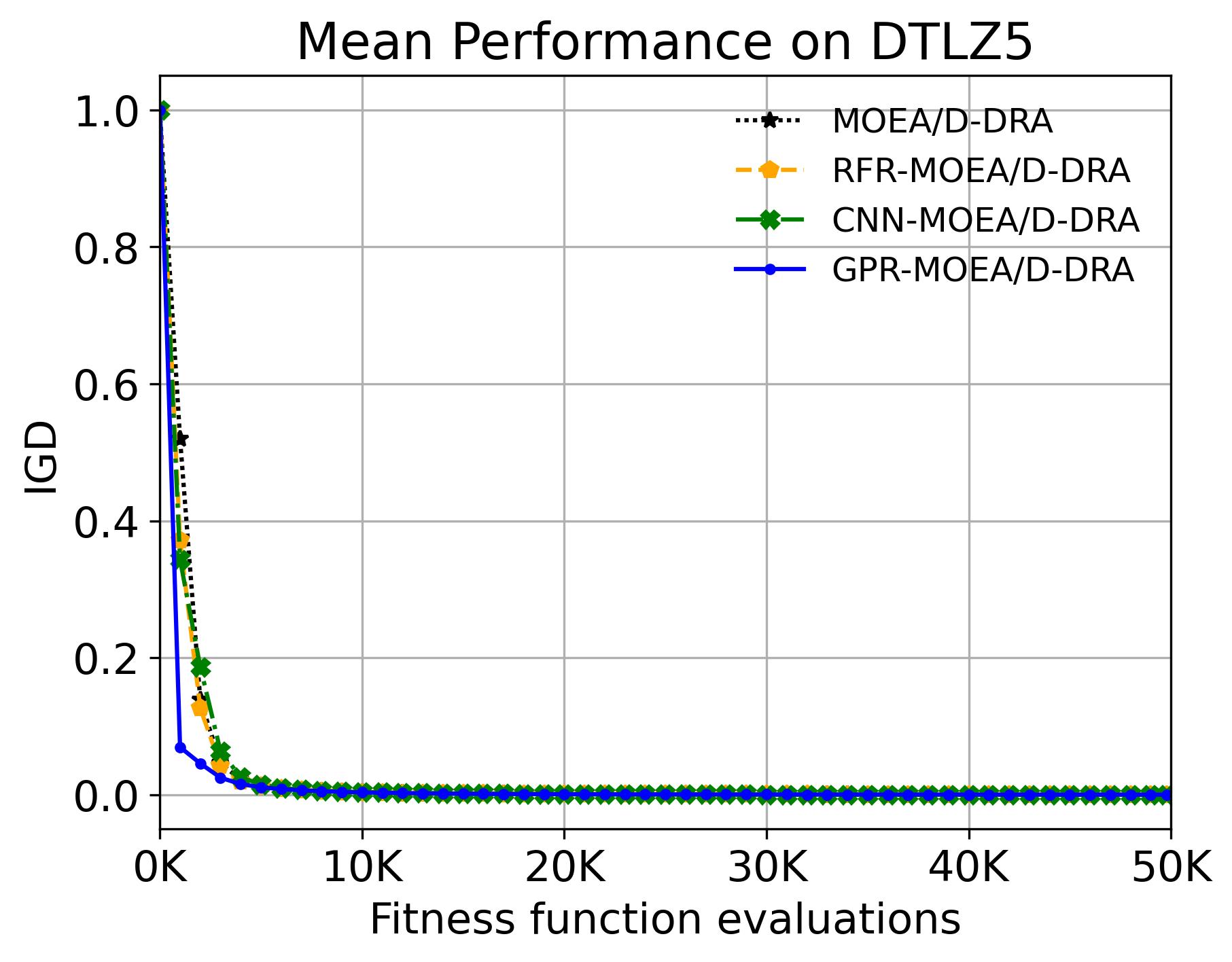}
    \end{subfigure}
    \begin{subfigure}[t]{\figwid\textwidth}
        \centering
        \includegraphics[width=\linewidth]{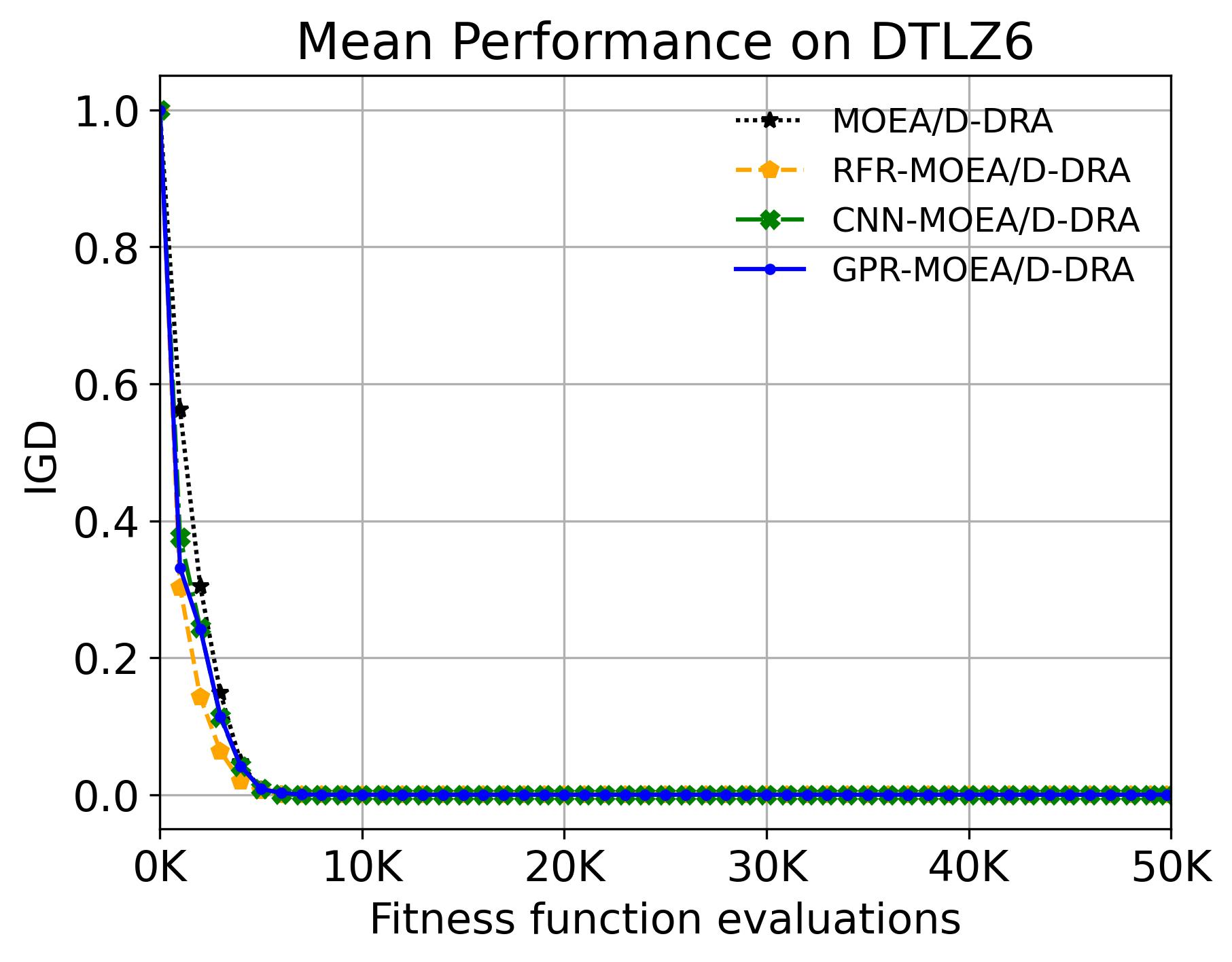}
    \end{subfigure}
    \begin{subfigure}[t]{\figwid\textwidth}
        \centering
        \includegraphics[width=\linewidth]{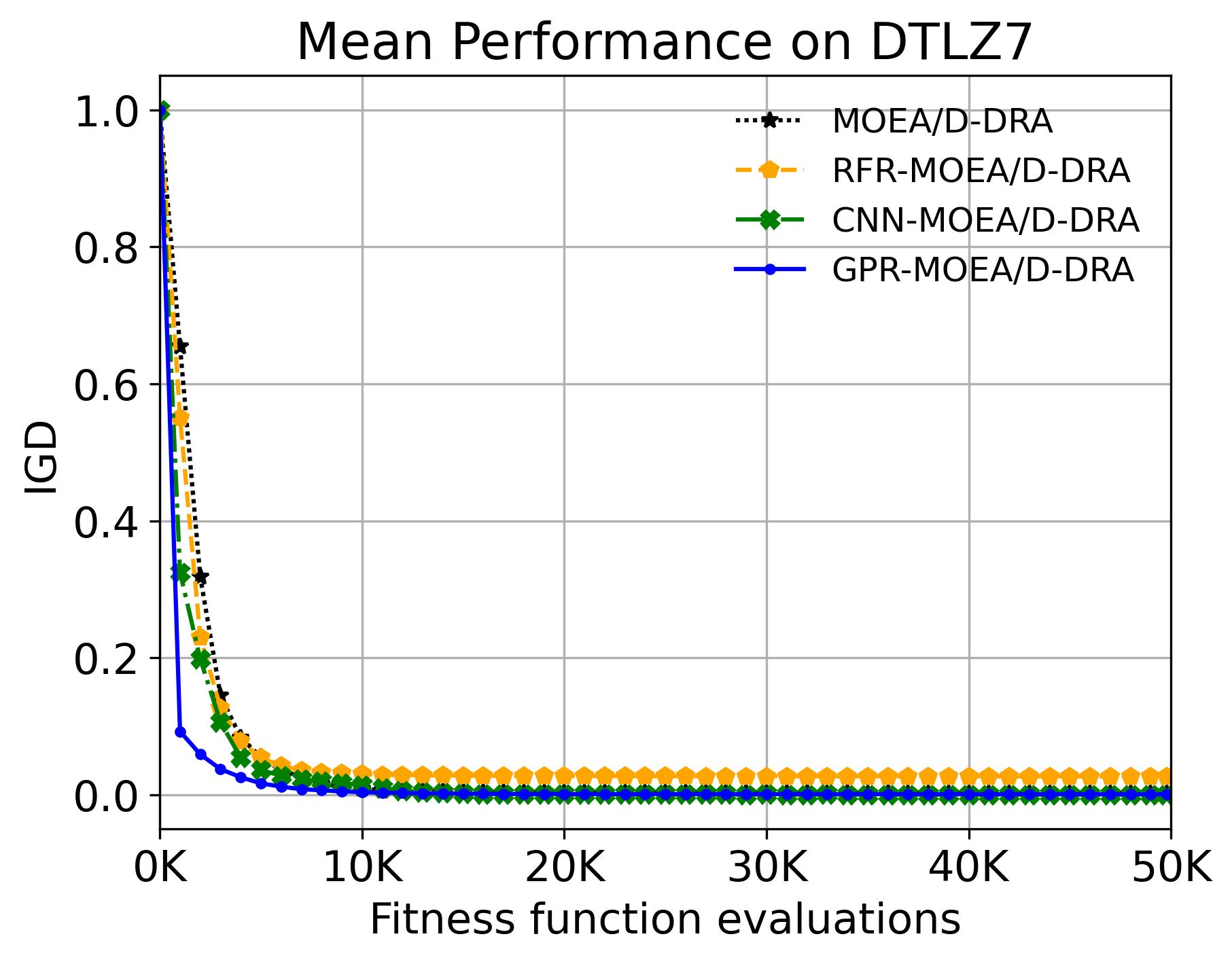}
    \end{subfigure}
    \begin{subfigure}[t]{\figwid\textwidth}
        \centering
        \includegraphics[width=\linewidth]{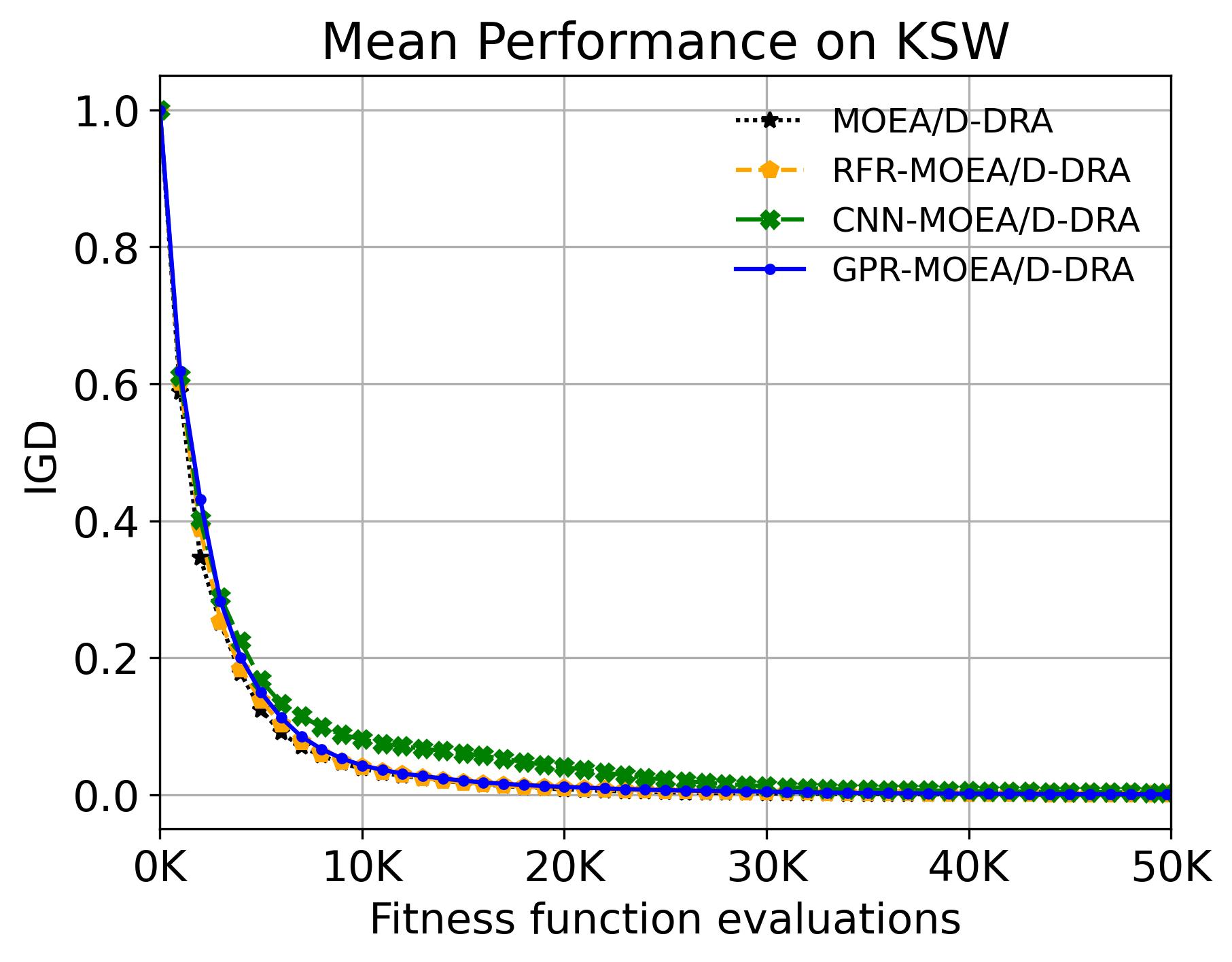}
    \end{subfigure}
    \begin{subfigure}[t]{\figwid\textwidth}
        \centering
        \includegraphics[width=\linewidth]{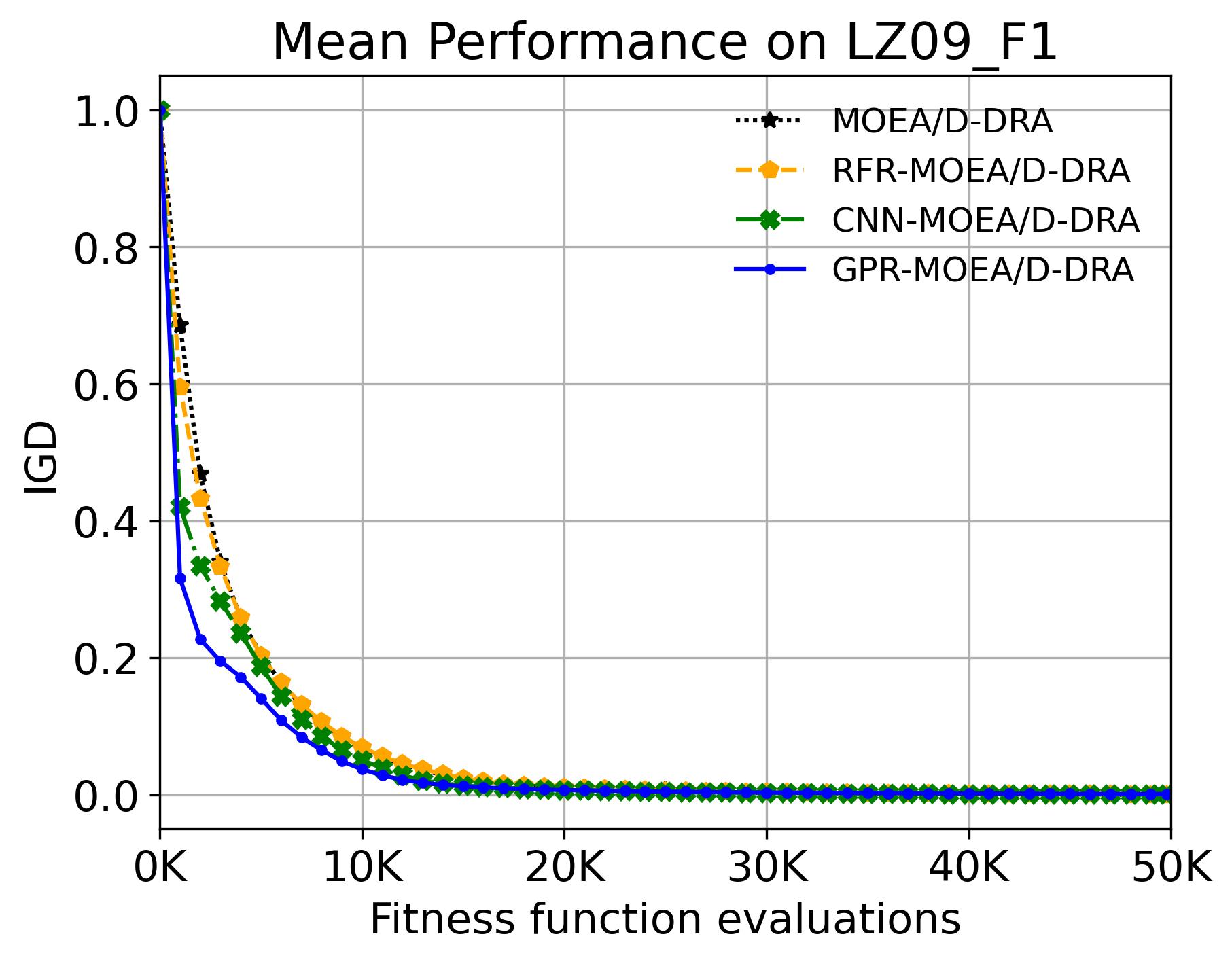}
    \end{subfigure}
    \begin{subfigure}[t]{\figwid\textwidth}
        \centering
        \includegraphics[width=\linewidth]{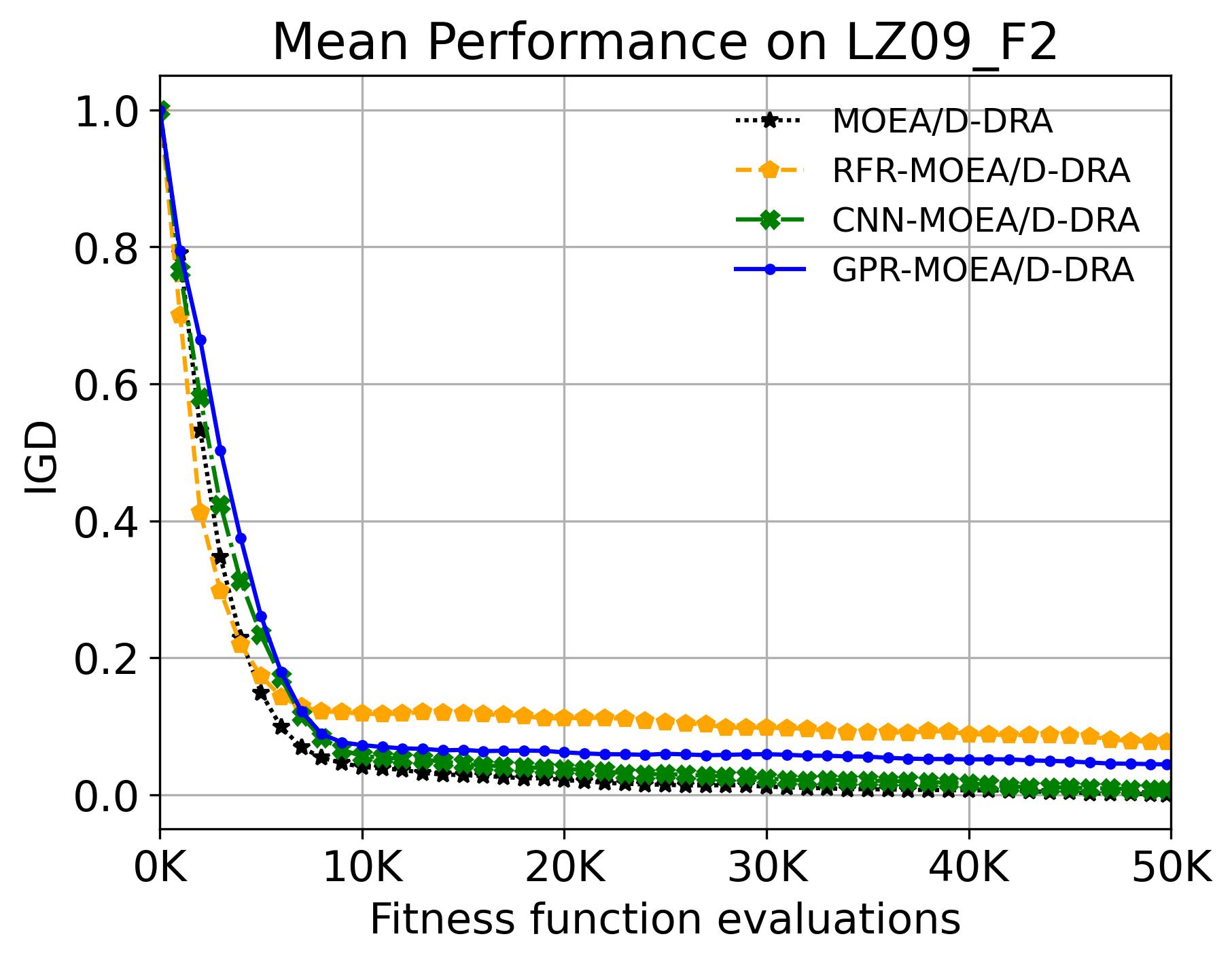}
    \end{subfigure}
    \begin{subfigure}[t]{\figwid\textwidth}
        \centering
        \includegraphics[width=\linewidth]{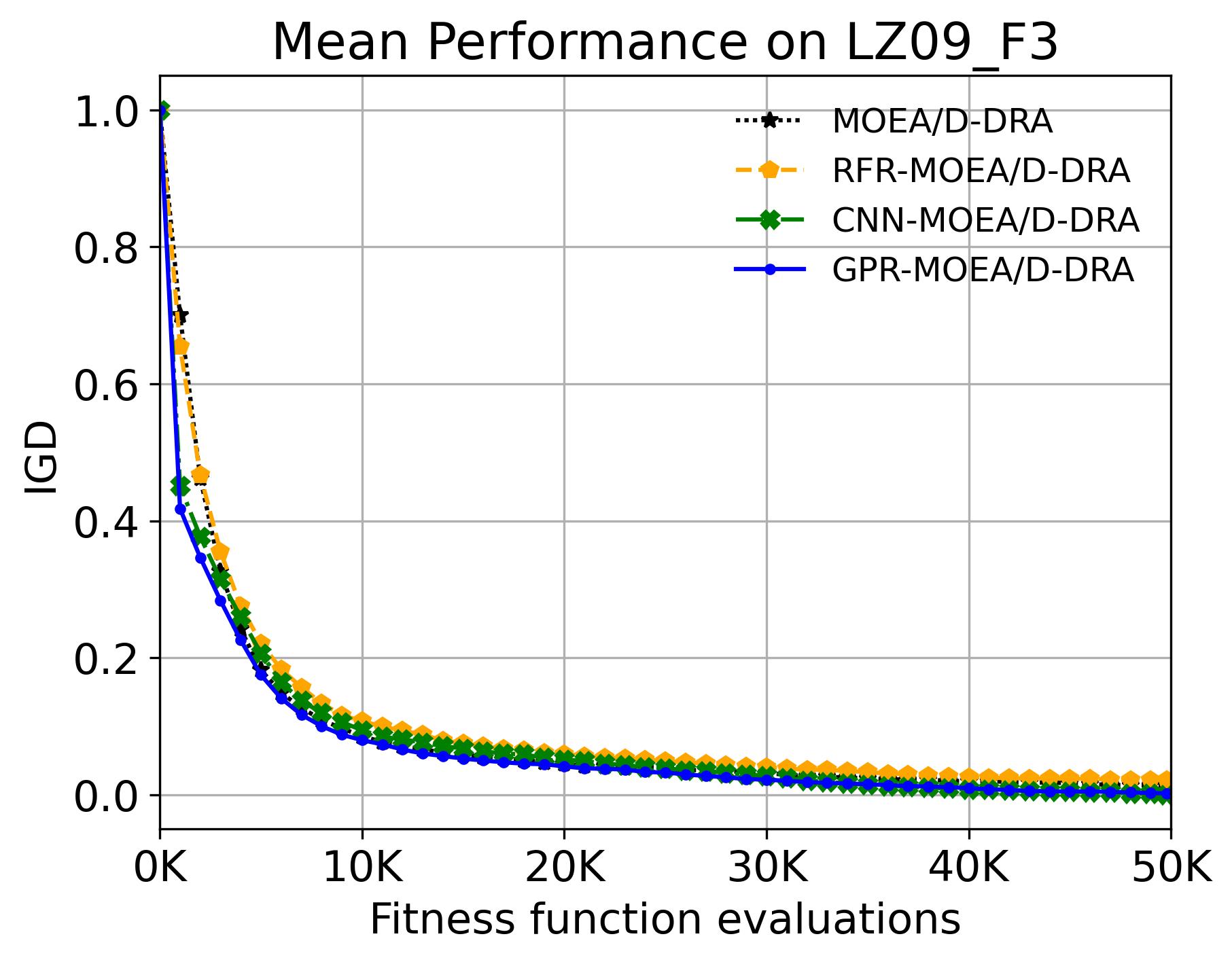}
    \end{subfigure}
    \begin{subfigure}[t]{\figwid\textwidth}
        \centering
        \includegraphics[width=\linewidth]{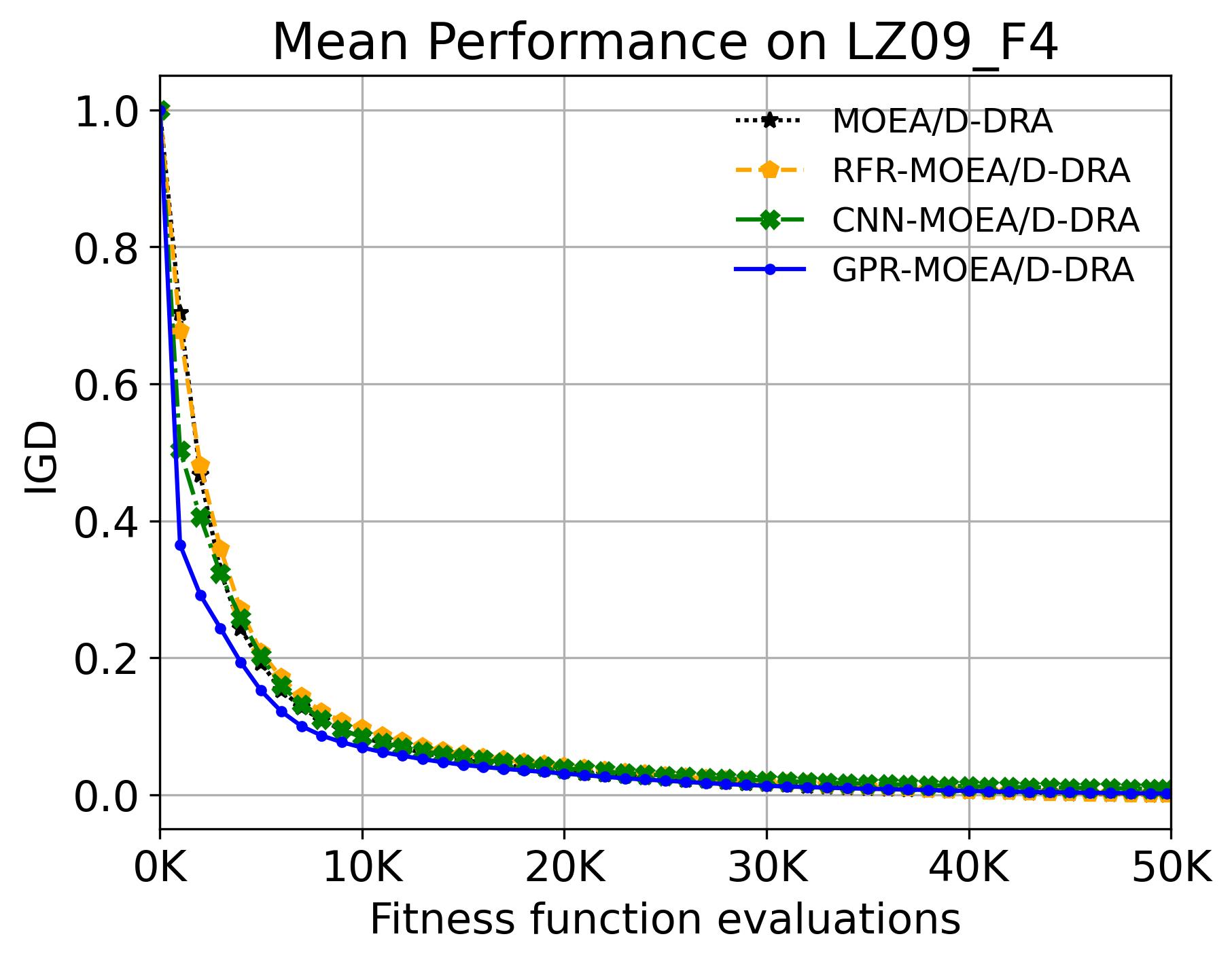}
    \end{subfigure}
    \begin{subfigure}[t]{\figwid\textwidth}
        \centering
        \includegraphics[width=\linewidth]{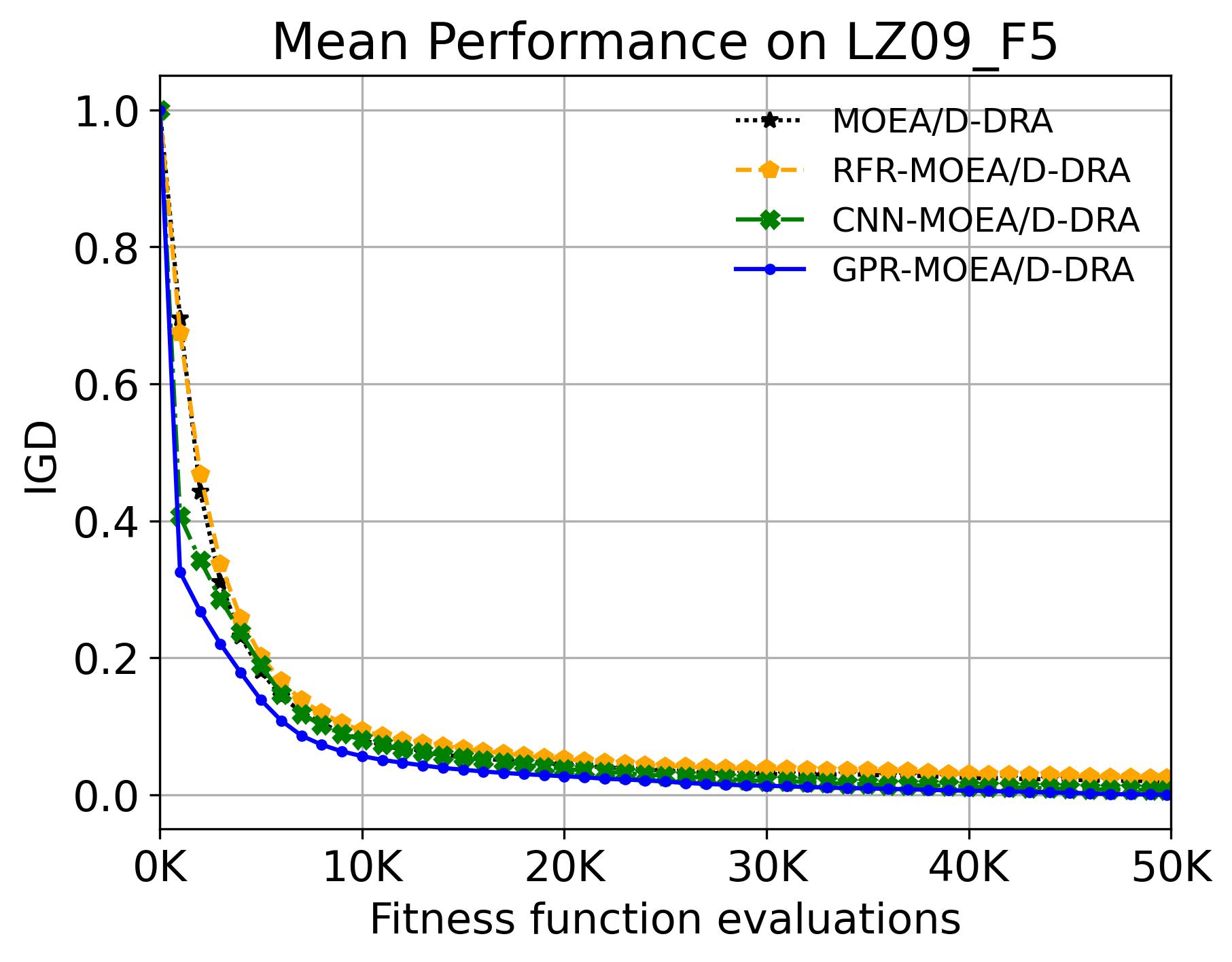}
    \end{subfigure}
    \begin{subfigure}[t]{\figwid\textwidth}
        \centering
        \includegraphics[width=\linewidth]{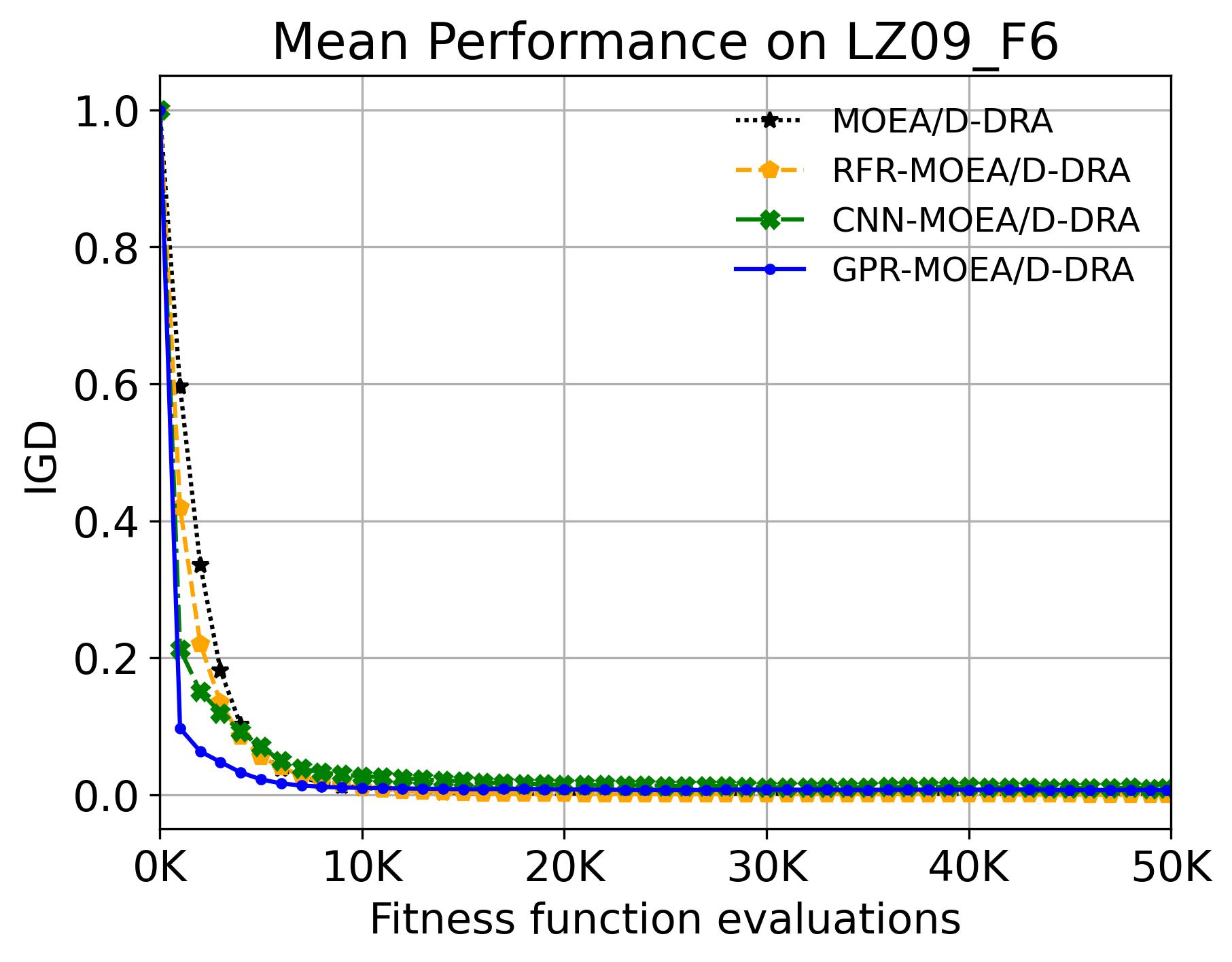}
    \end{subfigure}
    \begin{subfigure}[t]{\figwid\textwidth}
        \centering
        \includegraphics[width=\linewidth]{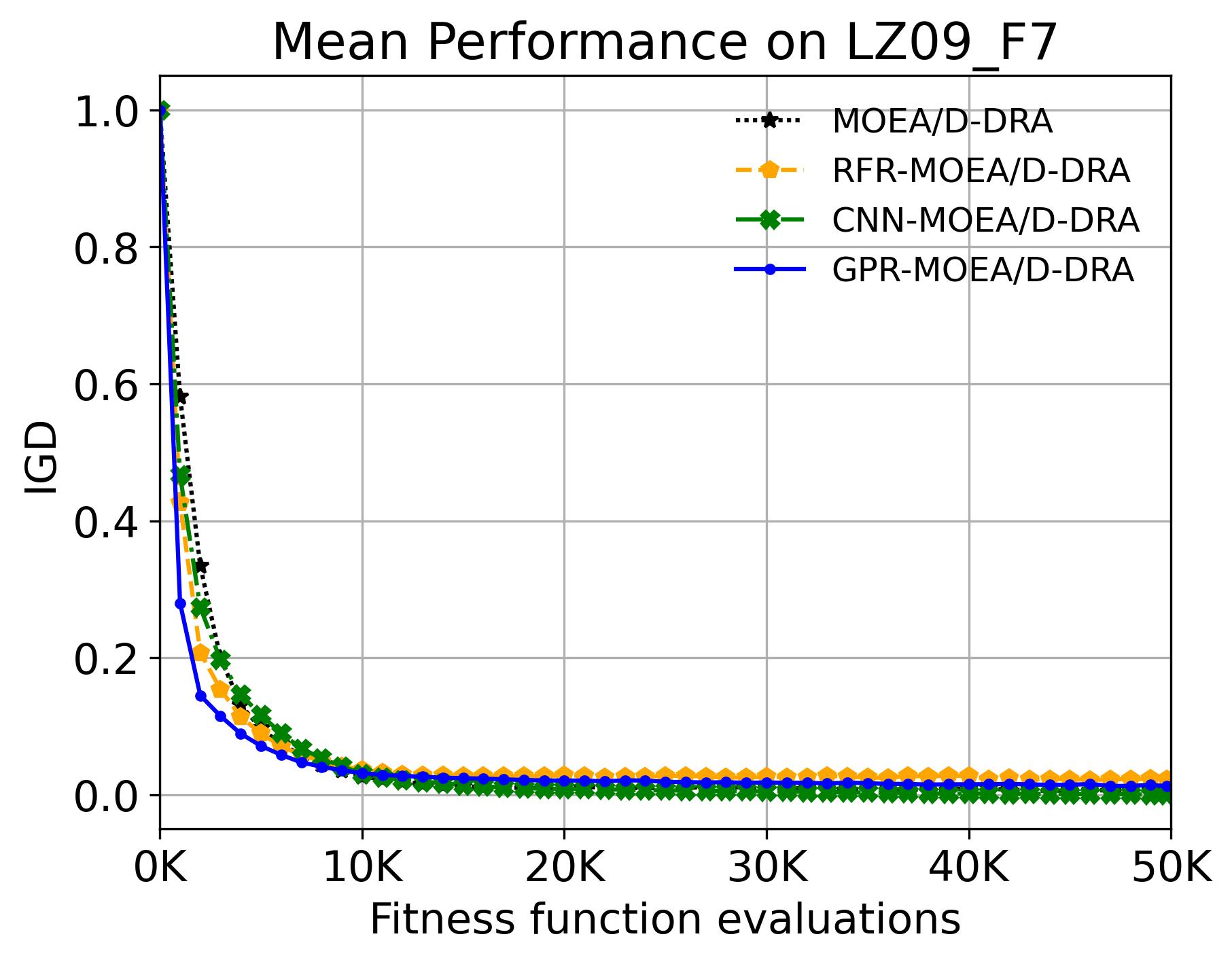}
    \end{subfigure}
    \begin{subfigure}[t]{\figwid\textwidth}
        \centering
        \includegraphics[width=\linewidth]{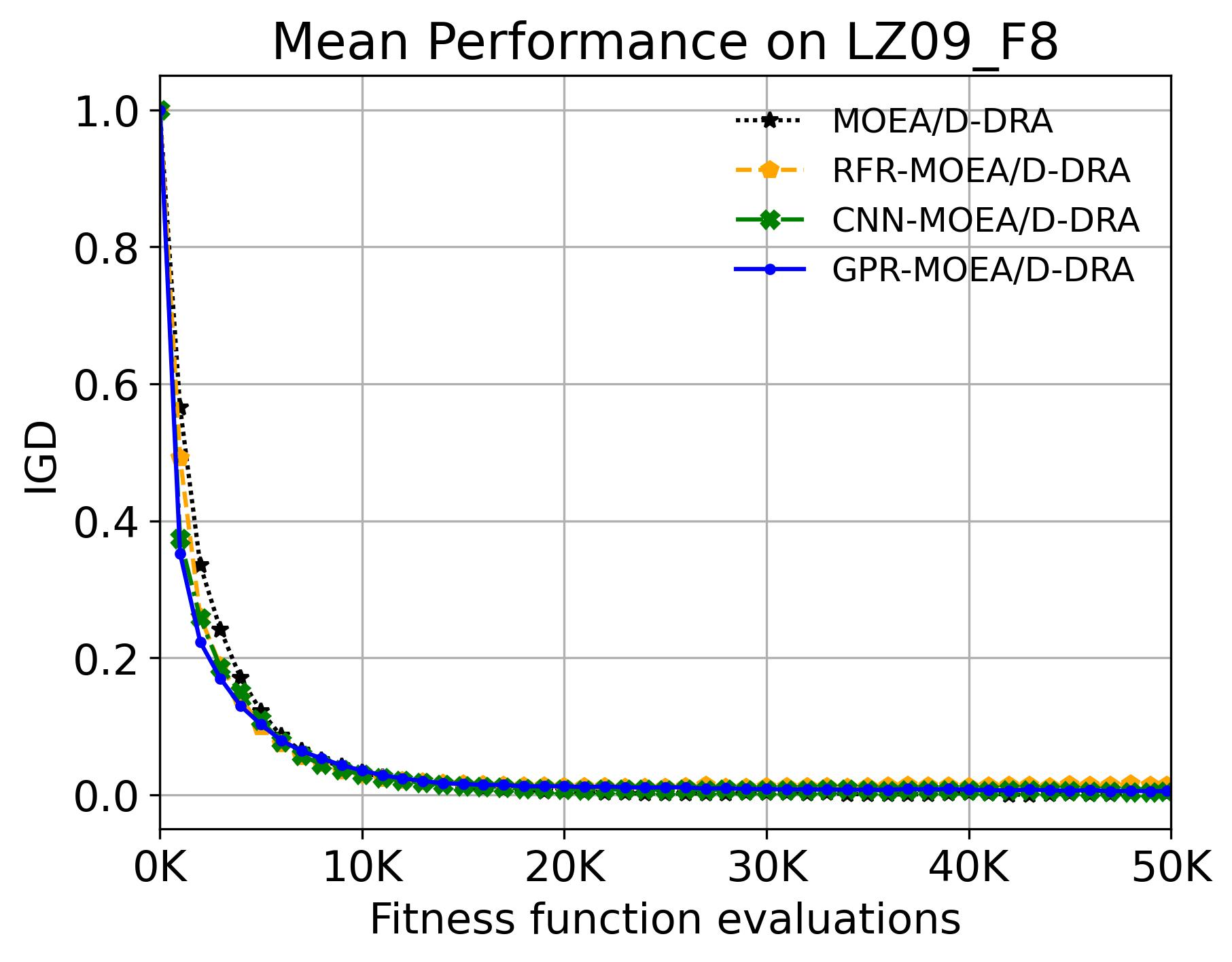}
    \end{subfigure}
    \begin{subfigure}[t]{\figwid\textwidth}
        \centering
        \includegraphics[width=\linewidth]{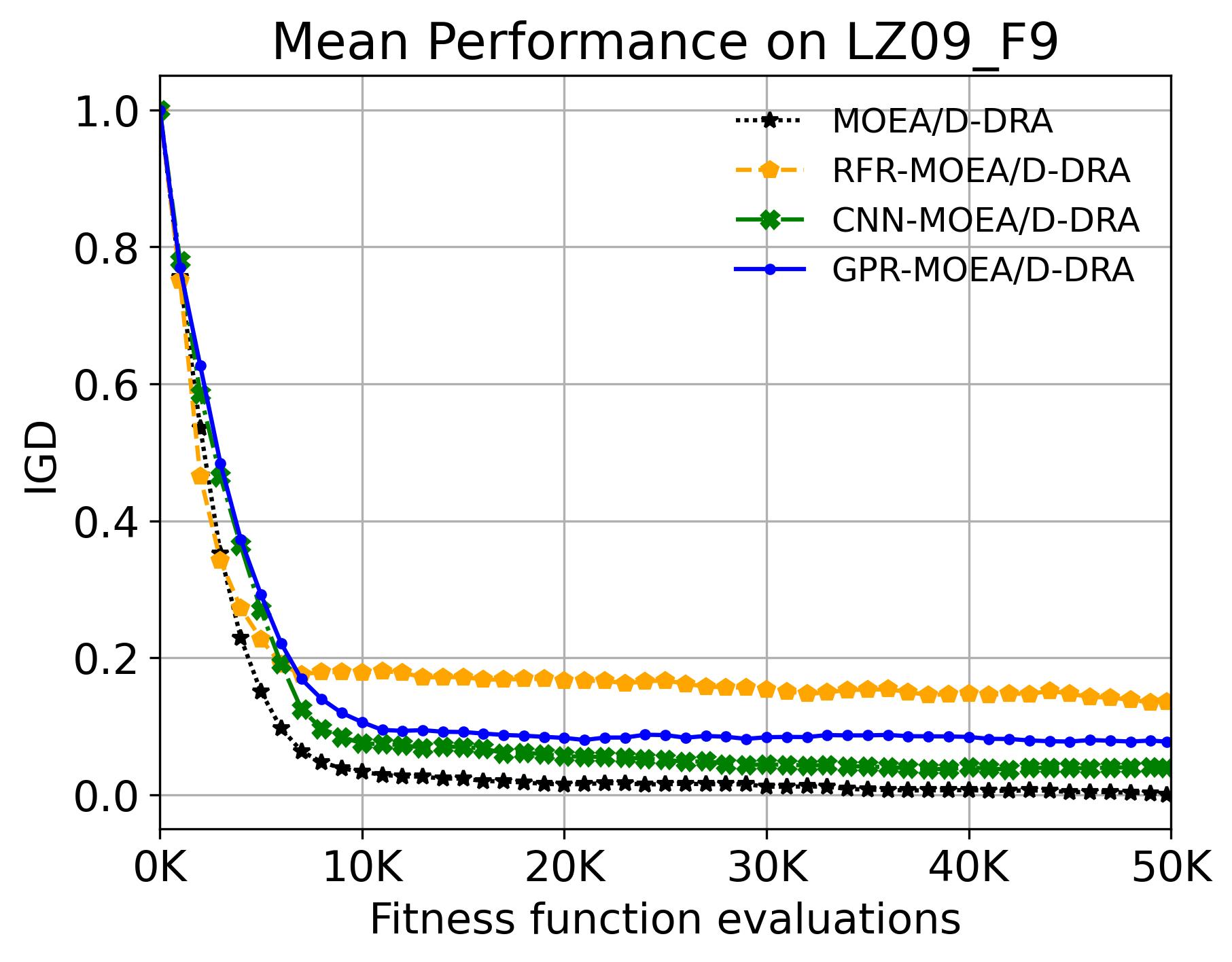}
    \end{subfigure}
    \begin{subfigure}[t]{\figwid\textwidth}
        \centering
        \includegraphics[width=\linewidth]{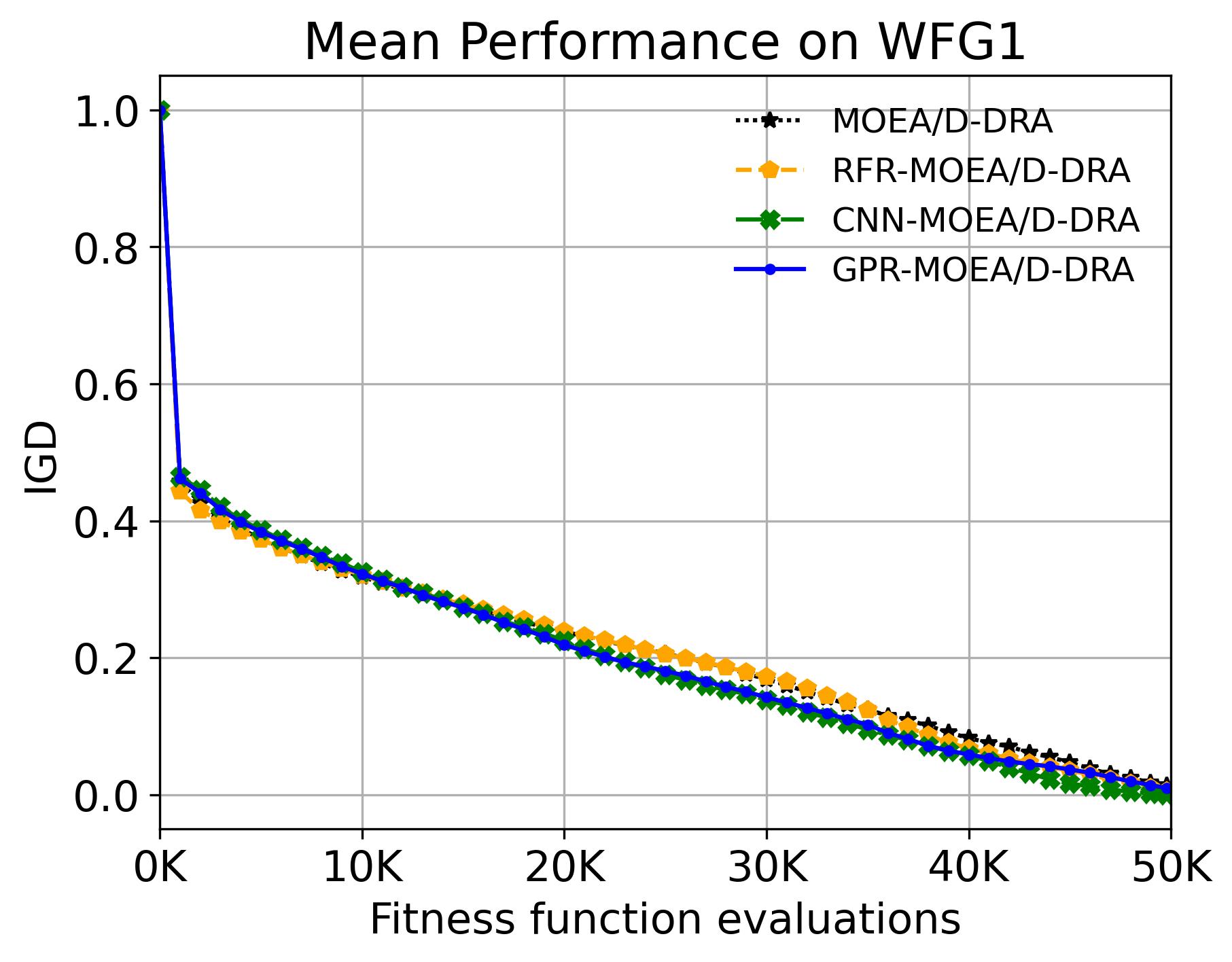}
    \end{subfigure}
    \begin{subfigure}[t]{\figwid\textwidth}
        \centering
        \includegraphics[width=\linewidth]{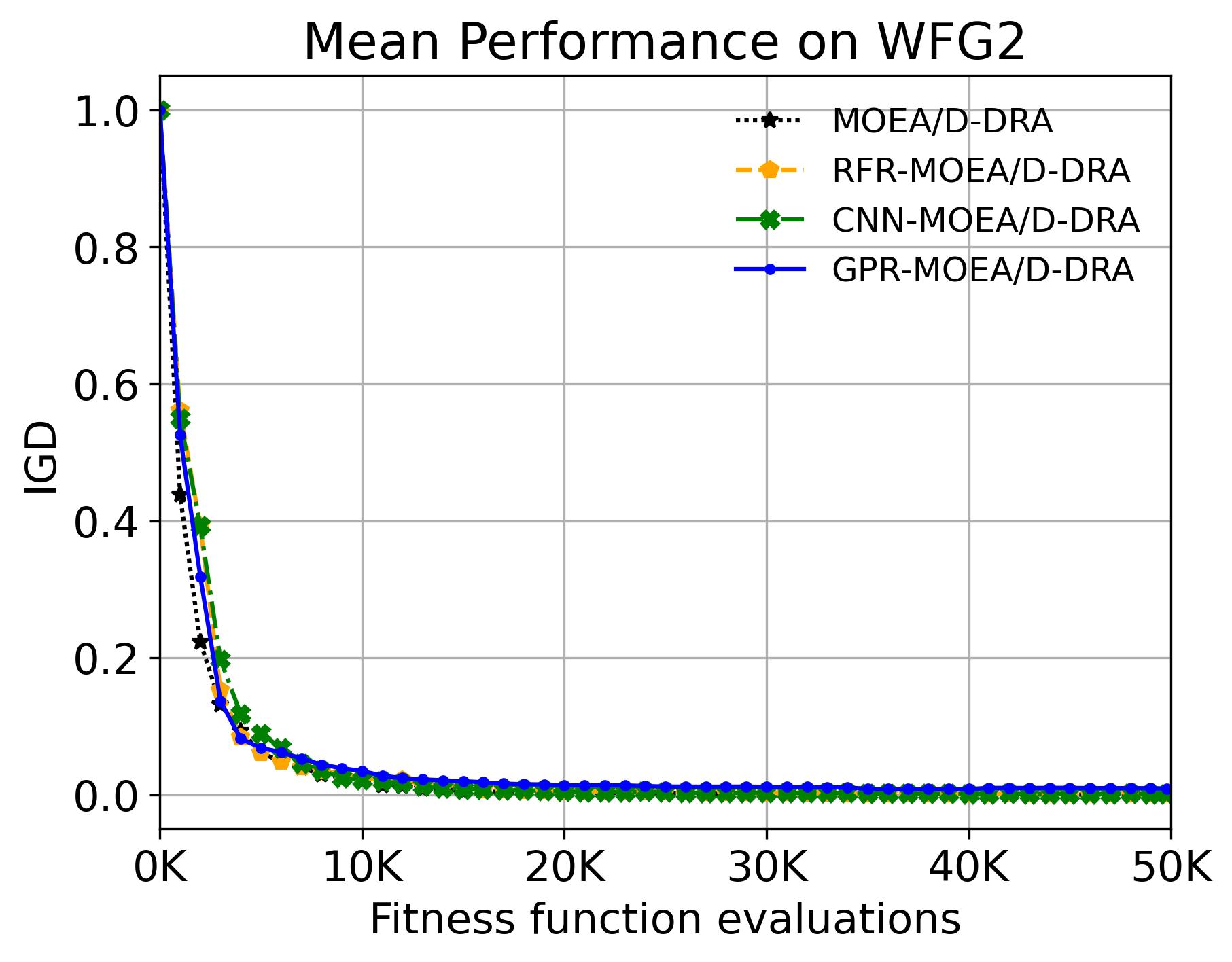}
    \end{subfigure}
    \begin{subfigure}[t]{\figwid\textwidth}
        \centering
        \includegraphics[width=\linewidth]{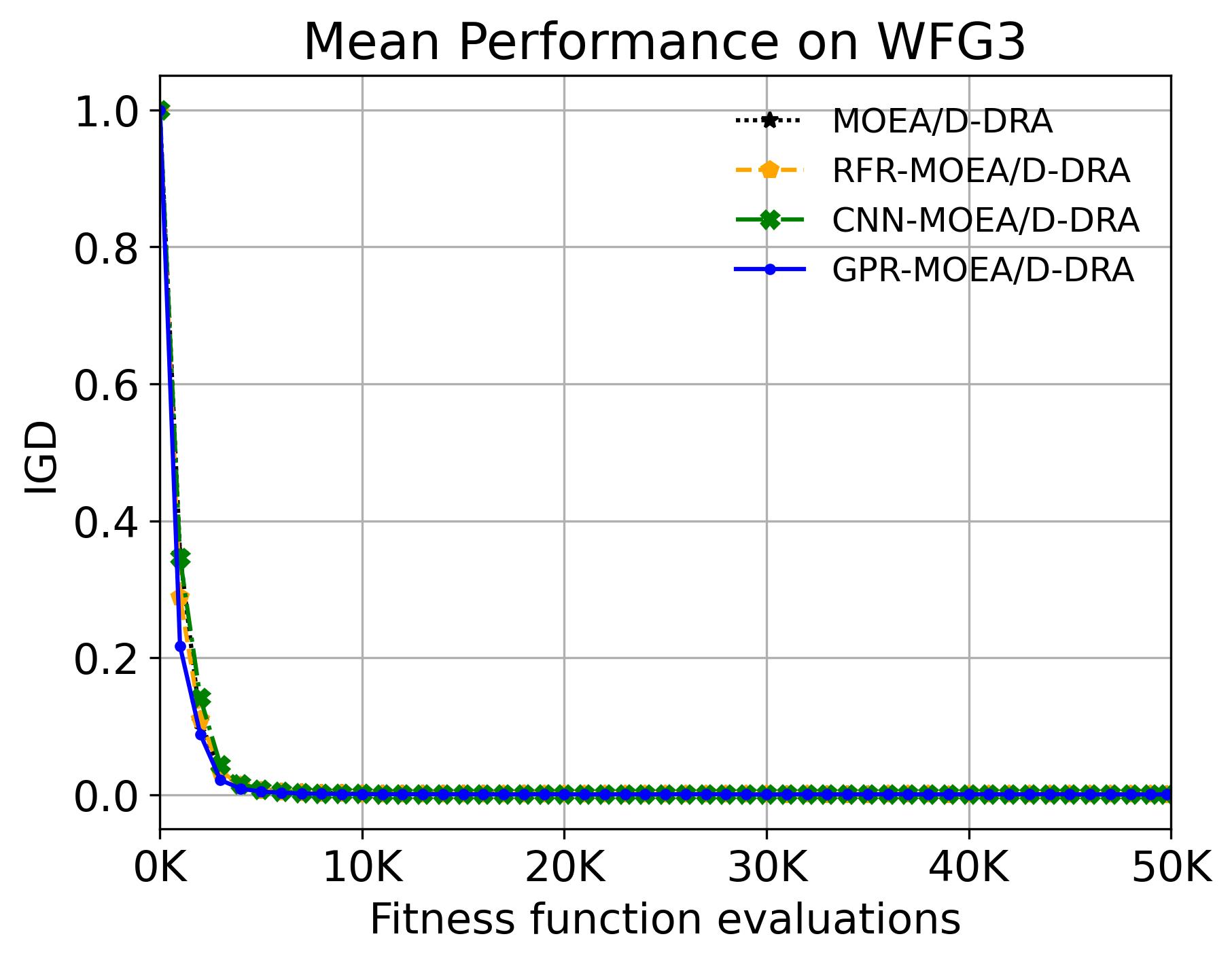}
    \end{subfigure}
    \begin{subfigure}[t]{\figwid\textwidth}
        \centering
        \includegraphics[width=\linewidth]{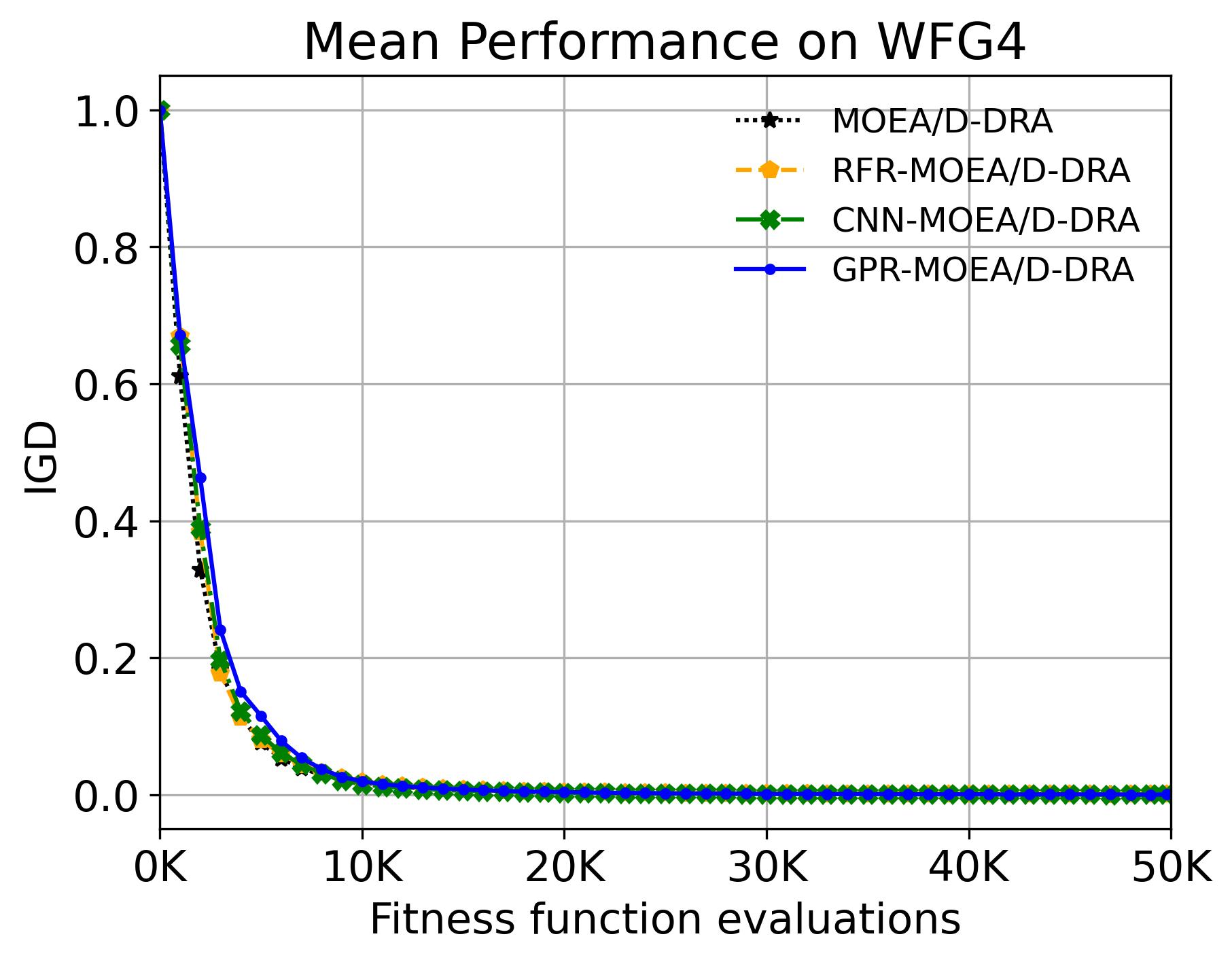}
    \end{subfigure}
    \begin{subfigure}[t]{\figwid\textwidth}
        \centering
        \includegraphics[width=\linewidth]{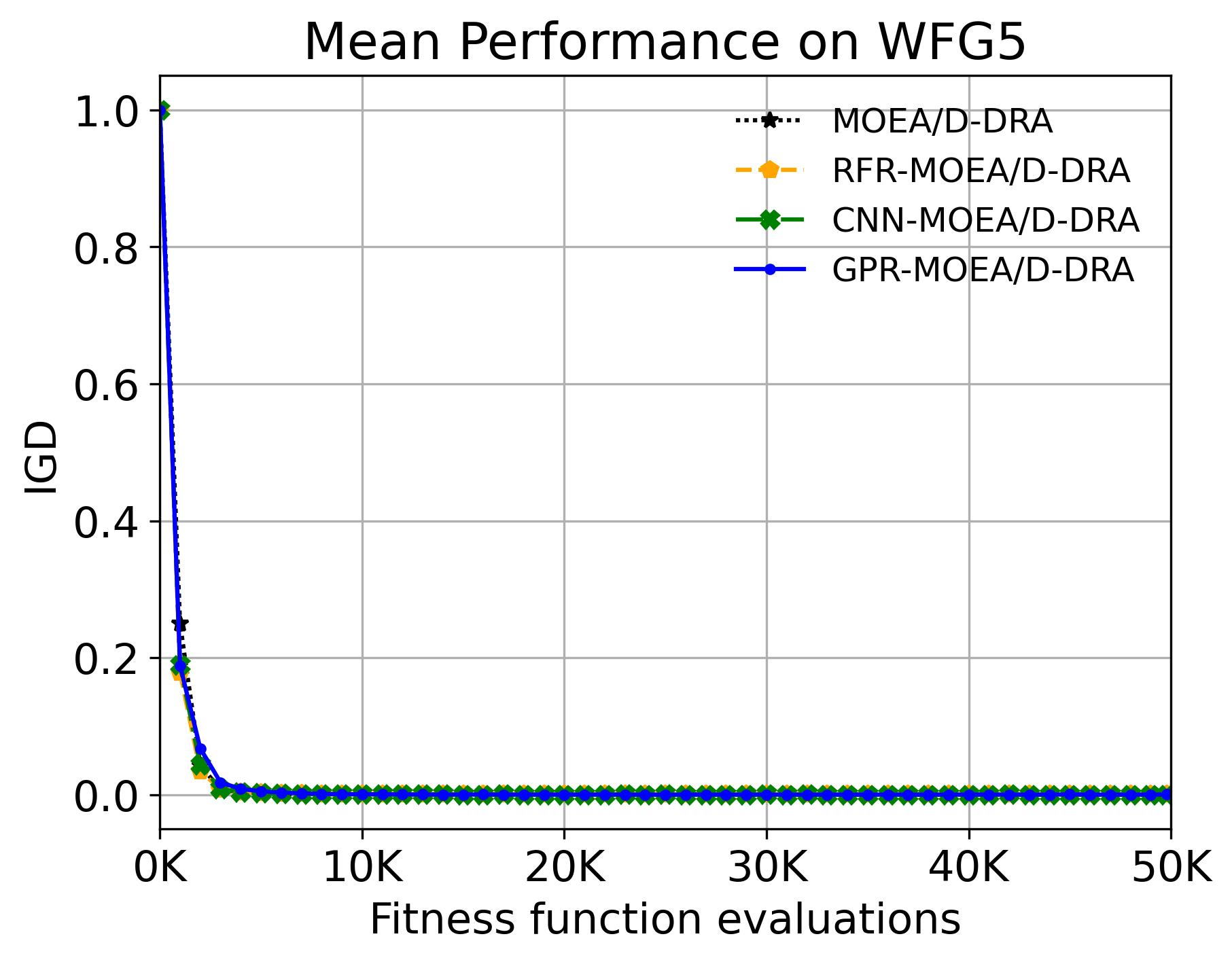}
    \end{subfigure}
    \begin{subfigure}[t]{\figwid\textwidth}
        \centering
        \includegraphics[width=\linewidth]{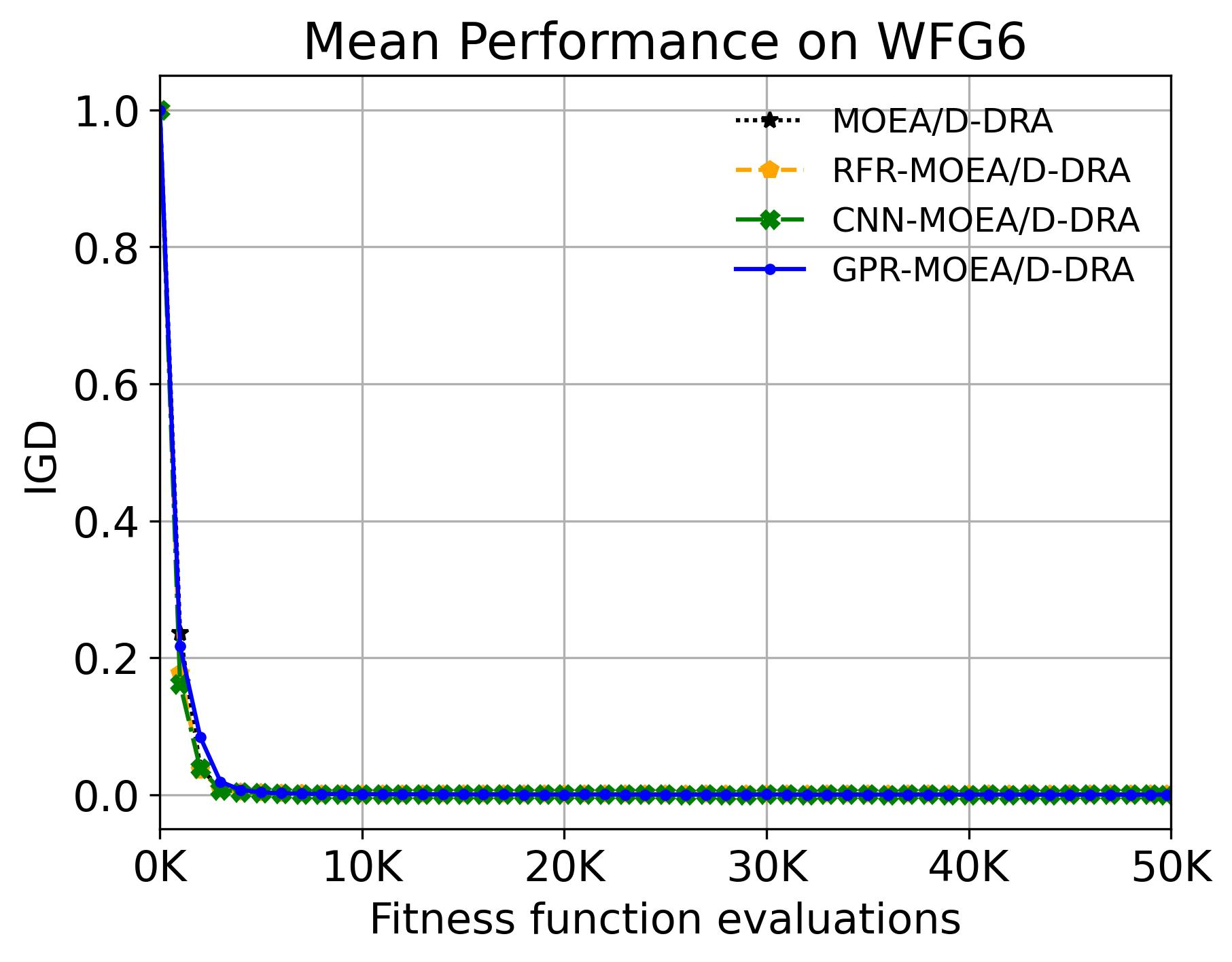}
    \end{subfigure}
    \begin{subfigure}[t]{\figwid\textwidth}
        \centering
        \includegraphics[width=\linewidth]{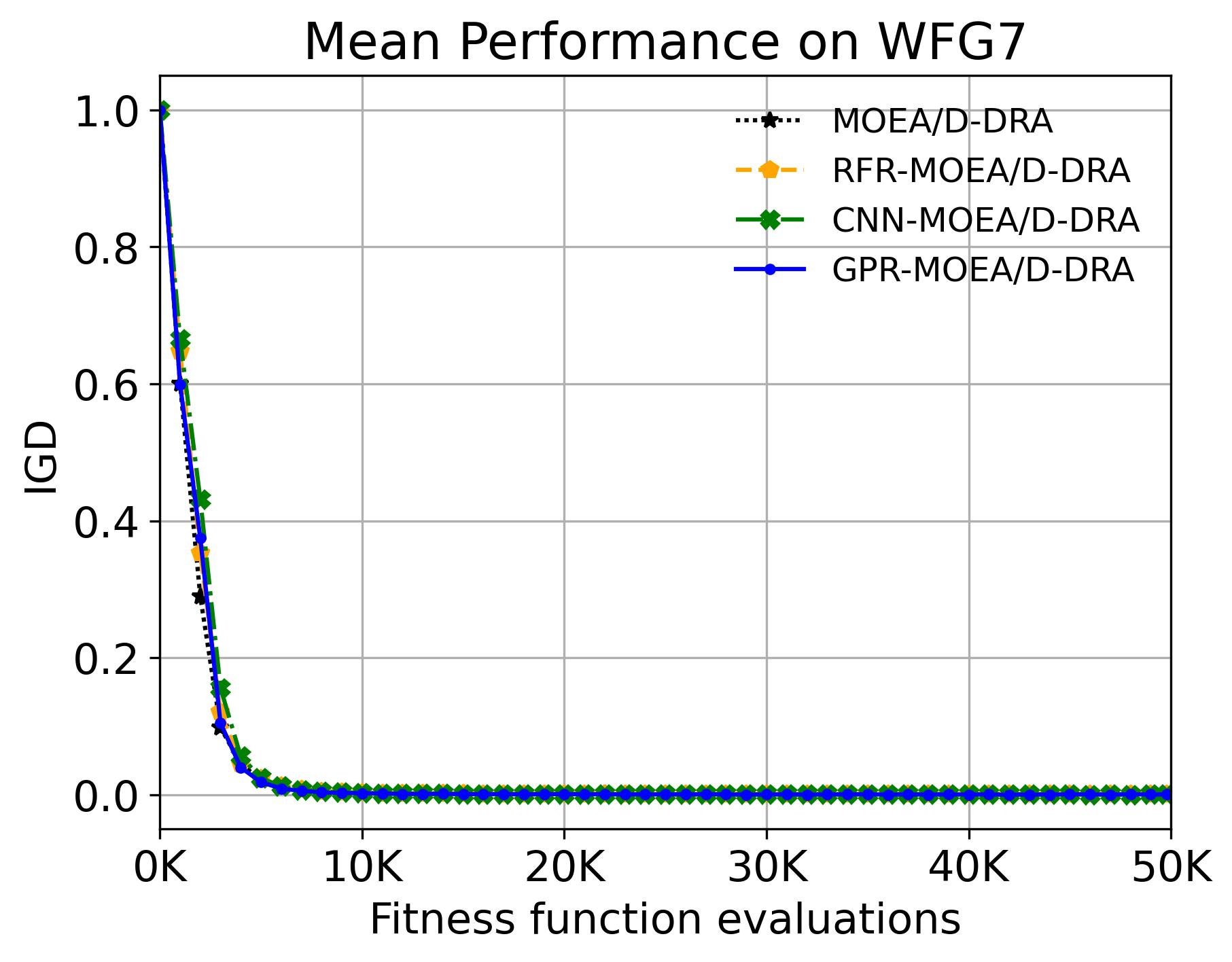}
    \end{subfigure}
    \begin{subfigure}[t]{\figwid\textwidth}
        \centering
        \includegraphics[width=\linewidth]{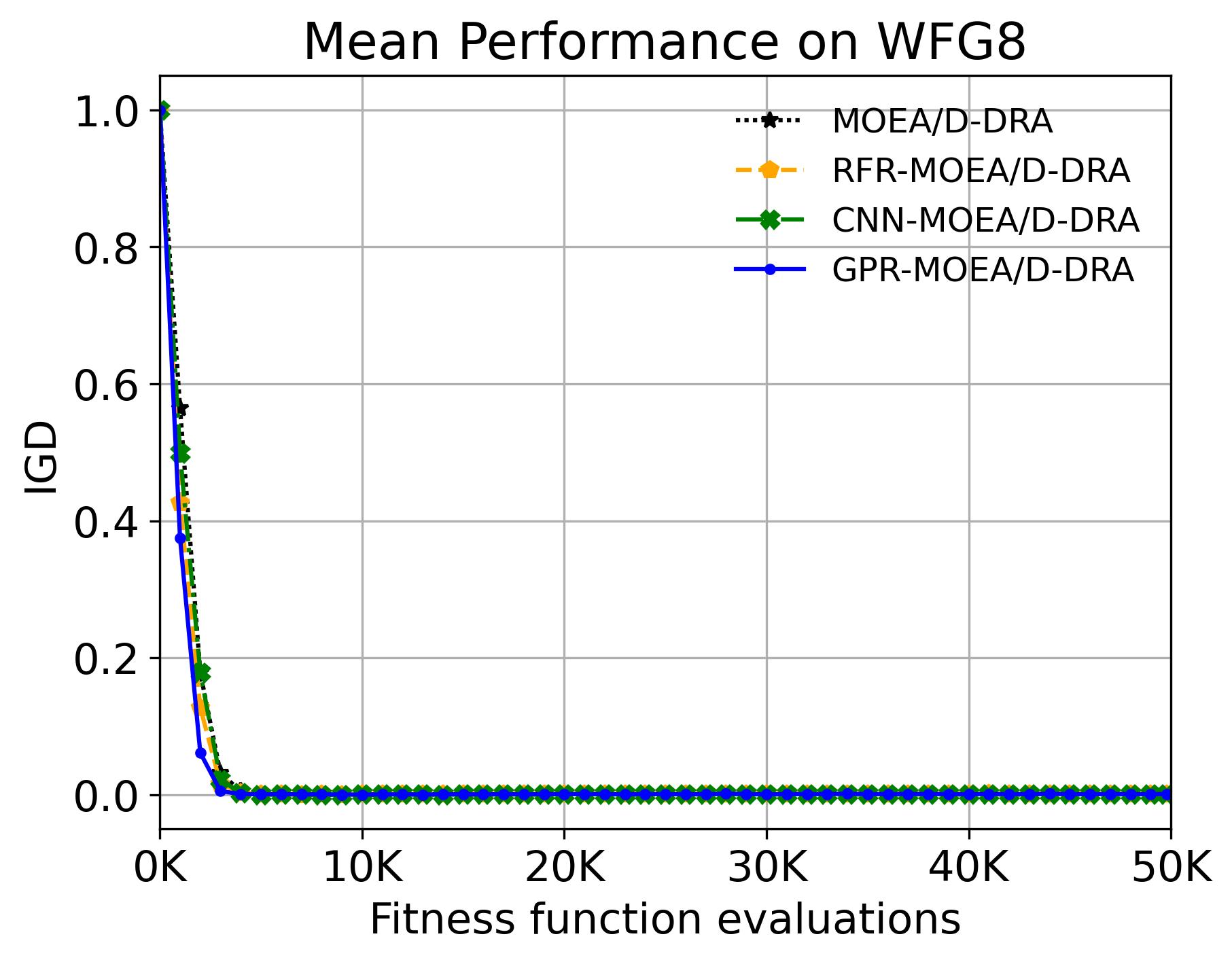}
    \end{subfigure}
    \begin{subfigure}[t]{\figwid\textwidth}
        \centering
        \includegraphics[width=\linewidth]{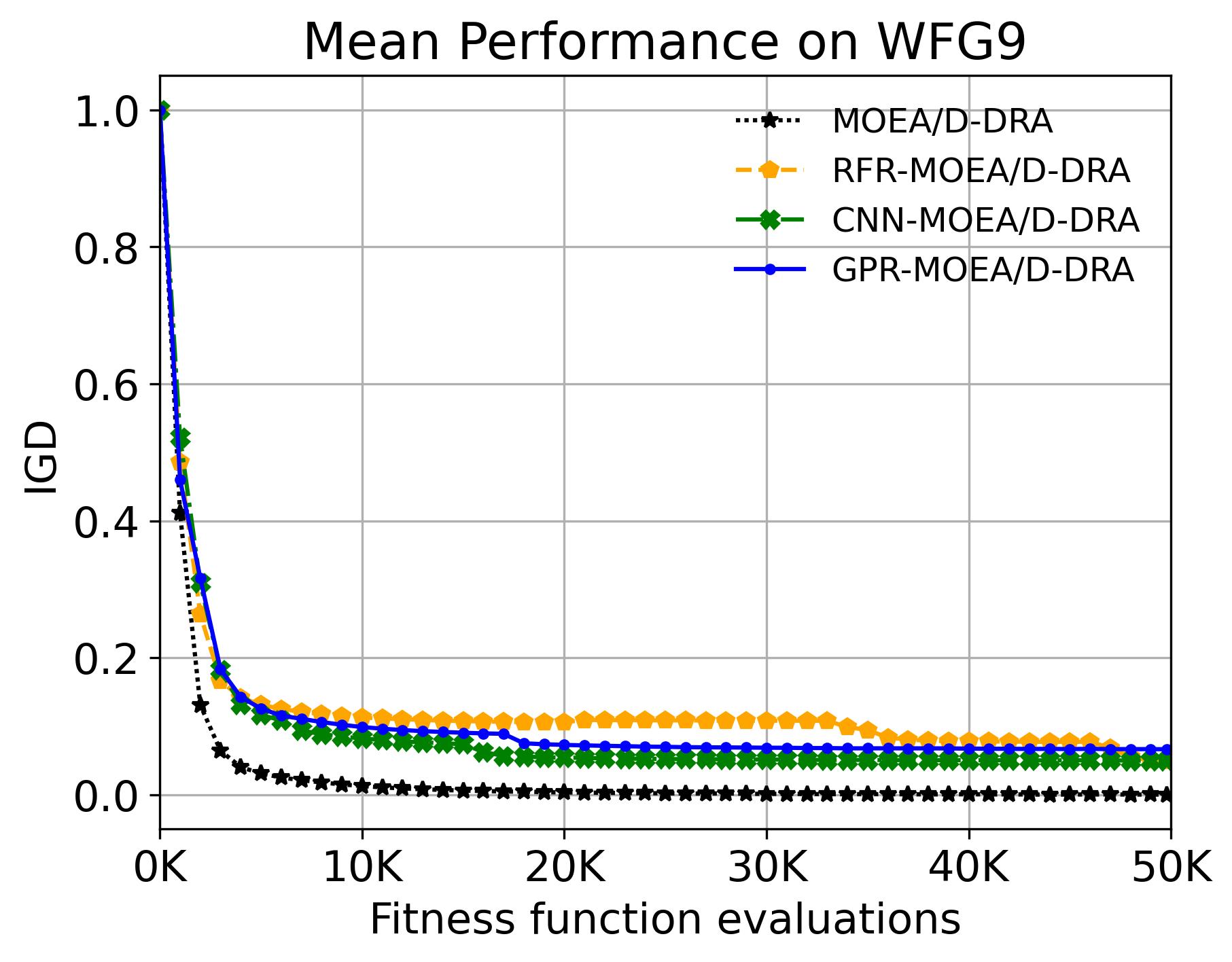}
    \end{subfigure}
    \begin{subfigure}[t]{\figwid\textwidth}
        \centering
        \includegraphics[width=\linewidth]{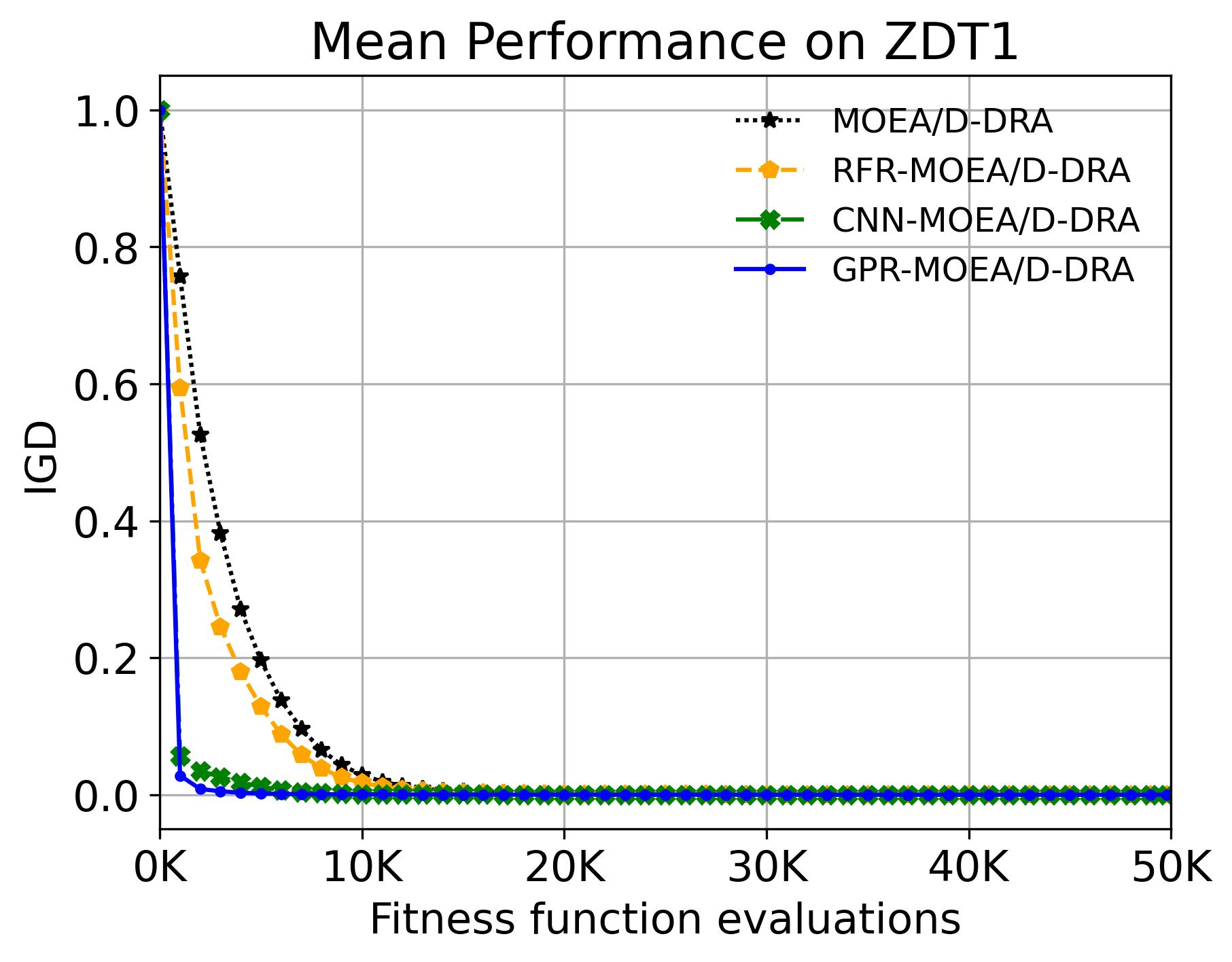}
    \end{subfigure}
    \begin{subfigure}[t]{\figwid\textwidth}
        \centering
        \includegraphics[width=\linewidth]{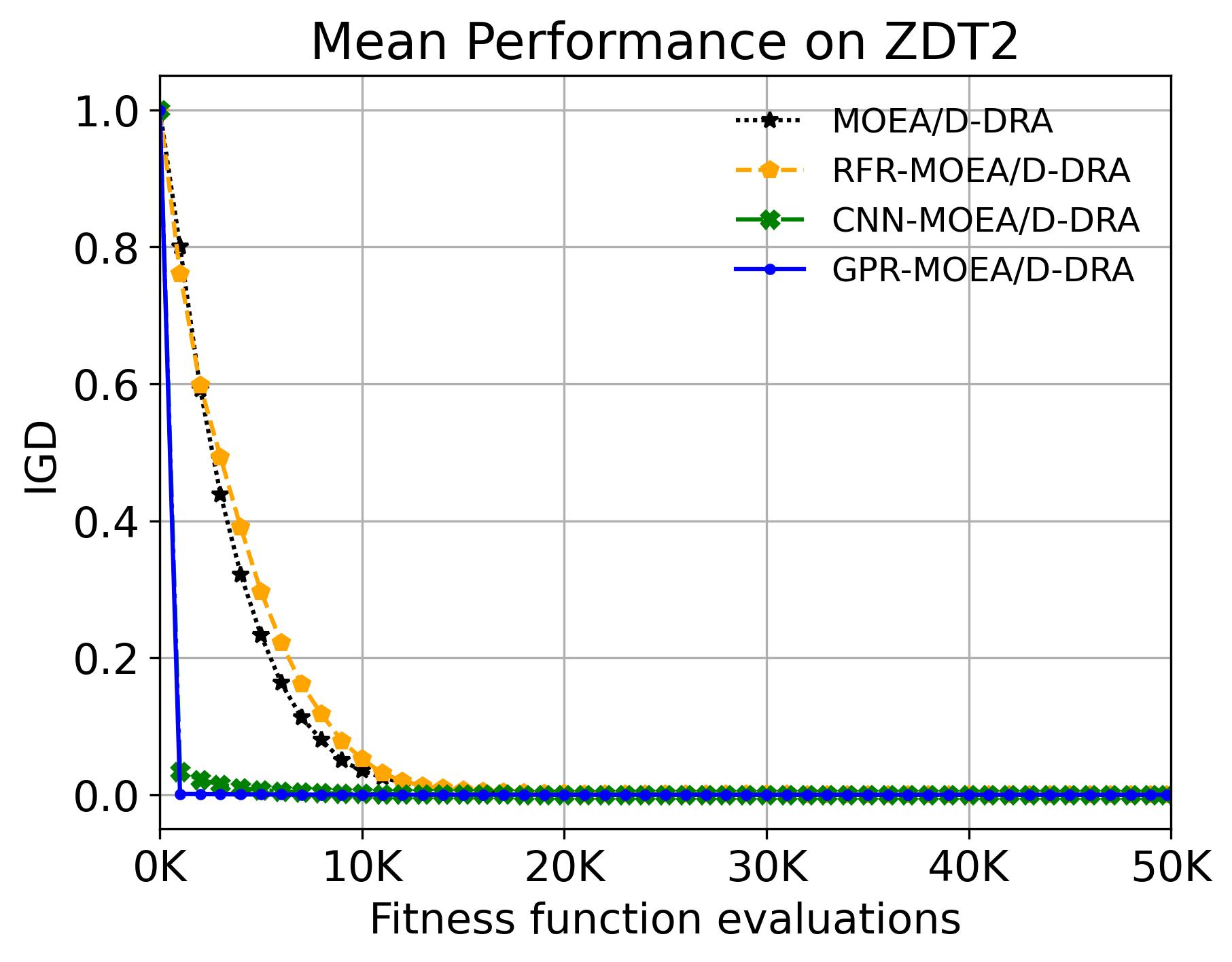}
    \end{subfigure}
    \begin{subfigure}[t]{\figwid\textwidth}
        \centering
        \includegraphics[width=\linewidth]{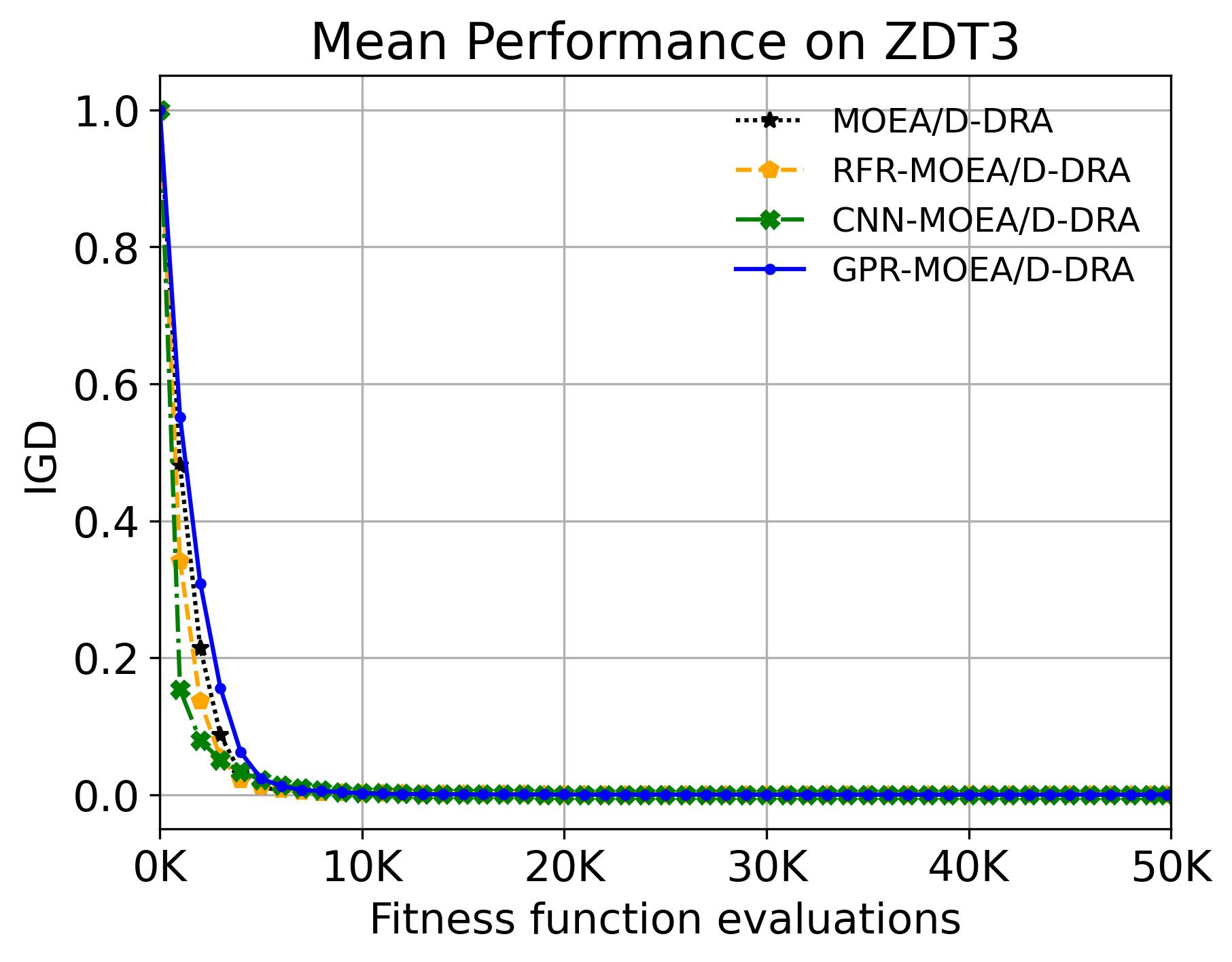}
    \end{subfigure}
    \begin{subfigure}[t]{\figwid\textwidth}
        \centering
        \includegraphics[width=\linewidth]{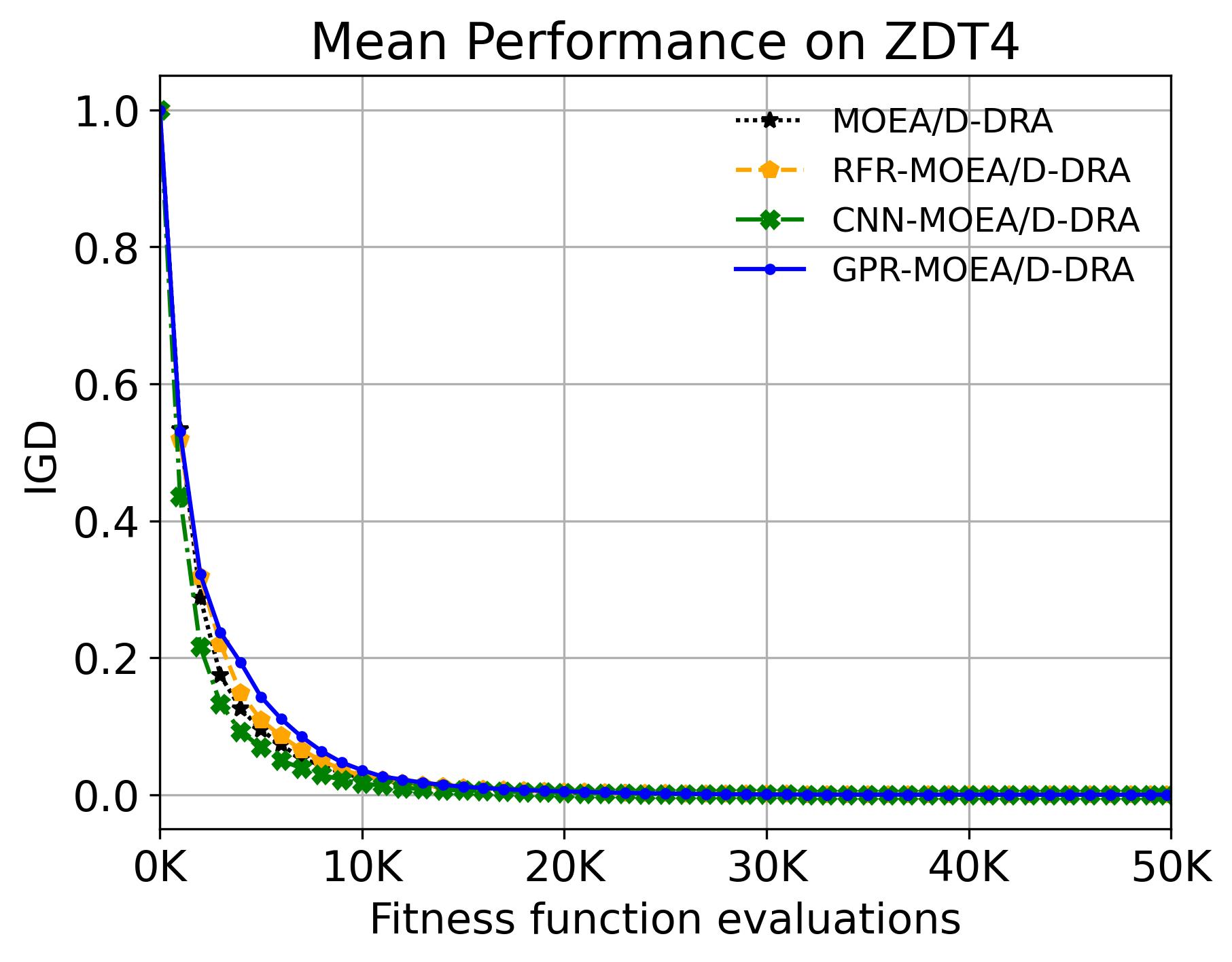}
    \end{subfigure}
    \begin{subfigure}[t]{\figwid\textwidth}
        \centering
        \includegraphics[width=\linewidth]{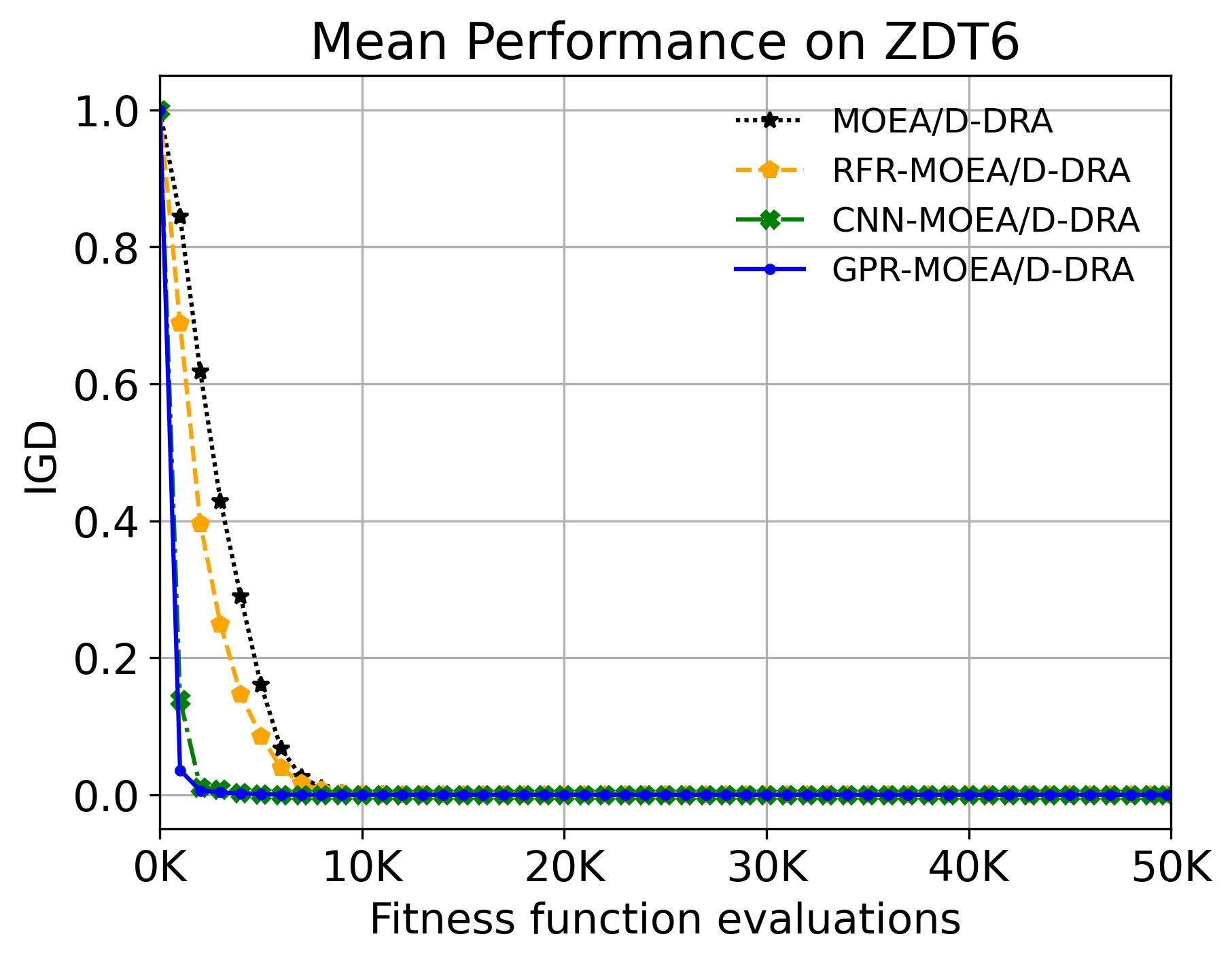}
    \end{subfigure}

        \caption{Comparison of MOEA/D and its associated surrogate-enhanced solvers on individual benchmark problems using  $IGD(PF_c)$ -- i.e., the normalised inverse generational distance -- as a performance indicator.}
        \label{fig:annexMOEAD_igd}
\end{figure*}

\end{document}